\documentclass[runningheads]{llncs}

\usepackage[mobile]{accv}

\usepackage{accvabbrv}
\usepackage{svg}

\usepackage[accsupp]{axessibility}  %

\usepackage{xr-hyper} 
\usepackage{hyperref}

\usepackage{orcidlink}

\usepackage{stfloats}
\usepackage{subcaption}
\usepackage{xcolor}
\usepackage{xspace}
\usepackage{graphicx}
\usepackage{booktabs}
\usepackage[table]{xcolor}

\newcommand{\cc}[2]{\multicolumn{1}{>{\columncolor[HTML]{#1}\centering\arraybackslash}c}{#2}}

\newcommand{\PAR}[1]{\vskip4pt \noindent{\bf #1~}}

\begin{document}

\title{The Role of Initialization in 3D Gaussian Splatting}

\author{Ivan Desiatov\inst{1,2}\orcidlink{0009-0000-8416-2098} \and
Torsten Sattler\inst{2}\orcidlink{0000-0001-9760-4553}}

\authorrunning{I.~Desiatov, T.~Sattler}

\institute{
Czech Technical University in Prague, Faculty of Electrical Engineering \and Czech Technical University in Prague, Czech Institute of Informatics, Robotics and Cybernetics }

\maketitle

\begin{abstract}
3D Gaussian Splatting (3DGS) has become the method of choice for photo-realistic novel view synthesis (NVS), due to its efficiency and compelling visual quality. 3DGS represents the scene as a set of 3D Gaussians, parameterized by their position, spatial extent, and view-dependent color. Starting from an initial point cloud, 3DGS refines the Gaussians' parameters to reconstruct a set of training images as accurately as possible. Typically, a sparse Structure-from-Motion point cloud is used as initialization. Thus, in order to obtain a full scene representation, 3DGS methods rely on a densification stage. In this paper, we systematically study how initialization affects 3DGS NVS performance and geometric quality, using several densification strategies. We show that dense initialization does \textit{not} lead to consistent visual improvements when paired with strong densification. Despite that, it helps in generalization to off-trajectory views and significantly improves geometric accuracy of the scenes. Our code is available at https://github.com/deivse/ivd\_splat.
\end{abstract}

\section{Introduction}
\label{sec:intro}
Accurately reconstructing the 3D structure of a scene from a set of images or a video stream is a long-standing problem in computer vision, with applications in cultural heritage documentation, asset generation for entertainment, and robotics (\eg, for path planning and navigation). 
The field of 3D reconstruction from images has received significant attention in the last few years due to the emergence of efficient learning-based reconstruction techniques, especially neural radiance fields (NeRFs)~\cite{nerf} and 3D Gaussian Splatting (3DGS)~\cite{3dgs}. 

3DGS represents a scene as a collection of 3D Gaussians, each defined by a position, covariance matrix, opacity, and a view-dependent color function.
To bootstrap optimization, 3DGS relies on an initial estimate of the Gaussian positions, typically provided by a sparse point cloud from Structure-from-Motion (SfM).
A crucial part of any 3DGS approach is thus a densification stage, which creates new Gaussians to improve fidelity and scene coverage~\cite{3dgs, 3dgs_mcmc, absgs, minisplatting, scaffold_gs, sugar}. Consequently, improving the densification stage has been a research focus~\cite{3dgs_mcmc, absgs, bogauss, idhfr, minisplatting}. 
In most cases, these densification strategies have been designed and evaluated with SfM or random initialization.
Another line of research, especially in the context of few-view reconstruction~\cite{instantsplat, initialize_to_generalize_sparse_view, mvpgs_sparse_view, hbsplat_sparse_view}, where only a few sparsely sampled images of the scene are available, has focused on improving the initialization stage of 3DGS~\cite{edgs, foroutan_nerf_init, gdgs_mlp_init_and_some_densification}.  
Examples include using stereo matching~\cite{edgs, instantsplat, mvpgs_sparse_view, hbsplat_sparse_view, sparse2dgs} and neural radiance fields~\cite{foroutan_nerf_init} to obtain denser point clouds for 3DGS initialization.
Beyond novel view synthesis, 3DGS-based methods have been successfully used for geometry extraction~\cite{2dgs, pgsr, sugar, gof}. As photometric supervision is often insufficient to obtain high-quality geometry, many of these methods encourage geometric consistency through regularization~\cite{2dgs,sugar,pgsr, gof}. Despite its potential to provide geometrically accurate priors, improved initialization has not been a major research focus in this area.

To the best of our knowledge, prior work has largely focused on improving and evaluating densification and initialization in relative isolation, \eg densification methods are often tested only with SfM and random initialization. Furthermore, the potential of dense initialization in improving geometric consistency remains underexplored. We aim to close this gap in the literature and perform comprehensive evaluation using 6 initialization methods and 5 densification strategies. We test initialization with SfM, precise laser scans, EDGS~\cite{edgs}, SfM-aligned monocular depth predictions, as well as feed-forward 3DGS predictions and point clouds produced by Depth Anything~3~\cite{depthanything3}. Besides novel view synthesis (NVS) performance, we investigate geometric accuracy of the initializations and of the final trained scene representations against laser scan ground truth.

Our main contribution is a set of insights drawn from our experiments: 
(1) For NVS, dense initialization does not lead to major improvements on well-constrained scenes, and can even decrease quality in some cases. Despite that, we see benefits in generalization to views far from the training images, which is often desirable in practice.
(2) Using dense initialization leads to a major jump in geometric accuracy of the trained scenes. With SfM, densification methods with the best NVS results often produce very
inaccurate geometry, but dense initialization helps mitigate this issue.
(3) When paired with advanced densification strategies, inexpensive practical initialization methods often rival laser scan initialization in improving geometric accuracy and NVS generalization.

From these insights, two recommendations for the field emerge: 
(1) Initialization and densification are often treated as isolated components outside sparse-view NVS literature~\cite{instantsplat,sparse2dgs,initialize_to_generalize_sparse_view}.
Based on our experiments, we find initialization priors to be valuable for many 3DGS applications, and suggest that future works strive towards utilizing them fully.
(2) As dense initialization using inexpensive methods  provides large benefits to the geometric accuracy of the trained scenes, it can likely be applied to improve performance of existing 3DGS-based geometry extraction pipelines and be incorporated as a core component of future methods.

\section{Related Work}
The success of the original 3DGS formulation~\cite{3dgs} has sparked significant interest in expanding and improving the method, %
targeting key areas such as core rendering and representation improvements~\cite{mip_splatting_2024,multi_scale_3dgs_antialiasing_2024, stop_the_pop, analytic_splatting}, the reconstruction of dynamic scenes~\cite{4d_gs_dynamic_2024, deformable_3d_gaussians_2024}, relighting techniques~\cite{gs_ir_2024, relightable_gaussians_2025, gs3_relighting}, feed-forward networks~\cite{pixelsplat, mvsplat, mlsharp}, geometrically consistent 3DGS and mesh reconstruction~\cite{2dgs, pgsr, sugar, gof}, and hybrid neural–Gaussian models~\cite{scaffold_gs, neuralgs,kulhanek2024wildgaussians}. In this work, we focus on the impact of initialization on the 3D Gaussian Splatting training process, and how it interacts with various densification procedures.

\PAR{Initializing 3DGS.} 
For training, 3DGS requires known camera intrinsics and extrinsics for the input images. These are typically computed via SfM, which, as a side product of the estimation process, produces a
sparse point cloud that can be used for initialization.
Hence, initialization from SfM point clouds is the de-facto standard approach, and densification strategies are typically tuned for this type of initialization. 
Still, a range of works focus on initialization improvements, including
using dense feature matching between images to obtain a dense point cloud~\cite{edgs},
predicting Gaussian parameters from SfM points with an MLP~\cite{gdgs_mlp_init_and_some_densification},
initializing from a partially trained NeRF model~\cite{foroutan_nerf_init},
or investigating alternative random initialization regimens~\cite{foroutan_nerf_init, rain_gs}.
Additionally, modern feed-forward geometry prediction networks~\cite{vggt,depthanything3,vggt_omega} can produce dense 3D point clouds suitable for 3DGS initialization, and can even be used as a backbone for feed-forward 3DGS~\cite{depthanything3}.
Unlike these works, our main contribution is not a new method, but a systematic evaluation of the effect of initialization
(laser scans, EDGS~\cite{edgs}, dense point clouds obtained from monocular depth estimates~\cite{metric3dv2} aligned to SfM points, point clouds and feed-forward 3DGS predictions produced by DA3~\cite{depthanything3})
on NVS quality and geometry accuracy under different densification strategies.

\PAR{Densifying 3DGS.} 
The original 3DGS formulation relies on Adaptive Density Control (ADC) heuristics for creating new Gaussians~\cite{3dgs}. 
Subsequent work has addressed a flaw in the original densification signal, replacing a per-pixel sum of gradients (which can cancel out) with a sum of gradient magnitudes~\cite{absgs, gof}, reducing the occurrence of under-reconstructed blurry regions produced by the original 3DGS formulation.
Rota Bulò \etal~\cite{revdgs} propose densifying based on per-pixel photometric errors distributed among Gaussians based on their contributions to the final pixel color. 
They also propose opacity correction after cloning, replace opacity resets with regularization, and limit the number of newly added Gaussians in each densification iteration to a fraction of the current model size. 
Other works~\cite{rain_gs,3dgs_mcmc} improve the densification stage through encouraging scene exploration. 
Kheradmand \etal~\cite{3dgs_mcmc} view the scene as a Markov Chain Monte Carlo (MCMC) sample, and add random noise to the Gaussians' means during training. They replace pruning and cloning with an alternative that minimizes changes in scene probability across densification iterations.
Deng \etal~\cite{idhfr} prioritize the creation of new Gaussians via an ``Edge Aware'' score which is based on edge detector response in the input images.
Other proposed improvements include
splitting Gaussians along their longest axis~\cite{bogauss,idhfr},
densifying based on photometric error relative to the errors of the other primitives in a given view~\cite{bogauss},
and using depth re-initialization~\cite{minisplatting}, which resets the scene by rendering depth images from the current representation and replacing the Gaussians with new ones, positioned based on the rendered depths. 
To evaluate initialization under different densification approaches, besides base ADC~\cite{3dgs} and AbsGS~\cite{absgs}, we test with~\cite{revdgs}, as a representative of pixel error driven densification, and with \cite{3dgs_mcmc,idhfr} as representatives of state-of-the-art methods.

\section{Preliminaries}\label{sec:preliminaries}
\label{par:preliminaries:3dgs}3DGS~\cite{3dgs} represents a scene as a set of $N$ anisotropic Gaussian primitives $\mathcal{G} = \{ G_i \}_{i=1}^{N}$,
each parameterized as
\begin{equation}
G_i = (\boldsymbol{\mu}_i, \mathbf{\Sigma}_i, \alpha_i, \mathbf{f}_i) \enspace ,
\end{equation}
where $\boldsymbol{\mu}_i \in \mathbb{R}^3$ denotes the mean,
$\mathbf{\Sigma}_i \in \mathbb{R}^{3 \times 3}$ is a positive semi-definite covariance matrix,
$\alpha_i \in \mathbb{R}$ is the opacity, and
$\mathbf{f}_i$ represents a view-dependent color function. In practice, $\mathbf{\Sigma}_i$ is represented as an $\mathbb{R}^3$ scale vector and a rotation quaternion, and $\mathbf{f}_i$ is encoded through a set of spherical harmonics (SH) coefficients. The rendered color is obtained through alpha compositing with a differentiable rasterizer.

Training optimizes the parameters $\{G_i\}_{i=1}^{N}$ using Stochastic Gradient Descent (SGD), minimizing a photometric loss between rendered and ground-truth (GT) images. To bootstrap the optimization, an initial set of Gaussians $\mathcal{G}_\text{init}$ is created from a point cloud provided as input. We use $\mathcal{G}_{\text{init}}^{\text{method}}$ to denote the initial set of Gaussians obtained using a given initialization (init) method.
During optimization, new Gaussians are added, and redundant Gaussians are removed or repurposed through a process called densification. The original 3DGS formulation~\cite{3dgs} uses a heuristic mechanism called Adaptive Density Control (ADC), which many follow-up densification works modify or replace~\cite{absgs, 3dgs_mcmc, idhfr}.

To obtain the initial set of Gaussians $\mathcal{G}_\text{init}$, 3DGS~\cite{3dgs} and most follow-up works use the sparse point cloud obtained from Structure-from-Motion (SfM).
Using this input, the means $\mu_i$ are initialized to point positions, opacities $\alpha_i$ to $0.1$, and the 0-th SH coefficients to the points' colors. Covariances $\mathbf{\Sigma}_i$ are initialized isotropically using the average distance to the four nearest neighbors of each point, and higher order SH coefficients are initially set to zero.

\section{Evaluation protocol}
\label{sec:protocol}
Our goal is to study the impact of initialization on NVS and geometric accuracy, when trained using different densification strategies.
To ensure a fair comparison, we evaluate all initialization and densification strategies within the same %
3DGS implementation,\footnote{Our code is available on https://github.com/deivse/ivd\_splat} using identical hyper-parameters whenever applicable.

When comparing models trained with different densification strategies, it is crucial to ensure that changes in reconstruction quality are explained by changes in the investigated variables, rather than changes in model capacity stemming from densification differences. Inspired by prior work~\cite{3dgs_mcmc,idhfr}, we address this by setting a per-scene model capacity limit, defined as the number of Gaussians reached after 30,000 training iterations using AbsGS densification~\cite{absgs} and SfM initialization.
While this is a scene-dependent value, we refer to it simply as $G_m$ for brevity. We confirm that the chosen limits are sufficient to represent the evaluated scenes in an experiment presented in \Cref{scene_size_limit_verif} of the supp.\ mat.

\subsection{Laser Scan Initialization}\label{sec:protocol:laser_scan_init}
We start by establishing a high-precision initialization baseline using point clouds obtained with laser scanners. To analyze the contribution of initialization \textit{size}, we train with point clouds subsampled to different fractions of the maximum number of Gaussians for each scene $G_m$. Additionally, to provide a direct comparison against SfM initialization, we also evaluate with the laser scans subsampled to the size $|\mathcal{G}_\text{init}^\text{SfM}|$ of the SfM point clouds. For a given target size $N_\textit{init}$, we uniformly sample $N_\textit{init}$ points from the laser scan point cloud without replacement.\footnote{This is always possible, as all the laser scan point clouds are larger than $G_m$.} To transform this point cloud into an initial set of Gaussians, we follow the procedure in~\cite{3dgs}, which is described in \cref{par:preliminaries:3dgs}. Additionally, on ScanNet++, we use hybrid initialization (laser scan + SfM points), please see~\Cref{sec:protocol:datasets} for details.

\subsection{Practical Initialization Methods}\label{sec:protocol:practical_init}
While laser scan point clouds serve as a useful baseline for high-accuracy initialization, access to such equipment is limited and capture with separate devices complicates data acquisition and processing. Thus, methods for producing dense initialization without using additional sensors 
are of interest. We evaluate several such methods, as described below. To establish a common initialization size, we uniformly subsample all point clouds to $\min\{ \min_{i \in I}{|\mathcal{G}_\text{init}^{i}|}, G_m\}$ Gaussians, where $I$ is the set of all init methods. On most scenes, this corresponds to $G_m$. For the methods that produce point clouds, we initialize the Gaussian parameters using the default 3DGS procedure~\cite{3dgs}. For the methods that directly output initial Gaussian parameters, we leave them unchanged unless specified otherwise.

\PAR{EDGS.} The EDGS initialization method~\cite{edgs} creates initial Gaussians by triangulating correspondences across image pairs predicted by a dense feature matching network. We base our implementation on the code publicly available at the time of writing. As confirmed by the authors, it only differs from the published paper in its lack of higher order SH initialization and in ``minor hyperparameter adjustments''. According to~\cite{edgs}, their full SH initialization improves performance but we saw the opposite result implementing it ourselves based on code snippets provided by the authors. Unfortunately, there's no public code to verify against at the time of writing, so we only initialize the DC SH coefficient in our evaluation. We use RoMa~\cite{roma} with the default settings from the public code, using the indoors or outdoors RoMa weights according to the scene. We denote our implementation $\text{EDGS}^*$. We additionally increase Gaussians' scales for $\text{EDGS}^*$ if subsampling, since they are not naturally adjusted as with point cloud init. As we show in~\Cref{hyperparam_changes} of the supp.\ mat., this moderately improves results.

\PAR{Monocular depth.}
To provide another baseline, we implement an initialization method which produces a point cloud with the aid of per-image monocular depth predictions. Specifically, we use Metric3Dv2~\cite{metric3dv2} with the DINOv2-reg ViT Large backbone~\cite{vit, dinov2, vits_need_registers}. 
Our pipeline is as follows: (1) the depth predictor is invoked for each image; (2) the depth maps are aligned to the SfM point cloud using LO-RANSAC~\cite{lo-ransac}, this alignment is then refined via linear interpolation of the depth range, following~\cite{depth_densifier_github}; (3) world-space points are created for a subset of the image pixels; and (4) we apply the floater removal method from~\cite{depth_densifier_github} to filter out noise in front of the cameras. For each scene, we select at most 300 cameras, using the K-means clustering approach from~\cite{edgs} where needed. We provide details in \Cref{monodepth_details} of the supp.\ mat. We refer to this method as ``Monodepth'' or ``M.D.''

\PAR{Depth Anything 3.} In recent years, feed-forward 3D reconstruction methods have gained traction. We use Depth Anything 3~(DA3)~\cite{depthanything3} to investigate its applicability to 3DGS initialization.
We use the ViT-Giant version of the model and provide camera parameters as input.
For point cloud initialization, we apply the same floater removal method as for Monodepth, as this improves results (see \cref{hyperparam_changes} of the supp.\ mat.), and makes the comparison between DA3 and Monodepth more direct.
We limit the number of images to 300, using the same method as for Monodepth.
Besides point cloud initialization, we also use DA3's feed-forward 3DGS head to evaluate the effect of using predicted initial values for all parameters.
In that case we limit the number of images to 150 due to VRAM constraints. We denote this version ``$\text{DA3}^\text{GS}$''. In later stages of our experiments, we have realized that 8 out of 15 ScanNet++\cite{scannetpp} scenes we use may overlap with DA3 
training data (the DA3 train set is not public).
Upon further investigation (\cref{da3_scannet_scene_selection} of the supp.\ mat.), we do not see improved performance on the potentially contaminated scenes. We thus use all 15 scenes in \cref{sec:experiments}.

\subsection{Densification strategies}\label{sec:protocol:densification_strats}

Since the inception of 3DGS, countless methods have been proposed to improve its densification heuristics. To evaluate how initialization interacts with these different strategies, we select five, which are described below. While this selection is not exhaustive, it should provide a representative sample of current approaches with strategies based on primitives' photometric error contributions, positional gradient magnitudes, stochastic exploration, and prioritization of image edges.

We adopt AbsGS~\cite{absgs} as our baseline (used to compute $G_m$) as it is very close to 3DGS ADC, but mitigates the blurriness issue.
We also report results with 3DGS ADC~\cite{3dgs} (labeled ``INRIA'' to avoid confusion).
Another popular method we employ is 3DGS MCMC~\cite{3dgs_mcmc}, which adds stochastic exploration and probabilistic sampling to the optimization and aims to increase robustness to initialization.
Furthermore, we adopt the work of Deng \etal~\cite{idhfr}, which we refer to as ``IDHFR'' (also known as ``ImprovedGS'').
It prioritizes splitting Gaussians which contribute to edges detected in the input images. Among its other changes are those that affect pruning and the backpropagation schedule in later optimization stages.
Finally, we use the method by Rota Bulò \etal~\cite{revdgs}, which we label ``RevDGS''.
It uses pixel-wise photometric errors redistributed to Gaussians in proportion to their contributions to the rendered image as the densification signal.
Finally, we also test disabling densification\footnote{We leave default 3DGS pruning~\cite{3dgs} enabled.} (labeled ``No. D.'' in tables).

\PAR{Densification hyper-parameters.} To enable direct comparison with the other strategies, we do not use the adjusted $\mu_i$ learning rate schedule for IDHFR as it did not lead to significant improvements (see supp.\ mat.,~\cref{hyperparam_changes}). For AbsGS, we use a gradient threshold of $0.0004$ and do not adjust the scale threshold to keep it as close as possible to INRIA while resolving the blurriness issue. We use default parameters for all other methods.

\subsection{Geometry accuracy evaluation} \label{sec:protocol:geometric}
Geometrically accurate scene representations aid generalization and are key for 3DGS-based geometry reconstruction~\cite{2dgs, pgsr, sugar, gof}.
In our experiments, we thus evaluate geometric accuracy of the initialization itself, as well as of the final trained scenes, using laser scans as ground truth (GT) data. We report $\text{F}_1$ scores, precision and recall, following a standard protocol~\cite{tanksandtemples, dtu_dataset, depthanything3}. 
Following~\cite{depthanything3}, we compute these scores with an inlier threshold of $0.05m$.
As these metrics are defined for point sets, we need to convert the trained scenes into a suitable representation.
To this end, we render expected depths\footnote{$(\sum_i w_i z_i)(\sum_i w_i)^{-1}$, where $w_i$ are the splats' relative contributions to a given pixel, and $z_i$ their projected depths.} for all training images, perform TSDF fusion~\cite{tsdf_fusion_og}, and extract a point cloud by identifying zero-crossings within the TSDF volume.
We apply voxel downsampling to all GT and init point clouds using the same voxel size $v=0.02m$ (following~\cite{depthanything3}) as for TSDF fusion.
Because $\text{EDGS}^*$ produces small Gaussians with uniform opacities, unsuitable for depth rendering, we treat its output as a point cloud, discarding opacities and scales. This approximation introduces minimal bias as EDGS output is very dense. $\text{DA3}^\text{GS}$, on the other hand, produces Gaussians with varying shapes and opacities, so we use the same procedure as for post-training geometry extraction.

\subsection{Datasets} \label{sec:protocol:datasets}
We evaluate on four diverse datasets to capture a wide range of geometric and viewing conditions. For our laser scan baseline we use 7 scenes from ETH3D~\cite{eth3d_mvs} (MVS) and 15 scenes from ScanNet++~\cite{scannetpp} (DSLR captures), of which the latter is also used for geometric evaluation.
For the other experiments we use the 15 ScanNet++ scenes, as well as all scenes from MipNerf360~\cite{mipnerf360} and 14 scenes from Tanks\&Temples~\cite{tanksandtemples}, which are widely used for 3DGS evaluation. We do not use the laser scan data on Tanks\&Temples due to its limited scene coverage.

\PAR{Dataset Characteristics.} All datasets contain a mix of indoors and outdoors scenes, except ScanNet++ which is indoors-only. 
ScanNet++ training sets are dense, whereas the default NVS test set consists of ``hard'' out-of distribution images. 
ETH3D is representative of under-constrained training as it contains only 10-50 images per scene.
MipNerf360 is a well-behaved dataset, created specifically for NVS, with sufficient image counts and little to no visual artifacts.
Tanks\&Temples, on the other hand, is more challenging as it contains overexposed images, as well as exposure and white balance variations~\cite[Appendix~D]{mipnerf360}.

\PAR{Evaluating generalization.} On ScanNet++, we use two separate evaluation regimes to test generalization -- ``default split'', which uses the official train/test split provided by the dataset authors, and ``on-trajectory'', for which we use every 8th image from the official training set as a test view. We provide examples for off-trajectory and on-trajectory camera positions in \Cref{scannetpp_cameras} of the supp.\ mat.

\PAR{Dataset-specific changes.} We find that ScanNet++ laser scans on most scenes contain missing geometry for reflective and transparent surfaces, which hurts NVS quality. To mitigate this issue, we employ a hybrid approach for some of the NVS experiments, where we include all SfM points in addition to the laser scan data (maintaining total init size). We use the ``+'' symbol to indicate such cases, \eg ``$\text{Laser}^+$'' or ``$0.5G_m^+$''. While ETH3D also contains some reflective surfaces, we do not use the hybrid approach there, as ETH3D's SfM point clouds are especially sparse and noisy due to low image counts, and including them led to worse results. \cref{hybrid_init_ablation} of the supp.\ mat. provides data supporting this choice.

\section{Experiments and Analysis}
We base our implementation on gsplat~\cite{gsplat}. To ensure consistent evaluation and data loading,\footnote{The way a dataset's images are downscaled can have an effect on NVS metrics.~\cite{nerfbaselines}}
we use NerfBaselines~\cite{nerfbaselines} to calculate NVS metrics and collect runtime statistics. For MCMC, AbsGS, and INRIA we use the implementations provided by gsplat, though we extend them to allow limiting scene size for AbsGS and INRIA. For consistent evaluation, we re-implement IDHFR in our framework based on the public code. We implement RevDGS following the paper~\cite{revdgs} as there's no reference implementation. All training was performed on 40GB Tesla A100 GPUs.
For the main experiments (\cref{tab:laser_scan_main,tab:noise_resiliency,tab:practical_main}) we report means over 3 runs with different (but fixed) random seeds. All other data is using one seed.

For \cref{tab:laser_scan_main,tab:practical_main,tab:final_recon_fscore}, we perform non-parametric statistical tests following~\cite{demsar_stat_comparisons_2006}.
For each metric, we test whether dense init improves upon SfM \textit{for a given densification}:
(1) Friedman test (Iman-Davenport variant) across scenes to confirm overall effect, and 
(2) post-hoc comparisons to detect significant improvements over SfM, controlling family-wise error rate (FWER) at $\alpha=.05$ using Holm's step-down procedure. We group data into families by metric and densification method. The power of the test is limited by the number of scenes, especially on ETH3D with only 7 scenes.
We describe the statistical tests in detail and provide per-scene data plots in \Cref{statistical_analysis,per_scene_data_analysis} of the supp.\ mat.

\label{sec:experiments}

\subsection{Novel View Synthesis with Laser Scan Initialization}\label{subsec:eval:laser_init}

\begin{table}[tb]
\centering
\caption{Novel view synthesis performance using laser scan initialization at various sizes w.r.t. $G_m$. Columns marked with ``+'' use \textit{hybrid initialization} (see~\cref{sec:protocol:datasets}). Cells with significant improvements over SfM are marked with ``$^*$''.}
\label{tab:laser_scan_main}
\begin{subtable}[t]{\linewidth}
\centering
\caption{ScanNet++}
\label{tab:laser_scan_main_scannet++}
{\small
\begingroup \setlength{\tabcolsep}{2.5\tabcolsep}
\resizebox{\linewidth}{!}{\begin{tabular}{l|ccccc|ccccc|ccccc}
\toprule
& \multicolumn{5}{c|}{\textbf{PSNR ↑}} & \multicolumn{5}{c|}{\textbf{SSIM ↑}} & \multicolumn{5}{c}{\textbf{LPIPS ↓}} \\
& SfM & $|\mathcal{G}_\mathit{init}^\text{SfM}|$ & $0.5\text  {G}_\mathit{m}$ & $0.75\text  {G}_\mathit{m}$ & $1.0\text  {G}_\mathit{m}$ & SfM & $|\mathcal{G}_\mathit{init}^\text{SfM}|$ & $0.5\text  {G}_\mathit{m}$ & $0.75\text  {G}_\mathit{m}$ & $1.0\text  {G}_\mathit{m}$ & SfM & $|\mathcal{G}_\mathit{init}^\text{SfM}|$ & $0.5\text  {G}_\mathit{m}$ & $0.75\text  {G}_\mathit{m}$ & $1.0\text  {G}_\mathit{m}$ \\
\midrule
AbsGS & \cc{BCDBEA}{\color[HTML]{000000} $22.36$} & \cc{D7E9F3}{\color[HTML]{000000} $22.50$} & \cc{FCE2D5}{\color[HTML]{000000} $22.90$\rlap{\textsuperscript{*}}} & \cc{FBD6C4}{\color[HTML]{000000} $22.98$\rlap{\textsuperscript{*}}} & \cc{FACCB5}{\color[HTML]{000000} $23.05$\rlap{\textsuperscript{*}}} & \cc{B2D6E7}{\color[HTML]{000000} $0.870$} & \cc{87B8D7}{\color[HTML]{000000} $0.869$} & \cc{F6FAFC}{\color[HTML]{000000} $0.873$\rlap{\textsuperscript{*}}} & \cc{FEF9F6}{\color[HTML]{000000} $0.874$\rlap{\textsuperscript{*}}} & \cc{FEF2EC}{\color[HTML]{000000} $0.874$\rlap{\textsuperscript{*}}} & \cc{DFEDF5}{\color[HTML]{000000} $0.249$} & \cc{B2D6E7}{\color[HTML]{000000} $0.254$} & \cc{FCE2D5}{\color[HTML]{000000} $0.241$\rlap{\textsuperscript{*}}} & \cc{FBD8C7}{\color[HTML]{000000} $0.240$\rlap{\textsuperscript{*}}} & \cc{FAD2BE}{\color[HTML]{000000} $0.239$\rlap{\textsuperscript{*}}} \\
INRIA & \cc{A7CEE3}{\color[HTML]{000000} $22.26$} & \cc{D8EAF3}{\color[HTML]{000000} $22.50$} & \cc{FCE1D3}{\color[HTML]{000000} $22.91$\rlap{\textsuperscript{*}}} & \cc{FBDBCB}{\color[HTML]{000000} $22.95$\rlap{\textsuperscript{*}}} & \cc{FBD9C8}{\color[HTML]{000000} $22.96$\rlap{\textsuperscript{*}}} & \cc{9DC7E0}{\color[HTML]{000000} $0.870$} & \cc{A5CDE3}{\color[HTML]{000000} $0.870$} & \cc{FEFFFF}{\color[HTML]{000000} $0.874$\rlap{\textsuperscript{*}}} & \cc{FEF8F4}{\color[HTML]{000000} $0.874$\rlap{\textsuperscript{*}}} & \cc{FFFDFB}{\color[HTML]{000000} $0.874$\rlap{\textsuperscript{*}}} & \cc{BCDBEA}{\color[HTML]{000000} $0.253$} & \cc{99C4DE}{\color[HTML]{000000} $0.256$} & \cc{FCE5D9}{\color[HTML]{000000} $0.242$\rlap{\textsuperscript{*}}} & \cc{FBDCCC}{\color[HTML]{000000} $0.240$\rlap{\textsuperscript{*}}} & \cc{FBDDCE}{\color[HTML]{000000} $0.240$\rlap{\textsuperscript{*}}} \\
MCMC & \cc{5C99C7}{\color[HTML]{F1F1F1} $22.02$} & \cc{DFEDF5}{\color[HTML]{000000} $22.54$\rlap{\textsuperscript{*}}} & \cc{C2DEEC}{\color[HTML]{000000} $22.39$\rlap{\textsuperscript{*}}} & \cc{E1EFF6}{\color[HTML]{000000} $22.55$\rlap{\textsuperscript{*}}} & \cc{CCE3EF}{\color[HTML]{000000} $22.44$\rlap{\textsuperscript{*}}} & \cc{ADD2E6}{\color[HTML]{000000} $0.870$} & \cc{FBDBCB}{\color[HTML]{000000} $0.876$\rlap{\textsuperscript{*}}} & \cc{F2F8FB}{\color[HTML]{000000} $0.873$} & \cc{FBD6C4}{\color[HTML]{000000} $0.876$\rlap{\textsuperscript{*}}} & \cc{FEF1EA}{\color[HTML]{000000} $0.874$\rlap{\textsuperscript{*}}} & \cc{E0EEF5}{\color[HTML]{000000} $0.249$} & \cc{FEF7F3}{\color[HTML]{000000} $0.244$\rlap{\textsuperscript{*}}} & \cc{FDE9DF}{\color[HTML]{000000} $0.242$\rlap{\textsuperscript{*}}} & \cc{F9C5AB}{\color[HTML]{000000} $0.237$\rlap{\textsuperscript{*}}} & \cc{FBD4C1}{\color[HTML]{000000} $0.239$\rlap{\textsuperscript{*}}} \\
IDHFR & \cc{FACDB7}{\color[HTML]{000000} $23.04$} & \cc{FACFBA}{\color[HTML]{000000} $23.02$} & \cc{E48C7B}{\color[HTML]{F1F1F1} $23.29$} & \cc{D5675E}{\color[HTML]{F1F1F1} $23.40$\rlap{\textsuperscript{*}}} & \cc{DA7368}{\color[HTML]{F1F1F1} $23.36$} & \cc{F9C2A7}{\color[HTML]{000000} $0.877$} & \cc{FAD3BF}{\color[HTML]{000000} $0.876$} & \cc{E89784}{\color[HTML]{000000} $0.878$} & \cc{D86F65}{\color[HTML]{F1F1F1} $0.879$\rlap{\textsuperscript{*}}} & \cc{D5675E}{\color[HTML]{F1F1F1} $0.879$\rlap{\textsuperscript{*}}} & \cc{E69280}{\color[HTML]{F1F1F1} $0.233$} & \cc{FCE1D3}{\color[HTML]{000000} $0.241$} & \cc{DF8072}{\color[HTML]{F1F1F1} $0.232$} & \cc{D86D63}{\color[HTML]{F1F1F1} $0.230$} & \cc{D5675E}{\color[HTML]{F1F1F1} $0.230$\rlap{\textsuperscript{*}}} \\
RevDGS & \cc{C8E1EE}{\color[HTML]{000000} $22.42$} & \cc{D5E8F2}{\color[HTML]{000000} $22.49$} & \cc{FDF0E9}{\color[HTML]{000000} $22.81$} & \cc{FDEBE2}{\color[HTML]{000000} $22.84$} & \cc{FBD7C5}{\color[HTML]{000000} $22.97$\rlap{\textsuperscript{*}}} & \cc{95C2DD}{\color[HTML]{000000} $0.869$} & \cc{5C99C7}{\color[HTML]{F1F1F1} $0.868$} & \cc{D3E7F2}{\color[HTML]{000000} $0.872$} & \cc{D7E9F3}{\color[HTML]{000000} $0.872$\rlap{\textsuperscript{*}}} & \cc{F3F9FB}{\color[HTML]{000000} $0.873$\rlap{\textsuperscript{*}}} & \cc{97C3DD}{\color[HTML]{000000} $0.257$} & \cc{72A9CF}{\color[HTML]{F1F1F1} $0.259$} & \cc{CBE3EF}{\color[HTML]{000000} $0.252$} & \cc{DDEDF5}{\color[HTML]{000000} $0.249$\rlap{\textsuperscript{*}}} & \cc{F6FAFC}{\color[HTML]{000000} $0.246$\rlap{\textsuperscript{*}}} \\
No D. & \cc{FFFFFF}{\color[HTML]{F1F1F1} \color{white} --} & \cc{FFFFFF}{\color[HTML]{F1F1F1} \color{white} --} & \cc{BDDBEB}{\color[HTML]{000000} $22.37$} & \cc{CDE4F0}{\color[HTML]{000000} $22.45$} & \cc{E3F0F6}{\color[HTML]{000000} $22.56$} & \cc{FFFFFF}{\color[HTML]{F1F1F1} \color{white} --} & \cc{FFFFFF}{\color[HTML]{F1F1F1} \color{white} --} & \cc{A9CFE4}{\color[HTML]{000000} $0.870$} & \cc{B1D5E7}{\color[HTML]{000000} $0.870$} & \cc{C7E1EE}{\color[HTML]{000000} $0.871$} & \cc{FFFFFF}{\color[HTML]{F1F1F1} \color{white} --} & \cc{FFFFFF}{\color[HTML]{F1F1F1} \color{white} --} & \cc{5C99C7}{\color[HTML]{F1F1F1} $0.261$} & \cc{93C0DC}{\color[HTML]{000000} $0.257$} & \cc{BFDCEB}{\color[HTML]{000000} $0.253$} \\
\bottomrule
\end{tabular}}
\endgroup
}
\end{subtable}
\begin{subtable}[t]{\linewidth}
\centering
\caption{ScanNet++ (On-Trajectory)}
\label{tab:laser_scan_main_eval_on_train_set_scannet++}
{\small
\begingroup \setlength{\tabcolsep}{2.5\tabcolsep}
\resizebox{\linewidth}{!}{\begin{tabular}{l|ccccc|ccccc|ccccc}
\toprule
& \multicolumn{5}{c|}{\textbf{PSNR ↑}} & \multicolumn{5}{c|}{\textbf{SSIM ↑}} & \multicolumn{5}{c}{\textbf{LPIPS ↓}} \\
& SfM & $|\mathcal{G}_\mathit{init}^\text{SfM}|$ & $0.5\text  {G}_\mathit{m}$ & $0.75\text  {G}_\mathit{m}$ & $1.0\text  {G}_\mathit{m}$ & SfM & $|\mathcal{G}_\mathit{init}^\text{SfM}|$ & $0.5\text  {G}_\mathit{m}$ & $0.75\text  {G}_\mathit{m}$ & $1.0\text  {G}_\mathit{m}$ & SfM & $|\mathcal{G}_\mathit{init}^\text{SfM}|$ & $0.5\text  {G}_\mathit{m}$ & $0.75\text  {G}_\mathit{m}$ & $1.0\text  {G}_\mathit{m}$ \\
\midrule
AbsGS & \cc{FDEAE0}{\color[HTML]{000000} $33.20$} & \cc{FEFFFF}{\color[HTML]{000000} $32.97$} & \cc{FAD0BB}{\color[HTML]{000000} $33.47$\rlap{\textsuperscript{*}}} & \cc{FAD3BF}{\color[HTML]{000000} $33.45$\rlap{\textsuperscript{*}}} & \cc{FBD5C2}{\color[HTML]{000000} $33.43$\rlap{\textsuperscript{*}}} & \cc{FBD9C8}{\color[HTML]{000000} $0.951$} & \cc{FEF9F6}{\color[HTML]{000000} $0.950$} & \cc{F9C4AA}{\color[HTML]{000000} $0.952$\rlap{\textsuperscript{*}}} & \cc{F9C3A8}{\color[HTML]{000000} $0.952$\rlap{\textsuperscript{*}}} & \cc{F9C4AA}{\color[HTML]{000000} $0.952$\rlap{\textsuperscript{*}}} & \cc{FCE6DB}{\color[HTML]{000000} $0.101$} & \cc{FEFFFF}{\color[HTML]{000000} $0.104$} & \cc{F5B89E}{\color[HTML]{000000} $0.095$\rlap{\textsuperscript{*}}} & \cc{EEA58F}{\color[HTML]{000000} $0.094$\rlap{\textsuperscript{*}}} & \cc{EA9B87}{\color[HTML]{000000} $0.093$\rlap{\textsuperscript{*}}} \\
INRIA & \cc{FEF4EF}{\color[HTML]{000000} $33.10$} & \cc{EBF4F9}{\color[HTML]{000000} $32.81$} & \cc{FBD8C7}{\color[HTML]{000000} $33.40$\rlap{\textsuperscript{*}}} & \cc{FBD8C7}{\color[HTML]{000000} $33.39$\rlap{\textsuperscript{*}}} & \cc{FCE0D2}{\color[HTML]{000000} $33.31$\rlap{\textsuperscript{*}}} & \cc{FEF8F4}{\color[HTML]{000000} $0.950$} & \cc{CFE5F0}{\color[HTML]{000000} $0.948$} & \cc{FAD0BB}{\color[HTML]{000000} $0.952$\rlap{\textsuperscript{*}}} & \cc{FAD1BC}{\color[HTML]{000000} $0.951$\rlap{\textsuperscript{*}}} & \cc{FBD8C7}{\color[HTML]{000000} $0.951$\rlap{\textsuperscript{*}}} & \cc{BBDAEA}{\color[HTML]{000000} $0.112$} & \cc{8DBCDA}{\color[HTML]{000000} $0.115$} & \cc{FAD3BF}{\color[HTML]{000000} $0.098$\rlap{\textsuperscript{*}}} & \cc{F9C2A7}{\color[HTML]{000000} $0.096$\rlap{\textsuperscript{*}}} & \cc{F8BEA3}{\color[HTML]{000000} $0.095$\rlap{\textsuperscript{*}}} \\
MCMC & \cc{FEFFFF}{\color[HTML]{000000} $32.97$} & \cc{FCFDFE}{\color[HTML]{000000} $32.95$} & \cc{FDF0E9}{\color[HTML]{000000} $33.14$} & \cc{FDEEE6}{\color[HTML]{000000} $33.16$} & \cc{FDEFE7}{\color[HTML]{000000} $33.16$} & \cc{FBD9C8}{\color[HTML]{000000} $0.951$} & \cc{FAD1BC}{\color[HTML]{000000} $0.951$} & \cc{F0AB94}{\color[HTML]{000000} $0.953$\rlap{\textsuperscript{*}}} & \cc{EFA791}{\color[HTML]{000000} $0.953$\rlap{\textsuperscript{*}}} & \cc{F1AD96}{\color[HTML]{000000} $0.953$\rlap{\textsuperscript{*}}} & \cc{FFFFFE}{\color[HTML]{000000} $0.104$} & \cc{FEF5F0}{\color[HTML]{000000} $0.103$} & \cc{F9C5AB}{\color[HTML]{000000} $0.096$\rlap{\textsuperscript{*}}} & \cc{F6BA9F}{\color[HTML]{000000} $0.095$\rlap{\textsuperscript{*}}} & \cc{F6BA9F}{\color[HTML]{000000} $0.095$\rlap{\textsuperscript{*}}} \\
IDHFR & \cc{F0A992}{\color[HTML]{000000} $33.76$} & \cc{EFA791}{\color[HTML]{000000} $33.78$} & \cc{DA7368}{\color[HTML]{F1F1F1} $34.04$\rlap{\textsuperscript{*}}} & \cc{D97166}{\color[HTML]{F1F1F1} $34.05$\rlap{\textsuperscript{*}}} & \cc{D5675E}{\color[HTML]{F1F1F1} $34.10$\rlap{\textsuperscript{*}}} & \cc{F0A992}{\color[HTML]{000000} $0.953$} & \cc{F4B49A}{\color[HTML]{000000} $0.953$} & \cc{D97166}{\color[HTML]{F1F1F1} $0.954$\rlap{\textsuperscript{*}}} & \cc{D97166}{\color[HTML]{F1F1F1} $0.954$\rlap{\textsuperscript{*}}} & \cc{D5675E}{\color[HTML]{F1F1F1} $0.954$\rlap{\textsuperscript{*}}} & \cc{F5B89E}{\color[HTML]{000000} $0.095$} & \cc{F9C4AA}{\color[HTML]{000000} $0.096$} & \cc{DE7E70}{\color[HTML]{F1F1F1} $0.091$\rlap{\textsuperscript{*}}} & \cc{D97166}{\color[HTML]{F1F1F1} $0.090$\rlap{\textsuperscript{*}}} & \cc{D5675E}{\color[HTML]{F1F1F1} $0.089$\rlap{\textsuperscript{*}}} \\
RevDGS & \cc{F9FCFD}{\color[HTML]{000000} $32.93$} & \cc{D8EAF3}{\color[HTML]{000000} $32.64$} & \cc{FFFBF9}{\color[HTML]{000000} $33.02$} & \cc{FEF3ED}{\color[HTML]{000000} $33.11$} & \cc{FDF0E9}{\color[HTML]{000000} $33.14$} & \cc{E8F3F8}{\color[HTML]{000000} $0.949$} & \cc{B6D8E8}{\color[HTML]{000000} $0.947$} & \cc{F8FBFD}{\color[HTML]{000000} $0.949$} & \cc{FDEBE2}{\color[HTML]{000000} $0.950$\rlap{\textsuperscript{*}}} & \cc{FCE4D8}{\color[HTML]{000000} $0.951$\rlap{\textsuperscript{*}}} & \cc{F6FAFC}{\color[HTML]{000000} $0.105$} & \cc{D6E9F2}{\color[HTML]{000000} $0.109$} & \cc{FDE8DD}{\color[HTML]{000000} $0.101$} & \cc{FACDB7}{\color[HTML]{000000} $0.097$\rlap{\textsuperscript{*}}} & \cc{F9C1A6}{\color[HTML]{000000} $0.096$\rlap{\textsuperscript{*}}} \\
No D. & \cc{FFFFFF}{\color[HTML]{F1F1F1} \color{white} --} & \cc{FFFFFF}{\color[HTML]{F1F1F1} \color{white} --} & \cc{5C99C7}{\color[HTML]{F1F1F1} $31.85$} & \cc{8FBDDA}{\color[HTML]{000000} $32.13$} & \cc{B5D7E8}{\color[HTML]{000000} $32.33$} & \cc{FFFFFF}{\color[HTML]{F1F1F1} \color{white} --} & \cc{FFFFFF}{\color[HTML]{F1F1F1} \color{white} --} & \cc{5C99C7}{\color[HTML]{F1F1F1} $0.945$} & \cc{9FC9E0}{\color[HTML]{000000} $0.946$} & \cc{C4DFED}{\color[HTML]{000000} $0.947$} & \cc{FFFFFF}{\color[HTML]{F1F1F1} \color{white} --} & \cc{FFFFFF}{\color[HTML]{F1F1F1} \color{white} --} & \cc{5C99C7}{\color[HTML]{F1F1F1} $0.119$} & \cc{BBDAEA}{\color[HTML]{000000} $0.112$} & \cc{E6F1F7}{\color[HTML]{000000} $0.107$} \\
\bottomrule
\end{tabular}}
\endgroup
}
\end{subtable}
\begin{subtable}[t]{\linewidth}
\centering
\caption{ETH3D}
\label{tab:laser_scan_main_eth3d}
{\small
\begingroup \setlength{\tabcolsep}{2.5\tabcolsep}
\resizebox{\linewidth}{!}{\begin{tabular}{l|ccccc|ccccc|ccccc}
\toprule
& \multicolumn{5}{c|}{\textbf{PSNR ↑}} & \multicolumn{5}{c|}{\textbf{SSIM ↑}} & \multicolumn{5}{c}{\textbf{LPIPS ↓}} \\
& SfM & $|\mathcal{G}_\mathit{init}^\text{SfM}|$ & $0.5\text  {G}_\mathit{m}$ & $0.75\text  {G}_\mathit{m}$ & $1.0\text  {G}_\mathit{m}$ & SfM & $|\mathcal{G}_\mathit{init}^\text{SfM}|$ & $0.5\text  {G}_\mathit{m}$ & $0.75\text  {G}_\mathit{m}$ & $1.0\text  {G}_\mathit{m}$ & SfM & $|\mathcal{G}_\mathit{init}^\text{SfM}|$ & $0.5\text  {G}_\mathit{m}$ & $0.75\text  {G}_\mathit{m}$ & $1.0\text  {G}_\mathit{m}$ \\
\midrule
AbsGS & \cc{CBE3EF}{\color[HTML]{000000} $21.06$} & \cc{E6F1F7}{\color[HTML]{000000} $21.58$} & \cc{FCE7DC}{\color[HTML]{000000} $22.61$} & \cc{FBD7C5}{\color[HTML]{000000} $22.98$\rlap{\textsuperscript{*}}} & \cc{FBD5C2}{\color[HTML]{000000} $23.03$\rlap{\textsuperscript{*}}} & \cc{E7F2F8}{\color[HTML]{000000} $0.796$} & \cc{D6E9F2}{\color[HTML]{000000} $0.788$} & \cc{FAD2BE}{\color[HTML]{000000} $0.836$} & \cc{F8C0A4}{\color[HTML]{000000} $0.847$\rlap{\textsuperscript{*}}} & \cc{F6BA9F}{\color[HTML]{000000} $0.849$\rlap{\textsuperscript{*}}} & \cc{F3F9FB}{\color[HTML]{000000} $0.305$} & \cc{ECF5F9}{\color[HTML]{000000} $0.312$} & \cc{FACBB4}{\color[HTML]{000000} $0.235$} & \cc{F9C2A7}{\color[HTML]{000000} $0.224$\rlap{\textsuperscript{*}}} & \cc{F7BCA1}{\color[HTML]{000000} $0.219$\rlap{\textsuperscript{*}}} \\
INRIA & \cc{F9FCFD}{\color[HTML]{000000} $21.95$} & \cc{FDFEFE}{\color[HTML]{000000} $22.01$} & \cc{FBD9C8}{\color[HTML]{000000} $22.94$\rlap{\textsuperscript{*}}} & \cc{FACFBA}{\color[HTML]{000000} $23.16$\rlap{\textsuperscript{*}}} & \cc{F9C7AE}{\color[HTML]{000000} $23.34$\rlap{\textsuperscript{*}}} & \cc{FCE0D2}{\color[HTML]{000000} $0.827$} & \cc{FEF2EC}{\color[HTML]{000000} $0.816$} & \cc{F9C6AD}{\color[HTML]{000000} $0.843$} & \cc{EFA791}{\color[HTML]{000000} $0.854$} & \cc{E99985}{\color[HTML]{000000} $0.858$} & \cc{FFFCFA}{\color[HTML]{000000} $0.289$} & \cc{FFFDFB}{\color[HTML]{000000} $0.291$} & \cc{F9C1A6}{\color[HTML]{000000} $0.223$} & \cc{F3B199}{\color[HTML]{000000} $0.214$\rlap{\textsuperscript{*}}} & \cc{EFA791}{\color[HTML]{000000} $0.208$\rlap{\textsuperscript{*}}} \\
MCMC & \cc{F5B89E}{\color[HTML]{000000} $23.60$} & \cc{D86D63}{\color[HTML]{F1F1F1} $24.42$} & \cc{DA7368}{\color[HTML]{F1F1F1} $24.36$} & \cc{D97166}{\color[HTML]{F1F1F1} $24.40$} & \cc{D5675E}{\color[HTML]{F1F1F1} $24.51$} & \cc{F5B89E}{\color[HTML]{000000} $0.849$} & \cc{DA7368}{\color[HTML]{F1F1F1} $0.869$} & \cc{D76B61}{\color[HTML]{F1F1F1} $0.872$} & \cc{D76B61}{\color[HTML]{F1F1F1} $0.872$} & \cc{D5675E}{\color[HTML]{F1F1F1} $0.873$} & \cc{F9C8B0}{\color[HTML]{000000} $0.231$} & \cc{DF8072}{\color[HTML]{F1F1F1} $0.186$} & \cc{D76B61}{\color[HTML]{F1F1F1} $0.175$\rlap{\textsuperscript{*}}} & \cc{D5675E}{\color[HTML]{F1F1F1} $0.174$\rlap{\textsuperscript{*}}} & \cc{D5675E}{\color[HTML]{F1F1F1} $0.173$\rlap{\textsuperscript{*}}} \\
IDHFR & \cc{FDEFE7}{\color[HTML]{000000} $22.44$} & \cc{FFFCFA}{\color[HTML]{000000} $22.13$} & \cc{FCE6DB}{\color[HTML]{000000} $22.63$} & \cc{FACCB5}{\color[HTML]{000000} $23.23$\rlap{\textsuperscript{*}}} & \cc{F8C0A4}{\color[HTML]{000000} $23.51$\rlap{\textsuperscript{*}}} & \cc{FDEEE6}{\color[HTML]{000000} $0.819$} & \cc{FBFDFE}{\color[HTML]{000000} $0.806$} & \cc{FBD7C5}{\color[HTML]{000000} $0.833$} & \cc{F7BCA1}{\color[HTML]{000000} $0.848$} & \cc{EEA58F}{\color[HTML]{000000} $0.855$\rlap{\textsuperscript{*}}} & \cc{FDE8DD}{\color[HTML]{000000} $0.267$} & \cc{FDFEFE}{\color[HTML]{000000} $0.295$} & \cc{FAD1BC}{\color[HTML]{000000} $0.241$} & \cc{F2AF97}{\color[HTML]{000000} $0.213$} & \cc{E89784}{\color[HTML]{000000} $0.200$\rlap{\textsuperscript{*}}} \\
RevDGS & \cc{5C99C7}{\color[HTML]{F1F1F1} $19.58$} & \cc{7AAED2}{\color[HTML]{F1F1F1} $19.92$} & \cc{FCDFD1}{\color[HTML]{000000} $22.80$} & \cc{FBD8C7}{\color[HTML]{000000} $22.96$\rlap{\textsuperscript{*}}} & \cc{FCE6DB}{\color[HTML]{000000} $22.63$} & \cc{5E9BC8}{\color[HTML]{F1F1F1} $0.745$} & \cc{5C99C7}{\color[HTML]{F1F1F1} $0.744$} & \cc{FCDFD1}{\color[HTML]{000000} $0.828$\rlap{\textsuperscript{*}}} & \cc{FAD3BF}{\color[HTML]{000000} $0.835$\rlap{\textsuperscript{*}}} & \cc{FCE1D3}{\color[HTML]{000000} $0.827$\rlap{\textsuperscript{*}}} & \cc{649FCA}{\color[HTML]{F1F1F1} $0.410$} & \cc{5C99C7}{\color[HTML]{F1F1F1} $0.414$} & \cc{FBDBCB}{\color[HTML]{000000} $0.252$\rlap{\textsuperscript{*}}} & \cc{FBD5C2}{\color[HTML]{000000} $0.246$\rlap{\textsuperscript{*}}} & \cc{FCDECF}{\color[HTML]{000000} $0.256$\rlap{\textsuperscript{*}}} \\
No D. & \cc{FFFFFF}{\color[HTML]{F1F1F1} \color{white} --} & \cc{FFFFFF}{\color[HTML]{F1F1F1} \color{white} --} & \cc{FCE4D8}{\color[HTML]{000000} $22.67$} & \cc{FAD0BB}{\color[HTML]{000000} $23.14$} & \cc{FACDB7}{\color[HTML]{000000} $23.22$} & \cc{FFFFFF}{\color[HTML]{F1F1F1} \color{white} --} & \cc{FFFFFF}{\color[HTML]{F1F1F1} \color{white} --} & \cc{F6BA9F}{\color[HTML]{000000} $0.849$} & \cc{F0AB94}{\color[HTML]{000000} $0.853$} & \cc{F0A992}{\color[HTML]{000000} $0.853$} & \cc{FFFFFF}{\color[HTML]{F1F1F1} \color{white} --} & \cc{FFFFFF}{\color[HTML]{F1F1F1} \color{white} --} & \cc{FBDCCC}{\color[HTML]{000000} $0.253$} & \cc{F9CAB2}{\color[HTML]{000000} $0.234$} & \cc{F9C2A7}{\color[HTML]{000000} $0.225$} \\
\bottomrule
\end{tabular}}
\endgroup
}
\end{subtable}
\end{table}

We begin by evaluating NVS performance using SfM and laser scan initialization on ScanNet++ (default, on-trajectory) and ETH3D, following the protocol described in~\cref{sec:protocol:laser_scan_init}.
The purpose of this experiment is to establish an accurate initialization baseline and isolate the effects of initialization size. 

Analyzing the results in~\cref{tab:laser_scan_main}, we find that laser scan initialization does lead to consistent improvements in NVS performance.
However, they are most pronounced for densification strategies that show lower baseline results using SfM.
Comparing \cref{tab:laser_scan_main_scannet++,tab:laser_scan_main_eval_on_train_set_scannet++}, we see that dense initialization helps generalization to new views more than it helps for views close to the training trajectory.
In most cases, we see very limited performance scaling with increasing initialization size. The effect is greatest on ETH3D, where larger initialization likely limits overfitting to the limited number of training views.
Comparing results using SfM and laser scan initalization at the same size (where we of course can't use hybrid initialization), the performance delta is less uniform, likely partly due to the missing regions in the laser scan data. The improvements we do see are on off-trajectory ScanNet++ and ETH3D, and only some of the strategies benefit.

\begin{figure}[tb]
  \centering
  \centering
  
  \includegraphics[width=0.2485\linewidth]{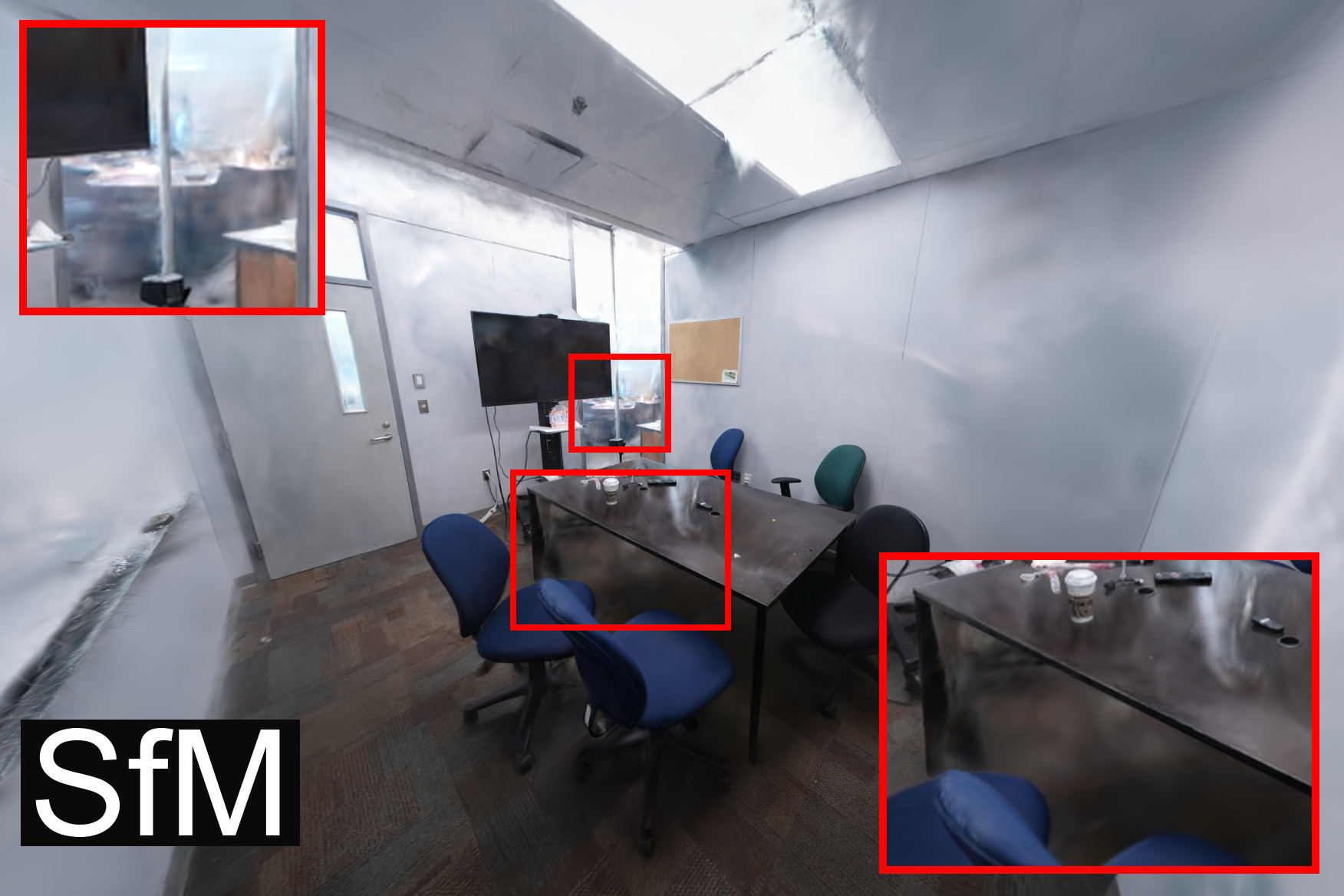}\hfill
  \includegraphics[width=0.2485\linewidth]{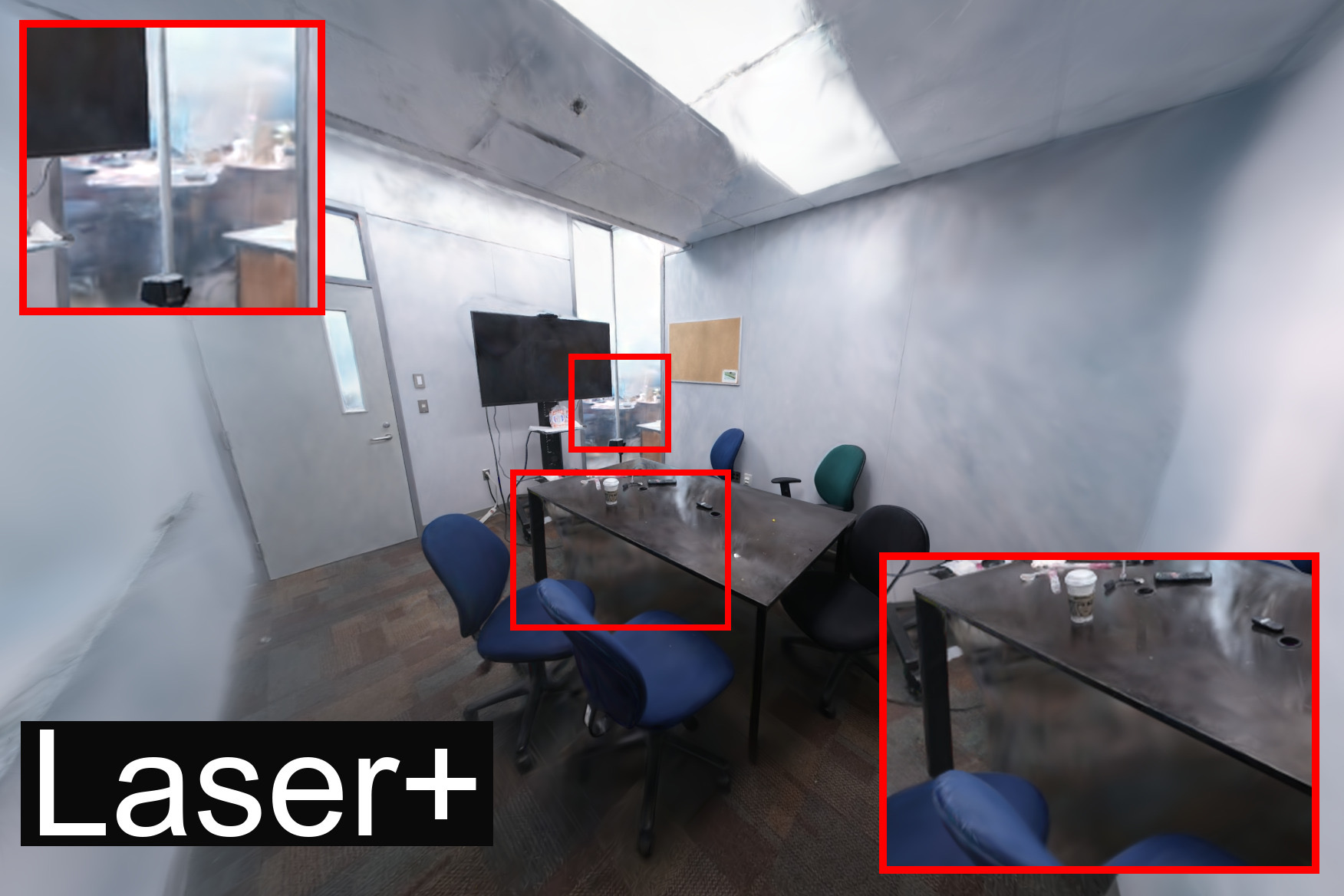}\hfill
  \includegraphics[width=0.2485\linewidth]{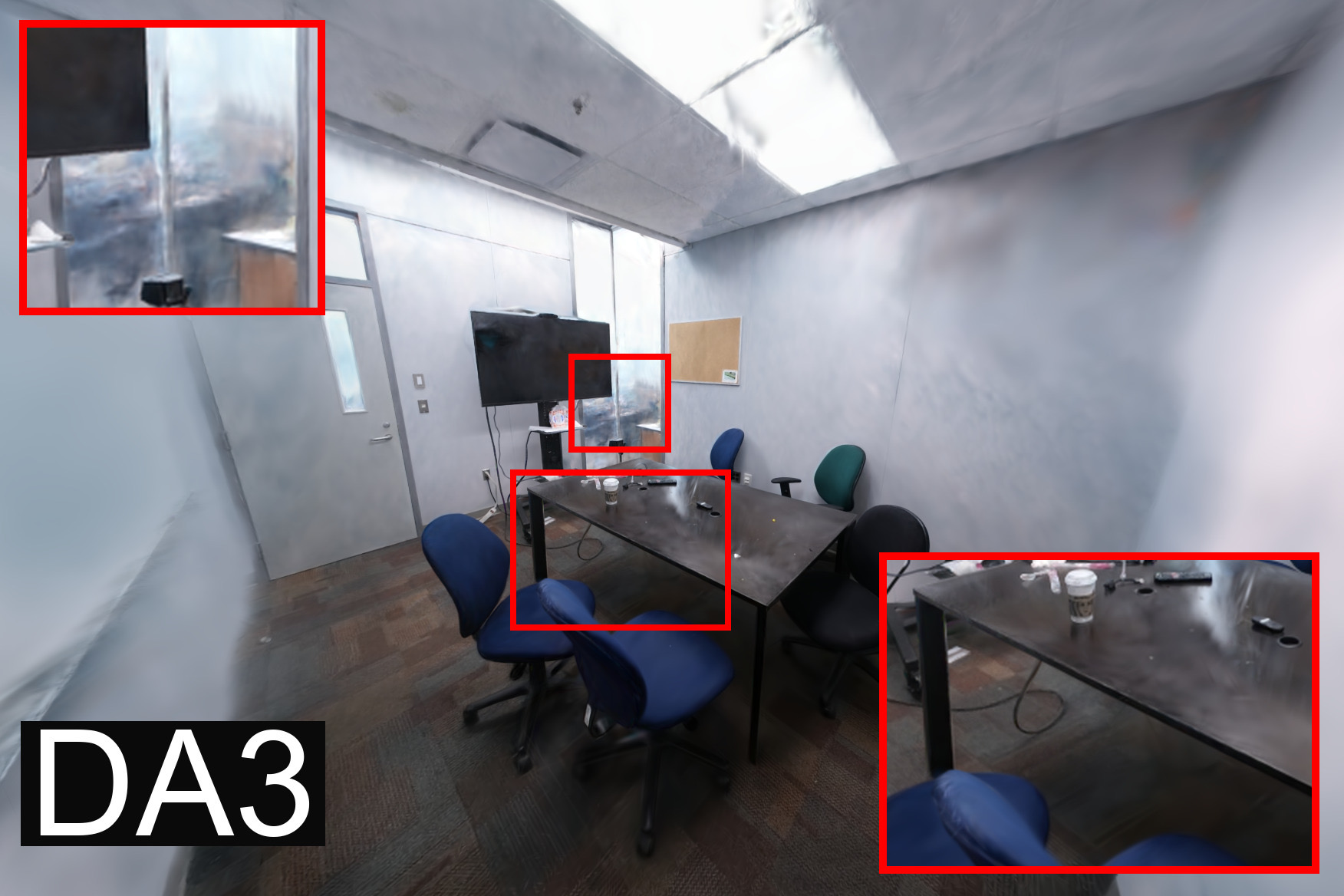}\hfill
  \includegraphics[width=0.2485\linewidth]{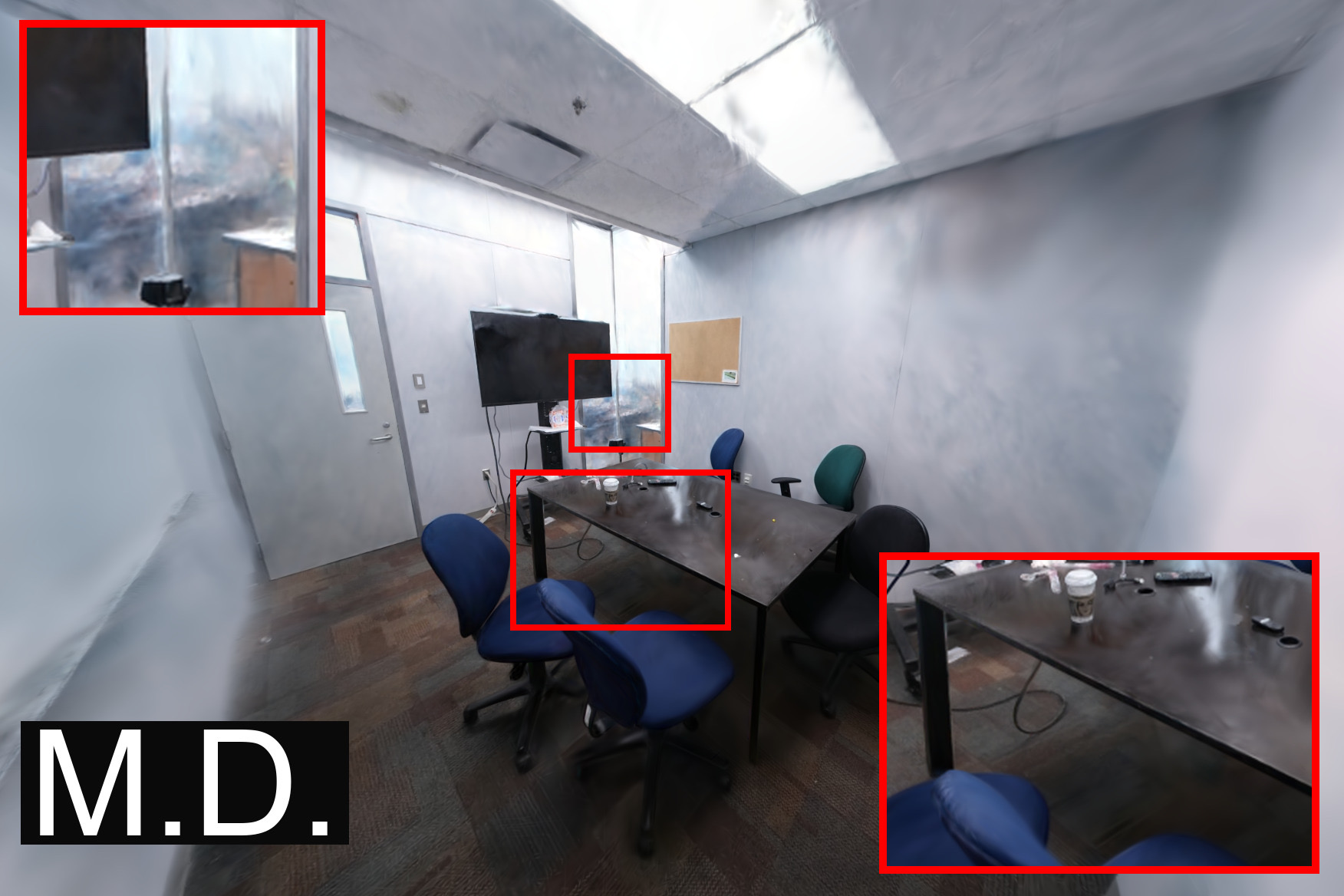}\hfill

  \caption{Qualitative results on an off-trajectory view of the ``5748ce6f01'' ScanNet++ scene using IDHFR densification. Note how adding Laser points to SfM points reduces artifacts under the reflective table, whereas not using SfM at all (DA3, M.D.) resolves them entirely, but reduces quality in the area behind the glass window.}
  \label{fig:nvs_example}
\end{figure}

Overall, we find that using laser scan initialization has a limited effect on NVS compared to picking a better densification method. In the vast majority of cases, even the best results achieved by a given strategy with dense initialization are worse than what is achieved with SfM by the best strategy for the given dataset.
Despite that, we do see statistically significant improvements, and, as showcased in~\cref{fig:nvs_example}, dense initialization can provide real generalization benefits.

\begin{table}[tb]
\centering
\caption{Noise resiliency with non-hybrid laser scan init with $0.5G_m$ initial points.}
\label{tab:noise_resiliency}
\begin{subtable}[t]{0.5374\linewidth}
\centering
\caption{ScanNet++}
\label{tab:noise_resiliency_scannet++}
{\small
\begingroup \setlength{\tabcolsep}{2.5\tabcolsep}
\resizebox{\linewidth}{!}{\begin{tabular}{l|ccc|ccc|ccc}
\toprule
& \multicolumn{3}{c|}{\textbf{PSNR ↑}} & \multicolumn{3}{c|}{\textbf{SSIM ↑}} & \multicolumn{3}{c}{\textbf{LPIPS ↓}} \\
Noise std.& 0\% & 1\% & 10\% & 0\% & 1\% & 10\% & 0\% & 1\% & 10\% \\
\midrule
AbsGS & \cc{E18676}{\color[HTML]{F1F1F1} $22.75$} & \cc{E58E7D}{\color[HTML]{F1F1F1} $22.61$} & \cc{FACFBA}{\color[HTML]{000000} $21.32$} & \cc{E18475}{\color[HTML]{F1F1F1} $0.871$} & \cc{E38A7A}{\color[HTML]{F1F1F1} $0.870$} & \cc{F8BEA3}{\color[HTML]{000000} $0.858$} & \cc{E08273}{\color[HTML]{F1F1F1} $0.249$} & \cc{E48C7B}{\color[HTML]{F1F1F1} $0.254$} & \cc{F9C4AA}{\color[HTML]{000000} $0.284$} \\
INRIA & \cc{E18676}{\color[HTML]{F1F1F1} $22.75$} & \cc{E58E7D}{\color[HTML]{F1F1F1} $22.59$} & \cc{F9CAB2}{\color[HTML]{000000} $21.47$} & \cc{E08273}{\color[HTML]{F1F1F1} $0.872$} & \cc{E28878}{\color[HTML]{F1F1F1} $0.870$} & \cc{F6BA9F}{\color[HTML]{000000} $0.859$} & \cc{E08273}{\color[HTML]{F1F1F1} $0.249$} & \cc{E58E7D}{\color[HTML]{F1F1F1} $0.255$} & \cc{F9C5AB}{\color[HTML]{000000} $0.285$} \\
MCMC & \cc{E79482}{\color[HTML]{F1F1F1} $22.52$} & \cc{E79482}{\color[HTML]{F1F1F1} $22.50$} & \cc{EB9F8A}{\color[HTML]{000000} $22.34$} & \cc{DC796D}{\color[HTML]{F1F1F1} $0.873$} & \cc{DC776B}{\color[HTML]{F1F1F1} $0.874$} & \cc{DD7C6E}{\color[HTML]{F1F1F1} $0.873$} & \cc{DC776B}{\color[HTML]{F1F1F1} $0.244$} & \cc{DD7C6E}{\color[HTML]{F1F1F1} $0.246$} & \cc{DE7E70}{\color[HTML]{F1F1F1} $0.248$} \\
IDHFR & \cc{D5675E}{\color[HTML]{F1F1F1} $23.24$} & \cc{D86F65}{\color[HTML]{F1F1F1} $23.11$} & \cc{EEA58F}{\color[HTML]{000000} $22.23$} & \cc{D5675E}{\color[HTML]{F1F1F1} $0.878$} & \cc{D76B61}{\color[HTML]{F1F1F1} $0.877$} & \cc{E38A7A}{\color[HTML]{F1F1F1} $0.870$} & \cc{D5675E}{\color[HTML]{F1F1F1} $0.236$} & \cc{D86F65}{\color[HTML]{F1F1F1} $0.240$} & \cc{E69280}{\color[HTML]{F1F1F1} $0.258$} \\
RevDGS & \cc{E48C7B}{\color[HTML]{F1F1F1} $22.64$} & \cc{E89784}{\color[HTML]{000000} $22.49$} & \cc{5C99C7}{\color[HTML]{F1F1F1} $16.16$} & \cc{E28878}{\color[HTML]{F1F1F1} $0.870$} & \cc{E6907F}{\color[HTML]{F1F1F1} $0.868$} & \cc{5C99C7}{\color[HTML]{F1F1F1} $0.775$} & \cc{E6907F}{\color[HTML]{F1F1F1} $0.256$} & \cc{EB9D88}{\color[HTML]{000000} $0.262$} & \cc{5C99C7}{\color[HTML]{F1F1F1} $0.453$} \\
\bottomrule
\end{tabular}}
\endgroup
}
\end{subtable}
\hfill
\begin{subtable}[t]{0.4526\linewidth}
\centering
\caption{ScanNet++ (On-Trajectory)}
\label{tab:noise_resiliency_eval_on_train_set_scannet++}
{\small
\begingroup \setlength{\tabcolsep}{2.5\tabcolsep}
\resizebox{\linewidth}{!}{\begin{tabular}{ccc|ccc|ccc}
\toprule
\multicolumn{3}{c|}{\textbf{PSNR ↑}} & \multicolumn{3}{c|}{\textbf{SSIM ↑}} & \multicolumn{3}{c}{\textbf{LPIPS ↓}} \\
0\% & 1\% & 10\% & 0\% & 1\% & 10\% & 0\% & 1\% & 10\% \\
\midrule
 \cc{E6907F}{\color[HTML]{F1F1F1} $33.00$} & \cc{E6907F}{\color[HTML]{F1F1F1} $32.99$} & \cc{F9C3A8}{\color[HTML]{000000} $31.94$} & \cc{DB7569}{\color[HTML]{F1F1F1} $0.950$} & \cc{DC796D}{\color[HTML]{F1F1F1} $0.949$} & \cc{E99985}{\color[HTML]{000000} $0.944$} & \cc{DA7368}{\color[HTML]{F1F1F1} $0.100$} & \cc{DE7E70}{\color[HTML]{F1F1F1} $0.104$} & \cc{F0A992}{\color[HTML]{000000} $0.123$} \\
 \cc{E79482}{\color[HTML]{F1F1F1} $32.91$} & \cc{E79482}{\color[HTML]{F1F1F1} $32.91$} & \cc{F5B89E}{\color[HTML]{000000} $32.22$} & \cc{DC796D}{\color[HTML]{F1F1F1} $0.949$} & \cc{DE7E70}{\color[HTML]{F1F1F1} $0.948$} & \cc{EB9D88}{\color[HTML]{000000} $0.943$} & \cc{DD7C6E}{\color[HTML]{F1F1F1} $0.103$} & \cc{E28878}{\color[HTML]{F1F1F1} $0.109$} & \cc{F7BCA1}{\color[HTML]{000000} $0.131$} \\
 \cc{E89784}{\color[HTML]{000000} $32.86$} & \cc{E89784}{\color[HTML]{000000} $32.87$} & \cc{E6907F}{\color[HTML]{F1F1F1} $32.98$} & \cc{D86D63}{\color[HTML]{F1F1F1} $0.952$} & \cc{D86D63}{\color[HTML]{F1F1F1} $0.951$} & \cc{D86F65}{\color[HTML]{F1F1F1} $0.951$} & \cc{DA7368}{\color[HTML]{F1F1F1} $0.100$} & \cc{DB7569}{\color[HTML]{F1F1F1} $0.100$} & \cc{DC776B}{\color[HTML]{F1F1F1} $0.101$} \\
 \cc{D66960}{\color[HTML]{F1F1F1} $33.75$} & \cc{D5675E}{\color[HTML]{F1F1F1} $33.81$} & \cc{DC776B}{\color[HTML]{F1F1F1} $33.46$} & \cc{D5675E}{\color[HTML]{F1F1F1} $0.953$} & \cc{D5675E}{\color[HTML]{F1F1F1} $0.953$} & \cc{D86F65}{\color[HTML]{F1F1F1} $0.951$} & \cc{D5675E}{\color[HTML]{F1F1F1} $0.094$} & \cc{D76B61}{\color[HTML]{F1F1F1} $0.096$} & \cc{DD7C6E}{\color[HTML]{F1F1F1} $0.104$} \\
 \cc{EB9D88}{\color[HTML]{000000} $32.77$} & \cc{EEA58F}{\color[HTML]{000000} $32.60$} & \cc{5C99C7}{\color[HTML]{F1F1F1} $25.26$} & \cc{DF8072}{\color[HTML]{F1F1F1} $0.948$} & \cc{E28878}{\color[HTML]{F1F1F1} $0.946$} & \cc{5C99C7}{\color[HTML]{F1F1F1} $0.871$} & \cc{DD7C6E}{\color[HTML]{F1F1F1} $0.104$} & \cc{E38A7A}{\color[HTML]{F1F1F1} $0.110$} & \cc{5C99C7}{\color[HTML]{F1F1F1} $0.288$} \\
\bottomrule
\end{tabular}}
\endgroup
}
\end{subtable}
\end{table}

\PAR{Resiliency to Noise.}\label{par:noise_resiliency} To quantify the effects of initialization \textit{accuracy} on NVS, we train with laser scan point clouds
perturbed with Gaussian noise at standard deviations of $1\%$ and $10\%$ of the scene extent $S$.
Following the implementation of~\cite{3dgs}, given $N_C$ cameras with centers $C_i$, we define $S$ as $\max_{i \in \hat{N_C}} \left\lVert C_i - C_{\text{mean}} \right\rVert$, where $C_\text{mean}$ is the centroid of all $C_i$.
We provide visual examples of these noise levels in \cref{noise_examples} of the supp.\ mat.
Looking at \cref{tab:noise_resiliency}, all densification methods show only minor degradation at $\sigma=0.01{}S_\text{scene}$.
At $\sigma=0.1S_\text{scene}$, performance drops quite significantly, especially for under-constrained views.
However even at this noise level, IDHFR and MCMC maintain very strong on-trajectory NVS performance, showcasing their initialization invariance.
Nevertheless, even here, initialization precision affects generalization, as even these strategies exhibit degradation in off-trajectory NVS at $\sigma=0.1S_\text{scene}$.
We also report ETH3D results in \Cref{excluded_eth3d_results} of the supp.\ mat., where the behavior is similar to~\Cref{tab:noise_resiliency_scannet++}.

\subsection{Novel View Synthesis with Practical Initialization Methods}\label{subsec:eval:practical_init}

\begin{table}[tb]
\centering
\caption{NVS results using all evaluated initialization and densification methods.
Cells with significant improvements over SfM are marked with ``$^*$''}
\label{tab:practical_main}
\begin{subtable}[t]{\linewidth}
\centering
\caption{ScanNet++}
\label{tab:practical_main_scannet++}
{\small
\begingroup \setlength{\tabcolsep}{2.5\tabcolsep}
\resizebox{\linewidth}{!}{\begin{tabular}{l|cccccc|cccccc|cccccc}
\toprule
& \multicolumn{6}{c|}{\textbf{PSNR ↑}} & \multicolumn{6}{c|}{\textbf{SSIM ↑}} & \multicolumn{6}{c}{\textbf{LPIPS ↓}} \\
& SfM & $\text{EDGS}^*$ & M. D. & DA3 & $\text{DA3}^\text{G.S.}$ & $\text{Laser}^{+}$ & SfM & $\text{EDGS}^*$ & M. D. & DA3 & $\text{DA3}^\text{G.S.}$ & $\text{Laser}^{+}$ & SfM & $\text{EDGS}^*$ & M. D. & DA3 & $\text{DA3}^\text{G.S.}$ & $\text{Laser}^{+}$ \\
\midrule
AbsGS & \cc{B2D6E7}{\color[HTML]{000000} $22.36$} & \cc{E1EFF6}{\color[HTML]{000000} $22.64$} & \cc{FEF5F0}{\color[HTML]{000000} $22.91$} & \cc{FEF2EC}{\color[HTML]{000000} $22.94$} & \cc{F7BCA1}{\color[HTML]{000000} $23.33$\rlap{\textsuperscript{*}}} & \cc{FCE3D6}{\color[HTML]{000000} $23.05$\rlap{\textsuperscript{*}}} & \cc{FDEEE6}{\color[HTML]{000000} $0.870$} & \cc{C7E1EE}{\color[HTML]{000000} $0.863$} & \cc{FAD1BC}{\color[HTML]{000000} $0.873$} & \cc{FACFBA}{\color[HTML]{000000} $0.873$} & \cc{FDE9DF}{\color[HTML]{000000} $0.871$} & \cc{F9C6AD}{\color[HTML]{000000} $0.874$\rlap{\textsuperscript{*}}} & \cc{FEFFFF}{\color[HTML]{000000} $0.249$} & \cc{E5F1F7}{\color[HTML]{000000} $0.254$} & \cc{FEF1EA}{\color[HTML]{000000} $0.246$} & \cc{FDEEE6}{\color[HTML]{000000} $0.245$} & \cc{F8C0A4}{\color[HTML]{000000} $0.235$\rlap{\textsuperscript{*}}} & \cc{FAD1BC}{\color[HTML]{000000} $0.239$\rlap{\textsuperscript{*}}} \\
INRIA & \cc{9BC6DF}{\color[HTML]{000000} $22.26$} & \cc{DCECF4}{\color[HTML]{000000} $22.62$} & \cc{FEFAF7}{\color[HTML]{000000} $22.87$\rlap{\textsuperscript{*}}} & \cc{FEFAF7}{\color[HTML]{000000} $22.87$\rlap{\textsuperscript{*}}} & \cc{FACEB8}{\color[HTML]{000000} $23.21$\rlap{\textsuperscript{*}}} & \cc{FDEFE7}{\color[HTML]{000000} $22.96$\rlap{\textsuperscript{*}}} & \cc{FEF3ED}{\color[HTML]{000000} $0.870$} & \cc{C5DFED}{\color[HTML]{000000} $0.863$} & \cc{FAD3BF}{\color[HTML]{000000} $0.873$} & \cc{FAD1BC}{\color[HTML]{000000} $0.873$} & \cc{FEF8F4}{\color[HTML]{000000} $0.869$} & \cc{FACBB4}{\color[HTML]{000000} $0.874$\rlap{\textsuperscript{*}}} & \cc{E8F3F8}{\color[HTML]{000000} $0.253$} & \cc{DDEDF5}{\color[HTML]{000000} $0.255$} & \cc{FEF2EC}{\color[HTML]{000000} $0.246$} & \cc{FEF4EF}{\color[HTML]{000000} $0.246$} & \cc{FAD2BE}{\color[HTML]{000000} $0.239$\rlap{\textsuperscript{*}}} & \cc{FBD9C8}{\color[HTML]{000000} $0.240$\rlap{\textsuperscript{*}}} \\
MCMC & \cc{5C99C7}{\color[HTML]{F1F1F1} $22.02$} & \cc{E0EEF5}{\color[HTML]{000000} $22.64$\rlap{\textsuperscript{*}}} & \cc{D2E7F1}{\color[HTML]{000000} $22.56$\rlap{\textsuperscript{*}}} & \cc{D3E7F2}{\color[HTML]{000000} $22.56$\rlap{\textsuperscript{*}}} & \cc{ECF5F9}{\color[HTML]{000000} $22.71$\rlap{\textsuperscript{*}}} & \cc{C0DDEB}{\color[HTML]{000000} $22.44$\rlap{\textsuperscript{*}}} & \cc{FDEFE7}{\color[HTML]{000000} $0.870$} & \cc{F8BEA3}{\color[HTML]{000000} $0.875$} & \cc{F4B49A}{\color[HTML]{000000} $0.876$\rlap{\textsuperscript{*}}} & \cc{F9C2A7}{\color[HTML]{000000} $0.875$\rlap{\textsuperscript{*}}} & \cc{F9C1A6}{\color[HTML]{000000} $0.875$\rlap{\textsuperscript{*}}} & \cc{F9C5AB}{\color[HTML]{000000} $0.874$} & \cc{FEFFFF}{\color[HTML]{000000} $0.249$} & \cc{FBD9C8}{\color[HTML]{000000} $0.240$\rlap{\textsuperscript{*}}} & \cc{FCDECF}{\color[HTML]{000000} $0.241$\rlap{\textsuperscript{*}}} & \cc{FCDECF}{\color[HTML]{000000} $0.242$\rlap{\textsuperscript{*}}} & \cc{F9C4AA}{\color[HTML]{000000} $0.236$\rlap{\textsuperscript{*}}} & \cc{FAD3BF}{\color[HTML]{000000} $0.239$\rlap{\textsuperscript{*}}} \\
IDHFR & \cc{FCE4D8}{\color[HTML]{000000} $23.04$} & \cc{FCE2D5}{\color[HTML]{000000} $23.06$} & \cc{F4B49A}{\color[HTML]{000000} $23.36$} & \cc{F4B69C}{\color[HTML]{000000} $23.36$} & \cc{D5675E}{\color[HTML]{F1F1F1} $23.65$\rlap{\textsuperscript{*}}} & \cc{F4B49A}{\color[HTML]{000000} $23.36$} & \cc{E89784}{\color[HTML]{000000} $0.877$} & \cc{FBD7C5}{\color[HTML]{000000} $0.872$} & \cc{D86D63}{\color[HTML]{F1F1F1} $0.879$} & \cc{D66960}{\color[HTML]{F1F1F1} $0.879$} & \cc{F0AB94}{\color[HTML]{000000} $0.876$} & \cc{D5675E}{\color[HTML]{F1F1F1} $0.879$} & \cc{F2AF97}{\color[HTML]{000000} $0.233$} & \cc{FCE1D3}{\color[HTML]{000000} $0.242$} & \cc{F3B199}{\color[HTML]{000000} $0.233$} & \cc{F4B49A}{\color[HTML]{000000} $0.234$} & \cc{D5675E}{\color[HTML]{F1F1F1} $0.225$\rlap{\textsuperscript{*}}} & \cc{E69280}{\color[HTML]{F1F1F1} $0.230$} \\
RevDGS & \cc{BCDBEA}{\color[HTML]{000000} $22.42$} & \cc{AFD4E6}{\color[HTML]{000000} $22.34$} & \cc{FEFAF7}{\color[HTML]{000000} $22.88$} & \cc{FFFBF9}{\color[HTML]{000000} $22.87$} & \cc{EFA791}{\color[HTML]{000000} $23.41$\rlap{\textsuperscript{*}}} & \cc{FDEDE5}{\color[HTML]{000000} $22.97$} & \cc{FEF5F0}{\color[HTML]{000000} $0.869$} & \cc{A5CDE3}{\color[HTML]{000000} $0.861$} & \cc{FCE0D2}{\color[HTML]{000000} $0.872$} & \cc{FCDFD1}{\color[HTML]{000000} $0.872$} & \cc{FEFAF7}{\color[HTML]{000000} $0.869$} & \cc{FAD2BE}{\color[HTML]{000000} $0.873$\rlap{\textsuperscript{*}}} & \cc{D6E9F2}{\color[HTML]{000000} $0.257$} & \cc{B6D8E8}{\color[HTML]{000000} $0.262$} & \cc{F8FBFD}{\color[HTML]{000000} $0.250$} & \cc{F7FBFD}{\color[HTML]{000000} $0.251$} & \cc{FAD0BB}{\color[HTML]{000000} $0.238$\rlap{\textsuperscript{*}}} & \cc{FEF4EF}{\color[HTML]{000000} $0.246$\rlap{\textsuperscript{*}}} \\
No D. & \cc{FFFFFF}{\color[HTML]{F1F1F1} \color{white} --} & \cc{BCDBEA}{\color[HTML]{000000} $22.41$} & \cc{D8EAF3}{\color[HTML]{000000} $22.59$} & \cc{DDEDF5}{\color[HTML]{000000} $22.62$} & \cc{89B9D8}{\color[HTML]{000000} $22.20$} & \cc{D3E7F2}{\color[HTML]{000000} $22.56$} & \cc{FFFFFF}{\color[HTML]{F1F1F1} \color{white} --} & \cc{85B6D7}{\color[HTML]{000000} $0.859$} & \cc{FEF2EC}{\color[HTML]{000000} $0.870$} & \cc{FDEFE7}{\color[HTML]{000000} $0.870$} & \cc{5C99C7}{\color[HTML]{F1F1F1} $0.857$} & \cc{FCE4D8}{\color[HTML]{000000} $0.871$} & \cc{FFFFFF}{\color[HTML]{F1F1F1} \color{white} --} & \cc{9DC7E0}{\color[HTML]{000000} $0.266$} & \cc{D5E8F2}{\color[HTML]{000000} $0.257$} & \cc{D6E9F2}{\color[HTML]{000000} $0.257$} & \cc{5C99C7}{\color[HTML]{F1F1F1} $0.273$} & \cc{EAF3F8}{\color[HTML]{000000} $0.253$} \\
\bottomrule
\end{tabular}}
\endgroup
}
\end{subtable}
\begin{subtable}[t]{\linewidth}
\centering
\caption{ScanNet++ (On-Trajectory)}
\label{tab:practical_main_eval_on_train_set_scannet++}
{\small
\begingroup \setlength{\tabcolsep}{2.5\tabcolsep}
\resizebox{\linewidth}{!}{\begin{tabular}{l|cccccc|cccccc|cccccc}
\toprule
& \multicolumn{6}{c|}{\textbf{PSNR ↑}} & \multicolumn{6}{c|}{\textbf{SSIM ↑}} & \multicolumn{6}{c}{\textbf{LPIPS ↓}} \\
& SfM & $\text{EDGS}^*$ & M. D. & DA3 & $\text{DA3}^\text{G.S.}$ & $\text{Laser}^{+}$ & SfM & $\text{EDGS}^*$ & M. D. & DA3 & $\text{DA3}^\text{G.S.}$ & $\text{Laser}^{+}$ & SfM & $\text{EDGS}^*$ & M. D. & DA3 & $\text{DA3}^\text{G.S.}$ & $\text{Laser}^{+}$ \\
\midrule
AbsGS & \cc{FAD3BF}{\color[HTML]{000000} $33.20$} & \cc{FDE9DF}{\color[HTML]{000000} $32.88$} & \cc{FDE8DD}{\color[HTML]{000000} $32.88$} & \cc{FDE9DF}{\color[HTML]{000000} $32.88$} & \cc{F9C2A7}{\color[HTML]{000000} $33.45$} & \cc{F9C4AA}{\color[HTML]{000000} $33.43$} & \cc{F3B199}{\color[HTML]{000000} $0.951$} & \cc{FBD4C1}{\color[HTML]{000000} $0.949$} & \cc{F9C9B1}{\color[HTML]{000000} $0.950$} & \cc{F9C8B0}{\color[HTML]{000000} $0.950$} & \cc{F9C4AA}{\color[HTML]{000000} $0.950$} & \cc{EA9B87}{\color[HTML]{000000} $0.952$\rlap{\textsuperscript{*}}} & \cc{FACBB4}{\color[HTML]{000000} $0.101$} & \cc{EFA791}{\color[HTML]{000000} $0.096$\rlap{\textsuperscript{*}}} & \cc{F9C2A7}{\color[HTML]{000000} $0.099$} & \cc{F9C1A6}{\color[HTML]{000000} $0.099$} & \cc{F3B199}{\color[HTML]{000000} $0.097$\rlap{\textsuperscript{*}}} & \cc{E28878}{\color[HTML]{F1F1F1} $0.093$\rlap{\textsuperscript{*}}} \\
INRIA & \cc{FBDAC9}{\color[HTML]{000000} $33.10$} & \cc{FDEEE6}{\color[HTML]{000000} $32.80$} & \cc{FEF1EA}{\color[HTML]{000000} $32.77$} & \cc{FDF0E9}{\color[HTML]{000000} $32.78$} & \cc{FACFBA}{\color[HTML]{000000} $33.26$} & \cc{FACCB5}{\color[HTML]{000000} $33.31$\rlap{\textsuperscript{*}}} & \cc{F9C9B1}{\color[HTML]{000000} $0.950$} & \cc{FBDCCC}{\color[HTML]{000000} $0.948$} & \cc{FAD1BC}{\color[HTML]{000000} $0.949$} & \cc{FACFBA}{\color[HTML]{000000} $0.949$} & \cc{FBDBCB}{\color[HTML]{000000} $0.948$} & \cc{F2AF97}{\color[HTML]{000000} $0.951$\rlap{\textsuperscript{*}}} & \cc{FDFEFE}{\color[HTML]{000000} $0.112$} & \cc{F4B69C}{\color[HTML]{000000} $0.097$\rlap{\textsuperscript{*}}} & \cc{FACBB4}{\color[HTML]{000000} $0.101$\rlap{\textsuperscript{*}}} & \cc{FACBB4}{\color[HTML]{000000} $0.101$\rlap{\textsuperscript{*}}} & \cc{FAD1BC}{\color[HTML]{000000} $0.102$\rlap{\textsuperscript{*}}} & \cc{EB9F8A}{\color[HTML]{000000} $0.095$\rlap{\textsuperscript{*}}} \\
MCMC & \cc{FCE3D6}{\color[HTML]{000000} $32.97$} & \cc{FBDAC9}{\color[HTML]{000000} $33.09$} & \cc{FDEEE6}{\color[HTML]{000000} $32.80$} & \cc{FDEEE6}{\color[HTML]{000000} $32.81$} & \cc{FACFBA}{\color[HTML]{000000} $33.27$} & \cc{FBD6C4}{\color[HTML]{000000} $33.15$} & \cc{F3B199}{\color[HTML]{000000} $0.951$} & \cc{EEA58F}{\color[HTML]{000000} $0.952$\rlap{\textsuperscript{*}}} & \cc{EFA791}{\color[HTML]{000000} $0.952$} & \cc{EFA791}{\color[HTML]{000000} $0.952$} & \cc{ECA18C}{\color[HTML]{000000} $0.952$} & \cc{E48C7B}{\color[HTML]{F1F1F1} $0.953$\rlap{\textsuperscript{*}}} & \cc{FBDCCC}{\color[HTML]{000000} $0.104$} & \cc{F0A992}{\color[HTML]{000000} $0.096$\rlap{\textsuperscript{*}}} & \cc{F9C3A8}{\color[HTML]{000000} $0.099$\rlap{\textsuperscript{*}}} & \cc{F9C3A8}{\color[HTML]{000000} $0.099$\rlap{\textsuperscript{*}}} & \cc{F9CAB2}{\color[HTML]{000000} $0.100$} & \cc{EB9D88}{\color[HTML]{000000} $0.095$\rlap{\textsuperscript{*}}} \\
IDHFR & \cc{E99985}{\color[HTML]{000000} $33.76$} & \cc{E89784}{\color[HTML]{000000} $33.77$} & \cc{ECA18C}{\color[HTML]{000000} $33.71$} & \cc{EB9F8A}{\color[HTML]{000000} $33.72$} & \cc{D5675E}{\color[HTML]{F1F1F1} $34.12$\rlap{\textsuperscript{*}}} & \cc{D66960}{\color[HTML]{F1F1F1} $34.11$} & \cc{E38A7A}{\color[HTML]{F1F1F1} $0.953$} & \cc{E38A7A}{\color[HTML]{F1F1F1} $0.953$} & \cc{E38A7A}{\color[HTML]{F1F1F1} $0.953$} & \cc{E48C7B}{\color[HTML]{F1F1F1} $0.953$} & \cc{DF8072}{\color[HTML]{F1F1F1} $0.953$} & \cc{D5675E}{\color[HTML]{F1F1F1} $0.954$\rlap{\textsuperscript{*}}} & \cc{EB9D88}{\color[HTML]{000000} $0.095$} & \cc{DA7368}{\color[HTML]{F1F1F1} $0.091$\rlap{\textsuperscript{*}}} & \cc{E6907F}{\color[HTML]{F1F1F1} $0.094$} & \cc{E6907F}{\color[HTML]{F1F1F1} $0.094$} & \cc{DC796D}{\color[HTML]{F1F1F1} $0.091$\rlap{\textsuperscript{*}}} & \cc{D5675E}{\color[HTML]{F1F1F1} $0.090$\rlap{\textsuperscript{*}}} \\
RevDGS & \cc{FCE5D9}{\color[HTML]{000000} $32.93$} & \cc{FDFEFE}{\color[HTML]{000000} $32.52$} & \cc{FDEEE6}{\color[HTML]{000000} $32.81$} & \cc{FDEEE6}{\color[HTML]{000000} $32.81$} & \cc{FBD8C7}{\color[HTML]{000000} $33.12$} & \cc{FBD7C5}{\color[HTML]{000000} $33.14$} & \cc{FBD7C5}{\color[HTML]{000000} $0.949$} & \cc{FDEAE0}{\color[HTML]{000000} $0.947$} & \cc{FAD2BE}{\color[HTML]{000000} $0.949$} & \cc{FAD3BF}{\color[HTML]{000000} $0.949$} & \cc{FBDCCC}{\color[HTML]{000000} $0.948$} & \cc{F7BCA1}{\color[HTML]{000000} $0.951$\rlap{\textsuperscript{*}}} & \cc{FCE1D3}{\color[HTML]{000000} $0.105$} & \cc{F9CAB2}{\color[HTML]{000000} $0.101$} & \cc{F9C9B1}{\color[HTML]{000000} $0.100$} & \cc{F9C9B1}{\color[HTML]{000000} $0.100$} & \cc{F9C6AD}{\color[HTML]{000000} $0.100$} & \cc{EDA38D}{\color[HTML]{000000} $0.096$\rlap{\textsuperscript{*}}} \\
No D. & \cc{FFFFFF}{\color[HTML]{F1F1F1} \color{white} --} & \cc{DCECF4}{\color[HTML]{000000} $32.12$} & \cc{D0E5F0}{\color[HTML]{000000} $31.98$} & \cc{D0E5F0}{\color[HTML]{000000} $31.97$} & \cc{5C99C7}{\color[HTML]{F1F1F1} $30.96$} & \cc{EEF6FA}{\color[HTML]{000000} $32.34$} & \cc{FFFFFF}{\color[HTML]{F1F1F1} \color{white} --} & \cc{F9FCFD}{\color[HTML]{000000} $0.945$} & \cc{FFFFFE}{\color[HTML]{000000} $0.945$} & \cc{FFFFFE}{\color[HTML]{000000} $0.945$} & \cc{5C99C7}{\color[HTML]{F1F1F1} $0.936$} & \cc{FCE6DB}{\color[HTML]{000000} $0.947$} & \cc{FFFFFF}{\color[HTML]{F1F1F1} \color{white} --} & \cc{FCDECF}{\color[HTML]{000000} $0.104$} & \cc{FDFEFE}{\color[HTML]{000000} $0.112$} & \cc{FEFFFF}{\color[HTML]{000000} $0.112$} & \cc{5C99C7}{\color[HTML]{F1F1F1} $0.133$} & \cc{FDE9DF}{\color[HTML]{000000} $0.107$} \\
\bottomrule
\end{tabular}}
\endgroup
}
\end{subtable}
\begin{subtable}[t]{\linewidth}
\centering
\caption{Mip-NeRF 360}
\label{tab:practical_main_mipnerf360}
{\small
\begingroup \setlength{\tabcolsep}{2.5\tabcolsep}
\resizebox{\linewidth}{!}{\begin{tabular}{l|ccccc|ccccc|ccccc}
\toprule
& \multicolumn{5}{c|}{\textbf{PSNR ↑}} & \multicolumn{5}{c|}{\textbf{SSIM ↑}} & \multicolumn{5}{c}{\textbf{LPIPS ↓}} \\
& SfM & $\text{EDGS}^*$ & M. D. & DA3 & $\text{DA3}^\text{G.S.}$ & SfM & $\text{EDGS}^*$ & M. D. & DA3 & $\text{DA3}^\text{G.S.}$ & SfM & $\text{EDGS}^*$ & M. D. & DA3 & $\text{DA3}^\text{G.S.}$ \\
\midrule
AbsGS & \cc{F2AF97}{\color[HTML]{000000} $27.48$} & \cc{F7BCA1}{\color[HTML]{000000} $27.39$} & \cc{E79482}{\color[HTML]{F1F1F1} $27.68$} & \cc{F7FBFD}{\color[HTML]{000000} $26.31$} & \cc{F4B69C}{\color[HTML]{000000} $27.43$} & \cc{E69280}{\color[HTML]{F1F1F1} $0.824$} & \cc{E69280}{\color[HTML]{F1F1F1} $0.824$} & \cc{E08273}{\color[HTML]{F1F1F1} $0.828$} & \cc{FBD7C5}{\color[HTML]{000000} $0.802$} & \cc{EFA791}{\color[HTML]{000000} $0.819$} & \cc{E48C7B}{\color[HTML]{F1F1F1} $0.140$} & \cc{DA7368}{\color[HTML]{F1F1F1} $0.134$\rlap{\textsuperscript{*}}} & \cc{DD7C6E}{\color[HTML]{F1F1F1} $0.136$} & \cc{F9C5AB}{\color[HTML]{000000} $0.156$} & \cc{E08273}{\color[HTML]{F1F1F1} $0.138$} \\
INRIA & \cc{F2AF97}{\color[HTML]{000000} $27.47$} & \cc{F3B199}{\color[HTML]{000000} $27.46$} & \cc{E58E7D}{\color[HTML]{F1F1F1} $27.73$\rlap{\textsuperscript{*}}} & \cc{FEF7F3}{\color[HTML]{000000} $26.53$} & \cc{F8BEA3}{\color[HTML]{000000} $27.38$} & \cc{F4B69C}{\color[HTML]{000000} $0.816$} & \cc{E48C7B}{\color[HTML]{F1F1F1} $0.826$\rlap{\textsuperscript{*}}} & \cc{E28878}{\color[HTML]{F1F1F1} $0.826$\rlap{\textsuperscript{*}}} & \cc{FACFBA}{\color[HTML]{000000} $0.806$} & \cc{F9C1A6}{\color[HTML]{000000} $0.813$} & \cc{FDEBE2}{\color[HTML]{000000} $0.176$} & \cc{DF8072}{\color[HTML]{F1F1F1} $0.137$\rlap{\textsuperscript{*}}} & \cc{F0AB94}{\color[HTML]{000000} $0.148$\rlap{\textsuperscript{*}}} & \cc{FACCB5}{\color[HTML]{000000} $0.159$\rlap{\textsuperscript{*}}} & \cc{F2AF97}{\color[HTML]{000000} $0.149$\rlap{\textsuperscript{*}}} \\
MCMC & \cc{D5675E}{\color[HTML]{F1F1F1} $28.01$} & \cc{E48C7B}{\color[HTML]{F1F1F1} $27.73$} & \cc{E58E7D}{\color[HTML]{F1F1F1} $27.72$} & \cc{FDEEE6}{\color[HTML]{000000} $26.66$} & \cc{E89784}{\color[HTML]{000000} $27.66$} & \cc{D5675E}{\color[HTML]{F1F1F1} $0.834$} & \cc{E18475}{\color[HTML]{F1F1F1} $0.827$} & \cc{EB9D88}{\color[HTML]{000000} $0.822$} & \cc{FBD5C2}{\color[HTML]{000000} $0.803$} & \cc{EEA58F}{\color[HTML]{000000} $0.820$} & \cc{E79482}{\color[HTML]{F1F1F1} $0.142$} & \cc{D66960}{\color[HTML]{F1F1F1} $0.131$\rlap{\textsuperscript{*}}} & \cc{DA7368}{\color[HTML]{F1F1F1} $0.134$\rlap{\textsuperscript{*}}} & \cc{F7BCA1}{\color[HTML]{000000} $0.152$} & \cc{E18676}{\color[HTML]{F1F1F1} $0.139$} \\
IDHFR & \cc{DD7C6E}{\color[HTML]{F1F1F1} $27.86$} & \cc{E6907F}{\color[HTML]{F1F1F1} $27.70$} & \cc{DA7368}{\color[HTML]{F1F1F1} $27.92$} & \cc{FACDB7}{\color[HTML]{000000} $27.17$} & \cc{EFA791}{\color[HTML]{000000} $27.54$} & \cc{DF8072}{\color[HTML]{F1F1F1} $0.828$} & \cc{DF8072}{\color[HTML]{F1F1F1} $0.828$} & \cc{DB7569}{\color[HTML]{F1F1F1} $0.830$} & \cc{F0A992}{\color[HTML]{000000} $0.819$} & \cc{F0AB94}{\color[HTML]{000000} $0.818$} & \cc{DB7569}{\color[HTML]{F1F1F1} $0.134$} & \cc{D5675E}{\color[HTML]{F1F1F1} $0.131$\rlap{\textsuperscript{*}}} & \cc{D86F65}{\color[HTML]{F1F1F1} $0.133$} & \cc{E48C7B}{\color[HTML]{F1F1F1} $0.140$} & \cc{E08273}{\color[HTML]{F1F1F1} $0.138$} \\
RevDGS & \cc{F6BA9F}{\color[HTML]{000000} $27.41$} & \cc{F4B69C}{\color[HTML]{000000} $27.44$} & \cc{F2AF97}{\color[HTML]{000000} $27.47$} & \cc{DBEBF4}{\color[HTML]{000000} $25.95$} & \cc{F9C5AB}{\color[HTML]{000000} $27.29$} & \cc{F8BEA3}{\color[HTML]{000000} $0.814$} & \cc{E89784}{\color[HTML]{000000} $0.823$} & \cc{F0AB94}{\color[HTML]{000000} $0.818$} & \cc{FDECE3}{\color[HTML]{000000} $0.792$} & \cc{F9CAB2}{\color[HTML]{000000} $0.809$} & \cc{EB9F8A}{\color[HTML]{000000} $0.145$} & \cc{D5675E}{\color[HTML]{F1F1F1} $0.131$\rlap{\textsuperscript{*}}} & \cc{E28878}{\color[HTML]{F1F1F1} $0.139$} & \cc{FAD2BE}{\color[HTML]{000000} $0.163$} & \cc{E58E7D}{\color[HTML]{F1F1F1} $0.141$} \\
No D. & \cc{FFFFFF}{\color[HTML]{F1F1F1} \color{white} --} & \cc{FAD2BE}{\color[HTML]{000000} $27.08$} & \cc{FEF3ED}{\color[HTML]{000000} $26.59$} & \cc{5C99C7}{\color[HTML]{F1F1F1} $24.79$} & \cc{B4D6E8}{\color[HTML]{000000} $25.47$} & \cc{FFFFFF}{\color[HTML]{F1F1F1} \color{white} --} & \cc{F5B89E}{\color[HTML]{000000} $0.815$} & \cc{F8FBFD}{\color[HTML]{000000} $0.780$} & \cc{CCE3EF}{\color[HTML]{000000} $0.763$} & \cc{5C99C7}{\color[HTML]{F1F1F1} $0.733$} & \cc{FFFFFF}{\color[HTML]{F1F1F1} \color{white} --} & \cc{FACEB8}{\color[HTML]{000000} $0.161$} & \cc{93C0DC}{\color[HTML]{000000} $0.228$} & \cc{5C99C7}{\color[HTML]{F1F1F1} $0.242$} & \cc{70A7CE}{\color[HTML]{F1F1F1} $0.237$} \\
\bottomrule
\end{tabular}}
\endgroup
}
\end{subtable}
\begin{subtable}[t]{\linewidth}
\centering
\caption{Tanks and Temples}
\label{tab:practical_main_tanksandtemples}
{\small
\begingroup \setlength{\tabcolsep}{2.5\tabcolsep}
\resizebox{\linewidth}{!}{\begin{tabular}{l|ccccc|ccccc|ccccc}
\toprule
& \multicolumn{5}{c|}{\textbf{PSNR ↑}} & \multicolumn{5}{c|}{\textbf{SSIM ↑}} & \multicolumn{5}{c}{\textbf{LPIPS ↓}} \\
& SfM & $\text{EDGS}^*$ & M. D. & DA3 & $\text{DA3}^\text{G.S.}$ & SfM & $\text{EDGS}^*$ & M. D. & DA3 & $\text{DA3}^\text{G.S.}$ & SfM & $\text{EDGS}^*$ & M. D. & DA3 & $\text{DA3}^\text{G.S.}$ \\
\midrule
AbsGS & \cc{F5F9FC}{\color[HTML]{000000} $22.85$} & \cc{FEF6F1}{\color[HTML]{000000} $23.09$} & \cc{FBDDCE}{\color[HTML]{000000} $23.43$\rlap{\textsuperscript{*}}} & \cc{FBFDFE}{\color[HTML]{000000} $22.91$} & \cc{FACCB5}{\color[HTML]{000000} $23.65$\rlap{\textsuperscript{*}}} & \cc{FCE2D5}{\color[HTML]{000000} $0.825$} & \cc{F9C9B1}{\color[HTML]{000000} $0.833$} & \cc{F9C3A8}{\color[HTML]{000000} $0.835$\rlap{\textsuperscript{*}}} & \cc{FCE4D8}{\color[HTML]{000000} $0.824$} & \cc{F9C5AB}{\color[HTML]{000000} $0.834$} & \cc{FBD9C8}{\color[HTML]{000000} $0.171$} & \cc{E58E7D}{\color[HTML]{F1F1F1} $0.145$\rlap{\textsuperscript{*}}} & \cc{F4B69C}{\color[HTML]{000000} $0.155$\rlap{\textsuperscript{*}}} & \cc{F9C8B0}{\color[HTML]{000000} $0.162$} & \cc{E79482}{\color[HTML]{F1F1F1} $0.147$\rlap{\textsuperscript{*}}} \\
INRIA & \cc{FAD1BC}{\color[HTML]{000000} $23.58$} & \cc{FCE1D3}{\color[HTML]{000000} $23.37$} & \cc{F9C3A8}{\color[HTML]{000000} $23.77$\rlap{\textsuperscript{*}}} & \cc{FDE8DD}{\color[HTML]{000000} $23.28$} & \cc{F9C5AB}{\color[HTML]{000000} $23.75$} & \cc{F9C4AA}{\color[HTML]{000000} $0.835$} & \cc{F8BEA3}{\color[HTML]{000000} $0.836$} & \cc{F1AD96}{\color[HTML]{000000} $0.839$\rlap{\textsuperscript{*}}} & \cc{FAD3BF}{\color[HTML]{000000} $0.829$} & \cc{F9C7AE}{\color[HTML]{000000} $0.833$} & \cc{F9C6AD}{\color[HTML]{000000} $0.161$} & \cc{DF8072}{\color[HTML]{F1F1F1} $0.142$\rlap{\textsuperscript{*}}} & \cc{F1AD96}{\color[HTML]{000000} $0.153$\rlap{\textsuperscript{*}}} & \cc{F6BA9F}{\color[HTML]{000000} $0.156$\rlap{\textsuperscript{*}}} & \cc{EB9F8A}{\color[HTML]{000000} $0.149$\rlap{\textsuperscript{*}}} \\
MCMC & \cc{D5675E}{\color[HTML]{F1F1F1} $24.39$} & \cc{F0A992}{\color[HTML]{000000} $23.96$} & \cc{E69280}{\color[HTML]{F1F1F1} $24.11$} & \cc{FACBB4}{\color[HTML]{000000} $23.66$} & \cc{DA7368}{\color[HTML]{F1F1F1} $24.32$} & \cc{D5675E}{\color[HTML]{F1F1F1} $0.850$} & \cc{DF8072}{\color[HTML]{F1F1F1} $0.846$} & \cc{E18475}{\color[HTML]{F1F1F1} $0.845$} & \cc{F9C2A7}{\color[HTML]{000000} $0.835$} & \cc{DC776B}{\color[HTML]{F1F1F1} $0.847$} & \cc{DC796D}{\color[HTML]{F1F1F1} $0.140$} & \cc{D5675E}{\color[HTML]{F1F1F1} $0.135$\rlap{\textsuperscript{*}}} & \cc{D76B61}{\color[HTML]{F1F1F1} $0.136$\rlap{\textsuperscript{*}}} & \cc{EB9D88}{\color[HTML]{000000} $0.149$} & \cc{DC796D}{\color[HTML]{F1F1F1} $0.140$} \\
IDHFR & \cc{F9C3A8}{\color[HTML]{000000} $23.77$} & \cc{FACBB4}{\color[HTML]{000000} $23.66$} & \cc{F2AF97}{\color[HTML]{000000} $23.92$} & \cc{FAD2BE}{\color[HTML]{000000} $23.57$} & \cc{F5B89E}{\color[HTML]{000000} $23.86$} & \cc{EA9B87}{\color[HTML]{000000} $0.842$} & \cc{EB9F8A}{\color[HTML]{000000} $0.841$} & \cc{E48C7B}{\color[HTML]{F1F1F1} $0.844$\rlap{\textsuperscript{*}}} & \cc{F6BA9F}{\color[HTML]{000000} $0.837$} & \cc{F0A992}{\color[HTML]{000000} $0.839$} & \cc{DE7E70}{\color[HTML]{F1F1F1} $0.141$} & \cc{D86F65}{\color[HTML]{F1F1F1} $0.137$\rlap{\textsuperscript{*}}} & \cc{DC776B}{\color[HTML]{F1F1F1} $0.140$} & \cc{E38A7A}{\color[HTML]{F1F1F1} $0.144$} & \cc{DA7368}{\color[HTML]{F1F1F1} $0.138$\rlap{\textsuperscript{*}}} \\
RevDGS & \cc{FEF9F6}{\color[HTML]{000000} $23.05$} & \cc{FFFDFB}{\color[HTML]{000000} $22.99$} & \cc{FEF8F4}{\color[HTML]{000000} $23.06$} & \cc{EAF3F8}{\color[HTML]{000000} $22.73$} & \cc{FDE8DD}{\color[HTML]{000000} $23.27$} & \cc{FDECE3}{\color[HTML]{000000} $0.821$} & \cc{FCDECF}{\color[HTML]{000000} $0.826$} & \cc{FDEDE5}{\color[HTML]{000000} $0.821$} & \cc{F7FBFD}{\color[HTML]{000000} $0.813$} & \cc{FDE8DD}{\color[HTML]{000000} $0.822$} & \cc{FDEBE2}{\color[HTML]{000000} $0.180$} & \cc{FACDB7}{\color[HTML]{000000} $0.165$\rlap{\textsuperscript{*}}} & \cc{FEF1EA}{\color[HTML]{000000} $0.183$} & \cc{FEF6F1}{\color[HTML]{000000} $0.186$} & \cc{FBD6C4}{\color[HTML]{000000} $0.169$\rlap{\textsuperscript{*}}} \\
No D. & \cc{FFFFFF}{\color[HTML]{F1F1F1} \color{white} --} & \cc{FFFFFE}{\color[HTML]{000000} $22.96$} & \cc{D0E5F0}{\color[HTML]{000000} $22.44$} & \cc{5C99C7}{\color[HTML]{F1F1F1} $21.52$} & \cc{7DB1D4}{\color[HTML]{F1F1F1} $21.75$} & \cc{FFFFFF}{\color[HTML]{F1F1F1} \color{white} --} & \cc{FDE8DD}{\color[HTML]{000000} $0.822$} & \cc{C4DFED}{\color[HTML]{000000} $0.799$} & \cc{7FB2D4}{\color[HTML]{000000} $0.785$} & \cc{5C99C7}{\color[HTML]{F1F1F1} $0.779$} & \cc{FFFFFF}{\color[HTML]{F1F1F1} \color{white} --} & \cc{FBD8C7}{\color[HTML]{000000} $0.170$} & \cc{83B5D6}{\color[HTML]{000000} $0.236$} & \cc{5C99C7}{\color[HTML]{F1F1F1} $0.246$} & \cc{6AA3CC}{\color[HTML]{F1F1F1} $0.243$} \\
\bottomrule
\end{tabular}}
\endgroup
}
\end{subtable}
\end{table}

Now that we've established a high-accuracy initialization baseline, we switch to evaluating NVS performance with the other initialization methods, following the protocol described in~\cref{sec:protocol:practical_init}. For ScanNet++ we include the laser-scan baseline using the same initialization size as the other methods.
Results are presented in~\cref{tab:practical_main}. We include additional results (reduced initialization size, hybrid initialization, ETH3D results) in the supplementary material.

On the two ScanNet++ variants, the results largely follow the trends we've seen in~\cref{subsec:eval:laser_init} -- improvements are most pronounced on off-trajectory views, while on on-trajectory views, less accurate dense initialization often decreases metrics.
The notable exception is $\text{DA3}^\text{GS}$ which actually surpasses laser scan performance on off-trajectory ScanNet++, and practically matches it on on-trajectory ScanNet++, where the other practical init methods usually do not provide any meaningful improvement. We investigate $\text{DA3}^\text{GS}$ performance below.

Looking at results on MipNerf360 and Tanks\&Temples, we find that, using SfM, MCMC, and IDHFR provide the best overall performance on these datasets.
While we see some improvements from dense initialization for strategies with lower SfM baselines, with these two, none of the initialization methods provide large benefits.
Interestingly, here DA3 (no GS) performs much worse than Monodepth which is not what we observed on ScanNet++.\footnote{On MipNerf360, the delta is dominated by DA3's failure on ``Bonsai'', but there are no extreme outliers on Tanks\&Temples (see \cref{per_scene_data_analysis} of the supplementary material).}
Even here, $\text{DA3}^\text{GS}$ performs much better than DA3, despite using fewer images (see~\cref{sec:protocol:practical_init}).

For this experiment, we see fewer statistically significant improvements, as the family sizes are larger (more initializations), further limiting the power of the tests. Those we do see are largely limited to \Cref{tab:practical_main_scannet++}, supporting our observation that using dense initialization mostly benefits generalization.

\begin{table}[tb]
\centering
\caption{Ablation on $\text{DA3}^\text{GS}$ initialization components. Isotropic scale initialization uses the mean of the three anisotropic scale components. Other ablations follow~\cite{3dgs}.}
\label{tab:da3_gs_components_ablation_compact}
\begin{subtable}[t]{0.9\linewidth}
\centering
\caption{Mip-NeRF 360}
\label{tab:da3_gs_components_ablation_mipnerf360}
{\small
\begingroup \setlength{\tabcolsep}{2.5\tabcolsep}
\resizebox{\linewidth}{!}{\begin{tabular}{l|ccc|ccc|ccc}
\toprule
& \multicolumn{3}{c|}{\textbf{PSNR ↑}} & \multicolumn{3}{c|}{\textbf{SSIM ↑}} & \multicolumn{3}{c}{\textbf{LPIPS ↓}} \\
& AbsGS & MCMC & IDHFR & AbsGS & MCMC & IDHFR & AbsGS & MCMC & IDHFR \\
\midrule
Base & \cc{FFFFFE}{\color[HTML]{000000} $27.45$} & \cc{FFFFFE}{\color[HTML]{000000} $27.65$} & \cc{FFFFFE}{\color[HTML]{000000} $27.55$} & \cc{FFFFFE}{\color[HTML]{000000} $0.819$} & \cc{FFFFFE}{\color[HTML]{000000} $0.819$} & \cc{FFFFFE}{\color[HTML]{000000} $0.818$} & \cc{FFFFFE}{\color[HTML]{000000} $0.138$} & \cc{FFFFFE}{\color[HTML]{000000} $0.139$} & \cc{FFFFFE}{\color[HTML]{000000} $0.138$} \\
\midrule
As point cloud & \cc{F9C5AB}{\color[HTML]{000000} $+0.16$} & \cc{3880BA}{\color[HTML]{F1F1F1} $-0.34$} & \cc{D5675E}{\color[HTML]{F1F1F1} $+0.29$} & \cc{F0AB94}{\color[HTML]{000000} $+0.008$} & \cc{327CB7}{\color[HTML]{F1F1F1} $-0.014$} & \cc{D36159}{\color[HTML]{F1F1F1} $+0.012$} & \cc{FFFFFE}{\color[HTML]{000000} $+0.000$} & \cc{327CB7}{\color[HTML]{F1F1F1} $+0.007$} & \cc{F0AB94}{\color[HTML]{000000} $-0.004$} \\
kNN scale & \cc{F4B49A}{\color[HTML]{000000} $+0.19$} & \cc{6CA4CD}{\color[HTML]{F1F1F1} $-0.27$} & \cc{C43B3C}{\color[HTML]{F1F1F1} $+0.35$} & \cc{E99985}{\color[HTML]{000000} $+0.009$} & \cc{7AAED2}{\color[HTML]{F1F1F1} $-0.010$} & \cc{CB4E4A}{\color[HTML]{F1F1F1} $+0.013$} & \cc{FDEDE5}{\color[HTML]{000000} $-0.001$} & \cc{7AAED2}{\color[HTML]{F1F1F1} $+0.005$} & \cc{E18676}{\color[HTML]{F1F1F1} $-0.005$} \\
Isotropic scale & \cc{EDF5FA}{\color[HTML]{000000} $-0.04$} & \cc{FBFDFE}{\color[HTML]{000000} $-0.01$} & \cc{F6FAFC}{\color[HTML]{000000} $-0.02$} & \cc{FEF6F1}{\color[HTML]{000000} $+0.001$} & \cc{FFFFFE}{\color[HTML]{000000} $+0.000$} & \cc{FCE4D8}{\color[HTML]{000000} $+0.003$} & \cc{FFFFFE}{\color[HTML]{000000} $+0.000$} & \cc{FFFFFE}{\color[HTML]{000000} $+0.000$} & \cc{FFFFFE}{\color[HTML]{000000} $+0.000$} \\
Uniform opacity & \cc{F2F8FB}{\color[HTML]{000000} $-0.03$} & \cc{FFFCFA}{\color[HTML]{000000} $+0.01$} & \cc{F6FAFC}{\color[HTML]{000000} $-0.02$} & \cc{FFFFFE}{\color[HTML]{000000} $+0.000$} & \cc{FFFFFE}{\color[HTML]{000000} $+0.000$} & \cc{FFFFFE}{\color[HTML]{000000} $+0.000$} & \cc{FFFFFE}{\color[HTML]{000000} $+0.000$} & \cc{FFFFFE}{\color[HTML]{000000} $+0.000$} & \cc{FFFFFE}{\color[HTML]{000000} $+0.000$} \\
\bottomrule
\end{tabular}}
\endgroup
}
\end{subtable}
\begin{subtable}[t]{0.9\linewidth}
\centering
\caption{Tanks and Temples}
\label{tab:da3_gs_components_ablation_tanksandtemples}
{\small
\begingroup \setlength{\tabcolsep}{2.5\tabcolsep}
\resizebox{\linewidth}{!}{\begin{tabular}{l|ccc|ccc|ccc}
\toprule
& \multicolumn{3}{c|}{\textbf{PSNR ↑}} & \multicolumn{3}{c|}{\textbf{SSIM ↑}} & \multicolumn{3}{c}{\textbf{LPIPS ↓}} \\
& AbsGS & MCMC & IDHFR & AbsGS & MCMC & IDHFR & AbsGS & MCMC & IDHFR \\
\midrule
Base & \cc{FFFFFE}{\color[HTML]{000000} $23.63$} & \cc{FFFFFE}{\color[HTML]{000000} $24.33$} & \cc{FFFFFE}{\color[HTML]{000000} $23.85$} & \cc{FFFFFE}{\color[HTML]{000000} $0.834$} & \cc{FFFFFE}{\color[HTML]{000000} $0.847$} & \cc{FFFFFE}{\color[HTML]{000000} $0.839$} & \cc{FFFFFE}{\color[HTML]{000000} $0.148$} & \cc{FFFFFE}{\color[HTML]{000000} $0.140$} & \cc{FFFFFE}{\color[HTML]{000000} $0.138$} \\
\midrule
As point cloud & \cc{B4D6E8}{\color[HTML]{000000} $-0.14$} & \cc{327CB7}{\color[HTML]{F1F1F1} $-0.29$} & \cc{FDE9DF}{\color[HTML]{000000} $+0.05$} & \cc{FDEDE5}{\color[HTML]{000000} $+0.001$} & \cc{327CB7}{\color[HTML]{F1F1F1} $-0.007$} & \cc{F0AB94}{\color[HTML]{000000} $+0.004$} & \cc{B1D5E7}{\color[HTML]{000000} $+0.005$} & \cc{E0EEF5}{\color[HTML]{000000} $+0.002$} & \cc{E0EEF5}{\color[HTML]{000000} $+0.002$} \\
kNN scale & \cc{EEF6FA}{\color[HTML]{000000} $-0.03$} & \cc{66A0CB}{\color[HTML]{F1F1F1} $-0.23$} & \cc{FEF2EC}{\color[HTML]{000000} $+0.03$} & \cc{FBDBCB}{\color[HTML]{000000} $+0.002$} & \cc{7AAED2}{\color[HTML]{F1F1F1} $-0.005$} & \cc{F0AB94}{\color[HTML]{000000} $+0.004$} & \cc{D0E5F0}{\color[HTML]{000000} $+0.003$} & \cc{F0F7FA}{\color[HTML]{000000} $+0.001$} & \cc{F0F7FA}{\color[HTML]{000000} $+0.001$} \\
Isotropic scale & \cc{CFE5F0}{\color[HTML]{000000} $-0.09$} & \cc{EAF3F8}{\color[HTML]{000000} $-0.04$} & \cc{F5F9FC}{\color[HTML]{000000} $-0.02$} & \cc{FFFFFE}{\color[HTML]{000000} $+0.000$} & \cc{FFFFFE}{\color[HTML]{000000} $+0.000$} & \cc{FBDBCB}{\color[HTML]{000000} $+0.002$} & \cc{F0F7FA}{\color[HTML]{000000} $+0.001$} & \cc{FFFFFE}{\color[HTML]{000000} $+0.000$} & \cc{E0EEF5}{\color[HTML]{000000} $+0.002$} \\
Uniform opacity & \cc{EEF6FA}{\color[HTML]{000000} $-0.03$} & \cc{EEF6FA}{\color[HTML]{000000} $-0.03$} & \cc{FFFBF9}{\color[HTML]{000000} $+0.01$} & \cc{E8F3F8}{\color[HTML]{000000} $-0.001$} & \cc{FFFFFE}{\color[HTML]{000000} $+0.000$} & \cc{FFFFFE}{\color[HTML]{000000} $+0.000$} & \cc{FEF3ED}{\color[HTML]{000000} $-0.001$} & \cc{F0F7FA}{\color[HTML]{000000} $+0.001$} & \cc{FFFFFE}{\color[HTML]{000000} $+0.000$} \\
\bottomrule
\end{tabular}}
\endgroup
}
\end{subtable}
\end{table}

\begin{figure}[tb]
  \centering
  \centering
  
  \includegraphics[width=0.2485\linewidth]{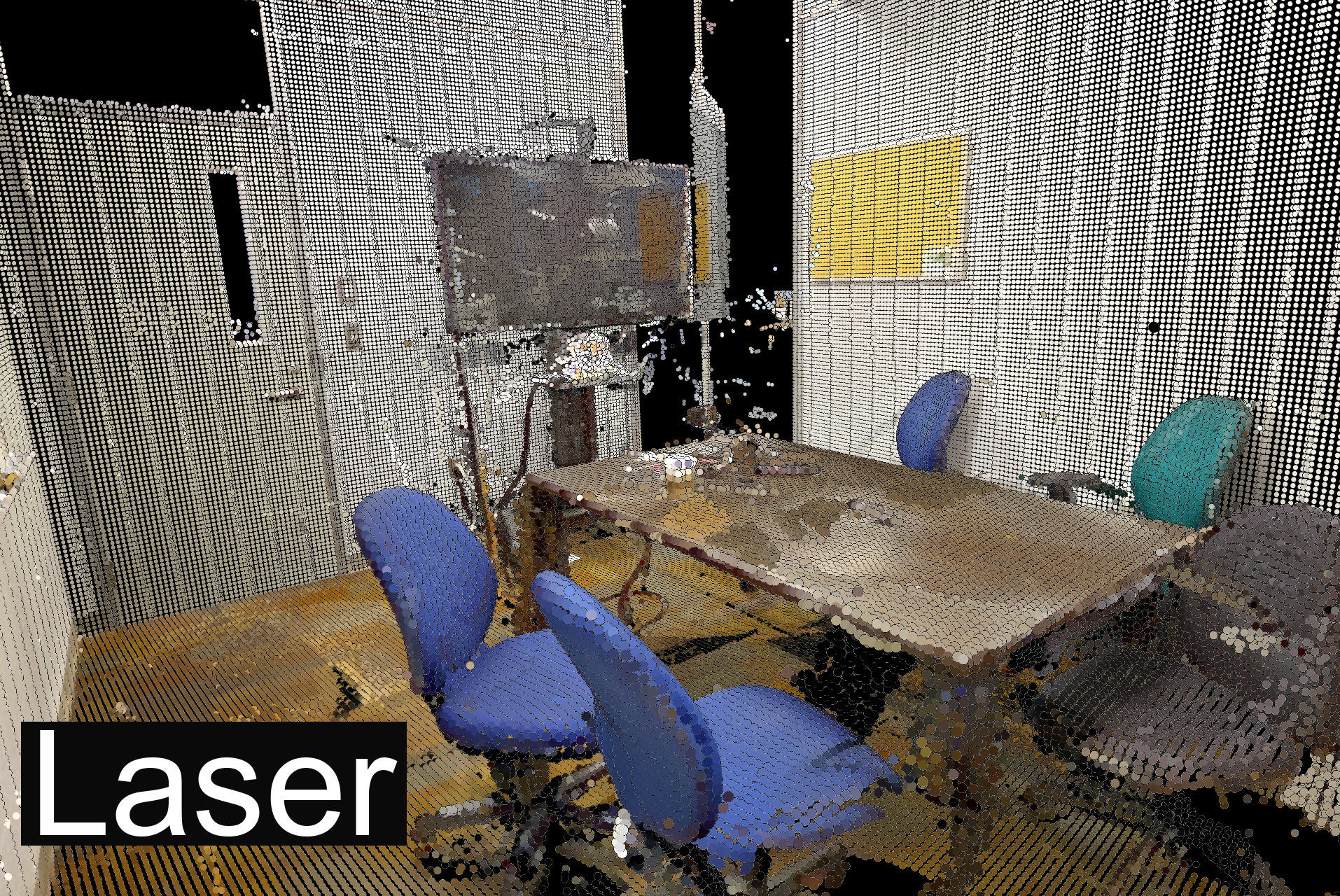}\hfill
  \includegraphics[width=0.2485\linewidth]{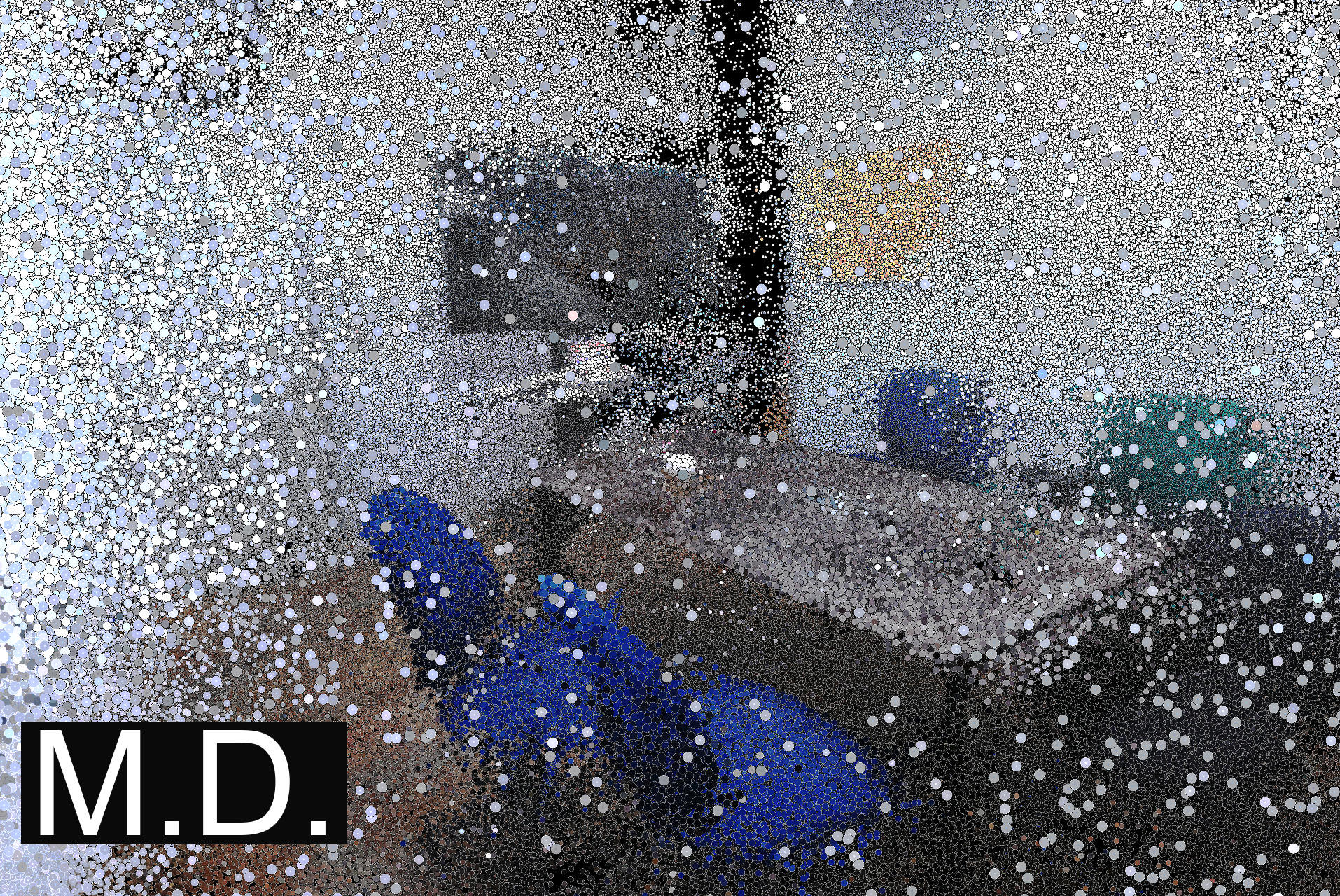}\hfill
  \includegraphics[width=0.2485\linewidth]{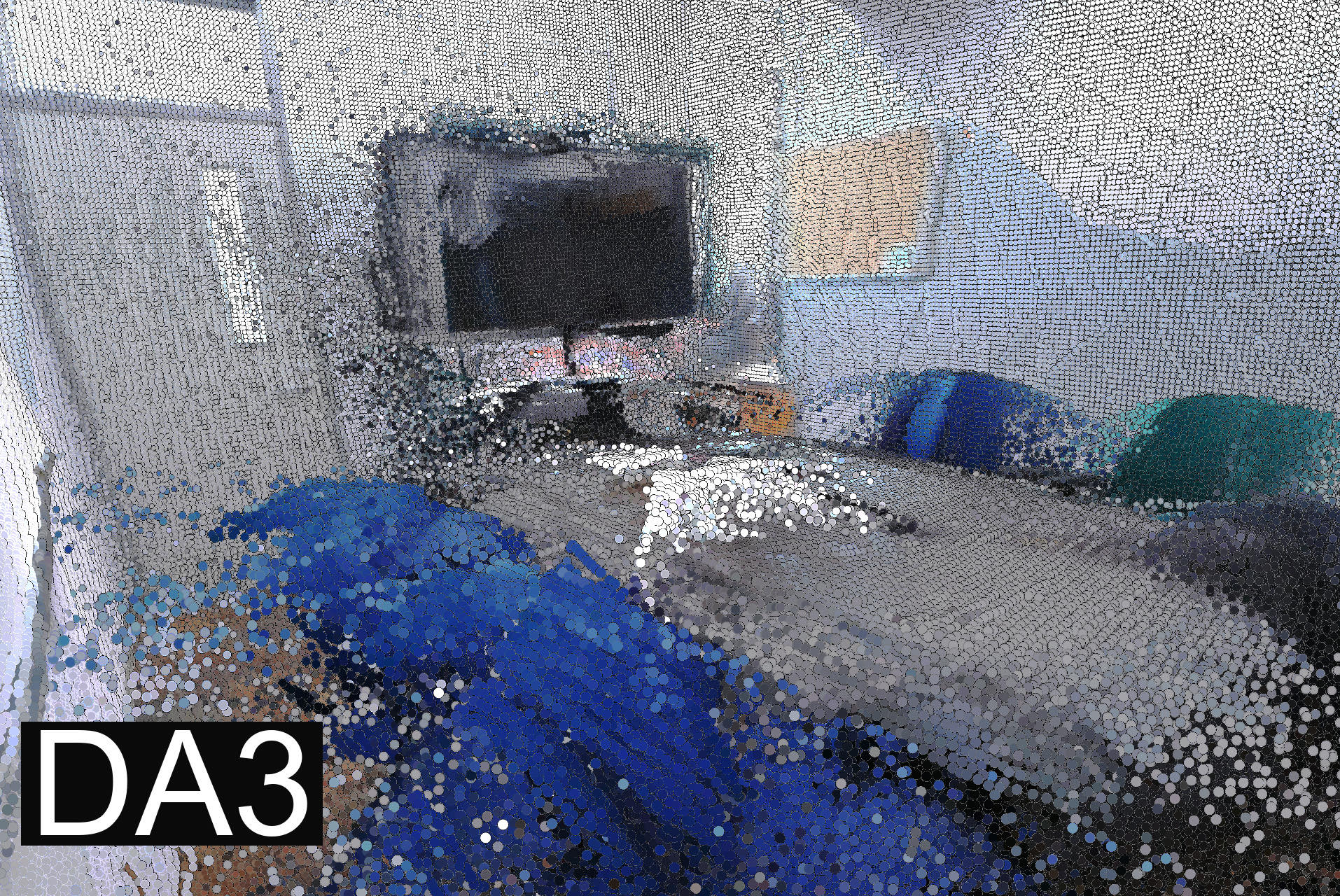}\hfill
  \includegraphics[width=0.2485\linewidth]{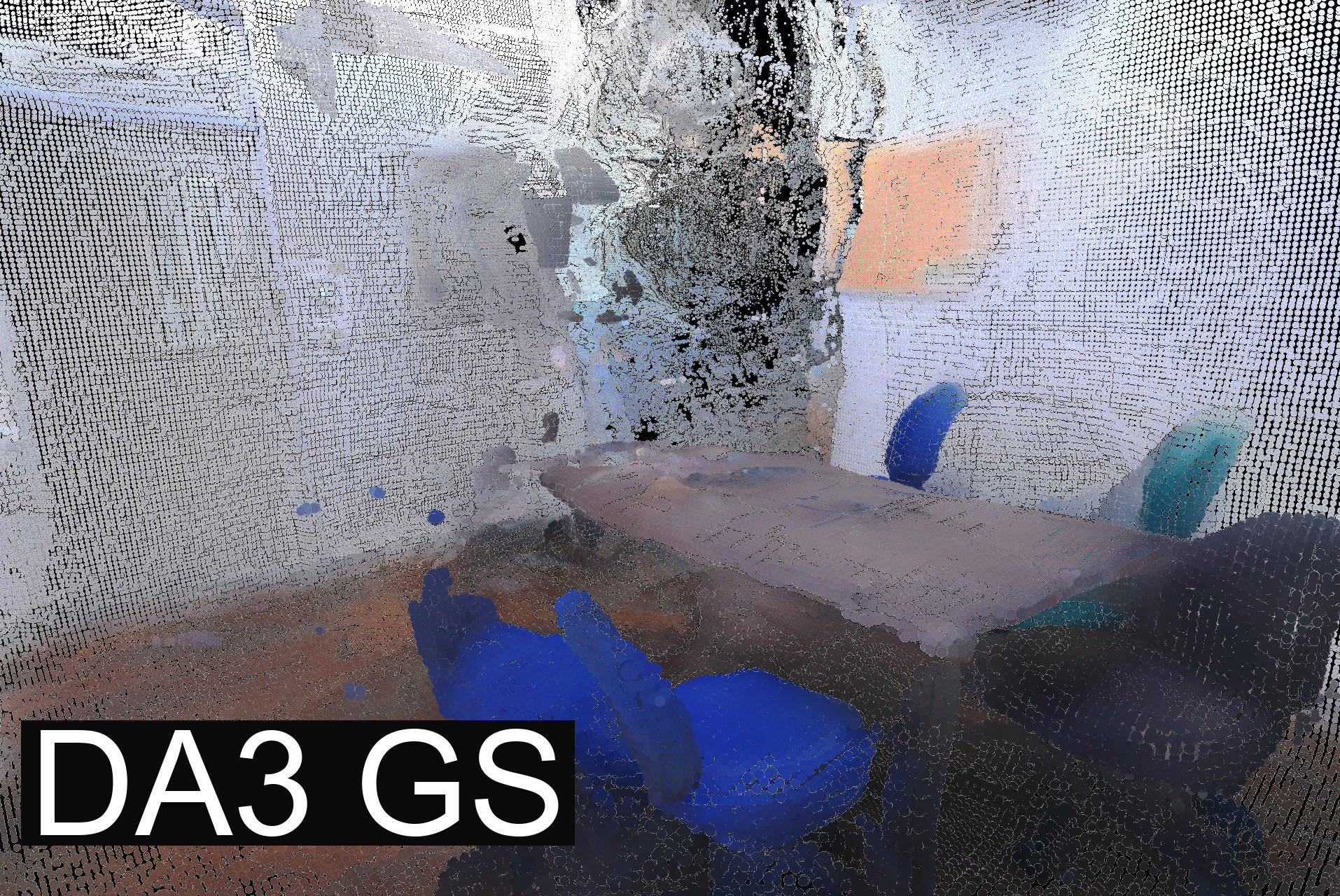}\hfill

  \caption{Outputs for some of the evaluated init methods on the ``5748ce6f01'' ScanNet++ scene (TSDF fusion was used for $\text{DA3}^\text{GS}$). Note how the glass surface in the back is missing from the laser scan, and how $\text{DA3}^\text{GS}$ places Gaussians \textit{behind} that surface.}
  \label{fig:init_outputs}
\end{figure}

\PAR{$\text{DA3}^\text{GS}$ performance analysis.} The effects of using the DA3 GS head seen in~\cref{tab:practical_main} might suggest that initializing all parameters to plausible values provides significant benefits.
To investigate this, we perform ablations to determine which $\text{DA3}^\text{GS}$-initialized parameters matter most (\cref{tab:da3_gs_components_ablation_compact}). 
However, even if we only use positions and colors, deriving all other parameters following~\cite{3dgs} (``As point cloud'' row), the performance delta is much smaller than in~\cref{tab:practical_main}.
Furthermore, in some cases, this actually improves results.
Instead, we find that the DA3 GS head actively uses its ability to offset depths predicted by the main model to provide point positions more suited for 3DGS. In general, the GS head seems to prioritize scene coverage at the cost of precision. In a limited number of cases, it \textit{significantly} improves upon the base model's predictions (see supp.\ mat., \cref{init_precision_no_gt,per_scene_data_analysis}). Additionally, it often puts points \textit{behind} transparent surfaces (as can be seen in~\cref{fig:init_outputs}), which is beneficial for 3DGS NVS.
Besides point positions, the initial scales seem to have the largest effect on NVS performance, though results are inconsistent across densification strategies and datasets. Initial values of the other parameters do not seem nearly as important. Results for ScanNet++ and further ablations are provided in \Cref{da3_gs_ablation} of the supp.\ mat.

\begin{figure}[tb]
  \centering
  \centering
  
  \includegraphics[width=0.2485\linewidth]{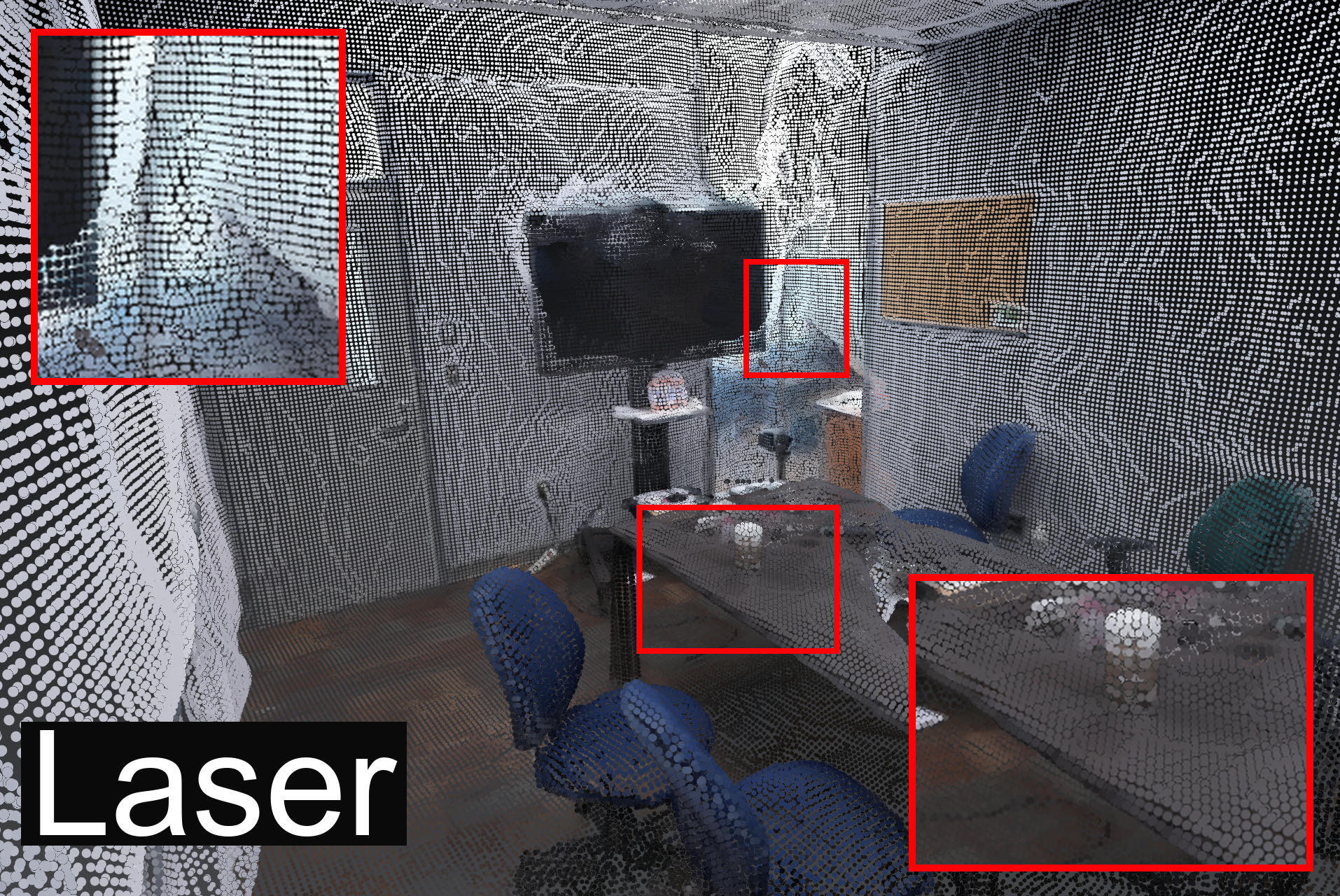}\hfill
  \includegraphics[width=0.2485\linewidth]{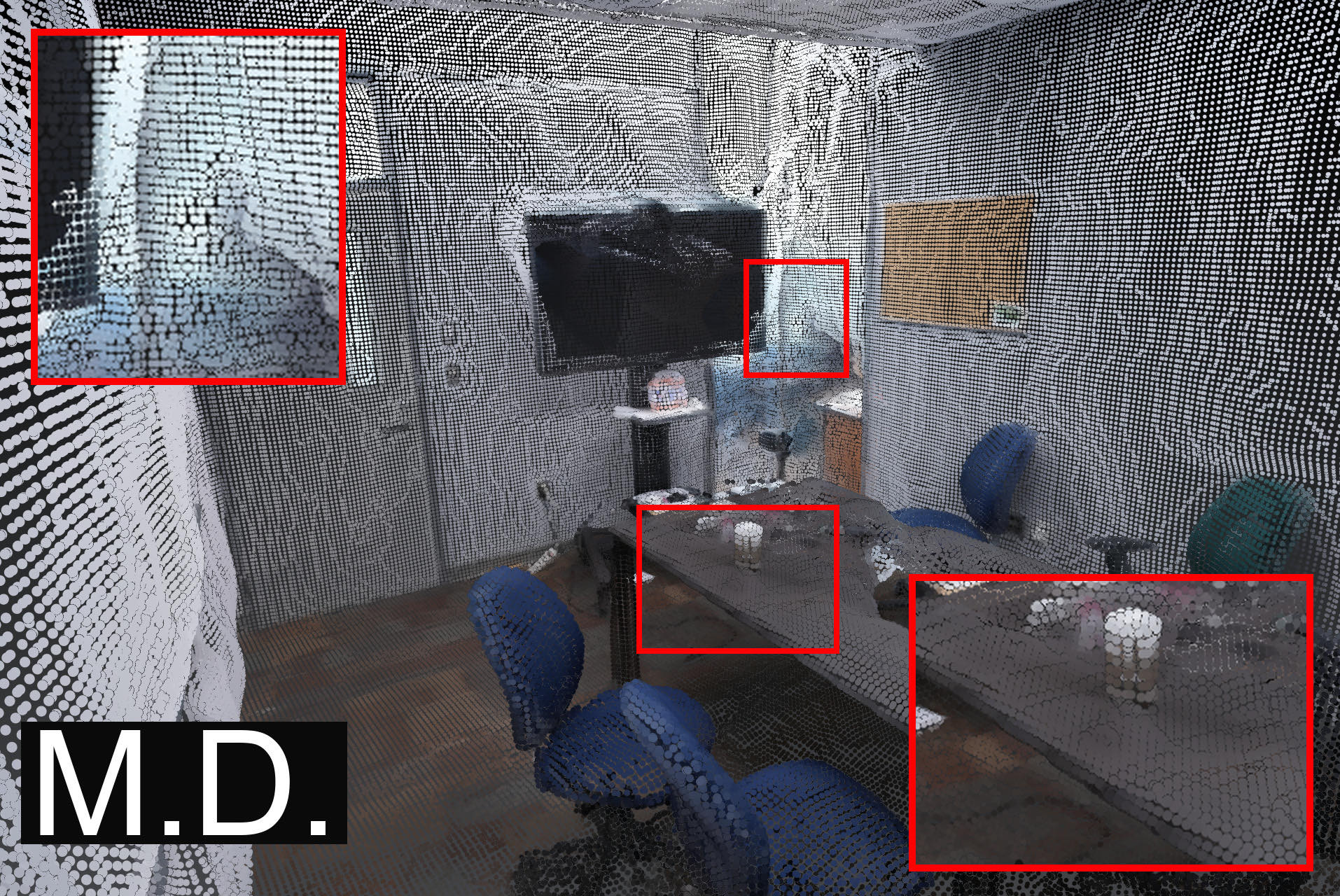}\hfill
  \includegraphics[width=0.2485\linewidth]{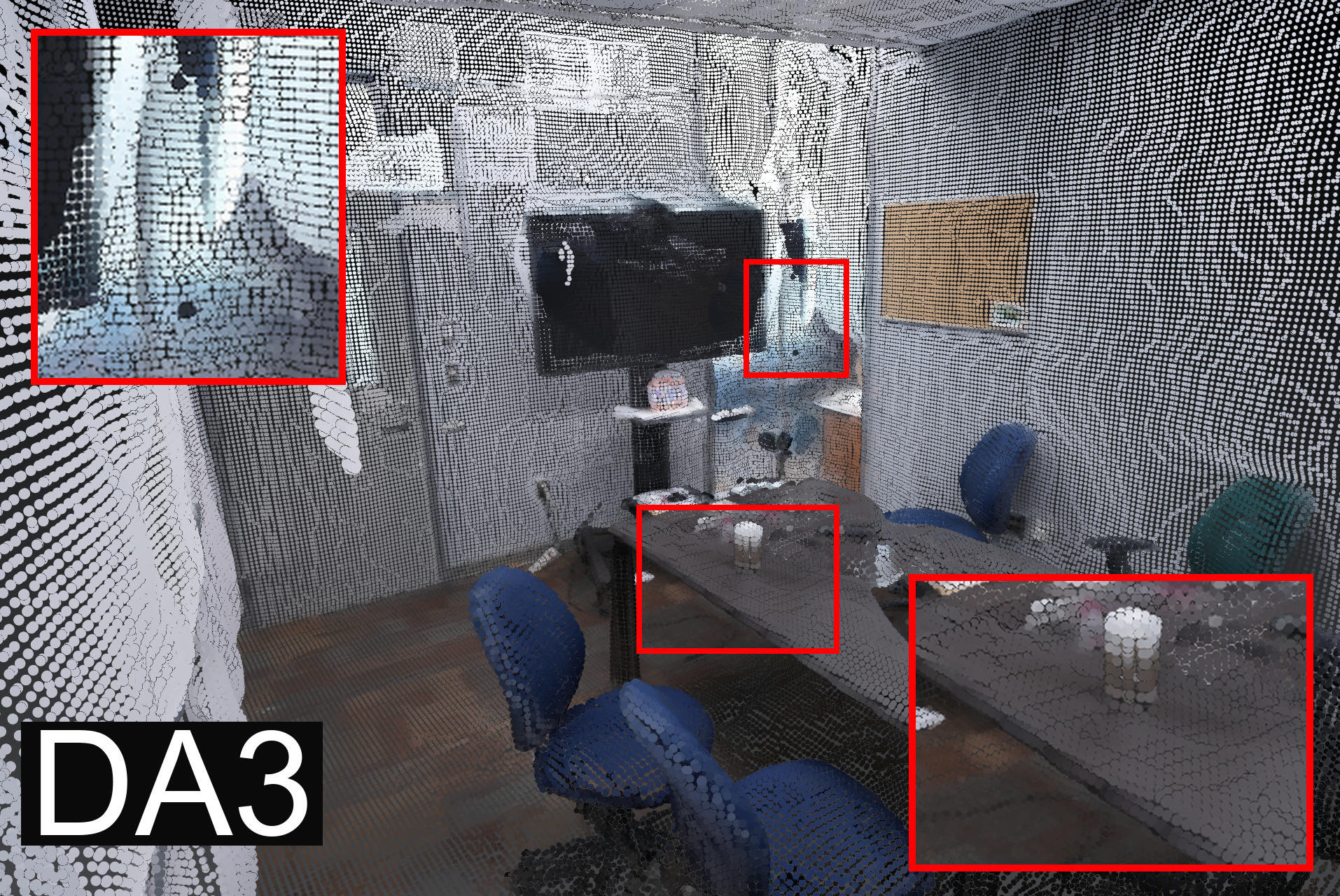}\hfill
  \includegraphics[width=0.2485\linewidth]{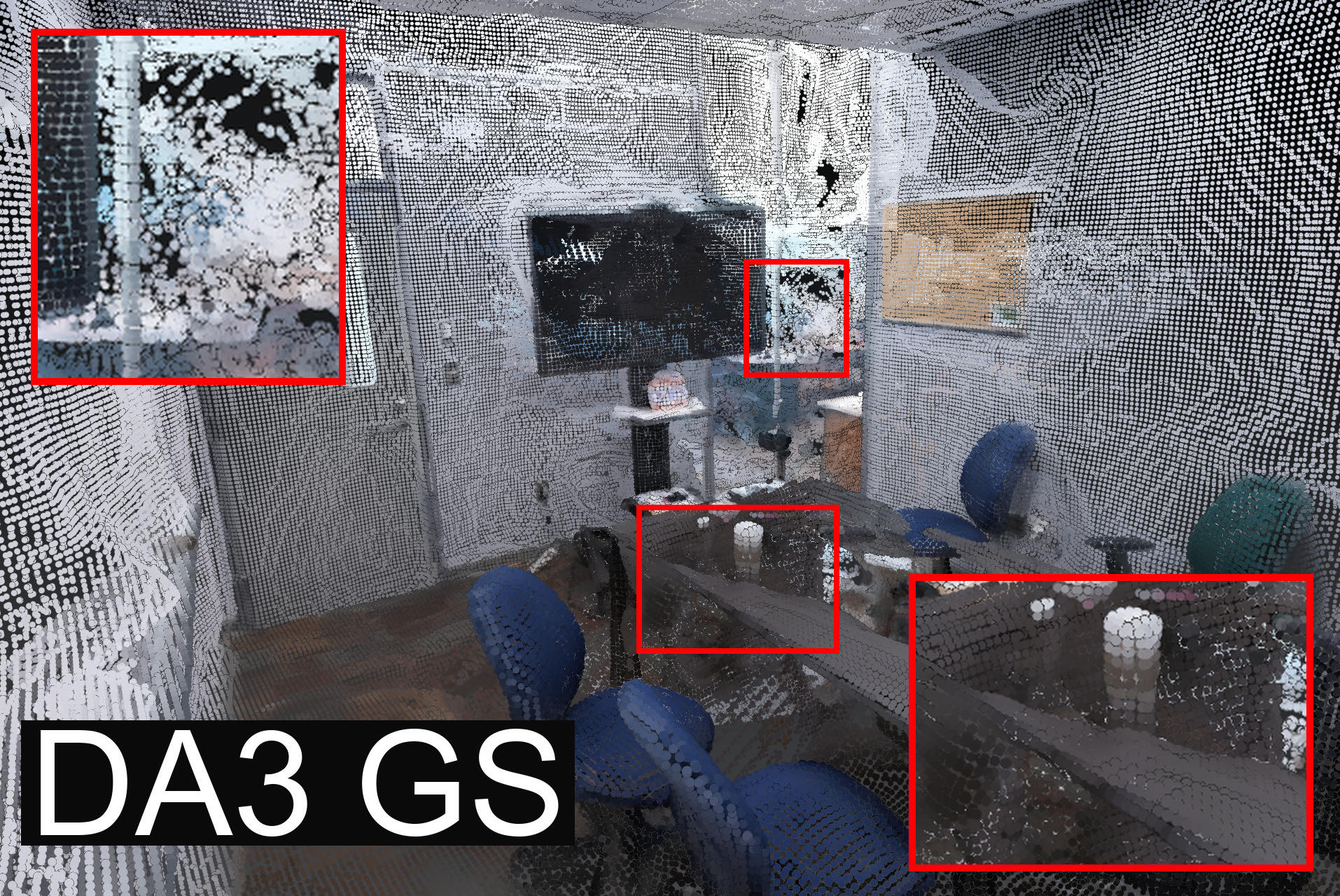}\hfill

  \caption{Point clouds extracted from scenes trained with IDHFR using TSDF fusion on the ``5748ce6f01'' ScanNet++ scene.
  }
  \label{fig:final_tsdfs}
\end{figure}

\subsection{Geometric Accuracy}
In~\Cref{subsec:eval:laser_init,subsec:eval:practical_init}, we have observed that dense initialization largely benefits NVS generalization, but does not lead to significant or consistent improvements on well-constrained scenes. 
However, until now, we have focused on the visual quality of the outputs, ignoring the underlying geometric structure.
In this experiment, following the protocol defined in~\cref{sec:protocol:geometric}, we investigate the geometric accuracy of both the initial point clouds (or Gaussians) and of the trained scene representations. We use ScanNet++, where we have precise laser scan data, and the scenes have good input view coverage (we use the default split as it has more training images).
We would like to note that the reported precisions (and thus $\text{F}_1$ scores) are pessimistic due to the lack of GT data for reflective surfaces. Nevertheless, this bias affects all configurations equally, and relative performance comparisons remain valid. See \cref{excluded_eth3d_results} of the supp.\ mat. for results on ETH3D.

Looking at the results in~\Cref{tab:final_recon_fscore}, Monodepth achieves the highest $\text{F}_1$ score prior to training, as it explicitly aligns predicted depths to the SfM point clouds.
Despite that, its outputs are noisy, and would require significant post-processing to be useful for tasks other than 3DGS initialization.
SfM has the highest precision but very low recall, as expected, given the nature of the method.
Both DA3 and $\text{EDGS}^*$ exhibit high recall but lower precision, whereas $\text{DA3}^\text{GS}$ drops both further. Please refer to~\Cref{fig:init_outputs} for examples of produced initializations.

In terms of post-training geometric accuracy, we see that dense initialization leads to major improvements in both precision and recall, whereas SfM-initialized training leads to geometrically inaccurate representations.
Interestingly, when using SfM, there seems to be an inverse relationship between NVS quality and $F_1$ scores -- densification methods that are worse at NVS tend to better adhere to the scene's true geometric structure, likely because they do not diverge from the initialization as much.
However, we do not observe the same behavior when using dense initialization, suggesting that accurate initialization priors enable the optimization to achieve similar NVS results as with SfM, while maintaining higher geometric consistency (see~\cref{fig:final_tsdfs}). For the laser scan results, introducing hybrid initialization decreases precision but maintains similar recall, as the added SfM points promote reconstructing regions \textit{behind} reflective and transparent scene surfaces, which are absent from the laser scans we use as GT. Across densifications, dense initialization generally yields statistically significant post-training \(F_1\) improvements, with FWER controlled at 5\% within each densification group.

\begin{table}[tb]
\captionsetup{skip=4pt}
\centering
\caption{Geometric accuracy of the initializations and corresponding trained 3DGS scenes, using ScanNet++ laser scans as GT. Metrics in cells (inlier threshold $d$=0.05\,m): $\text{F}_1$ Score / Precision / Recall. Cell colors are determined by $\text{F}_1$ scores; metrics with statistically significant improvements over SfM (for post-training rows) are in bold.}  
\label{tab:final_recon_fscore}
{\small
\begingroup \setlength{\tabcolsep}{3.2\tabcolsep}
\resizebox{\columnwidth}{!}{
\begin{tabular}{lccccccc}
\toprule
 \textbf{F/P/R ↑} & SfM & $\text{EDGS}^\text{*}$ & Monodepth & DA3 & $\text{DA3}^\text{GS}$ & $\text{Laser}^+$ & Laser \\
\midrule
At init. & \cc{FBDCCC}{\color[HTML]{000000} $51$ / $79$ / $38$} & \cc{FDECE3}{\color[HTML]{000000} $47$ / $32$ / $92$} & \cc{D5675E}{\color[HTML]{F1F1F1} $65$ / $50$ / $97$} & \cc{FBD7C5}{\color[HTML]{000000} $52$ / $38$ / $89$} & \cc{A7CEE3}{\color[HTML]{000000} $29$ / $21$ / $50$} & \cc{FFFFFF}{\color[HTML]{F1F1F1} \color{black}} & \cc{FFFFFF}{\color[HTML]{F1F1F1} \color{black}} \\
\midrule
AbsGS & \cc{C6E0ED}{\color[HTML]{000000} $34$ / $29$ / $43$} & \cc{E0EEF5}{\color[HTML]{000000} $\mathbf{38}$ / $30$ / $\mathbf{56}$} & \cc{FEF4EF}{\color[HTML]{000000} $\mathbf{46}$ / $\mathbf{45}$ / $\mathbf{47}$} & \cc{FEF2EC}{\color[HTML]{000000} $\mathbf{46}$ / $\mathbf{45}$ / $\mathbf{48}$} & \cc{FBD6C4}{\color[HTML]{000000} $\mathbf{52}$ / $\mathbf{44}$ / $\mathbf{66}$} & \cc{D3E7F2}{\color[HTML]{000000} $36$ / $32$ / $43$} & \cc{FEF2EC}{\color[HTML]{000000} $\mathbf{46}$ / $\mathbf{45}$ / $\mathbf{48}$} \\
INRIA & \cc{BFDCEB}{\color[HTML]{000000} $32$ / $30$ / $38$} & \cc{E0EEF5}{\color[HTML]{000000} $\mathbf{38}$ / $30$ / $\mathbf{55}$} & \cc{FEF4EF}{\color[HTML]{000000} $\mathbf{46}$ / $\mathbf{45}$ / $\mathbf{47}$} & \cc{FEF4EF}{\color[HTML]{000000} $\mathbf{46}$ / $\mathbf{45}$ / $\mathbf{47}$} & \cc{FBD7C5}{\color[HTML]{000000} $\mathbf{52}$ / $\mathbf{45}$ / $\mathbf{65}$} & \cc{CBE3EF}{\color[HTML]{000000} $34$ / $32$ / $39$} & \cc{FEF4EF}{\color[HTML]{000000} $\mathbf{46}$ / $\mathbf{45}$ / $\mathbf{47}$} \\
IDHFR & \cc{97C3DD}{\color[HTML]{000000} $27$ / $19$ / $59$} & \cc{E1EFF6}{\color[HTML]{000000} $\mathbf{38}$ / $\mathbf{29}$ / $64$} & \cc{FAD3BF}{\color[HTML]{000000} $\mathbf{53}$ / $\mathbf{50}$ / $55$} & \cc{FBD6C4}{\color[HTML]{000000} $\mathbf{52}$ / $\mathbf{50}$ / $55$} & \cc{F3F9FB}{\color[HTML]{000000} $\mathbf{41}$ / $\mathbf{32}$ / $\mathbf{70}$} & \cc{D2E7F1}{\color[HTML]{000000} $\mathbf{36}$ / $\mathbf{28}$ / $58$} & \cc{FBD5C2}{\color[HTML]{000000} $\mathbf{52}$ / $\mathbf{50}$ / $55$} \\
MCMC & \cc{89B9D8}{\color[HTML]{000000} $26$ / $18$ / $55$} & \cc{C4DFED}{\color[HTML]{000000} $\mathbf{33}$ / $24$ / $57$} & \cc{FEF8F4}{\color[HTML]{000000} $\mathbf{45}$ / $\mathbf{43}$ / $48$} & \cc{FEFAF7}{\color[HTML]{000000} $\mathbf{45}$ / $\mathbf{43}$ / $48$} & \cc{BAD9EA}{\color[HTML]{000000} $32$ / $23$ / $\mathbf{59}$} & \cc{7AAED2}{\color[HTML]{F1F1F1} $24$ / $16$ / $55$} & \cc{FEF9F6}{\color[HTML]{000000} $\mathbf{45}$ / $\mathbf{43}$ / $48$} \\
RevDGS & \cc{EDF5FA}{\color[HTML]{000000} $40$ / $34$ / $53$} & \cc{F2F8FB}{\color[HTML]{000000} $41$ / $33$ / $\mathbf{58}$} & \cc{FAD3BF}{\color[HTML]{000000} $\mathbf{53}$ / $\mathbf{51}$ / $\mathbf{55}$} & \cc{FBD4C1}{\color[HTML]{000000} $\mathbf{52}$ / $\mathbf{50}$ / $55$} & \cc{FDE8DD}{\color[HTML]{000000} $\mathbf{48}$ / $\mathbf{39}$ / $\mathbf{67}$} & \cc{F8FBFD}{\color[HTML]{000000} $42$ / $37$ / $53$} & \cc{FAD3BF}{\color[HTML]{000000} $\mathbf{53}$ / $\mathbf{51}$ / $55$} \\
No D. & \cc{5C99C7}{\color[HTML]{F1F1F1} $21$ / $19$ / $25$} & \cc{DDEDF5}{\color[HTML]{000000} $\mathbf{38}$ / $29$ / $\mathbf{56}$} & \cc{FEF7F3}{\color[HTML]{000000} $\mathbf{45}$ / $\mathbf{45}$ / $\mathbf{46}$} & \cc{FEF7F3}{\color[HTML]{000000} $\mathbf{45}$ / $\mathbf{45}$ / $\mathbf{46}$} & \cc{FCE2D5}{\color[HTML]{000000} $\mathbf{49}$ / $\mathbf{42}$ / $\mathbf{63}$} & \cc{D3E7F2}{\color[HTML]{000000} $\mathbf{36}$ / $\mathbf{35}$ / $38$} & \cc{FEF6F1}{\color[HTML]{000000} $\mathbf{45}$ / $\mathbf{45}$ / $\mathbf{46}$} \\
\bottomrule
\end{tabular}
}
\endgroup
}
\end{table}

\section{Conclusion}
In this work, we systematically studied the impact of initialization on 3DGS novel view synthesis (NVS) quality and geometric accuracy.
We designed a comprehensive evaluation protocol and tested the performance of several initialization methods paired with multiple densification procedures.
Our detailed experiments lead to two key insights:
(1) On well-textured scenes with near-trajectory test images, dense initialization leads to marginal NVS improvements at best. There's simply little room for improvement as, on such scenes, base 3DGS already produces near-photorealistic results. Nevertheless, we do see benefits in generalizing to off-trajectory views and increased robustness under sparse supervision.
(2) Because photometric supervision is often insufficient to fully constrain scene geometry, densification methods that achieve the best NVS results using SfM initialization often do so at the expense of geometric accuracy.
Dense initialization helps avoid geometrically inaccurate local minima, alleviating this issue.

We thus conclude that, despite marginal NVS improvements, dense initialization provides a valuable prior that is beneficial to NVS generalization and geometric accuracy. We see great potential in developing view synthesis and geometry extraction approaches that treat initialization as a core component, fully utilizing the available information and adjusting densification to account for the specifics of a given initialization method.

\PAR{Acknowledgments.}
This work was supported by the Czech Science Foundation (GAČR) EXPRO (project UNI-3D, grant no. 23-07973X) and the CEDMO 2.0 NPO project (designation MPO 60273/24/21300/21000). We gratefully acknowledge access to the computational infrastructure of the OP VVV funded project CZ.02.1.01/0.0/0.0/16 019/0000765 “Research Center for Informatics”.

\bibliographystyle{splncs04}
\bibliography{main}

\clearpage
\appendix  %

\renewcommand{\theHsection}{A\arabic{section}}

\setcounter{equation}{0}
\setcounter{figure}{0}
\setcounter{table}{0}

\renewcommand{\thetable}{S\arabic{table}} 
\renewcommand{\thefigure}{S\arabic{figure}} 

\phantomsection
\addcontentsline{toc}{section}{Supplementary Material}

\suppressfloats[t]
\begin{center}
    {\huge \textbf{Supplementary Material}}
\end{center}

\section{Geometry Extraction Proof of Concept Using PGSR}
Based on our findings, we assume that there is potential in using dense initialization from readily available sources, such as monocular depth predictions, to provide priors for 3DGS-based geometry extraction methods.
To put this claim to the test, we performed preliminary testing of using dense initialization from our Monodepth implementation with PGSR~\cite{pgsr}, without further modification to the method. We choose PGSR as it is well established, relatively fast, and maintains competitive geometric accuracy results even against very recent
methods~\cite[Tabs. 1 and 2]{va_gs_2026}. We trained with and without dense initialization on 2 scenes from the Tanks \& Temples~\cite{tanksandtemples} dataset with GT data available -- ``Meetingroom'' and  ``Ignatius''. We use the same parameters for Monodepth initialization as in our other experiments (ViT-Large, 300 image limit).
We compute $\text{F}_1$ scores using the scripts from the official PGSR implementation.\footnote{We successfully replicated the results reported by the authors on their GitHub (the authors further improved their method with hyper-parameter adjustments since the paper was published).}

\begin{figure}[tbp]
    \centering
    \setlength{\tabcolsep}{1.5pt}

    \begin{tabular}{cc}
        SfM & Monodepth \\[0.6em]

        \includegraphics[width=0.48\textwidth]{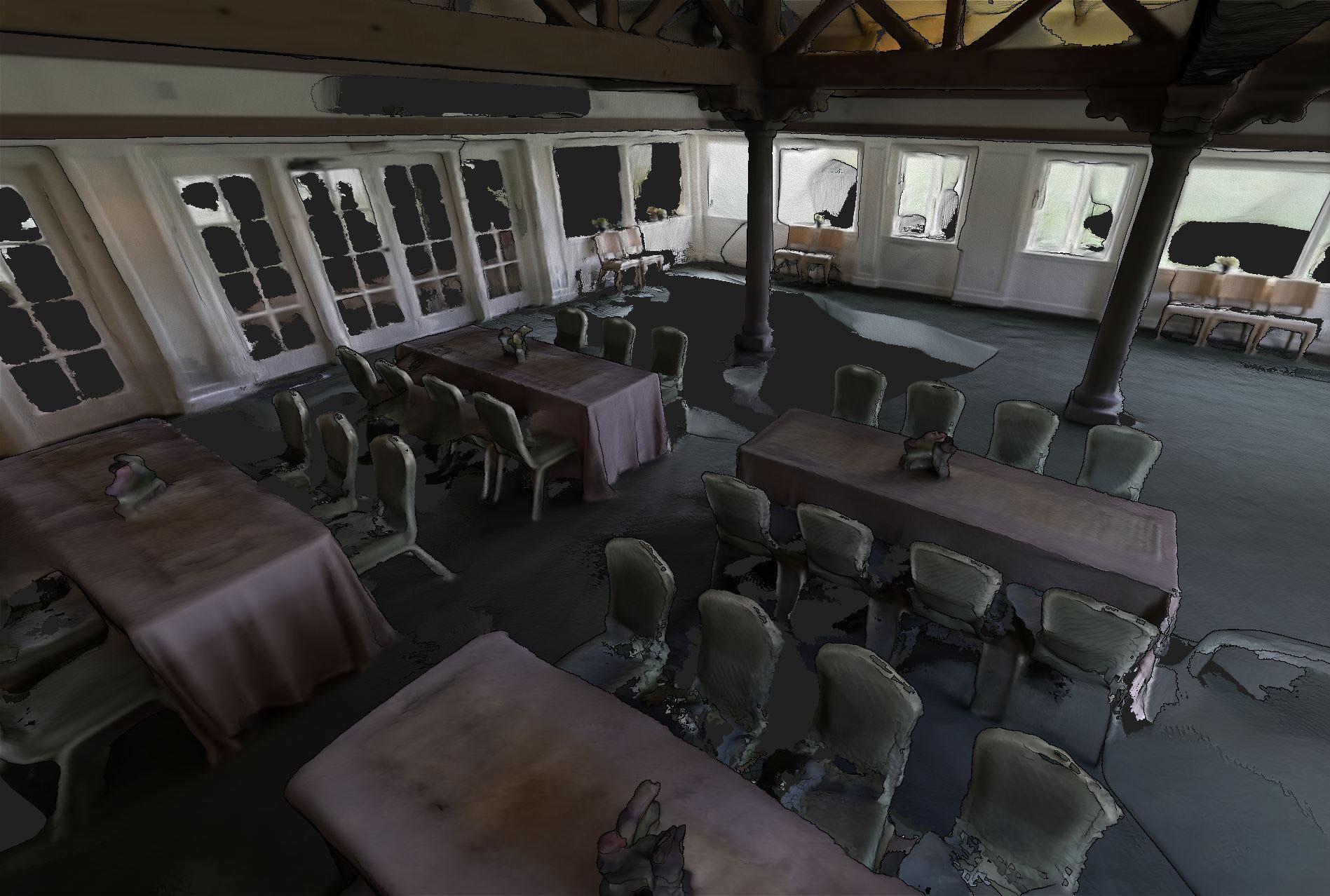} &
        \includegraphics[width=0.48\textwidth]{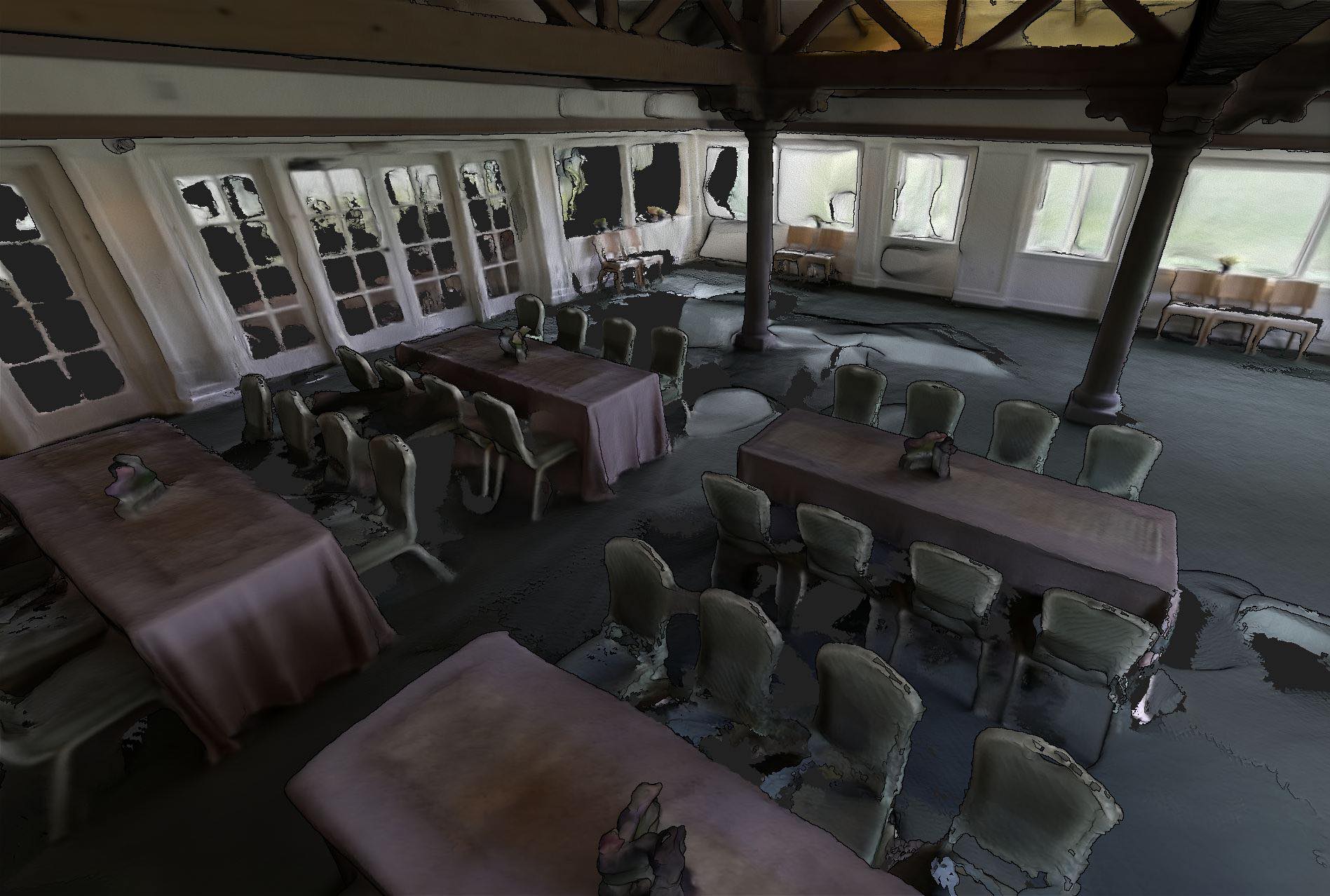} \\%
        \includegraphics[width=0.48\textwidth]{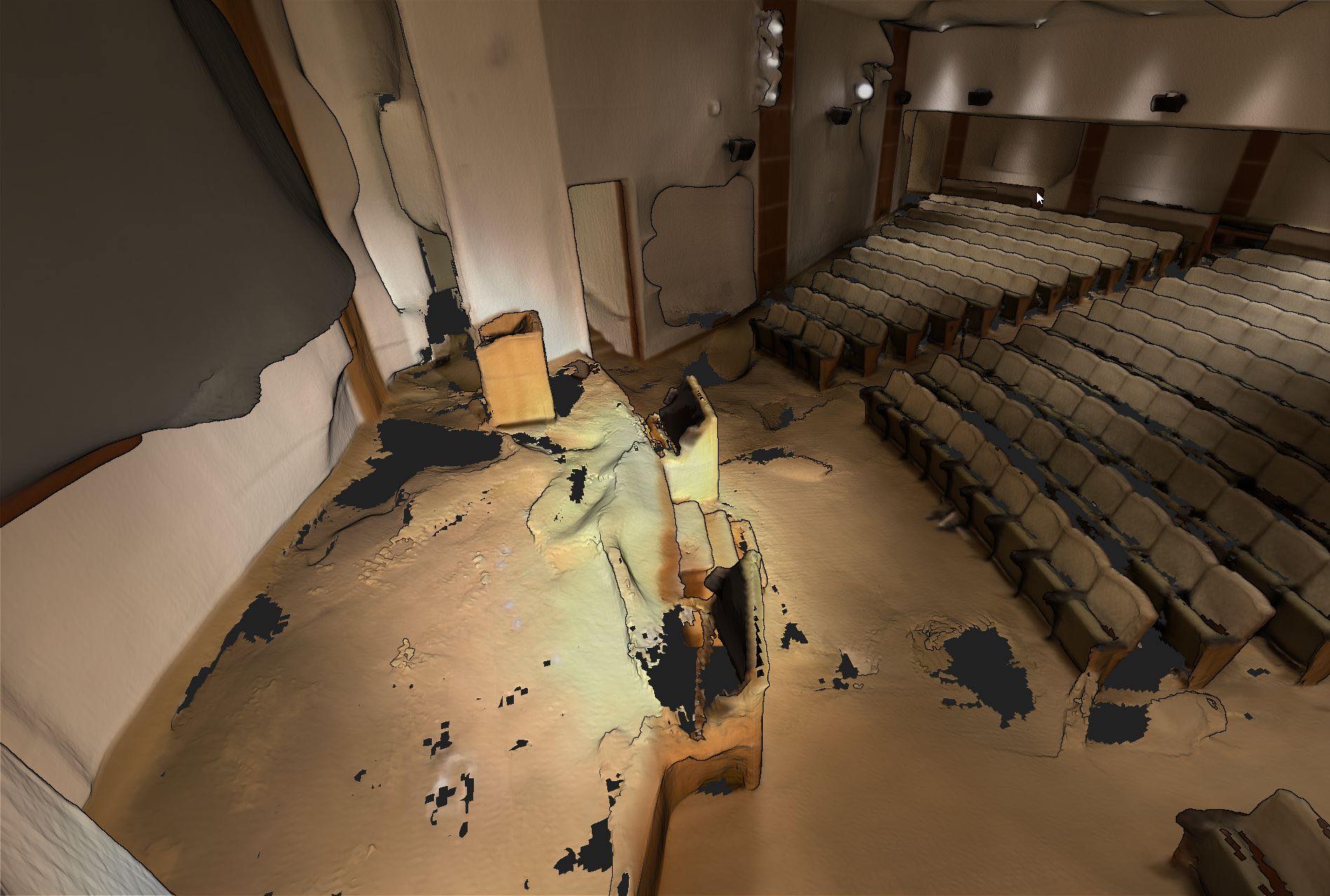} &
        \includegraphics[width=0.48\textwidth]{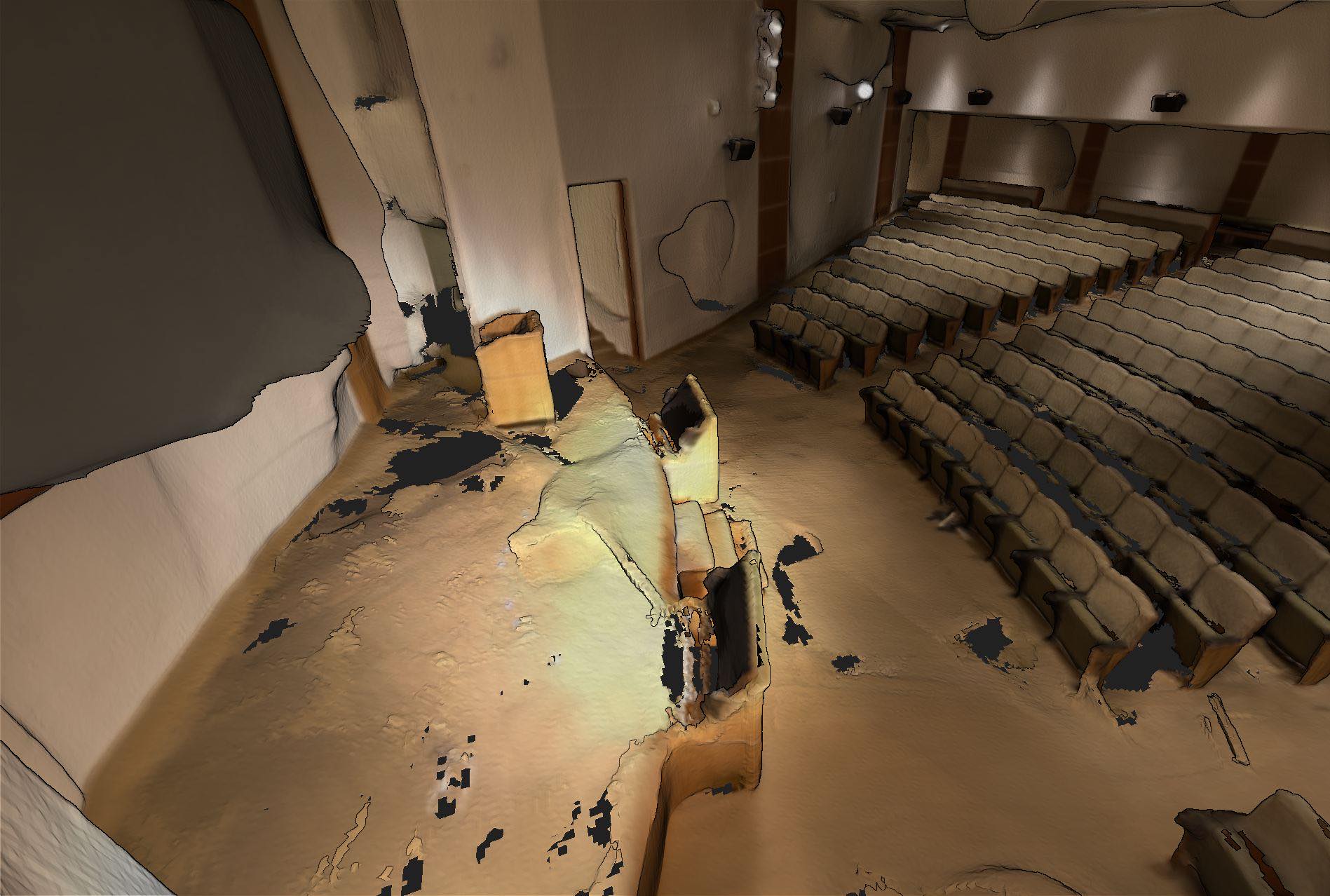} \\%
        \includegraphics[width=0.48\textwidth]{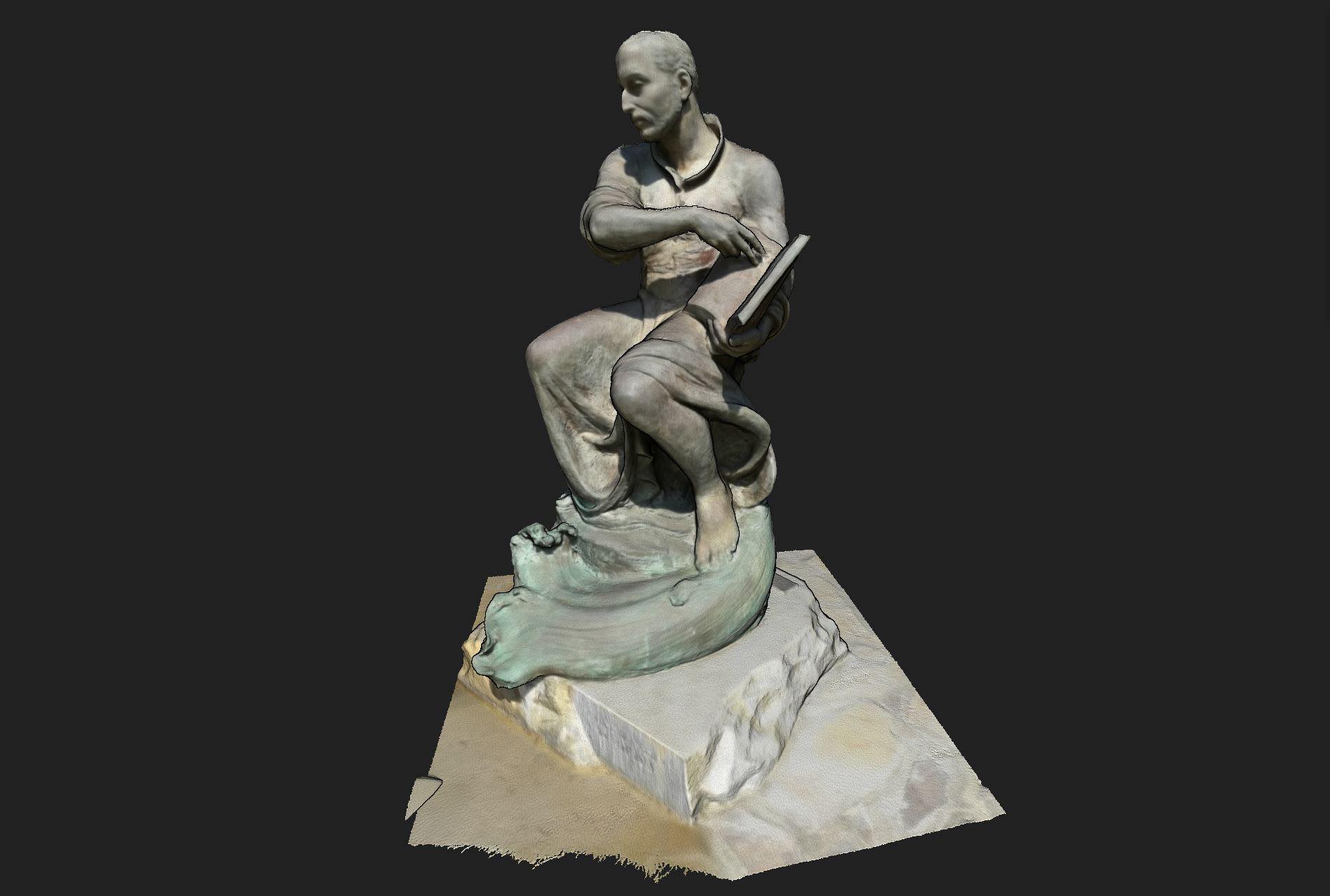} &
        \includegraphics[width=0.48\textwidth]{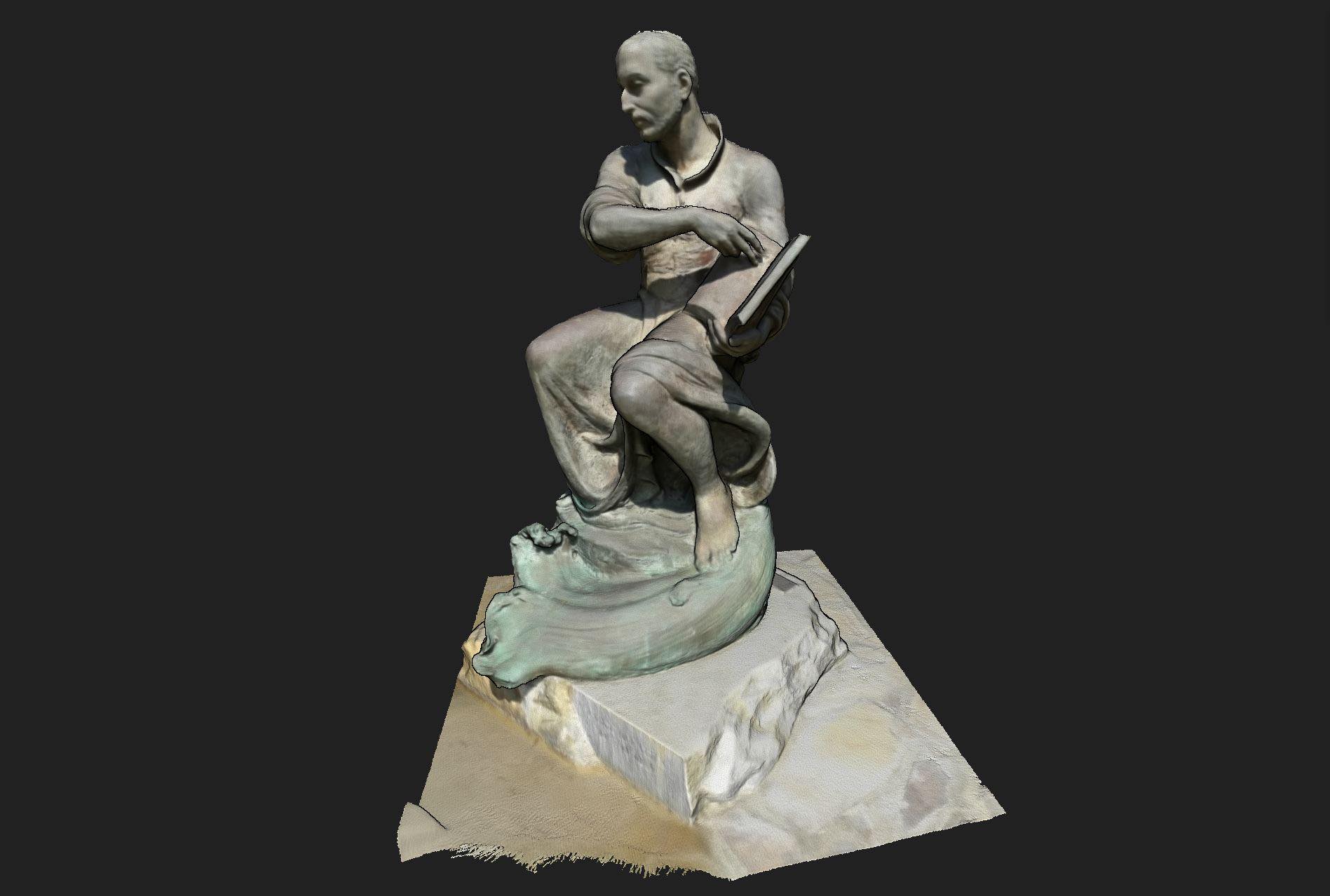}
    \end{tabular}

    \caption{Comparison of PGSR-extracted meshes with SfM and Monodepth initialization on scenes from Tanks\&Temples~\cite{tanksandtemples}. Eye Dome Lighting~\cite{eye_dome_lighting} was used to improve legibility of geometric detail. Depicted scenes (top to bottom): ``Meetingroom'', ``Auditorium'', ``Ignatius''.}
    \label{fig:pgsr_meshes}
\end{figure}

On ``Meetingroom'', the $\text{F}_1$ score increases from 0.32 to 0.34, and, as demonstrated in~\cref{fig:pgsr_meshes}, using dense initialization resolves some of the holes in the ground and on the walls (and brings back a missing chair). On ``Ignatius'', the statue's geometry is well constrained by the input views, so we do not see any change in the $\text{F}_1$ score. Qualitative evaluation also does not reveal any differences. As the Tanks \& Temples scenes with GT geometry do not contain many planar untextured surfaces, which are often a failure case for geometry extraction, we also provide qualitative results on the ``Auditorium'' scene (\cref{fig:pgsr_meshes}), where we see a similar reduction in discontinuities of the mesh, as well as improved accuracy in some areas of the scene. This shows that dense initialization can help in correctly resolving geometry under challenging conditions. However, this is just a preliminary test, and these effects could likely be enhanced by tweaking hyper-parameters or adjusting the optimization procedure to account for dense initialization.

\section{Verification of Chosen Scene Size Limits}\label{scene_size_limit_verif}

As discussed in~\cref{sec:protocol} of the paper, we use a per-scene limit $G_m$ on the number of Gaussians in the scene, which we set to be the final scene size when trained with SfM and AbsGS~\cite{absgs}. In \Cref{tab:gaussian_cap_ablation_sfm,tab:gaussian_cap_ablation_laser_scan}, we present results of an experiment where we train with scene size limits decreased to $0.75G_m$ and increased to $1.25G_m$. Neither increasing nor decreasing the limits provides large or consistent improvements in metric values across all densification strategies. We thus conclude that the used $G_m$ values are sufficient to represent the scenes, and our results are not systematically affected by low scene size limits.
\begin{table}[tbp]
\centering
\caption{Results of varying the scene size limit with SfM initialization.}
\label{tab:gaussian_cap_ablation_sfm}
\begin{subtable}[t]{0.8\linewidth}
\centering
\caption{ScanNet++}
\label{tab:gaussian_cap_ablation_sfm_scannet++}
{\small
\begingroup \setlength{\tabcolsep}{2.5\tabcolsep}
\resizebox{\linewidth}{!}{\begin{tabular}{l|ccc|ccc|ccc}
\toprule
& \multicolumn{3}{c|}{\textbf{PSNR ↑}} & \multicolumn{3}{c|}{\textbf{SSIM ↑}} & \multicolumn{3}{c}{\textbf{LPIPS ↓}} \\
Cap frac.& $0.75\text  {G}_\mathit{m}$ & $1.0\text  {G}_\mathit{m}$ & $1.25\text  {G}_\mathit{m}$ & $0.75\text  {G}_\mathit{m}$ & $1.0\text  {G}_\mathit{m}$ & $1.25\text  {G}_\mathit{m}$ & $0.75\text  {G}_\mathit{m}$ & $1.0\text  {G}_\mathit{m}$ & $1.25\text  {G}_\mathit{m}$ \\
\midrule
AbsGS & \cc{D2E7F1}{\color[HTML]{000000} $22.30$} & \cc{E0EEF5}{\color[HTML]{000000} $22.36$} & \cc{EBF4F9}{\color[HTML]{000000} $22.41$} & \cc{74AAD0}{\color[HTML]{F1F1F1} $0.869$} & \cc{A5CDE3}{\color[HTML]{000000} $0.870$} & \cc{89B9D8}{\color[HTML]{000000} $0.870$} & \cc{BAD9EA}{\color[HTML]{000000} $0.253$} & \cc{DCECF4}{\color[HTML]{000000} $0.249$} & \cc{D7E9F3}{\color[HTML]{000000} $0.250$} \\
INRIA & \cc{BBDAEA}{\color[HTML]{000000} $22.20$} & \cc{CAE2EF}{\color[HTML]{000000} $22.26$} & \cc{CDE4F0}{\color[HTML]{000000} $22.28$} & \cc{629DC9}{\color[HTML]{F1F1F1} $0.869$} & \cc{89B9D8}{\color[HTML]{000000} $0.870$} & \cc{74AAD0}{\color[HTML]{F1F1F1} $0.869$} & \cc{91BFDB}{\color[HTML]{000000} $0.255$} & \cc{B2D6E7}{\color[HTML]{000000} $0.253$} & \cc{A9CFE4}{\color[HTML]{000000} $0.254$} \\
MCMC & \cc{C0DDEB}{\color[HTML]{000000} $22.22$} & \cc{7AAED2}{\color[HTML]{F1F1F1} $22.02$} & \cc{5C99C7}{\color[HTML]{F1F1F1} $21.94$} & \cc{FFFCFA}{\color[HTML]{000000} $0.873$} & \cc{9FC9E0}{\color[HTML]{000000} $0.870$} & \cc{B1D5E7}{\color[HTML]{000000} $0.870$} & \cc{F9FCFD}{\color[HTML]{000000} $0.246$} & \cc{DDEDF5}{\color[HTML]{000000} $0.249$} & \cc{EDF5FA}{\color[HTML]{000000} $0.248$} \\
IDHFR & \cc{D5675E}{\color[HTML]{F1F1F1} $23.05$} & \cc{D66960}{\color[HTML]{F1F1F1} $23.04$} & \cc{DA7368}{\color[HTML]{F1F1F1} $23.02$} & \cc{D5675E}{\color[HTML]{F1F1F1} $0.877$} & \cc{DE7E70}{\color[HTML]{F1F1F1} $0.877$} & \cc{E38A7A}{\color[HTML]{F1F1F1} $0.877$} & \cc{E38A7A}{\color[HTML]{F1F1F1} $0.235$} & \cc{D5675E}{\color[HTML]{F1F1F1} $0.233$} & \cc{D86F65}{\color[HTML]{F1F1F1} $0.233$} \\
RevDGS & \cc{E2EFF6}{\color[HTML]{000000} $22.37$} & \cc{EDF5FA}{\color[HTML]{000000} $22.42$} & \cc{E1EFF6}{\color[HTML]{000000} $22.36$} & \cc{629DC9}{\color[HTML]{F1F1F1} $0.869$} & \cc{7FB2D4}{\color[HTML]{000000} $0.869$} & \cc{5C99C7}{\color[HTML]{F1F1F1} $0.869$} & \cc{5C99C7}{\color[HTML]{F1F1F1} $0.259$} & \cc{7FB2D4}{\color[HTML]{000000} $0.257$} & \cc{83B5D6}{\color[HTML]{000000} $0.256$} \\
\bottomrule
\end{tabular}}
\endgroup
}
\end{subtable}
\begin{subtable}[t]{0.8\linewidth}
\centering
\caption{ScanNet++ (On-Trajectory)}
\label{tab:gaussian_cap_ablation_sfm_eval_on_train_set_scannet++}
{\small
\begingroup \setlength{\tabcolsep}{2.5\tabcolsep}
\resizebox{\linewidth}{!}{\begin{tabular}{l|ccc|ccc|ccc}
\toprule
& \multicolumn{3}{c|}{\textbf{PSNR ↑}} & \multicolumn{3}{c|}{\textbf{SSIM ↑}} & \multicolumn{3}{c}{\textbf{LPIPS ↓}} \\
Cap frac.& $0.75\text  {G}_\mathit{m}$ & $1.0\text  {G}_\mathit{m}$ & $1.25\text  {G}_\mathit{m}$ & $0.75\text  {G}_\mathit{m}$ & $1.0\text  {G}_\mathit{m}$ & $1.25\text  {G}_\mathit{m}$ & $0.75\text  {G}_\mathit{m}$ & $1.0\text  {G}_\mathit{m}$ & $1.25\text  {G}_\mathit{m}$ \\
\midrule
AbsGS & \cc{C5DFED}{\color[HTML]{000000} $33.10$} & \cc{E3F0F6}{\color[HTML]{000000} $33.20$} & \cc{D8EAF3}{\color[HTML]{000000} $33.17$} & \cc{FFFCFA}{\color[HTML]{000000} $0.951$} & \cc{FCE0D2}{\color[HTML]{000000} $0.951$} & \cc{FDEDE5}{\color[HTML]{000000} $0.951$} & \cc{FFFFFE}{\color[HTML]{000000} $0.104$} & \cc{FBD7C5}{\color[HTML]{000000} $0.101$} & \cc{FBDCCC}{\color[HTML]{000000} $0.101$} \\
INRIA & \cc{BCDBEA}{\color[HTML]{000000} $33.07$} & \cc{C6E0ED}{\color[HTML]{000000} $33.10$} & \cc{C5DFED}{\color[HTML]{000000} $33.10$} & \cc{C2DEEC}{\color[HTML]{000000} $0.949$} & \cc{D3E7F2}{\color[HTML]{000000} $0.950$} & \cc{CDE4F0}{\color[HTML]{000000} $0.950$} & \cc{5C99C7}{\color[HTML]{F1F1F1} $0.113$} & \cc{83B5D6}{\color[HTML]{000000} $0.112$} & \cc{81B4D5}{\color[HTML]{000000} $0.112$} \\
MCMC & \cc{83B5D6}{\color[HTML]{000000} $32.93$} & \cc{97C3DD}{\color[HTML]{000000} $32.97$} & \cc{85B6D7}{\color[HTML]{000000} $32.93$} & \cc{FDEEE6}{\color[HTML]{000000} $0.951$} & \cc{FCE2D5}{\color[HTML]{000000} $0.951$} & \cc{FBD9C8}{\color[HTML]{000000} $0.951$} & \cc{E5F1F7}{\color[HTML]{000000} $0.106$} & \cc{FFFEFD}{\color[HTML]{000000} $0.104$} & \cc{FDEBE2}{\color[HTML]{000000} $0.102$} \\
IDHFR & \cc{DD7C6E}{\color[HTML]{F1F1F1} $33.72$} & \cc{D5675E}{\color[HTML]{F1F1F1} $33.76$} & \cc{E58E7D}{\color[HTML]{F1F1F1} $33.68$} & \cc{DC776B}{\color[HTML]{F1F1F1} $0.953$} & \cc{D5675E}{\color[HTML]{F1F1F1} $0.953$} & \cc{DF8072}{\color[HTML]{F1F1F1} $0.953$} & \cc{F0A992}{\color[HTML]{000000} $0.098$} & \cc{D5675E}{\color[HTML]{F1F1F1} $0.095$} & \cc{D66960}{\color[HTML]{F1F1F1} $0.095$} \\
RevDGS & \cc{5C99C7}{\color[HTML]{F1F1F1} $32.84$} & \cc{85B6D7}{\color[HTML]{000000} $32.93$} & \cc{83B5D6}{\color[HTML]{000000} $32.92$} & \cc{5C99C7}{\color[HTML]{F1F1F1} $0.948$} & \cc{7FB2D4}{\color[HTML]{000000} $0.949$} & \cc{70A7CE}{\color[HTML]{F1F1F1} $0.948$} & \cc{CBE3EF}{\color[HTML]{000000} $0.108$} & \cc{F2F8FB}{\color[HTML]{000000} $0.105$} & \cc{FDFEFE}{\color[HTML]{000000} $0.104$} \\
\bottomrule
\end{tabular}}
\endgroup
}
\end{subtable}
\begin{subtable}[t]{0.8\linewidth}
\centering
\caption{ETH3D}
\label{tab:gaussian_cap_ablation_sfm_eth3d}
{\small
\begingroup \setlength{\tabcolsep}{2.5\tabcolsep}
\resizebox{\linewidth}{!}{\begin{tabular}{l|ccc|ccc|ccc}
\toprule
& \multicolumn{3}{c|}{\textbf{PSNR ↑}} & \multicolumn{3}{c|}{\textbf{SSIM ↑}} & \multicolumn{3}{c}{\textbf{LPIPS ↓}} \\
Cap frac.& $0.75\text  {G}_\mathit{m}$ & $1.0\text  {G}_\mathit{m}$ & $1.25\text  {G}_\mathit{m}$ & $0.75\text  {G}_\mathit{m}$ & $1.0\text  {G}_\mathit{m}$ & $1.25\text  {G}_\mathit{m}$ & $0.75\text  {G}_\mathit{m}$ & $1.0\text  {G}_\mathit{m}$ & $1.25\text  {G}_\mathit{m}$ \\
\midrule
AbsGS & \cc{E7F2F8}{\color[HTML]{000000} $21.16$} & \cc{E1EFF6}{\color[HTML]{000000} $21.06$} & \cc{E8F3F8}{\color[HTML]{000000} $21.18$} & \cc{FFFDFB}{\color[HTML]{000000} $0.796$} & \cc{FFFBF9}{\color[HTML]{000000} $0.796$} & \cc{FFFEFD}{\color[HTML]{000000} $0.795$} & \cc{FEF8F4}{\color[HTML]{000000} $0.315$} & \cc{FDECE3}{\color[HTML]{000000} $0.305$} & \cc{FEF7F3}{\color[HTML]{000000} $0.315$} \\
INRIA & \cc{FDECE3}{\color[HTML]{000000} $21.94$} & \cc{FDEBE2}{\color[HTML]{000000} $21.95$} & \cc{FEF3ED}{\color[HTML]{000000} $21.80$} & \cc{F9C6AD}{\color[HTML]{000000} $0.825$} & \cc{F9C1A6}{\color[HTML]{000000} $0.827$} & \cc{FACDB7}{\color[HTML]{000000} $0.821$} & \cc{FCE7DC}{\color[HTML]{000000} $0.300$} & \cc{FBDBCB}{\color[HTML]{000000} $0.289$} & \cc{FDF0E9}{\color[HTML]{000000} $0.308$} \\
MCMC & \cc{D97166}{\color[HTML]{F1F1F1} $23.60$} & \cc{D97166}{\color[HTML]{F1F1F1} $23.60$} & \cc{D5675E}{\color[HTML]{F1F1F1} $23.71$} & \cc{D5675E}{\color[HTML]{F1F1F1} $0.851$} & \cc{D86D63}{\color[HTML]{F1F1F1} $0.849$} & \cc{D5675E}{\color[HTML]{F1F1F1} $0.851$} & \cc{DE7E70}{\color[HTML]{F1F1F1} $0.235$} & \cc{DA7368}{\color[HTML]{F1F1F1} $0.231$} & \cc{D5675E}{\color[HTML]{F1F1F1} $0.225$} \\
IDHFR & \cc{FBDAC9}{\color[HTML]{000000} $22.30$} & \cc{FBD4C1}{\color[HTML]{000000} $22.44$} & \cc{FAD2BE}{\color[HTML]{000000} $22.48$} & \cc{FACDB7}{\color[HTML]{000000} $0.821$} & \cc{FAD1BC}{\color[HTML]{000000} $0.819$} & \cc{FACFBA}{\color[HTML]{000000} $0.820$} & \cc{FACBB4}{\color[HTML]{000000} $0.274$} & \cc{F9C3A8}{\color[HTML]{000000} $0.267$} & \cc{FACDB7}{\color[HTML]{000000} $0.277$} \\
RevDGS & \cc{68A2CB}{\color[HTML]{F1F1F1} $19.50$} & \cc{70A7CE}{\color[HTML]{F1F1F1} $19.58$} & \cc{5C99C7}{\color[HTML]{F1F1F1} $19.38$} & \cc{8BBBD9}{\color[HTML]{000000} $0.750$} & \cc{78ADD1}{\color[HTML]{F1F1F1} $0.745$} & \cc{5C99C7}{\color[HTML]{F1F1F1} $0.737$} & \cc{5C99C7}{\color[HTML]{F1F1F1} $0.420$} & \cc{72A9CF}{\color[HTML]{F1F1F1} $0.410$} & \cc{6AA3CC}{\color[HTML]{F1F1F1} $0.414$} \\
\bottomrule
\end{tabular}}
\endgroup
}
\end{subtable}
\end{table}

\begin{table}[tbp]
\centering
\caption{Results of varying the scene size limit with laser scan initialization at $0.5G_m$.}
\label{tab:gaussian_cap_ablation_laser_scan}
\begin{subtable}[t]{0.8\linewidth}
\centering
\caption{ScanNet++}
\label{tab:gaussian_cap_ablation_laser_scan_scannet++}
{\small
\begingroup \setlength{\tabcolsep}{2.5\tabcolsep}
\resizebox{\linewidth}{!}{\begin{tabular}{l|ccc|ccc|ccc}
\toprule
& \multicolumn{3}{c|}{\textbf{PSNR ↑}} & \multicolumn{3}{c|}{\textbf{SSIM ↑}} & \multicolumn{3}{c}{\textbf{LPIPS ↓}} \\
Cap frac.& $0.75\text  {G}_\mathit{m}$ & $1.0\text  {G}_\mathit{m}$ & $1.25\text  {G}_\mathit{m}$ & $0.75\text  {G}_\mathit{m}$ & $1.0\text  {G}_\mathit{m}$ & $1.25\text  {G}_\mathit{m}$ & $0.75\text  {G}_\mathit{m}$ & $1.0\text  {G}_\mathit{m}$ & $1.25\text  {G}_\mathit{m}$ \\
\midrule
AbsGS & \cc{CFE5F0}{\color[HTML]{000000} $22.76$} & \cc{CBE3EF}{\color[HTML]{000000} $22.75$} & \cc{DCECF4}{\color[HTML]{000000} $22.80$} & \cc{DBEBF4}{\color[HTML]{000000} $0.872$} & \cc{C5DFED}{\color[HTML]{000000} $0.871$} & \cc{C5DFED}{\color[HTML]{000000} $0.871$} & \cc{DBEBF4}{\color[HTML]{000000} $0.250$} & \cc{E3F0F6}{\color[HTML]{000000} $0.249$} & \cc{F7FBFD}{\color[HTML]{000000} $0.247$} \\
INRIA & \cc{C5DFED}{\color[HTML]{000000} $22.73$} & \cc{C8E1EE}{\color[HTML]{000000} $22.75$} & \cc{DFEDF5}{\color[HTML]{000000} $22.81$} & \cc{DDEDF5}{\color[HTML]{000000} $0.872$} & \cc{CFE5F0}{\color[HTML]{000000} $0.872$} & \cc{DCECF4}{\color[HTML]{000000} $0.872$} & \cc{D5E8F2}{\color[HTML]{000000} $0.250$} & \cc{E7F2F8}{\color[HTML]{000000} $0.249$} & \cc{FFFBF9}{\color[HTML]{000000} $0.246$} \\
MCMC & \cc{609CC8}{\color[HTML]{F1F1F1} $22.53$} & \cc{5C99C7}{\color[HTML]{F1F1F1} $22.52$} & \cc{5C99C7}{\color[HTML]{F1F1F1} $22.52$} & \cc{FEF7F3}{\color[HTML]{000000} $0.874$} & \cc{F8FBFD}{\color[HTML]{000000} $0.873$} & \cc{FDECE3}{\color[HTML]{000000} $0.875$} & \cc{FEF1EA}{\color[HTML]{000000} $0.245$} & \cc{FDEAE0}{\color[HTML]{000000} $0.244$} & \cc{FBD4C1}{\color[HTML]{000000} $0.242$} \\
IDHFR & \cc{D5675E}{\color[HTML]{F1F1F1} $23.29$} & \cc{DF8072}{\color[HTML]{F1F1F1} $23.24$} & \cc{E08273}{\color[HTML]{F1F1F1} $23.24$} & \cc{D5675E}{\color[HTML]{F1F1F1} $0.879$} & \cc{E6907F}{\color[HTML]{F1F1F1} $0.878$} & \cc{E28878}{\color[HTML]{F1F1F1} $0.878$} & \cc{DD7C6E}{\color[HTML]{F1F1F1} $0.236$} & \cc{DC776B}{\color[HTML]{F1F1F1} $0.236$} & \cc{D5675E}{\color[HTML]{F1F1F1} $0.235$} \\
RevDGS & \cc{68A2CB}{\color[HTML]{F1F1F1} $22.54$} & \cc{9DC7E0}{\color[HTML]{000000} $22.64$} & \cc{B1D5E7}{\color[HTML]{000000} $22.68$} & \cc{5C99C7}{\color[HTML]{F1F1F1} $0.868$} & \cc{A1CAE1}{\color[HTML]{000000} $0.870$} & \cc{8FBDDA}{\color[HTML]{000000} $0.870$} & \cc{5C99C7}{\color[HTML]{F1F1F1} $0.258$} & \cc{85B6D7}{\color[HTML]{000000} $0.256$} & \cc{91BFDB}{\color[HTML]{000000} $0.255$} \\
\bottomrule
\end{tabular}}
\endgroup
}
\end{subtable}
\begin{subtable}[t]{0.8\linewidth}
\centering
\caption{ScanNet++ (On-Trajectory)}
\label{tab:gaussian_cap_ablation_laser_scan_eval_on_train_set_scannet++}
{\small
\begingroup \setlength{\tabcolsep}{2.5\tabcolsep}
\resizebox{\linewidth}{!}{\begin{tabular}{l|ccc|ccc|ccc}
\toprule
& \multicolumn{3}{c|}{\textbf{PSNR ↑}} & \multicolumn{3}{c|}{\textbf{SSIM ↑}} & \multicolumn{3}{c}{\textbf{LPIPS ↓}} \\
Cap frac.& $0.75\text  {G}_\mathit{m}$ & $1.0\text  {G}_\mathit{m}$ & $1.25\text  {G}_\mathit{m}$ & $0.75\text  {G}_\mathit{m}$ & $1.0\text  {G}_\mathit{m}$ & $1.25\text  {G}_\mathit{m}$ & $0.75\text  {G}_\mathit{m}$ & $1.0\text  {G}_\mathit{m}$ & $1.25\text  {G}_\mathit{m}$ \\
\midrule
AbsGS & \cc{99C4DE}{\color[HTML]{000000} $32.86$} & \cc{C2DEEC}{\color[HTML]{000000} $33.00$} & \cc{E2EFF6}{\color[HTML]{000000} $33.13$} & \cc{BFDCEB}{\color[HTML]{000000} $0.949$} & \cc{EBF4F9}{\color[HTML]{000000} $0.950$} & \cc{FEF9F6}{\color[HTML]{000000} $0.951$} & \cc{BDDBEB}{\color[HTML]{000000} $0.104$} & \cc{FFFEFD}{\color[HTML]{000000} $0.100$} & \cc{FBDBCB}{\color[HTML]{000000} $0.097$} \\
INRIA & \cc{74AAD0}{\color[HTML]{F1F1F1} $32.76$} & \cc{ADD2E6}{\color[HTML]{000000} $32.91$} & \cc{CBE3EF}{\color[HTML]{000000} $33.03$} & \cc{74AAD0}{\color[HTML]{F1F1F1} $0.948$} & \cc{C2DEEC}{\color[HTML]{000000} $0.949$} & \cc{DBEBF4}{\color[HTML]{000000} $0.950$} & \cc{5C99C7}{\color[HTML]{F1F1F1} $0.107$} & \cc{C2DEEC}{\color[HTML]{000000} $0.103$} & \cc{E1EFF6}{\color[HTML]{000000} $0.102$} \\
MCMC & \cc{5C99C7}{\color[HTML]{F1F1F1} $32.70$} & \cc{99C4DE}{\color[HTML]{000000} $32.86$} & \cc{B1D5E7}{\color[HTML]{000000} $32.92$} & \cc{FEF7F3}{\color[HTML]{000000} $0.951$} & \cc{FAD3BF}{\color[HTML]{000000} $0.952$} & \cc{F9C7AE}{\color[HTML]{000000} $0.952$} & \cc{C5DFED}{\color[HTML]{000000} $0.103$} & \cc{FFFDFB}{\color[HTML]{000000} $0.100$} & \cc{FCDECF}{\color[HTML]{000000} $0.098$} \\
IDHFR & \cc{ECA18C}{\color[HTML]{000000} $33.66$} & \cc{DC796D}{\color[HTML]{F1F1F1} $33.75$} & \cc{D5675E}{\color[HTML]{F1F1F1} $33.80$} & \cc{F5B89E}{\color[HTML]{000000} $0.952$} & \cc{E08273}{\color[HTML]{F1F1F1} $0.953$} & \cc{D5675E}{\color[HTML]{F1F1F1} $0.953$} & \cc{FDEDE5}{\color[HTML]{000000} $0.099$} & \cc{E89784}{\color[HTML]{000000} $0.094$} & \cc{D5675E}{\color[HTML]{F1F1F1} $0.093$} \\
RevDGS & \cc{68A2CB}{\color[HTML]{F1F1F1} $32.73$} & \cc{76ABD1}{\color[HTML]{F1F1F1} $32.77$} & \cc{97C3DD}{\color[HTML]{000000} $32.85$} & \cc{5C99C7}{\color[HTML]{F1F1F1} $0.948$} & \cc{7AAED2}{\color[HTML]{F1F1F1} $0.948$} & \cc{9FC9E0}{\color[HTML]{000000} $0.949$} & \cc{7CAFD3}{\color[HTML]{F1F1F1} $0.106$} & \cc{BBDAEA}{\color[HTML]{000000} $0.104$} & \cc{E7F2F8}{\color[HTML]{000000} $0.101$} \\
\bottomrule
\end{tabular}}
\endgroup
}
\end{subtable}
\begin{subtable}[t]{0.8\linewidth}
\centering
\caption{ETH3D}
\label{tab:gaussian_cap_ablation_laser_scan_eth3d}
{\small
\begingroup \setlength{\tabcolsep}{2.5\tabcolsep}
\resizebox{\linewidth}{!}{\begin{tabular}{l|ccc|ccc|ccc}
\toprule
& \multicolumn{3}{c|}{\textbf{PSNR ↑}} & \multicolumn{3}{c|}{\textbf{SSIM ↑}} & \multicolumn{3}{c}{\textbf{LPIPS ↓}} \\
Cap frac.& $0.75\text  {G}_\mathit{m}$ & $1.0\text  {G}_\mathit{m}$ & $1.25\text  {G}_\mathit{m}$ & $0.75\text  {G}_\mathit{m}$ & $1.0\text  {G}_\mathit{m}$ & $1.25\text  {G}_\mathit{m}$ & $0.75\text  {G}_\mathit{m}$ & $1.0\text  {G}_\mathit{m}$ & $1.25\text  {G}_\mathit{m}$ \\
\midrule
AbsGS & \cc{DFEDF5}{\color[HTML]{000000} $23.01$} & \cc{ADD2E6}{\color[HTML]{000000} $22.61$} & \cc{B1D5E7}{\color[HTML]{000000} $22.63$} & \cc{FEF2EC}{\color[HTML]{000000} $0.851$} & \cc{C6E0ED}{\color[HTML]{000000} $0.836$} & \cc{D7E9F3}{\color[HTML]{000000} $0.840$} & \cc{F8FBFD}{\color[HTML]{000000} $0.223$} & \cc{D8EAF3}{\color[HTML]{000000} $0.235$} & \cc{E0EEF5}{\color[HTML]{000000} $0.232$} \\
INRIA & \cc{FFFDFB}{\color[HTML]{000000} $23.31$} & \cc{D6E9F2}{\color[HTML]{000000} $22.94$} & \cc{E8F3F8}{\color[HTML]{000000} $23.10$} & \cc{FDE8DD}{\color[HTML]{000000} $0.854$} & \cc{EAF3F8}{\color[HTML]{000000} $0.843$} & \cc{FFFBF9}{\color[HTML]{000000} $0.849$} & \cc{FEF9F6}{\color[HTML]{000000} $0.218$} & \cc{F7FBFD}{\color[HTML]{000000} $0.223$} & \cc{FFFBF9}{\color[HTML]{000000} $0.218$} \\
MCMC & \cc{D5675E}{\color[HTML]{F1F1F1} $24.40$} & \cc{D86F65}{\color[HTML]{F1F1F1} $24.36$} & \cc{D97166}{\color[HTML]{F1F1F1} $24.34$} & \cc{D5675E}{\color[HTML]{F1F1F1} $0.874$} & \cc{DC796D}{\color[HTML]{F1F1F1} $0.872$} & \cc{E18676}{\color[HTML]{F1F1F1} $0.870$} & \cc{D5675E}{\color[HTML]{F1F1F1} $0.174$} & \cc{D86D63}{\color[HTML]{F1F1F1} $0.175$} & \cc{DB7569}{\color[HTML]{F1F1F1} $0.177$} \\
IDHFR & \cc{E3F0F6}{\color[HTML]{000000} $23.05$} & \cc{B2D6E7}{\color[HTML]{000000} $22.63$} & \cc{5C99C7}{\color[HTML]{F1F1F1} $22.17$} & \cc{FFFFFE}{\color[HTML]{000000} $0.848$} & \cc{B6D8E8}{\color[HTML]{000000} $0.833$} & \cc{66A0CB}{\color[HTML]{F1F1F1} $0.823$} & \cc{FEF3ED}{\color[HTML]{000000} $0.215$} & \cc{C6E0ED}{\color[HTML]{000000} $0.241$} & \cc{5C99C7}{\color[HTML]{F1F1F1} $0.267$} \\
RevDGS & \cc{B5D7E8}{\color[HTML]{000000} $22.66$} & \cc{C6E0ED}{\color[HTML]{000000} $22.80$} & \cc{B1D5E7}{\color[HTML]{000000} $22.63$} & \cc{9FC9E0}{\color[HTML]{000000} $0.830$} & \cc{93C0DC}{\color[HTML]{000000} $0.828$} & \cc{5C99C7}{\color[HTML]{F1F1F1} $0.821$} & \cc{89B9D8}{\color[HTML]{000000} $0.257$} & \cc{A1CAE1}{\color[HTML]{000000} $0.252$} & \cc{7CAFD3}{\color[HTML]{F1F1F1} $0.260$} \\
\bottomrule
\end{tabular}}
\endgroup
}
\end{subtable}
\end{table}

\section{DA3 Evaluation on ScanNet++}\label{da3_scannet_scene_selection}
As we saw DA3 perform significantly better on ScanNet++ than on MipNerf360 and Tanks\&Temples, we decided to establish whether its performance on ScanNet++ could be affected by overlap with scenes from the training set. We have found the list of DA3 \textit{test} scenes specified by the authors in a GitHub issue,\footnote{https://github.com/ByteDance-Seed/Depth-Anything-3/issues/29} and have verified that only the following 8 of the 15 scenes we use for evaluation are not on that list (and could thus potentially be in the DA3 training set): ``b0a08200c9'', ``6115eddb86'', ``3f15a9266d'', ``5eb31827b7'', ``3db0a1c8f3'', ``e7af285f7d'', ``5748ce6f01'', ``825d228aec''. In this section, we use $S_\text{full}$ to denote the set of all 15 scenes, $S_\text{test}$ to denote the 7 scenes that are confirmed to be part of the DA3 test set, and $S_\text{rest}$ to denote the other 8 scenes, for which we are unable to establish overlap.

As ScanNet++ has more than 1000 scenes in total, and DA3 only used 230 ScanNet++ scenes for training~\cite{depthanything3}, we estimate the probability of overlap with our 8 scenes to be low.
As the list of ScanNet++ scenes used for DA3 training is not public (we did not receive a response to our inquiry regarding this before this paper was finalized), we have decided to compare NVS and geometric performance on the 15 scenes used in our evaluation and on the 7 scenes that are confirmed to be part of the DA3 test set.

First, we compare DA3 and $\text{DA3}^\text{GS}$ NVS performance on $S_\text{full}$, $S_\text{test}$, and $S_\text{rest}$. As there could be inherent scene difficulty differences between $S_\text{test}$ and $S_\text{rest}$, we compare differences relative to the SfM metrics instead of the raw values. As can be seen in \cref{tab:da3_scene_selection_ablation}, the performance is actually slightly better on $S_\text{test}$. Aggregating over both datasets, all scenes, and all densification strategies, the mean difference in deltas over SfM between $S_\text{test}$ and $S_\text{rest}$ was +0.23 PSNR, 0 SSIM and -0.001 LPIPS for DA3, and +0.02 PSNR, -0.001 SSIM and -0.002 LPIPS for $\text{DA3}^\text{GS}$.

In \cref{tab:final_recon_fscore_da3_test_scenes}, we provide results for our geometric evaluation on ScanNet++, limited to $S_\text{test}$. Comparing the numbers to \Cref{tab:final_recon_fscore} in the main paper, we see that the results are generally better than what we see on $S_\text{full}$. For example, the F1 score for DA3 improved by 10\% at initialization time, and by 3\% to 6\% post-training. Furthermore, for the geometry evaluation we can also look at the ETH3D results (see \cref{excluded_eth3d_results} and \cref{tab:final_recon_fscore_eth3d}), where the performance dynamics between DA3 and the other methods are similar to our ScanNet++ results.

Based on this analysis, we conclude that it is unlikely that our results using DA3 on ScanNet++ are skewed by overlap with the DA3 training set.
Thus, in the paper, we present the results using all 15 scenes ($S_\text{full}$).
It should be noted that DA3 performance on ScanNet++ is likely still better than on out-of-distribution data,
as the dataset is relatively homogeneous in terms of scene content.

\begin{table}[tbp]
\centering
\caption{Improvement over SfM initialization for DA3 and $\text{DA3}^\text{G.S.}$ when evaluated over all ScanNet++ scenes versus only the scenes in the DA3 test set. Values are the mean and standard deviations of deltas over the SfM baseline across strategies.}
\label{tab:da3_scene_selection_ablation}
\begin{subtable}[t]{\linewidth}
\centering
\caption{ScanNet++}
\label{tab:da3_scene_selection_scannetpp}
{\small
\begingroup \setlength{\tabcolsep}{1.0\tabcolsep}
\resizebox{\linewidth}{!}{\begin{tabular}{l|ccc|ccc|ccc}
\toprule
& \multicolumn{3}{c|}{\textbf{All Scenes ($S_\text{full}$)}} & \multicolumn{3}{c|}{\textbf{DA3 Test Scenes ($S_\text{test}$)}} & \multicolumn{3}{c}{\textbf{Unconfirmed Scenes}} \\
\textbf{Init} & \textbf{$\Delta$PSNR $\uparrow$} & \textbf{$\Delta$SSIM $\uparrow$} & \textbf{$\Delta$LPIPS $\downarrow$} & \textbf{$\Delta$PSNR $\uparrow$} & \textbf{$\Delta$SSIM $\uparrow$} & \textbf{$\Delta$LPIPS $\downarrow$} & \textbf{$\Delta$PSNR $\uparrow$} & \textbf{$\Delta$SSIM $\uparrow$} & \textbf{$\Delta$LPIPS $\downarrow$} \\
\midrule
DA3 & \cc{F8C0A4}{\color[HTML]{000000} $0.50 \pm 0.11$} & \cc{D5675E}{\color[HTML]{F1F1F1} $0.003 \pm 0.001$} & \cc{FBD9C8}{\color[HTML]{000000} $-0.005 \pm 0.003$} & \cc{EDA38D}{\color[HTML]{000000} $0.61 \pm 0.22$} & \cc{D5675E}{\color[HTML]{F1F1F1} $0.003 \pm 0.001$} & \cc{F9CAB2}{\color[HTML]{000000} $-0.007 \pm 0.003$} & \cc{FACDB7}{\color[HTML]{000000} $0.40 \pm 0.24$} & \cc{D5675E}{\color[HTML]{F1F1F1} $0.003 \pm 0.003$} & \cc{FDE9DF}{\color[HTML]{000000} $-0.003 \pm 0.005$} \\
$\text{DA3}^\text{G.S.}$ & \cc{D5675E}{\color[HTML]{F1F1F1} $0.84 \pm 0.16$} & \cc{FBDCCC}{\color[HTML]{000000} $0.001 \pm 0.002$} & \cc{D5675E}{\color[HTML]{F1F1F1} $-0.014 \pm 0.003$} & \cc{D5675E}{\color[HTML]{F1F1F1} $0.84 \pm 0.31$} & \cc{FFFFFE}{\color[HTML]{000000} $-0.000 \pm 0.001$} & \cc{D5675E}{\color[HTML]{F1F1F1} $-0.014 \pm 0.005$} & \cc{D5675E}{\color[HTML]{F1F1F1} $0.84 \pm 0.11$} & \cc{F2AF97}{\color[HTML]{000000} $0.002 \pm 0.003$} & \cc{DB7569}{\color[HTML]{F1F1F1} $-0.013 \pm 0.004$} \\
\bottomrule
\end{tabular}}
\endgroup
}
\end{subtable}
\begin{subtable}[t]{\linewidth}
\centering
\caption{ScanNet++ (On-Trajectory)}
\label{tab:da3_scene_selection_eval_on_train_set_scannetpp}
{\small
\begingroup \setlength{\tabcolsep}{1.0\tabcolsep}
\resizebox{\linewidth}{!}{\begin{tabular}{l|ccc|ccc|ccc}
\toprule
& \multicolumn{3}{c|}{\textbf{All Scenes ($S_\text{full}$)}} & \multicolumn{3}{c|}{\textbf{DA3 Test Scenes ($S_\text{test}$)}} & \multicolumn{3}{c}{\textbf{Unconfirmed Scenes}} \\
\textbf{Init} & \textbf{$\Delta$PSNR $\uparrow$} & \textbf{$\Delta$SSIM $\uparrow$} & \textbf{$\Delta$LPIPS $\downarrow$} & \textbf{$\Delta$PSNR $\uparrow$} & \textbf{$\Delta$SSIM $\uparrow$} & \textbf{$\Delta$LPIPS $\downarrow$} & \textbf{$\Delta$PSNR $\uparrow$} & \textbf{$\Delta$SSIM $\uparrow$} & \textbf{$\Delta$LPIPS $\downarrow$} \\
\midrule
DA3 & \cc{A7CEE3}{\color[HTML]{000000} $-0.20 \pm 0.11$} & \cc{FFFFFE}{\color[HTML]{000000} $-0.000 \pm 0.001$} & \cc{EEA58F}{\color[HTML]{000000} $-0.005 \pm 0.003$} & \cc{E6F1F7}{\color[HTML]{000000} $-0.06 \pm 0.11$} & \cc{5C99C7}{\color[HTML]{F1F1F1} $-0.001 \pm 0.001$} & \cc{F9C3A8}{\color[HTML]{000000} $-0.004 \pm 0.003$} & \cc{5C99C7}{\color[HTML]{F1F1F1} $-0.31 \pm 0.14$} & \cc{FFFFFE}{\color[HTML]{000000} $0.000 \pm 0.001$} & \cc{E18676}{\color[HTML]{F1F1F1} $-0.006 \pm 0.004$} \\
$\text{DA3}^\text{G.S.}$ & \cc{E6907F}{\color[HTML]{F1F1F1} $0.25 \pm 0.07$} & \cc{FFFFFE}{\color[HTML]{000000} $-0.000 \pm 0.001$} & \cc{EEA58F}{\color[HTML]{000000} $-0.005 \pm 0.002$} & \cc{E18475}{\color[HTML]{F1F1F1} $0.27 \pm 0.14$} & \cc{FFFFFE}{\color[HTML]{000000} $-0.000 \pm 0.001$} & \cc{D5675E}{\color[HTML]{F1F1F1} $-0.007 \pm 0.003$} & \cc{E99985}{\color[HTML]{000000} $0.24 \pm 0.03$} & \cc{FFFFFE}{\color[HTML]{000000} $-0.000 \pm 0.001$} & \cc{FAD2BE}{\color[HTML]{000000} $-0.003 \pm 0.002$} \\
\bottomrule
\end{tabular}}
\endgroup
}
\end{subtable}
\end{table}

\begin{table}[t]
\centering
\caption{Geometric evaluation on ScanNet++ on the subset of scenes contained in the DA3 test set.}
\label{tab:final_recon_fscore_da3_test_scenes}
\resizebox{\linewidth}{!}{\small
\begingroup \setlength{\tabcolsep}{2.5\tabcolsep}
\begin{tabular}{lccccccc}
\toprule
 \textbf{F / P / R (\%) ↑} & SfM & $\text{EDGS}^\text{*}$ & Monodepth & DA3 & $\text{DA3}^\text{GS}$ & $\text{Laser}^+$ & Laser \\
\midrule
Initialization & \cc{FCE2D5}{\color[HTML]{000000} $48 / 79 / 36$} & \cc{FBD6C4}{\color[HTML]{000000} $50 / 35 / 92$} & \cc{D66960}{\color[HTML]{F1F1F1} $62 / 46 / 96$} & \cc{D5675E}{\color[HTML]{F1F1F1} $62 / 47 / 96$} & \cc{E1EFF6}{\color[HTML]{000000} $38 / 29 / 57$} & \cc{FFFFFF}{\color[HTML]{F1F1F1} \color{black} } & \cc{FFFFFF}{\color[HTML]{F1F1F1} \color{black} } \\
\midrule
AbsGS & \cc{C7E1EE}{\color[HTML]{000000} $34 / 28 / 48$} & \cc{E7F2F8}{\color[HTML]{000000} $39 / 29 / 61$} & \cc{FBDAC9}{\color[HTML]{000000} $49 / 46 / 54$} & \cc{FBD9C8}{\color[HTML]{000000} $50 / 46 / 54$} & \cc{FACFBA}{\color[HTML]{000000} $51 / 42 / 72$} & \cc{DAEBF3}{\color[HTML]{000000} $37 / 31 / 50$} & \cc{FBD7C5}{\color[HTML]{000000} $50 / 47 / 54$} \\
INRIA & \cc{C4DFED}{\color[HTML]{000000} $34 / 29 / 44$} & \cc{E8F3F8}{\color[HTML]{000000} $39 / 30 / 61$} & \cc{FBD5C2}{\color[HTML]{000000} $50 / 48 / 53$} & \cc{FBD6C4}{\color[HTML]{000000} $50 / 48 / 53$} & \cc{FACCB5}{\color[HTML]{000000} $52 / 43 / 71$} & \cc{D3E7F2}{\color[HTML]{000000} $36 / 32 / 45$} & \cc{FBD5C2}{\color[HTML]{000000} $50 / 48 / 53$} \\
IDHFR & \cc{A7CEE3}{\color[HTML]{000000} $30 / 20 / 65$} & \cc{E0EEF5}{\color[HTML]{000000} $38 / 27 / 69$} & \cc{F0A992}{\color[HTML]{000000} $56 / 52 / 62$} & \cc{F4B49A}{\color[HTML]{000000} $55 / 51 / 61$} & \cc{FFFCFA}{\color[HTML]{000000} $43 / 32 / 74$} & \cc{D3E7F2}{\color[HTML]{000000} $36 / 26 / 66$} & \cc{F3B199}{\color[HTML]{000000} $55 / 51 / 61$} \\
MCMC & \cc{74AAD0}{\color[HTML]{F1F1F1} $25 / 17 / 59$} & \cc{B9D9E9}{\color[HTML]{000000} $32 / 22 / 62$} & \cc{FBDAC9}{\color[HTML]{000000} $49 / 44 / 56$} & \cc{FBDBCB}{\color[HTML]{000000} $49 / 44 / 56$} & \cc{BFDCEB}{\color[HTML]{000000} $33 / 23 / 64$} & \cc{5E9BC8}{\color[HTML]{F1F1F1} $23 / 15 / 57$} & \cc{FBDAC9}{\color[HTML]{000000} $49 / 44 / 57$} \\
RevDGS & \cc{E3F0F6}{\color[HTML]{000000} $38 / 31 / 57$} & \cc{EEF6FA}{\color[HTML]{000000} $40 / 31 / 62$} & \cc{F8C0A4}{\color[HTML]{000000} $54 / 50 / 59$} & \cc{F9C4AA}{\color[HTML]{000000} $54 / 49 / 59$} & \cc{FCE7DC}{\color[HTML]{000000} $47 / 37 / 70$} & \cc{F1F7FB}{\color[HTML]{000000} $40 / 33 / 58$} & \cc{F7BCA1}{\color[HTML]{000000} $55 / 51 / 59$} \\
No D. & \cc{5C99C7}{\color[HTML]{F1F1F1} $23 / 19 / 29$} & \cc{E6F1F7}{\color[HTML]{000000} $39 / 29 / 61$} & \cc{FAD0BB}{\color[HTML]{000000} $51 / 49 / 53$} & \cc{FAD0BB}{\color[HTML]{000000} $51 / 49 / 53$} & \cc{F9C3A8}{\color[HTML]{000000} $54 / 44 / 71$} & \cc{EAF3F8}{\color[HTML]{000000} $39 / 36 / 44$} & \cc{FACFBA}{\color[HTML]{000000} $51 / 50 / 53$} \\
\bottomrule
\end{tabular}
\endgroup
}
\end{table}

\section{Ablations on Adjustments to Evaluated Methods}\label{hyperparam_changes}

\PAR{IDHFR Without Means LR Adjustment.} For IDHFR~\cite{idhfr}, we report results without the adjusted $\mu_i$  learning rate (LR) schedule used by the authors (specified in the supplementary material of~\cite{idhfr}). This is to provide a direct comparison with the other densification strategies. As shown in~\cref{tab:idhfr_means_lr}, using the adjusted LR schedule results in small differences in metric values, which would not affect our overall analysis.

\begin{table}[tbp]
    \captionsetup{skip=6pt}
    \caption{Mean metric values achieved by IDHFR with and without the adjusted $\mu_i$ LR schedule. These results are with SfM initialization. Columns marked with ``$^*$'' represent results \textit{with} the adjusted LR schedule used by the authors, but not in our experiments.}
    \label{tab:idhfr_means_lr}
    
    \centering
    \begin{tabular}{@{}lllllll@{}}
    \toprule
    Dataset & PSNR & $\text{PSNR}^* \uparrow$ & SSIM & $\text{SSIM}^* \uparrow$ & LPIPS & $\text{LPIPS}^* \downarrow$ \\
    \midrule
    ScanNet++ (default split) & \textbf{23.00} & 22.98 & 0.877 & 0.877 & 0.233 & 0.233 \\
    ScanNet++ (on-trajectory) & \textbf{33.80} & 33.74 & 0.953 & 0.953 & \textbf{0.095} & 0.096 \\
    ETH3D                     & \textbf{22.61} & 22.21 & \textbf{0.820} & 0.814 & \textbf{0.266} & 0.287 \\
    MipNerf360                & 27.85 & \textbf{27.96} & 0.829 & \textbf{0.834} & 0.134 & \textbf{0.133} \\
    Tanks \& Temples          & 23.78 & \textbf{23.84} & 0.842 & \textbf{0.844} & 0.141 & \textbf{0.139} \\
    \bottomrule
    \end{tabular}
\end{table}

\PAR{$\text{EDGS}^*$ Scale Increase.} As mentioned in the paper, we increase the scale of Gaussians produced by EDGS~\cite{edgs} when subsampling its output to fit a target initialization size. Specifically, given the original number of Gaussians produced by $\text{EDGS}^*$ $|\mathcal{G}_\text{init}^{\text{EDGS}^*}|$, and the target initialization size $N$, we scale the splats' extents by
\begin{equation}
    \frac{|\mathcal{G}_\text{init}^{\text{EDGS}^*}|}{N} \space.
\end{equation}
This is done to reduce the gaps between Gaussians left by the removed primitives. As can be seen in \cref{tab:edgs_scale_increase}, this adjustment slightly improves average metric values on all datasets except ScanNet++ with the on-trajectory split, where the only change is a very minor (-0.01) drop in PSNR.

\begin{table}[tbp]
    \captionsetup{skip=6pt}
    \caption{Ablation on increasing scales of Gaussians when subsampling $\text{EDGS}^*$ initialization. In rows, we report means across results with AbsGS, MCMC, and IDHFR, for a given dataset. Columns with ``$^*$'' represent results \textit{with} scale increase.}
    \label{tab:edgs_scale_increase}
    \centering
    \begin{tabular}{@{}lllllll@{}}
    \toprule
    Dataset & PSNR & $\text{PSNR}^* \uparrow$ & SSIM & $\text{SSIM}^* \uparrow$ & LPIPS & $\text{LPIPS}^* \downarrow$ \\
    \midrule
    ScanNet++ (default split) & 22.74 & \textbf{22.78} & 0.869 & \textbf{0.870} & 0.246 & \textbf{0.245} \\
    ScanNet++ (on-trajectory) & \textbf{33.25} & 33.24 & 0.951 & 0.951 & 0.094 & 0.094 \\
    MipNerf360 & 27.57 & \textbf{27.59} & 0.825 & \textbf{0.826} & 0.132 & 0.132 \\
    Tanks \& Temples & 23.55 & 23.55 & 0.838 & \textbf{0.839} & 0.139 & 0.139 \\
    \bottomrule
    \end{tabular}
\end{table}

\PAR{DA3 Floater Removal.}
As specified in~\cref{sec:protocol:practical_init}, we apply the same floater removal process we use for Monodepth (see~\cref{monodepth_details}) to Depth Anything 3 (when not using the GS head).
This makes comparisons with Monodepth more direct, and slightly improves results on MipNerf360 and Tanks \& Temples, as reported in~\cref{tab:da3_floater_removal_ablation}.

\begin{table}[tbp]
    \captionsetup{skip=6pt}
    \caption{Ablation on applying floater removal to DA3. In rows, we report means across results with all densification strategies (and without densification), for a given dataset. Columns with ``$^*$'' represent results \textit{with} floater removal.}
    \label{tab:da3_floater_removal_ablation}
    \centering
    \begin{tabular}{@{}lllllll@{}}
    \toprule
    Dataset & PSNR & $\text{PSNR}^* \uparrow$ & SSIM & $\text{SSIM}^* \uparrow$ & LPIPS & $\text{LPIPS}^* \downarrow$ \\
    \midrule
    ScanNet++ (default split) & 22.86 & 22.86 & 0.873 & 0.873 & 0.245 & 0.245 \\
    ScanNet++ (on-trajectory) & 32.82 & 32.82 & 0.949 & 0.949 & 0.100 & 0.100 \\
    MipNerf360 & 26.20 & \textbf{26.23} & 0.796 & \textbf{0.797} & 0.170 & \textbf{0.168} \\
    Tanks \& Temples & 23.81 & \textbf{23.94} & 0.818 & \textbf{0.820} & 0.175 & \textbf{0.173} \\
    \bottomrule
    \end{tabular}
\end{table}

\PAR{EDGS without full SH init.} As mentioned in the paper, the public EDGS~\cite{edgs} code does not currently include the higher-order SH
coefficient initialization described in the paper.\footnote{Higher-order SH coefficients are established based on the colors of the two keypoints from which the given Gaussian's world space position was triangulated.}
As it is claimed as one of the method's contributions, we reached out to the authors, who graciously provided code snippets for the core parts of the feature, which we integrated into our version.
Unfortunately, we were not able to replicate their results, and actually saw decreased performance using this approach, which is why we report results without full SH init in the paper. While we made all efforts to verify our implementation and find any potential issues causing this discrepancy, we were unfortunately unable to find any specific bugs, and, as current public code does not contain this feature, we are unable to test against a known good implementation.
\Cref{tab:edgs_full_sh} provides a comparison of our results with and without full SH init. 

\begin{table}[tbp]
\centering
\caption{Results for EDGS without full SH init, and with full SH init, which we implemented based on code snippets from the authors. $\text{EDGS}^*$ is our version \textit{without} full SH init.}
\label{tab:edgs_full_sh}
\begin{subtable}[t]{0.49\linewidth}
\centering
\caption{ScanNet++}
{\small
\begingroup \setlength{\tabcolsep}{2.5\tabcolsep}
\resizebox{\linewidth}{!}{\begin{tabular}{l|cc|cc|cc}
\toprule
& \multicolumn{2}{c|}{\textbf{PSNR ↑}} & \multicolumn{2}{c|}{\textbf{SSIM ↑}} & \multicolumn{2}{c}{\textbf{LPIPS ↓}} \\
& $\text{EDGS}^*$ & $\text{EDGS}$ & $\text{EDGS}^*$ & $\text{EDGS}$ & $\text{EDGS}^*$ & $\text{EDGS}$ \\
\midrule
AbsGS & \cc{EA9B87}{\color[HTML]{000000} $22.64$} & \cc{A1CAE1}{\color[HTML]{000000} $20.13$} & \cc{F9CAB2}{\color[HTML]{000000} $0.863$} & \cc{C5DFED}{\color[HTML]{000000} $0.842$} & \cc{EEA58F}{\color[HTML]{000000} $0.254$} & \cc{BAD9EA}{\color[HTML]{000000} $0.314$} \\
INRIA & \cc{EB9F8A}{\color[HTML]{000000} $22.62$} & \cc{ADD2E6}{\color[HTML]{000000} $20.23$} & \cc{FACBB4}{\color[HTML]{000000} $0.863$} & \cc{CCE3EF}{\color[HTML]{000000} $0.843$} & \cc{F0AB94}{\color[HTML]{000000} $0.255$} & \cc{C5DFED}{\color[HTML]{000000} $0.310$} \\
MCMC & \cc{EA9B87}{\color[HTML]{000000} $22.64$} & \cc{D7E9F3}{\color[HTML]{000000} $20.78$} & \cc{D5675E}{\color[HTML]{F1F1F1} $0.875$} & \cc{FCE3D6}{\color[HTML]{000000} $0.858$} & \cc{D5675E}{\color[HTML]{F1F1F1} $0.240$} & \cc{FCE4D8}{\color[HTML]{000000} $0.276$} \\
IDHFR & \cc{D5675E}{\color[HTML]{F1F1F1} $23.06$} & \cc{EEF6FA}{\color[HTML]{000000} $21.09$} & \cc{DF8072}{\color[HTML]{F1F1F1} $0.872$} & \cc{FDEFE7}{\color[HTML]{000000} $0.856$} & \cc{D86F65}{\color[HTML]{F1F1F1} $0.242$} & \cc{FFFFFE}{\color[HTML]{000000} $0.288$} \\
RevDGS & \cc{F9C1A6}{\color[HTML]{000000} $22.34$} & \cc{5C99C7}{\color[HTML]{F1F1F1} $19.56$} & \cc{FBD6C4}{\color[HTML]{000000} $0.861$} & \cc{5C99C7}{\color[HTML]{F1F1F1} $0.829$} & \cc{F9C6AD}{\color[HTML]{000000} $0.262$} & \cc{5C99C7}{\color[HTML]{F1F1F1} $0.336$} \\
No D. & \cc{F5B89E}{\color[HTML]{000000} $22.41$} & \cc{9BC6DF}{\color[HTML]{000000} $20.09$} & \cc{FCDECF}{\color[HTML]{000000} $0.859$} & \cc{A1CAE1}{\color[HTML]{000000} $0.837$} & \cc{FACDB7}{\color[HTML]{000000} $0.266$} & \cc{8FBDDA}{\color[HTML]{000000} $0.324$} \\
\bottomrule
\end{tabular}}
\endgroup
}
\end{subtable}
\begin{subtable}[t]{0.49\linewidth}
\centering
\caption{ScanNet++ (On-Trajectory)}
{\small
\begingroup \setlength{\tabcolsep}{2.5\tabcolsep}
\resizebox{\linewidth}{!}{\begin{tabular}{l|cc|cc|cc}
\toprule
& \multicolumn{2}{c|}{\textbf{PSNR ↑}} & \multicolumn{2}{c|}{\textbf{SSIM ↑}} & \multicolumn{2}{c}{\textbf{LPIPS ↓}} \\
& $\text{EDGS}^*$ & $\text{EDGS}$ & $\text{EDGS}^*$ & $\text{EDGS}$ & $\text{EDGS}^*$ & $\text{EDGS}$ \\
\midrule
AbsGS & \cc{FBDBCB}{\color[HTML]{000000} $32.88$} & \cc{F3F9FB}{\color[HTML]{000000} $32.29$} & \cc{FACDB7}{\color[HTML]{000000} $0.949$} & \cc{FCE7DC}{\color[HTML]{000000} $0.947$} & \cc{EB9F8A}{\color[HTML]{000000} $0.096$} & \cc{FDE9DF}{\color[HTML]{000000} $0.107$} \\
INRIA & \cc{FCE1D3}{\color[HTML]{000000} $32.80$} & \cc{F3F9FB}{\color[HTML]{000000} $32.29$} & \cc{FBD6C4}{\color[HTML]{000000} $0.948$} & \cc{FDEBE2}{\color[HTML]{000000} $0.947$} & \cc{F2AF97}{\color[HTML]{000000} $0.097$} & \cc{FDEBE2}{\color[HTML]{000000} $0.107$} \\
MCMC & \cc{F9CAB2}{\color[HTML]{000000} $33.09$} & \cc{FCE5D9}{\color[HTML]{000000} $32.75$} & \cc{E18676}{\color[HTML]{F1F1F1} $0.952$} & \cc{ECA18C}{\color[HTML]{000000} $0.951$} & \cc{ECA18C}{\color[HTML]{000000} $0.096$} & \cc{FACDB7}{\color[HTML]{000000} $0.101$} \\
IDHFR & \cc{D5675E}{\color[HTML]{F1F1F1} $33.77$} & \cc{E28878}{\color[HTML]{F1F1F1} $33.56$} & \cc{D5675E}{\color[HTML]{F1F1F1} $0.953$} & \cc{E6907F}{\color[HTML]{F1F1F1} $0.951$} & \cc{D5675E}{\color[HTML]{F1F1F1} $0.091$} & \cc{F4B49A}{\color[HTML]{000000} $0.098$} \\
RevDGS & \cc{FEF7F3}{\color[HTML]{000000} $32.52$} & \cc{66A0CB}{\color[HTML]{F1F1F1} $31.12$} & \cc{FDE9DF}{\color[HTML]{000000} $0.947$} & \cc{5C99C7}{\color[HTML]{F1F1F1} $0.938$} & \cc{F9C8B0}{\color[HTML]{000000} $0.101$} & \cc{5C99C7}{\color[HTML]{F1F1F1} $0.131$} \\
No D. & \cc{E3F0F6}{\color[HTML]{000000} $32.12$} & \cc{5C99C7}{\color[HTML]{F1F1F1} $31.05$} & \cc{F5F9FC}{\color[HTML]{000000} $0.945$} & \cc{83B5D6}{\color[HTML]{000000} $0.939$} & \cc{FBDDCE}{\color[HTML]{000000} $0.104$} & \cc{ADD2E6}{\color[HTML]{000000} $0.123$} \\
\bottomrule
\end{tabular}}
\endgroup
}
\end{subtable}
\begin{subtable}[t]{0.49\linewidth}
\centering
\caption{Mip-NeRF 360}
{\small
\begingroup \setlength{\tabcolsep}{2.5\tabcolsep}
\resizebox{\linewidth}{!}{\begin{tabular}{l|cc|cc|cc}
\toprule
& \multicolumn{2}{c|}{\textbf{PSNR ↑}} & \multicolumn{2}{c|}{\textbf{SSIM ↑}} & \multicolumn{2}{c}{\textbf{LPIPS ↓}} \\
& $\text{EDGS}^*$ & $\text{EDGS}$ & $\text{EDGS}^*$ & $\text{EDGS}$ & $\text{EDGS}^*$ & $\text{EDGS}$ \\
\midrule
AbsGS & \cc{FEFAF7}{\color[HTML]{000000} $27.39$} & \cc{A3CBE2}{\color[HTML]{000000} $27.13$} & \cc{FACDB7}{\color[HTML]{000000} $0.824$} & \cc{F0F7FA}{\color[HTML]{000000} $0.819$} & \cc{E38A7A}{\color[HTML]{F1F1F1} $0.134$} & \cc{FBDCCC}{\color[HTML]{000000} $0.144$} \\
INRIA & \cc{FCE4D8}{\color[HTML]{000000} $27.46$} & \cc{DFEDF5}{\color[HTML]{000000} $27.28$} & \cc{F1AD96}{\color[HTML]{000000} $0.826$} & \cc{FDEAE0}{\color[HTML]{000000} $0.822$} & \cc{F0AB94}{\color[HTML]{000000} $0.137$} & \cc{FCE7DC}{\color[HTML]{000000} $0.146$} \\
MCMC & \cc{D5675E}{\color[HTML]{F1F1F1} $27.73$} & \cc{DB7569}{\color[HTML]{F1F1F1} $27.71$} & \cc{E18676}{\color[HTML]{F1F1F1} $0.827$} & \cc{F9CAB2}{\color[HTML]{000000} $0.824$} & \cc{D76B61}{\color[HTML]{F1F1F1} $0.131$} & \cc{E69280}{\color[HTML]{F1F1F1} $0.135$} \\
IDHFR & \cc{DD7C6E}{\color[HTML]{F1F1F1} $27.70$} & \cc{D97166}{\color[HTML]{F1F1F1} $27.72$} & \cc{D5675E}{\color[HTML]{F1F1F1} $0.828$} & \cc{F8BEA3}{\color[HTML]{000000} $0.825$} & \cc{D66960}{\color[HTML]{F1F1F1} $0.131$} & \cc{EDA38D}{\color[HTML]{000000} $0.136$} \\
RevDGS & \cc{FDEBE2}{\color[HTML]{000000} $27.44$} & \cc{D1E6F1}{\color[HTML]{000000} $27.24$} & \cc{FBD6C4}{\color[HTML]{000000} $0.823$} & \cc{DAEBF3}{\color[HTML]{000000} $0.817$} & \cc{D5675E}{\color[HTML]{F1F1F1} $0.131$} & \cc{FACEB8}{\color[HTML]{000000} $0.141$} \\
No D. & \cc{85B6D7}{\color[HTML]{000000} $27.08$} & \cc{5C99C7}{\color[HTML]{F1F1F1} $27.01$} & \cc{BAD9EA}{\color[HTML]{000000} $0.815$} & \cc{5C99C7}{\color[HTML]{F1F1F1} $0.811$} & \cc{BCDBEA}{\color[HTML]{000000} $0.161$} & \cc{5C99C7}{\color[HTML]{F1F1F1} $0.170$} \\
\bottomrule
\end{tabular}}
\endgroup
}
\end{subtable}
\begin{subtable}[t]{0.49\linewidth}
\centering
\caption{Tanks and Temples}
{\small
\begingroup \setlength{\tabcolsep}{2.5\tabcolsep}
\resizebox{\linewidth}{!}{\begin{tabular}{l|cc|cc|cc}
\toprule
& \multicolumn{2}{c|}{\textbf{PSNR ↑}} & \multicolumn{2}{c|}{\textbf{SSIM ↑}} & \multicolumn{2}{c}{\textbf{LPIPS ↓}} \\
& $\text{EDGS}^*$ & $\text{EDGS}$ & $\text{EDGS}^*$ & $\text{EDGS}$ & $\text{EDGS}^*$ & $\text{EDGS}$ \\
\midrule
AbsGS & \cc{D7E9F3}{\color[HTML]{000000} $23.09$} & \cc{5C99C7}{\color[HTML]{F1F1F1} $22.57$} & \cc{FDEAE0}{\color[HTML]{000000} $0.833$} & \cc{A5CDE3}{\color[HTML]{000000} $0.819$} & \cc{F1AD96}{\color[HTML]{000000} $0.145$} & \cc{D0E5F0}{\color[HTML]{000000} $0.177$} \\
INRIA & \cc{FEF7F3}{\color[HTML]{000000} $23.37$} & \cc{F5F9FC}{\color[HTML]{000000} $23.26$} & \cc{FBD6C4}{\color[HTML]{000000} $0.836$} & \cc{FDEEE6}{\color[HTML]{000000} $0.832$} & \cc{E79482}{\color[HTML]{F1F1F1} $0.142$} & \cc{FBD9C8}{\color[HTML]{000000} $0.155$} \\
MCMC & \cc{E18676}{\color[HTML]{F1F1F1} $23.96$} & \cc{D5675E}{\color[HTML]{F1F1F1} $24.06$} & \cc{D5675E}{\color[HTML]{F1F1F1} $0.846$} & \cc{D86D63}{\color[HTML]{F1F1F1} $0.846$} & \cc{D5675E}{\color[HTML]{F1F1F1} $0.135$} & \cc{E58E7D}{\color[HTML]{F1F1F1} $0.141$} \\
IDHFR & \cc{FACDB7}{\color[HTML]{000000} $23.66$} & \cc{FAD3BF}{\color[HTML]{000000} $23.63$} & \cc{F0A992}{\color[HTML]{000000} $0.841$} & \cc{F9C2A7}{\color[HTML]{000000} $0.839$} & \cc{DB7569}{\color[HTML]{F1F1F1} $0.137$} & \cc{EFA791}{\color[HTML]{000000} $0.144$} \\
RevDGS & \cc{C6E0ED}{\color[HTML]{000000} $22.99$} & \cc{609CC8}{\color[HTML]{F1F1F1} $22.59$} & \cc{E2EFF6}{\color[HTML]{000000} $0.826$} & \cc{6EA6CE}{\color[HTML]{F1F1F1} $0.814$} & \cc{FFFBF9}{\color[HTML]{000000} $0.165$} & \cc{85B6D7}{\color[HTML]{000000} $0.190$} \\
No D. & \cc{C1DDEC}{\color[HTML]{000000} $22.96$} & \cc{7AAED2}{\color[HTML]{F1F1F1} $22.67$} & \cc{C7E1EE}{\color[HTML]{000000} $0.822$} & \cc{5C99C7}{\color[HTML]{F1F1F1} $0.813$} & \cc{ECF5F9}{\color[HTML]{000000} $0.170$} & \cc{5C99C7}{\color[HTML]{F1F1F1} $0.196$} \\
\bottomrule
\end{tabular}}
\endgroup
}
\end{subtable}
\end{table}

\section{Full NVS Results and Geometry Accuracy on ETH3D}\label{excluded_eth3d_results}
As we were unable to include some ETH3D results in the paper due to length constraints, we provide both the data and analysis in this section instead.

\PAR{Noise resiliency.}
\Cref{tab:noise_resiliency_eth3d_only} shows the results of our noise resiliency experiment (\cref{par:noise_resiliency}) with ETH3D data. On this under-constrained dataset with relatively sparse input view coverage, initialization matters slightly more than even on the off-trajectory version of ScanNet++. We see small but noticeable degradation even at $\sigma=0.01S$, whereas at 
$\sigma=0.1S$, the metrics drop sharply, as noise dominates over structure, and photometric optimization struggles to recover scene geometry under limited supervision.
\begin{table}[tbp]
\captionsetup{skip=6pt}
\centering
\caption{Noise resiliency on ETH3D with non-hybrid laser scan init with $0.5G_m$ initial points.}
\label{tab:noise_resiliency_eth3d_only}

\resizebox{0.8\textwidth}{!}{
\setlength{\tabcolsep}{4pt}
\begin{tabular}{l|ccc|ccc|ccc}
\toprule
& \multicolumn{3}{c|}{\textbf{PSNR ↑}} & \multicolumn{3}{c|}{\textbf{SSIM ↑}} & \multicolumn{3}{c}{\textbf{LPIPS ↓}} \\
Noise std.& 0\% & 1\% & 10\% & 0\% & 1\% & 10\% & 0\% & 1\% & 10\% \\
\midrule
AbsGS & \cc{FAD3BF}{\color[HTML]{000000} $22.61$} & \cc{FBD8C7}{\color[HTML]{000000} $22.46$} & \cc{ADD2E6}{\color[HTML]{000000} $19.50$} & \cc{F9C3A8}{\color[HTML]{000000} $0.836$} & \cc{F9C6AD}{\color[HTML]{000000} $0.834$} & \cc{B9D9E9}{\color[HTML]{000000} $0.745$} & \cc{F9C1A6}{\color[HTML]{000000} $0.235$} & \cc{F9C9B1}{\color[HTML]{000000} $0.246$} & \cc{B6D8E8}{\color[HTML]{000000} $0.400$} \\
INRIA & \cc{F9C7AE}{\color[HTML]{000000} $22.94$} & \cc{F9C7AE}{\color[HTML]{000000} $22.95$} & \cc{E0EEF5}{\color[HTML]{000000} $20.61$} & \cc{F4B49A}{\color[HTML]{000000} $0.843$} & \cc{F2AF97}{\color[HTML]{000000} $0.845$} & \cc{FEF8F4}{\color[HTML]{000000} $0.795$} & \cc{F2AF97}{\color[HTML]{000000} $0.223$} & \cc{F5B89E}{\color[HTML]{000000} $0.229$} & \cc{F3F9FB}{\color[HTML]{000000} $0.332$} \\
MCMC & \cc{D5675E}{\color[HTML]{F1F1F1} $24.36$} & \cc{DD7C6E}{\color[HTML]{F1F1F1} $24.06$} & \cc{FBD8C7}{\color[HTML]{000000} $22.46$} & \cc{D5675E}{\color[HTML]{F1F1F1} $0.872$} & \cc{DF8072}{\color[HTML]{F1F1F1} $0.862$} & \cc{FEF2EC}{\color[HTML]{000000} $0.800$} & \cc{D5675E}{\color[HTML]{F1F1F1} $0.175$} & \cc{DE7E70}{\color[HTML]{F1F1F1} $0.190$} & \cc{FCE1D3}{\color[HTML]{000000} $0.278$} \\
IDHFR & \cc{FAD2BE}{\color[HTML]{000000} $22.63$} & \cc{FBD5C2}{\color[HTML]{000000} $22.53$} & \cc{E7F2F8}{\color[HTML]{000000} $20.79$} & \cc{F9C8B0}{\color[HTML]{000000} $0.833$} & \cc{FACCB5}{\color[HTML]{000000} $0.830$} & \cc{F1F7FB}{\color[HTML]{000000} $0.781$} & \cc{F9C5AB}{\color[HTML]{000000} $0.241$} & \cc{FACCB5}{\color[HTML]{000000} $0.250$} & \cc{F1F7FB}{\color[HTML]{000000} $0.335$} \\
RevDGS & \cc{FACCB5}{\color[HTML]{000000} $22.80$} & \cc{FACCB5}{\color[HTML]{000000} $22.81$} & \cc{5C99C7}{\color[HTML]{F1F1F1} $18.32$} & \cc{FACEB8}{\color[HTML]{000000} $0.828$} & \cc{FBDBCB}{\color[HTML]{000000} $0.818$} & \cc{5C99C7}{\color[HTML]{F1F1F1} $0.708$} & \cc{FACEB8}{\color[HTML]{000000} $0.252$} & \cc{FBDAC9}{\color[HTML]{000000} $0.268$} & \cc{5C99C7}{\color[HTML]{F1F1F1} $0.464$} \\
\bottomrule
\end{tabular}
}
\end{table}

\begin{figure}[tbp]
    \centering
    
    \includegraphics[width=0.8\linewidth]{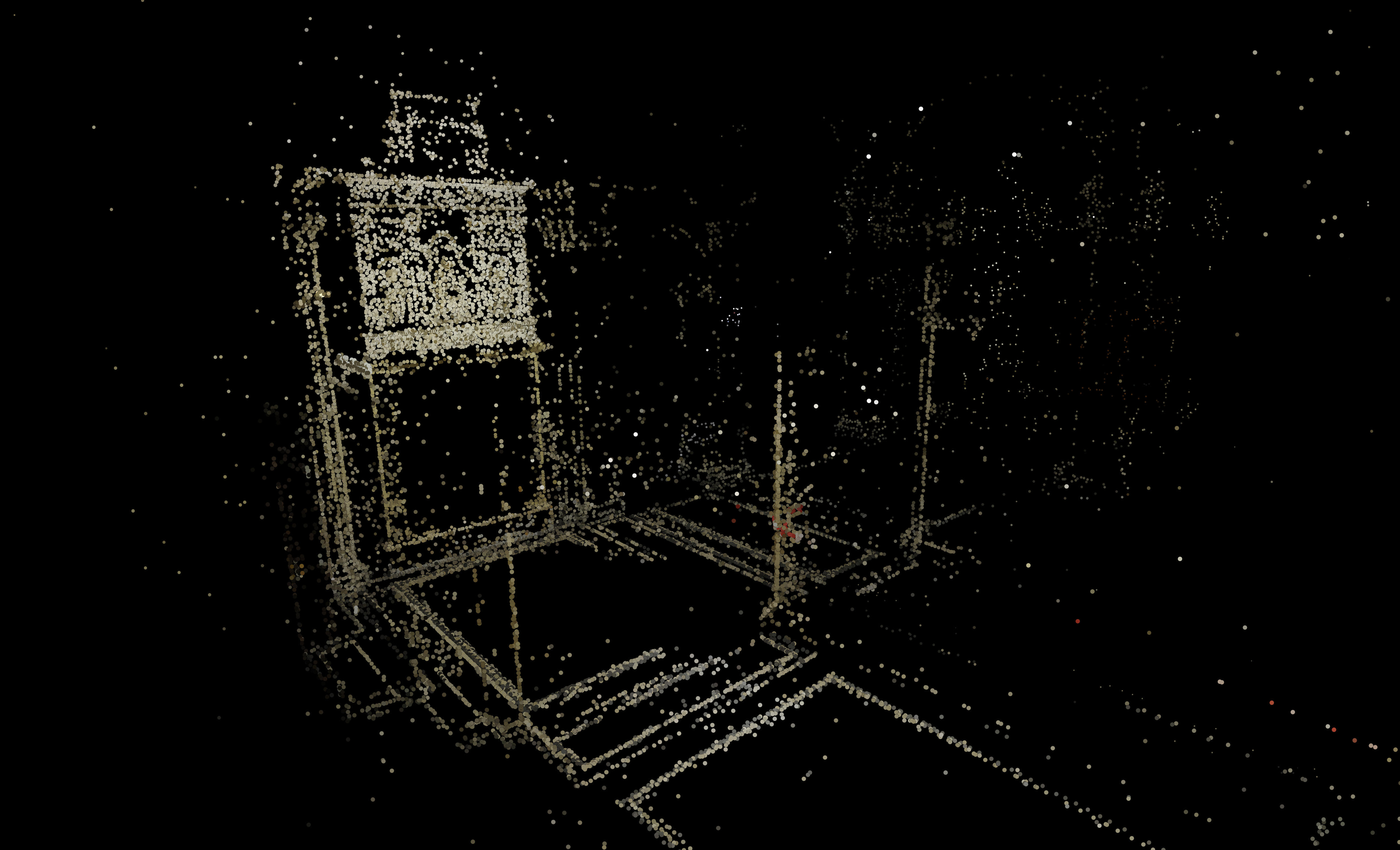} 
    
    \caption{SfM point cloud on the ``relief\_2'' scene from the ETH3D dataset. There are very few points on surfaces without pronounced textures, and there is significant noise in the reconstruction in the form of points placed in space that should be empty.}
    \label{fig:eth3d_init_sfm_relief_2}
\end{figure}

\PAR{NVS with all initialization methods.}
\Cref{tab:practical_eth3d} supplements the NVS results presented in~\cref{tab:practical_main} in the paper.
On ETH3D, using dense initialization with any of the evaluated methods generally improves NVS metrics compared to the SfM baseline, even more so than on off-trajectory ScanNet++.
However, the number of statistically significant improvements over SfM is very low here compared to our other results, in part due to the low number of scenes limiting the statistical power of the test, but also because performance is truly not as consistent here as on the other datasets (see \cref{per_scene_data_analysis}).
While on some scenes, the improvements are much larger than what we see on other datasets (can be attributed to both the low number of training views and the sparsity of ETH3D's SfM point clouds, see \cref{fig:eth3d_init_sfm_relief_2}), other scenes see significant performance degradation.
Overall this increased variance across scenes is likely caused by the low number of input views, as the geometric precision of the initializations is not significantly below what we see on ScanNet++ (see next paragraph and \cref{tab:final_recon_fscore_eth3d}).

\begin{table}[t]
\captionsetup{skip=6pt}
\centering
\caption{NVS performance of practical initialization methods on ETH3D. Cells with statistically significant improvements over SfM are marked with ``$^{*}$'' (see \cref{statistical_analysis} for methodology).}
\label{tab:practical_eth3d}
{\small
\begingroup \setlength{\tabcolsep}{2.5\tabcolsep}
\resizebox{\linewidth}{!}{\begin{tabular}{l|cccccc|cccccc|cccccc}
\toprule
& \multicolumn{6}{c|}{\textbf{PSNR ↑}} & \multicolumn{6}{c|}{\textbf{SSIM ↑}} & \multicolumn{6}{c}{\textbf{LPIPS ↓}} \\
& SfM & $\text{EDGS}^*$ & M. D. & DA3 & $\text{DA3}^\text{G.S.}$ & Laser & SfM & $\text{EDGS}^*$ & M. D. & DA3 & $\text{DA3}^\text{G.S.}$ & Laser & SfM & $\text{EDGS}^*$ & M. D. & DA3 & $\text{DA3}^\text{G.S.}$ & Laser \\
\midrule
AbsGS & \cc{C7E1EE}{\color[HTML]{000000} $21.06$} & \cc{FFFCFA}{\color[HTML]{000000} $22.26$\rlap{\textsuperscript{*}}} & \cc{FEF4EF}{\color[HTML]{000000} $22.44$\rlap{\textsuperscript{*}}} & \cc{FEFAF7}{\color[HTML]{000000} $22.30$\rlap{\textsuperscript{*}}} & \cc{FBD5C2}{\color[HTML]{000000} $23.19$\rlap{\textsuperscript{*}}} & \cc{FEFAF7}{\color[HTML]{000000} $22.31$\rlap{\textsuperscript{*}}} & \cc{E2EFF6}{\color[HTML]{000000} $0.796$} & \cc{FCE7DC}{\color[HTML]{000000} $0.826$} & \cc{FCE6DB}{\color[HTML]{000000} $0.827$} & \cc{FCFDFE}{\color[HTML]{000000} $0.809$} & \cc{FEF5F0}{\color[HTML]{000000} $0.818$} & \cc{FDEAE0}{\color[HTML]{000000} $0.824$} & \cc{F0F7FA}{\color[HTML]{000000} $0.305$} & \cc{FCE0D2}{\color[HTML]{000000} $0.256$} & \cc{FCE1D3}{\color[HTML]{000000} $0.258$} & \cc{FCE7DC}{\color[HTML]{000000} $0.264$} & \cc{FBDDCE}{\color[HTML]{000000} $0.253$} & \cc{FCE3D6}{\color[HTML]{000000} $0.259$} \\
INRIA & \cc{F5F9FC}{\color[HTML]{000000} $21.95$} & \cc{FCDFD1}{\color[HTML]{000000} $22.94$} & \cc{FDE9DF}{\color[HTML]{000000} $22.72$} & \cc{FDE9DF}{\color[HTML]{000000} $22.71$} & \cc{FACFBA}{\color[HTML]{000000} $23.35$} & \cc{FBD7C5}{\color[HTML]{000000} $23.14$} & \cc{FCE5D9}{\color[HTML]{000000} $0.827$} & \cc{FACEB8}{\color[HTML]{000000} $0.842$} & \cc{FACEB8}{\color[HTML]{000000} $0.842$} & \cc{FCDECF}{\color[HTML]{000000} $0.832$} & \cc{FDE8DD}{\color[HTML]{000000} $0.825$} & \cc{F9C9B1}{\color[HTML]{000000} $0.845$} & \cc{FFFEFD}{\color[HTML]{000000} $0.289$} & \cc{F9CAB2}{\color[HTML]{000000} $0.232$} & \cc{FBDBCB}{\color[HTML]{000000} $0.250$} & \cc{FBD4C1}{\color[HTML]{000000} $0.243$} & \cc{FBDAC9}{\color[HTML]{000000} $0.249$} & \cc{FACDB7}{\color[HTML]{000000} $0.235$} \\
MCMC & \cc{F9C5AB}{\color[HTML]{000000} $23.60$} & \cc{DE7E70}{\color[HTML]{F1F1F1} $24.47$} & \cc{E48C7B}{\color[HTML]{F1F1F1} $24.30$} & \cc{F0A992}{\color[HTML]{000000} $23.97$} & \cc{E08273}{\color[HTML]{F1F1F1} $24.43$} & \cc{D5675E}{\color[HTML]{F1F1F1} $24.75$} & \cc{F9C2A7}{\color[HTML]{000000} $0.849$} & \cc{DD7C6E}{\color[HTML]{F1F1F1} $0.871$} & \cc{E89784}{\color[HTML]{000000} $0.863$} & \cc{F3B199}{\color[HTML]{000000} $0.855$} & \cc{EA9B87}{\color[HTML]{000000} $0.862$} & \cc{D5675E}{\color[HTML]{F1F1F1} $0.877$\rlap{\textsuperscript{*}}} & \cc{F9C9B1}{\color[HTML]{000000} $0.231$} & \cc{DC796D}{\color[HTML]{F1F1F1} $0.182$} & \cc{E38A7A}{\color[HTML]{F1F1F1} $0.191$} & \cc{E38A7A}{\color[HTML]{F1F1F1} $0.191$} & \cc{DE7E70}{\color[HTML]{F1F1F1} $0.184$} & \cc{D5675E}{\color[HTML]{F1F1F1} $0.172$\rlap{\textsuperscript{*}}} \\
IDHFR & \cc{FEF4EF}{\color[HTML]{000000} $22.44$} & \cc{FCDECF}{\color[HTML]{000000} $22.99$} & \cc{FBDCCC}{\color[HTML]{000000} $23.03$} & \cc{FDEFE7}{\color[HTML]{000000} $22.58$} & \cc{FBD9C8}{\color[HTML]{000000} $23.11$} & \cc{FDE9DF}{\color[HTML]{000000} $22.71$} & \cc{FEF2EC}{\color[HTML]{000000} $0.819$} & \cc{FCE5D9}{\color[HTML]{000000} $0.827$} & \cc{FBDDCE}{\color[HTML]{000000} $0.833$} & \cc{FEF8F4}{\color[HTML]{000000} $0.816$} & \cc{FFFFFE}{\color[HTML]{000000} $0.811$} & \cc{FDEAE0}{\color[HTML]{000000} $0.824$} & \cc{FDEAE0}{\color[HTML]{000000} $0.267$} & \cc{FAD1BC}{\color[HTML]{000000} $0.240$} & \cc{FBDCCC}{\color[HTML]{000000} $0.252$} & \cc{FCDFD1}{\color[HTML]{000000} $0.255$} & \cc{FBDBCB}{\color[HTML]{000000} $0.250$} & \cc{FCE0D2}{\color[HTML]{000000} $0.256$} \\
RevDGS & \cc{5C99C7}{\color[HTML]{F1F1F1} $19.58$} & \cc{F1F7FB}{\color[HTML]{000000} $21.90$} & \cc{FFFBF9}{\color[HTML]{000000} $22.28$} & \cc{EDF5FA}{\color[HTML]{000000} $21.82$} & \cc{FCE6DB}{\color[HTML]{000000} $22.79$\rlap{\textsuperscript{*}}} & \cc{FFFFFE}{\color[HTML]{000000} $22.17$} & \cc{5C99C7}{\color[HTML]{F1F1F1} $0.745$} & \cc{EEF6FA}{\color[HTML]{000000} $0.803$} & \cc{FCFDFE}{\color[HTML]{000000} $0.809$} & \cc{D7E9F3}{\color[HTML]{000000} $0.791$} & \cc{D6E9F2}{\color[HTML]{000000} $0.790$} & \cc{ECF5F9}{\color[HTML]{000000} $0.801$} & \cc{5C99C7}{\color[HTML]{F1F1F1} $0.410$} & \cc{FEF9F6}{\color[HTML]{000000} $0.284$\rlap{\textsuperscript{*}}} & \cc{FEF5F0}{\color[HTML]{000000} $0.279$\rlap{\textsuperscript{*}}} & \cc{FEF7F3}{\color[HTML]{000000} $0.281$\rlap{\textsuperscript{*}}} & \cc{FEF3ED}{\color[HTML]{000000} $0.278$\rlap{\textsuperscript{*}}} & \cc{FEFAF7}{\color[HTML]{000000} $0.285$\rlap{\textsuperscript{*}}} \\
No D. & \cc{FFFFFF}{\color[HTML]{F1F1F1} \color{white} --} & \cc{FDECE3}{\color[HTML]{000000} $22.64$} & \cc{FCFDFE}{\color[HTML]{000000} $22.10$} & \cc{E5F1F7}{\color[HTML]{000000} $21.65$} & \cc{EEF6FA}{\color[HTML]{000000} $21.84$} & \cc{D8EAF3}{\color[HTML]{000000} $21.40$} & \cc{FFFFFF}{\color[HTML]{F1F1F1} \color{white} --} & \cc{F9C5AB}{\color[HTML]{000000} $0.848$} & \cc{FEF1EA}{\color[HTML]{000000} $0.820$} & \cc{FEF7F3}{\color[HTML]{000000} $0.816$} & \cc{FBFDFE}{\color[HTML]{000000} $0.809$} & \cc{FEF8F4}{\color[HTML]{000000} $0.816$} & \cc{FFFFFF}{\color[HTML]{F1F1F1} \color{white} --} & \cc{FCE4D8}{\color[HTML]{000000} $0.260$} & \cc{D0E5F0}{\color[HTML]{000000} $0.334$} & \cc{E6F1F7}{\color[HTML]{000000} $0.314$} & \cc{D6E9F2}{\color[HTML]{000000} $0.329$} & \cc{97C3DD}{\color[HTML]{000000} $0.377$} \\
\bottomrule
\end{tabular}}
\endgroup
}
\end{table}

\PAR{Geometry accuracy.} The results of our geometry accuracy evaluation experiment on ETH3D are presented in \cref{tab:final_recon_fscore_eth3d}. Looking at the first row (accuracy of the initializations themselves), we see that the sparse SfM point clouds have very limited scene coverage compared to ScanNet++ (8\% vs 38\% recall). The other methods produce significantly more accurate results, with the exception of $\text{DA3}^\text{GS}$, which has significantly lower recall. Somewhat surprisingly, despite that, it produces the best post-training geometry with most densification strategies.

Upon investigating the $\text{DA3}^\text{GS}$-produced splats, we find a simple explanation for this behavior. In general, $\text{DA3}^\text{GS}$ often creates semi-transparent splats, and can often produce noisy primitives behind opaque scene surfaces. As our at-initialization geometry evaluation for $\text{DA3}^\text{GS}$ involves depth rendering based on alpha-compositing (see \cref{sec:protocol:geometric}), this can result in holes in the evaluated geometry at initialization time, if the opacity is not correctly concentrated near the true surface.
However, 3DGS optimization can trivially resolve this fault by increasing the opacity of the splats positioned near scene surfaces. This is likely the mechanism by which post-training geometric precision for $\text{DA3}^\text{GS
}$ increases compared to the initialization.
Because $\text{DA3}^\text{GS}$ produces splats independently for each image and does not adjust opacities based on the number of input images, the cumulative opacity of the output scene decreases with the number of images. This is why at-initialization recall is so low on ETH3D specifically -- because it has very few images per scene, the density of the semi-transparent splats near scene surfaces can often fall below what is required to produce correct depths. This effect is much less pronounced on ScanNet++ as it has more input images (though we still see recall increase after training in \Cref{tab:final_recon_fscore}).

\begin{table}[t]
\captionsetup{skip=6pt}
\centering
\caption{Reconstruction accuracy on ETH3D (inlier threshold 0.05\,m). Cells with statistically significant improvements over SfM are marked with ``$^{*}$'' (see \cref{statistical_analysis} for methodology).}
\label{tab:final_recon_fscore_eth3d}
\small
\begingroup \setlength{\tabcolsep}{3\tabcolsep}
\resizebox{\columnwidth}{!}{\begin{tabular}{lcccccc}
\toprule
 \textbf{$\text{F}_1$ / P / R} & SfM & $\text{EDGS}^\text{*}$ & M.D. & DA3 & $\text{DA3}^\text{GS}$ & Laser \\
\midrule
At Init & \cc{5C99C7}{\color[HTML]{F1F1F1} $14$ / $89$ / $8$} & \cc{D5675E}{\color[HTML]{F1F1F1} $63$ / $82$ / $53$} & \cc{F6BA9F}{\color[HTML]{000000} $54$ / $61$ / $51$} & \cc{DA7368}{\color[HTML]{F1F1F1} $61$ / $63$ / $60$} & \cc{7FB2D4}{\color[HTML]{000000} $19$ / $52$ / $12$} & \cc{FFFFFF}{\color[HTML]{F1F1F1} \color{black}} \\
\midrule
AbsGS & \cc{629DC9}{\color[HTML]{F1F1F1} $15$ / $17$ / $16$} & \cc{CBE3EF}{\color[HTML]{000000} $29$\rlap{\textsuperscript{*}} / $27$\rlap{\textsuperscript{*}} / $32$\rlap{\textsuperscript{*}}} & \cc{ABD1E5}{\color[HTML]{000000} $24$ / $25$ / $24$\rlap{\textsuperscript{*}}} & \cc{A1CAE1}{\color[HTML]{000000} $22$ / $20$ / $27$\rlap{\textsuperscript{*}}} & \cc{F8FBFD}{\color[HTML]{000000} $37$\rlap{\textsuperscript{*}} / $42$\rlap{\textsuperscript{*}} / $35$\rlap{\textsuperscript{*}}} & \cc{9FC9E0}{\color[HTML]{000000} $22$ / $22$ / $24$\rlap{\textsuperscript{*}}} \\
INRIA & \cc{66A0CB}{\color[HTML]{F1F1F1} $16$ / $18$ / $16$} & \cc{D1E6F1}{\color[HTML]{000000} $30$\rlap{\textsuperscript{*}} / $29$\rlap{\textsuperscript{*}} / $32$\rlap{\textsuperscript{*}}} & \cc{B2D6E7}{\color[HTML]{000000} $24$ / $25$ / $26$\rlap{\textsuperscript{*}}} & \cc{A3CBE2}{\color[HTML]{000000} $23$ / $21$ / $26$\rlap{\textsuperscript{*}}} & \cc{EBF4F9}{\color[HTML]{000000} $35$\rlap{\textsuperscript{*}} / $41$\rlap{\textsuperscript{*}} / $32$\rlap{\textsuperscript{*}}} & \cc{93C0DC}{\color[HTML]{000000} $21$ / $20$ / $24$\rlap{\textsuperscript{*}}} \\
IDHFR & \cc{CDE4F0}{\color[HTML]{000000} $29$ / $38$ / $25$} & \cc{EEF6FA}{\color[HTML]{000000} $35$\rlap{\textsuperscript{*}} / $36$ / $36$\rlap{\textsuperscript{*}}} & \cc{EDF5FA}{\color[HTML]{000000} $35$ / $39$ / $34$\rlap{\textsuperscript{*}}} & \cc{D5E8F2}{\color[HTML]{000000} $31$ / $31$ / $31$} & \cc{FEFFFF}{\color[HTML]{000000} $38$\rlap{\textsuperscript{*}} / $49$ / $33$} & \cc{C4DFED}{\color[HTML]{000000} $28$ / $30$ / $28$} \\
MCMC & \cc{ADD2E6}{\color[HTML]{000000} $24$ / $36$ / $19$} & \cc{D8EAF3}{\color[HTML]{000000} $31$ / $34$ / $31$\rlap{\textsuperscript{*}}} & \cc{D0E5F0}{\color[HTML]{000000} $30$ / $36$ / $26$\rlap{\textsuperscript{*}}} & \cc{CCE3EF}{\color[HTML]{000000} $29$ / $30$ / $29$\rlap{\textsuperscript{*}}} & \cc{CDE4F0}{\color[HTML]{000000} $29$ / $40$ / $24$} & \cc{C7E1EE}{\color[HTML]{000000} $28$ / $31$ / $27$\rlap{\textsuperscript{*}}} \\
RevDGS & \cc{7AAED2}{\color[HTML]{F1F1F1} $18$ / $24$ / $17$} & \cc{DBEBF4}{\color[HTML]{000000} $32$\rlap{\textsuperscript{*}} / $29$ / $36$\rlap{\textsuperscript{*}}} & \cc{C6E0ED}{\color[HTML]{000000} $28$\rlap{\textsuperscript{*}} / $29$ / $28$\rlap{\textsuperscript{*}}} & \cc{C2DEEC}{\color[HTML]{000000} $27$\rlap{\textsuperscript{*}} / $25$ / $31$\rlap{\textsuperscript{*}}} & \cc{FFFCFA}{\color[HTML]{000000} $39$\rlap{\textsuperscript{*}} / $46$\rlap{\textsuperscript{*}} / $36$\rlap{\textsuperscript{*}}} & \cc{D0E5F0}{\color[HTML]{000000} $30$\rlap{\textsuperscript{*}} / $29$ / $32$\rlap{\textsuperscript{*}}} \\
No D. & \cc{FFFFFF}{\color[HTML]{F1F1F1} \color{black}} & \cc{B5D7E8}{\color[HTML]{000000} $25$\rlap{\textsuperscript{*}} / $26$\rlap{\textsuperscript{*}} / $26$\rlap{\textsuperscript{*}}} & \cc{85B6D7}{\color[HTML]{000000} $19$ / $22$ / $20$\rlap{\textsuperscript{*}}} & \cc{83B5D6}{\color[HTML]{000000} $19$ / $19$ / $21$} & \cc{B1D5E7}{\color[HTML]{000000} $24$\rlap{\textsuperscript{*}} / $33$\rlap{\textsuperscript{*}} / $21$} & \cc{68A2CB}{\color[HTML]{F1F1F1} $16$ / $18$ / $16$} \\
\bottomrule
\end{tabular}}
\endgroup
\end{table}

\section{Ablation on Using Hybrid Initialization}\label{hybrid_init_ablation}
\Cref{tab:laser_scan_hybrid_init_ablation} shows the influence of using hybrid initialization (see \cref{sec:protocol:datasets}) on ScanNet++ and ETH3D. Hybrid initialization improves results on both ScanNet++ variants almost uniformly, with the exception of MCMC densification with initialization size of $0.5G_m$, though even there using more laser scan points results in improvements over the non-hybrid baseline. On ETH3D, however, the results are inconclusive, and lead to decreased performance in most cases, which is why we don't use hybrid initialization for this dataset in our main experiments.

For results of using hybrid initialization with initialization methods besides laser scans, please see \cref{additional_practical_results}. We also provide non-delta results for ScanNet++ \textit{without} hybrid initialization in \cref{tab:laser_scan_scannet_no_hybrid} and for ETH3D \textit{with} hybrid initialization in \cref{tab:laser_scan_eth3d_hybrid}.

\begin{table}[tbp]
\centering
\caption{Improvement from hybrid laser-scan initialization (adding sparse SfM points) across strategies and initialization sizes. Each cell is the mean and across-scene std of the per-scene metric delta.}
\label{tab:laser_scan_hybrid_init_ablation}
\begin{subtable}[t]{\linewidth}
\centering
\caption{ScanNet++}
\label{tab:laser_scan_hybrid_init_ablation_scannet++}
{\small
\begingroup \setlength{\tabcolsep}{2.5\tabcolsep}
\resizebox{0.8\linewidth}{!}{\begin{tabular}{l|ccc|ccc|ccc}
\toprule
& \multicolumn{3}{c|}{\textbf{PSNR ↑}} & \multicolumn{3}{c|}{\textbf{SSIM ↑}} & \multicolumn{3}{c}{\textbf{LPIPS ↓}} \\
Init size& $0.5\text  {G}_\mathit{m}$ & $0.75\text  {G}_\mathit{m}$ & $1.0\text  {G}_\mathit{m}$ & $0.5\text  {G}_\mathit{m}$ & $0.75\text  {G}_\mathit{m}$ & $1.0\text  {G}_\mathit{m}$ & $0.5\text  {G}_\mathit{m}$ & $0.75\text  {G}_\mathit{m}$ & $1.0\text  {G}_\mathit{m}$ \\
\midrule
AbsGS & \cc{FACDB7}{\color[HTML]{000000} $0.09$} & \cc{FBD8C7}{\color[HTML]{000000} $0.07$} & \cc{C74442}{\color[HTML]{F1F1F1} $0.23$} & \cc{E48C7B}{\color[HTML]{F1F1F1} $0.002$} & \cc{FFFBF9}{\color[HTML]{000000} $0.000$} & \cc{E18676}{\color[HTML]{F1F1F1} $0.002$} & \cc{C43B3C}{\color[HTML]{F1F1F1} $-0.008$} & \cc{E18676}{\color[HTML]{F1F1F1} $-0.006$} & \cc{C84644}{\color[HTML]{F1F1F1} $-0.008$} \\
INRIA & \cc{DF8072}{\color[HTML]{F1F1F1} $0.18$} & \cc{F4B49A}{\color[HTML]{000000} $0.13$} & \cc{FCE1D3}{\color[HTML]{000000} $0.06$} & \cc{D4655C}{\color[HTML]{F1F1F1} $0.002$} & \cc{F2AF97}{\color[HTML]{000000} $0.001$} & \cc{FDEEE6}{\color[HTML]{000000} $0.000$} & \cc{CA4C49}{\color[HTML]{F1F1F1} $-0.007$} & \cc{D5675E}{\color[HTML]{F1F1F1} $-0.007$} & \cc{E79482}{\color[HTML]{F1F1F1} $-0.005$} \\
MCMC & \cc{327CB7}{\color[HTML]{F1F1F1} $-0.24$} & \cc{F1F7FB}{\color[HTML]{000000} $-0.02$} & \cc{F8FBFD}{\color[HTML]{000000} $-0.01$} & \cc{327CB7}{\color[HTML]{F1F1F1} $-0.002$} & \cc{CA4C49}{\color[HTML]{F1F1F1} $0.002$} & \cc{FFFDFB}{\color[HTML]{000000} $0.000$} & \cc{C1DDEC}{\color[HTML]{000000} $0.003$} & \cc{D05A54}{\color[HTML]{F1F1F1} $-0.007$} & \cc{E08273}{\color[HTML]{F1F1F1} $-0.006$} \\
IDHFR & \cc{FEF9F6}{\color[HTML]{000000} $0.01$} & \cc{F9C8B0}{\color[HTML]{000000} $0.10$} & \cc{FCE2D5}{\color[HTML]{000000} $0.05$} & \cc{F3F9FB}{\color[HTML]{000000} $-0.000$} & \cc{FBD7C5}{\color[HTML]{000000} $0.001$} & \cc{F9C5AB}{\color[HTML]{000000} $0.001$} & \cc{F0AB94}{\color[HTML]{000000} $-0.005$} & \cc{F9C3A8}{\color[HTML]{000000} $-0.004$} & \cc{F5B89E}{\color[HTML]{000000} $-0.004$} \\
RevDGS & \cc{E79482}{\color[HTML]{F1F1F1} $0.16$} & \cc{F0A992}{\color[HTML]{000000} $0.14$} & \cc{FAD3BF}{\color[HTML]{000000} $0.08$} & \cc{E69280}{\color[HTML]{F1F1F1} $0.002$} & \cc{F9C9B1}{\color[HTML]{000000} $0.001$} & \cc{EEA58F}{\color[HTML]{000000} $0.001$} & \cc{F1AD96}{\color[HTML]{000000} $-0.004$} & \cc{F0AB94}{\color[HTML]{000000} $-0.005$} & \cc{FACEB8}{\color[HTML]{000000} $-0.003$} \\
\bottomrule
\end{tabular}}
\endgroup
}
\end{subtable}
\begin{subtable}[t]{\linewidth}
\centering
\caption{ScanNet++ (On-Trajectory)}
\label{tab:laser_scan_hybrid_init_ablation_eval_on_train_set_scannet++}
{\small
\begingroup \setlength{\tabcolsep}{2.5\tabcolsep}
\resizebox{0.8\linewidth}{!}{\begin{tabular}{l|ccc|ccc|ccc}
\toprule
& \multicolumn{3}{c|}{\textbf{PSNR ↑}} & \multicolumn{3}{c|}{\textbf{SSIM ↑}} & \multicolumn{3}{c}{\textbf{LPIPS ↓}} \\
Init size& $0.5\text  {G}_\mathit{m}$ & $0.75\text  {G}_\mathit{m}$ & $1.0\text  {G}_\mathit{m}$ & $0.5\text  {G}_\mathit{m}$ & $0.75\text  {G}_\mathit{m}$ & $1.0\text  {G}_\mathit{m}$ & $0.5\text  {G}_\mathit{m}$ & $0.75\text  {G}_\mathit{m}$ & $1.0\text  {G}_\mathit{m}$ \\
\midrule
AbsGS & \cc{CE544F}{\color[HTML]{F1F1F1} $0.48$} & \cc{D15C56}{\color[HTML]{F1F1F1} $0.46$} & \cc{C74442}{\color[HTML]{F1F1F1} $0.51$} & \cc{C74442}{\color[HTML]{F1F1F1} $0.002$} & \cc{C43D3D}{\color[HTML]{F1F1F1} $0.002$} & \cc{C43B3C}{\color[HTML]{F1F1F1} $0.002$} & \cc{CA4C49}{\color[HTML]{F1F1F1} $-0.005$} & \cc{CB4E4A}{\color[HTML]{F1F1F1} $-0.005$} & \cc{C43B3C}{\color[HTML]{F1F1F1} $-0.005$} \\
INRIA & \cc{D36159}{\color[HTML]{F1F1F1} $0.46$} & \cc{C43B3C}{\color[HTML]{F1F1F1} $0.53$} & \cc{C43B3C}{\color[HTML]{F1F1F1} $0.53$} & \cc{C43B3C}{\color[HTML]{F1F1F1} $0.002$} & \cc{C43B3C}{\color[HTML]{F1F1F1} $0.002$} & \cc{C94A47}{\color[HTML]{F1F1F1} $0.002$} & \cc{CC504C}{\color[HTML]{F1F1F1} $-0.005$} & \cc{C74442}{\color[HTML]{F1F1F1} $-0.005$} & \cc{C53F3F}{\color[HTML]{F1F1F1} $-0.005$} \\
MCMC & \cc{F9C3A8}{\color[HTML]{000000} $0.25$} & \cc{DE7E70}{\color[HTML]{F1F1F1} $0.40$} & \cc{D4655C}{\color[HTML]{F1F1F1} $0.45$} & \cc{FACDB7}{\color[HTML]{000000} $0.001$} & \cc{E58E7D}{\color[HTML]{F1F1F1} $0.001$} & \cc{CE5651}{\color[HTML]{F1F1F1} $0.002$} & \cc{F5B89E}{\color[HTML]{000000} $-0.003$} & \cc{D4655C}{\color[HTML]{F1F1F1} $-0.005$} & \cc{CE5651}{\color[HTML]{F1F1F1} $-0.005$} \\
IDHFR & \cc{F5B89E}{\color[HTML]{000000} $0.28$} & \cc{FAD0BB}{\color[HTML]{000000} $0.20$} & \cc{D86F65}{\color[HTML]{F1F1F1} $0.43$} & \cc{F4B49A}{\color[HTML]{000000} $0.001$} & \cc{FACFBA}{\color[HTML]{000000} $0.001$} & \cc{D97166}{\color[HTML]{F1F1F1} $0.002$} & \cc{EB9D88}{\color[HTML]{000000} $-0.003$} & \cc{F1AD96}{\color[HTML]{000000} $-0.003$} & \cc{DC796D}{\color[HTML]{F1F1F1} $-0.004$} \\
RevDGS & \cc{F9C7AE}{\color[HTML]{000000} $0.23$} & \cc{EB9F8A}{\color[HTML]{000000} $0.33$} & \cc{E28878}{\color[HTML]{F1F1F1} $0.38$} & \cc{F4B69C}{\color[HTML]{000000} $0.001$} & \cc{D3635B}{\color[HTML]{F1F1F1} $0.002$} & \cc{D66960}{\color[HTML]{F1F1F1} $0.002$} & \cc{F9CAB2}{\color[HTML]{000000} $-0.002$} & \cc{D5675E}{\color[HTML]{F1F1F1} $-0.005$} & \cc{D5675E}{\color[HTML]{F1F1F1} $-0.005$} \\
\bottomrule
\end{tabular}}
\endgroup
}
\end{subtable}
\begin{subtable}[t]{\linewidth}
\centering
\caption{ETH3D}
\label{tab:laser_scan_hybrid_init_ablation_eth3d}
{\small
\begingroup \setlength{\tabcolsep}{2.5\tabcolsep}
\resizebox{0.8\linewidth}{!}{\begin{tabular}{l|ccc|ccc|ccc}
\toprule
& \multicolumn{3}{c|}{\textbf{PSNR ↑}} & \multicolumn{3}{c|}{\textbf{SSIM ↑}} & \multicolumn{3}{c}{\textbf{LPIPS ↓}} \\
Init size& $0.5\text  {G}_\mathit{m}$ & $0.75\text  {G}_\mathit{m}$ & $1.0\text  {G}_\mathit{m}$ & $0.5\text  {G}_\mathit{m}$ & $0.75\text  {G}_\mathit{m}$ & $1.0\text  {G}_\mathit{m}$ & $0.5\text  {G}_\mathit{m}$ & $0.75\text  {G}_\mathit{m}$ & $1.0\text  {G}_\mathit{m}$ \\
\midrule
AbsGS & \cc{8BBBD9}{\color[HTML]{000000} $-0.71$} & \cc{EEF6FA}{\color[HTML]{000000} $-0.12$} & \cc{7CAFD3}{\color[HTML]{F1F1F1} $-0.78$} & \cc{F9C3A8}{\color[HTML]{000000} $0.010$} & \cc{FDEAE0}{\color[HTML]{000000} $0.004$} & \cc{FBFDFE}{\color[HTML]{000000} $-0.001$} & \cc{FBDBCB}{\color[HTML]{000000} $-0.014$} & \cc{FCE4D8}{\color[HTML]{000000} $-0.011$} & \cc{C4DFED}{\color[HTML]{000000} $0.018$} \\
INRIA & \cc{FFFEFD}{\color[HTML]{000000} $0.01$} & \cc{FDE9DF}{\color[HTML]{000000} $0.20$} & \cc{E7F2F8}{\color[HTML]{000000} $-0.17$} & \cc{FACEB8}{\color[HTML]{000000} $0.009$} & \cc{FEF2EC}{\color[HTML]{000000} $0.002$} & \cc{FFFBF9}{\color[HTML]{000000} $0.001$} & \cc{FDEFE7}{\color[HTML]{000000} $-0.006$} & \cc{FAD2BE}{\color[HTML]{000000} $-0.017$} & \cc{FEF8F4}{\color[HTML]{000000} $-0.003$} \\
MCMC & \cc{FDE9DF}{\color[HTML]{000000} $0.20$} & \cc{3C83BB}{\color[HTML]{F1F1F1} $-1.06$} & \cc{C0DDEB}{\color[HTML]{000000} $-0.45$} & \cc{EDF5FA}{\color[HTML]{000000} $-0.002$} & \cc{B7D8E9}{\color[HTML]{000000} $-0.010$} & \cc{E1EFF6}{\color[HTML]{000000} $-0.004$} & \cc{87B8D7}{\color[HTML]{000000} $0.032$} & \cc{327CB7}{\color[HTML]{F1F1F1} $0.048$} & \cc{8FBDDA}{\color[HTML]{000000} $0.030$} \\
IDHFR & \cc{FBD8C7}{\color[HTML]{000000} $0.34$} & \cc{D0E5F0}{\color[HTML]{000000} $-0.33$} & \cc{9BC6DF}{\color[HTML]{000000} $-0.65$} & \cc{F9C4AA}{\color[HTML]{000000} $0.010$} & \cc{FEF8F4}{\color[HTML]{000000} $0.001$} & \cc{BDDBEB}{\color[HTML]{000000} $-0.009$} & \cc{FBDAC9}{\color[HTML]{000000} $-0.014$} & \cc{FEF6F1}{\color[HTML]{000000} $-0.004$} & \cc{D8EAF3}{\color[HTML]{000000} $0.012$} \\
RevDGS & \cc{D6E9F2}{\color[HTML]{000000} $-0.29$} & \cc{327CB7}{\color[HTML]{F1F1F1} $-1.11$} & \cc{BCDBEA}{\color[HTML]{000000} $-0.48$} & \cc{F1F7FB}{\color[HTML]{000000} $-0.002$} & \cc{327CB7}{\color[HTML]{F1F1F1} $-0.022$} & \cc{F3F9FB}{\color[HTML]{000000} $-0.002$} & \cc{E0EEF5}{\color[HTML]{000000} $0.010$} & \cc{5E9BC8}{\color[HTML]{F1F1F1} $0.040$} & \cc{F8FBFD}{\color[HTML]{000000} $0.002$} \\
\bottomrule
\end{tabular}}
\endgroup
}
\end{subtable}
\end{table}

\begin{table}[tbp]
\centering
\caption{Results using laser scan initialization \textbf{without} hybrid initialization on ScanNet++. Cells with statistically significant improvements over SfM are marked with ``$^{*}$'' (see \cref{statistical_analysis} for methodology).}
\label{tab:laser_scan_scannet_no_hybrid}
\begin{subtable}[t]{\linewidth}
\centering
\caption{ScanNet++}
{\small
\begingroup \setlength{\tabcolsep}{2.5\tabcolsep}
\resizebox{\linewidth}{!}{\begin{tabular}{l|ccccc|ccccc|ccccc}
\toprule
& \multicolumn{5}{c|}{\textbf{PSNR ↑}} & \multicolumn{5}{c|}{\textbf{SSIM ↑}} & \multicolumn{5}{c}{\textbf{LPIPS ↓}} \\
& SfM & $|\mathcal{G}_\mathit{init}^\text{SfM}|$ & $0.5\text  {G}_\mathit{m}$ & $0.75\text  {G}_\mathit{m}$ & $1.0\text  {G}_\mathit{m}$ & SfM & $|\mathcal{G}_\mathit{init}^\text{SfM}|$ & $0.5\text  {G}_\mathit{m}$ & $0.75\text  {G}_\mathit{m}$ & $1.0\text  {G}_\mathit{m}$ & SfM & $|\mathcal{G}_\mathit{init}^\text{SfM}|$ & $0.5\text  {G}_\mathit{m}$ & $0.75\text  {G}_\mathit{m}$ & $1.0\text  {G}_\mathit{m}$ \\
\midrule
AbsGS & \cc{BFDCEB}{\color[HTML]{000000} $22.36$} & \cc{DBEBF4}{\color[HTML]{000000} $22.50$} & \cc{FEF6F1}{\color[HTML]{000000} $22.75$} & \cc{FCE0D2}{\color[HTML]{000000} $22.89$} & \cc{FBD8C7}{\color[HTML]{000000} $22.94$\rlap{\textsuperscript{*}}} & \cc{B2D6E7}{\color[HTML]{000000} $0.870$} & \cc{89B9D8}{\color[HTML]{000000} $0.869$} & \cc{CDE4F0}{\color[HTML]{000000} $0.871$} & \cc{F8FBFD}{\color[HTML]{000000} $0.873$\rlap{\textsuperscript{*}}} & \cc{FEFAF7}{\color[HTML]{000000} $0.874$\rlap{\textsuperscript{*}}} & \cc{FFFBF9}{\color[HTML]{000000} $0.249$} & \cc{DCECF4}{\color[HTML]{000000} $0.254$} & \cc{FFFBF9}{\color[HTML]{000000} $0.249$} & \cc{FCE7DC}{\color[HTML]{000000} $0.246$} & \cc{FCE1D3}{\color[HTML]{000000} $0.245$} \\
INRIA & \cc{A9CFE4}{\color[HTML]{000000} $22.26$} & \cc{DBEBF4}{\color[HTML]{000000} $22.50$} & \cc{FEF6F1}{\color[HTML]{000000} $22.75$} & \cc{FDEBE2}{\color[HTML]{000000} $22.82$} & \cc{FCDFD1}{\color[HTML]{000000} $22.89$\rlap{\textsuperscript{*}}} & \cc{9FC9E0}{\color[HTML]{000000} $0.870$} & \cc{A5CDE3}{\color[HTML]{000000} $0.870$} & \cc{D6E9F2}{\color[HTML]{000000} $0.872$} & \cc{E8F3F8}{\color[HTML]{000000} $0.872$} & \cc{F6FAFC}{\color[HTML]{000000} $0.873$\rlap{\textsuperscript{*}}} & \cc{E5F1F7}{\color[HTML]{000000} $0.253$} & \cc{CDE4F0}{\color[HTML]{000000} $0.256$} & \cc{FEF9F6}{\color[HTML]{000000} $0.249$} & \cc{FDEDE5}{\color[HTML]{000000} $0.247$} & \cc{FCE7DC}{\color[HTML]{000000} $0.246$\rlap{\textsuperscript{*}}} \\
MCMC & \cc{5C99C7}{\color[HTML]{F1F1F1} $22.02$} & \cc{E2EFF6}{\color[HTML]{000000} $22.54$} & \cc{DFEDF5}{\color[HTML]{000000} $22.52$} & \cc{E7F2F8}{\color[HTML]{000000} $22.57$} & \cc{E3F0F6}{\color[HTML]{000000} $22.54$} & \cc{AFD4E6}{\color[HTML]{000000} $0.870$} & \cc{FBDAC9}{\color[HTML]{000000} $0.876$\rlap{\textsuperscript{*}}} & \cc{FCFDFE}{\color[HTML]{000000} $0.873$} & \cc{FDEFE7}{\color[HTML]{000000} $0.874$\rlap{\textsuperscript{*}}} & \cc{FCDFD1}{\color[HTML]{000000} $0.875$\rlap{\textsuperscript{*}}} & \cc{FEFAF7}{\color[HTML]{000000} $0.249$} & \cc{FBDBCB}{\color[HTML]{000000} $0.244$} & \cc{FBDDCE}{\color[HTML]{000000} $0.244$} & \cc{FACFBA}{\color[HTML]{000000} $0.242$\rlap{\textsuperscript{*}}} & \cc{F9CAB2}{\color[HTML]{000000} $0.241$\rlap{\textsuperscript{*}}} \\
IDHFR & \cc{F9C8B0}{\color[HTML]{000000} $23.04$} & \cc{F9CAB2}{\color[HTML]{000000} $23.02$} & \cc{E58E7D}{\color[HTML]{F1F1F1} $23.24$} & \cc{DF8072}{\color[HTML]{F1F1F1} $23.29$} & \cc{D5675E}{\color[HTML]{F1F1F1} $23.36$} & \cc{F8C0A4}{\color[HTML]{000000} $0.877$} & \cc{FAD2BE}{\color[HTML]{000000} $0.876$} & \cc{E99985}{\color[HTML]{000000} $0.878$} & \cc{EA9B87}{\color[HTML]{000000} $0.878$} & \cc{D5675E}{\color[HTML]{F1F1F1} $0.879$\rlap{\textsuperscript{*}}} & \cc{D5675E}{\color[HTML]{F1F1F1} $0.233$} & \cc{F9C7AE}{\color[HTML]{000000} $0.241$} & \cc{E58E7D}{\color[HTML]{F1F1F1} $0.236$} & \cc{DD7C6E}{\color[HTML]{F1F1F1} $0.235$} & \cc{D86D63}{\color[HTML]{F1F1F1} $0.233$} \\
RevDGS & \cc{CBE3EF}{\color[HTML]{000000} $22.42$} & \cc{D8EAF3}{\color[HTML]{000000} $22.49$} & \cc{F6FAFC}{\color[HTML]{000000} $22.64$} & \cc{FEF7F3}{\color[HTML]{000000} $22.74$} & \cc{FCDECF}{\color[HTML]{000000} $22.90$} & \cc{95C2DD}{\color[HTML]{000000} $0.869$} & \cc{5C99C7}{\color[HTML]{F1F1F1} $0.868$} & \cc{B1D5E7}{\color[HTML]{000000} $0.870$} & \cc{D1E6F1}{\color[HTML]{000000} $0.871$} & \cc{E0EEF5}{\color[HTML]{000000} $0.872$\rlap{\textsuperscript{*}}} & \cc{CCE3EF}{\color[HTML]{000000} $0.257$} & \cc{B9D9E9}{\color[HTML]{000000} $0.259$} & \cc{D0E5F0}{\color[HTML]{000000} $0.256$} & \cc{EAF3F8}{\color[HTML]{000000} $0.253$} & \cc{FFFFFE}{\color[HTML]{000000} $0.250$} \\
No D. & \cc{FFFFFF}{\color[HTML]{F1F1F1} \color{white} --} & \cc{FFFFFF}{\color[HTML]{F1F1F1} \color{white} --} & \cc{CFE5F0}{\color[HTML]{000000} $22.44$} & \cc{DDEDF5}{\color[HTML]{000000} $22.52$} & \cc{F9FCFD}{\color[HTML]{000000} $22.66$} & \cc{FFFFFF}{\color[HTML]{F1F1F1} \color{white} --} & \cc{FFFFFF}{\color[HTML]{F1F1F1} \color{white} --} & \cc{649FCA}{\color[HTML]{F1F1F1} $0.868$} & \cc{9BC6DF}{\color[HTML]{000000} $0.869$} & \cc{C0DDEB}{\color[HTML]{000000} $0.871$} & \cc{FFFFFF}{\color[HTML]{F1F1F1} \color{white} --} & \cc{FFFFFF}{\color[HTML]{F1F1F1} \color{white} --} & \cc{5C99C7}{\color[HTML]{F1F1F1} $0.267$} & \cc{ABD1E5}{\color[HTML]{000000} $0.261$} & \cc{D2E7F1}{\color[HTML]{000000} $0.256$} \\
\bottomrule
\end{tabular}}
\endgroup
}
\end{subtable}
\begin{subtable}[t]{\linewidth}
\centering
\caption{ScanNet++ (On-Trajectory)}
{\small
\begingroup \setlength{\tabcolsep}{2.5\tabcolsep}
\resizebox{\linewidth}{!}{\begin{tabular}{l|ccccc|ccccc|ccccc}
\toprule
& \multicolumn{5}{c|}{\textbf{PSNR ↑}} & \multicolumn{5}{c|}{\textbf{SSIM ↑}} & \multicolumn{5}{c}{\textbf{LPIPS ↓}} \\
& SfM & $|\mathcal{G}_\mathit{init}^\text{SfM}|$ & $0.5\text  {G}_\mathit{m}$ & $0.75\text  {G}_\mathit{m}$ & $1.0\text  {G}_\mathit{m}$ & SfM & $|\mathcal{G}_\mathit{init}^\text{SfM}|$ & $0.5\text  {G}_\mathit{m}$ & $0.75\text  {G}_\mathit{m}$ & $1.0\text  {G}_\mathit{m}$ & SfM & $|\mathcal{G}_\mathit{init}^\text{SfM}|$ & $0.5\text  {G}_\mathit{m}$ & $0.75\text  {G}_\mathit{m}$ & $1.0\text  {G}_\mathit{m}$ \\
\midrule
AbsGS & \cc{F9C9B1}{\color[HTML]{000000} $33.20$} & \cc{FCDECF}{\color[HTML]{000000} $32.97$} & \cc{FBDCCC}{\color[HTML]{000000} $33.00$} & \cc{FCDECF}{\color[HTML]{000000} $32.97$} & \cc{FCE7DC}{\color[HTML]{000000} $32.87$} & \cc{EEA58F}{\color[HTML]{000000} $0.951$} & \cc{FACCB5}{\color[HTML]{000000} $0.950$} & \cc{F9C6AD}{\color[HTML]{000000} $0.950$} & \cc{F9C6AD}{\color[HTML]{000000} $0.950$} & \cc{FACCB5}{\color[HTML]{000000} $0.950$} & \cc{F9C1A6}{\color[HTML]{000000} $0.101$} & \cc{FBD7C5}{\color[HTML]{000000} $0.104$} & \cc{F5B89E}{\color[HTML]{000000} $0.100$} & \cc{F0AB94}{\color[HTML]{000000} $0.099$\rlap{\textsuperscript{*}}} & \cc{EFA791}{\color[HTML]{000000} $0.099$} \\
INRIA & \cc{FAD2BE}{\color[HTML]{000000} $33.10$} & \cc{FDEDE5}{\color[HTML]{000000} $32.81$} & \cc{FCE3D6}{\color[HTML]{000000} $32.91$} & \cc{FCE7DC}{\color[HTML]{000000} $32.87$} & \cc{FEF1EA}{\color[HTML]{000000} $32.76$} & \cc{FACCB5}{\color[HTML]{000000} $0.950$} & \cc{FEF2EC}{\color[HTML]{000000} $0.948$} & \cc{FBD4C1}{\color[HTML]{000000} $0.949$} & \cc{FAD2BE}{\color[HTML]{000000} $0.949$} & \cc{FBD6C4}{\color[HTML]{000000} $0.949$} & \cc{F6FAFC}{\color[HTML]{000000} $0.112$} & \cc{DAEBF3}{\color[HTML]{000000} $0.115$} & \cc{FAD2BE}{\color[HTML]{000000} $0.103$} & \cc{F9C6AD}{\color[HTML]{000000} $0.101$\rlap{\textsuperscript{*}}} & \cc{F9C2A7}{\color[HTML]{000000} $0.101$\rlap{\textsuperscript{*}}} \\
MCMC & \cc{FCDECF}{\color[HTML]{000000} $32.97$} & \cc{FCE0D2}{\color[HTML]{000000} $32.95$} & \cc{FDE8DD}{\color[HTML]{000000} $32.86$} & \cc{FDEDE5}{\color[HTML]{000000} $32.80$} & \cc{FDEFE7}{\color[HTML]{000000} $32.78$} & \cc{EEA58F}{\color[HTML]{000000} $0.951$} & \cc{E99985}{\color[HTML]{000000} $0.951$} & \cc{E79482}{\color[HTML]{F1F1F1} $0.952$} & \cc{E79482}{\color[HTML]{F1F1F1} $0.952$} & \cc{E89784}{\color[HTML]{000000} $0.951$} & \cc{FBD7C5}{\color[HTML]{000000} $0.104$} & \cc{FACEB8}{\color[HTML]{000000} $0.103$} & \cc{F5B89E}{\color[HTML]{000000} $0.100$\rlap{\textsuperscript{*}}} & \cc{F2AF97}{\color[HTML]{000000} $0.099$\rlap{\textsuperscript{*}}} & \cc{F1AD96}{\color[HTML]{000000} $0.099$\rlap{\textsuperscript{*}}} \\
IDHFR & \cc{D76B61}{\color[HTML]{F1F1F1} $33.76$} & \cc{D5675E}{\color[HTML]{F1F1F1} $33.78$} & \cc{D76B61}{\color[HTML]{F1F1F1} $33.75$} & \cc{D86F65}{\color[HTML]{F1F1F1} $33.73$} & \cc{D86F65}{\color[HTML]{F1F1F1} $33.73$} & \cc{D5675E}{\color[HTML]{F1F1F1} $0.953$} & \cc{D97166}{\color[HTML]{F1F1F1} $0.953$} & \cc{D66960}{\color[HTML]{F1F1F1} $0.953$} & \cc{D5675E}{\color[HTML]{F1F1F1} $0.953$} & \cc{D66960}{\color[HTML]{F1F1F1} $0.953$} & \cc{DC776B}{\color[HTML]{F1F1F1} $0.095$} & \cc{E18676}{\color[HTML]{F1F1F1} $0.096$} & \cc{D86F65}{\color[HTML]{F1F1F1} $0.094$} & \cc{D66960}{\color[HTML]{F1F1F1} $0.094$} & \cc{D5675E}{\color[HTML]{F1F1F1} $0.094$} \\
RevDGS & \cc{FCE2D5}{\color[HTML]{000000} $32.93$} & \cc{FFFCFA}{\color[HTML]{000000} $32.64$} & \cc{FDF0E9}{\color[HTML]{000000} $32.77$} & \cc{FDEDE5}{\color[HTML]{000000} $32.80$} & \cc{FDEFE7}{\color[HTML]{000000} $32.79$} & \cc{FCE0D2}{\color[HTML]{000000} $0.949$} & \cc{FDFEFE}{\color[HTML]{000000} $0.947$} & \cc{FCE7DC}{\color[HTML]{000000} $0.948$} & \cc{FBDAC9}{\color[HTML]{000000} $0.949$} & \cc{FBDAC9}{\color[HTML]{000000} $0.949$} & \cc{FBDDCE}{\color[HTML]{000000} $0.105$} & \cc{FEF4EF}{\color[HTML]{000000} $0.109$} & \cc{FBD5C2}{\color[HTML]{000000} $0.104$} & \cc{F9C6AD}{\color[HTML]{000000} $0.101$} & \cc{F8BEA3}{\color[HTML]{000000} $0.100$} \\
No D. & \cc{FFFFFF}{\color[HTML]{F1F1F1} \color{white} --} & \cc{FFFFFF}{\color[HTML]{F1F1F1} \color{white} --} & \cc{5C99C7}{\color[HTML]{F1F1F1} $31.42$} & \cc{95C2DD}{\color[HTML]{000000} $31.75$} & \cc{BBDAEA}{\color[HTML]{000000} $31.99$} & \cc{FFFFFF}{\color[HTML]{F1F1F1} \color{white} --} & \cc{FFFFFF}{\color[HTML]{F1F1F1} \color{white} --} & \cc{5C99C7}{\color[HTML]{F1F1F1} $0.941$} & \cc{B5D7E8}{\color[HTML]{000000} $0.943$} & \cc{DCECF4}{\color[HTML]{000000} $0.945$} & \cc{FFFFFF}{\color[HTML]{F1F1F1} \color{white} --} & \cc{FFFFFF}{\color[HTML]{F1F1F1} \color{white} --} & \cc{5C99C7}{\color[HTML]{F1F1F1} $0.127$} & \cc{C7E1EE}{\color[HTML]{000000} $0.118$} & \cc{F6FAFC}{\color[HTML]{000000} $0.112$} \\
\bottomrule
\end{tabular}}
\endgroup
}
\end{subtable}
\end{table}

\begin{table}[tbp]
\centering
\caption{Results using laser scan initialization \textbf{with} hybrid initialization on ETH3D. Cells with statistically significant improvements over SfM are marked with ``$^{*}$'' (see \cref{statistical_analysis} for methodology).}
\label{tab:laser_scan_eth3d_hybrid}
\begin{subtable}[t]{\linewidth}
\centering
\caption{ETH3D}
{\small
\begingroup \setlength{\tabcolsep}{2.5\tabcolsep}
\resizebox{\linewidth}{!}{\begin{tabular}{l|ccccc|ccccc|ccccc}
\toprule
& \multicolumn{5}{c|}{\textbf{PSNR ↑}} & \multicolumn{5}{c|}{\textbf{SSIM ↑}} & \multicolumn{5}{c}{\textbf{LPIPS ↓}} \\
& SfM & $|\mathcal{G}_\mathit{init}^\text{SfM}|$ & $0.5\text  {G}_\mathit{m}$ & $0.75\text  {G}_\mathit{m}$ & $1.0\text  {G}_\mathit{m}$ & SfM & $|\mathcal{G}_\mathit{init}^\text{SfM}|$ & $0.5\text  {G}_\mathit{m}$ & $0.75\text  {G}_\mathit{m}$ & $1.0\text  {G}_\mathit{m}$ & SfM & $|\mathcal{G}_\mathit{init}^\text{SfM}|$ & $0.5\text  {G}_\mathit{m}$ & $0.75\text  {G}_\mathit{m}$ & $1.0\text  {G}_\mathit{m}$ \\
\midrule
AbsGS & \cc{CBE3EF}{\color[HTML]{000000} $21.06$} & \cc{E7F2F8}{\color[HTML]{000000} $21.58$} & \cc{FFFCFA}{\color[HTML]{000000} $22.12$} & \cc{FBDDCE}{\color[HTML]{000000} $22.81$\rlap{\textsuperscript{*}}} & \cc{FEF4EF}{\color[HTML]{000000} $22.30$\rlap{\textsuperscript{*}}} & \cc{EAF3F8}{\color[HTML]{000000} $0.796$} & \cc{D8EAF3}{\color[HTML]{000000} $0.788$} & \cc{F9C5AB}{\color[HTML]{000000} $0.841$} & \cc{F4B49A}{\color[HTML]{000000} $0.847$\rlap{\textsuperscript{*}}} & \cc{F9C2A7}{\color[HTML]{000000} $0.843$\rlap{\textsuperscript{*}}} & \cc{F9FCFD}{\color[HTML]{000000} $0.305$} & \cc{F2F8FB}{\color[HTML]{000000} $0.312$} & \cc{F2AF97}{\color[HTML]{000000} $0.224$} & \cc{EA9B87}{\color[HTML]{000000} $0.213$\rlap{\textsuperscript{*}}} & \cc{F9C9B1}{\color[HTML]{000000} $0.243$\rlap{\textsuperscript{*}}} \\
INRIA & \cc{FBFDFE}{\color[HTML]{000000} $21.95$} & \cc{FEFFFF}{\color[HTML]{000000} $22.01$} & \cc{FBDAC9}{\color[HTML]{000000} $22.90$\rlap{\textsuperscript{*}}} & \cc{FACEB8}{\color[HTML]{000000} $23.16$\rlap{\textsuperscript{*}}} & \cc{FACEB8}{\color[HTML]{000000} $23.15$\rlap{\textsuperscript{*}}} & \cc{FBDCCC}{\color[HTML]{000000} $0.827$} & \cc{FDEFE7}{\color[HTML]{000000} $0.816$} & \cc{EEA58F}{\color[HTML]{000000} $0.852$} & \cc{E69280}{\color[HTML]{F1F1F1} $0.857$\rlap{\textsuperscript{*}}} & \cc{E28878}{\color[HTML]{F1F1F1} $0.859$\rlap{\textsuperscript{*}}} & \cc{FEF5F0}{\color[HTML]{000000} $0.289$} & \cc{FEF7F3}{\color[HTML]{000000} $0.291$} & \cc{E79482}{\color[HTML]{F1F1F1} $0.211$} & \cc{DE7E70}{\color[HTML]{F1F1F1} $0.198$\rlap{\textsuperscript{*}}} & \cc{DE7E70}{\color[HTML]{F1F1F1} $0.199$\rlap{\textsuperscript{*}}} \\
MCMC & \cc{F4B69C}{\color[HTML]{000000} $23.60$} & \cc{D76B61}{\color[HTML]{F1F1F1} $24.42$} & \cc{D5675E}{\color[HTML]{F1F1F1} $24.48$} & \cc{F8BEA3}{\color[HTML]{000000} $23.51$} & \cc{DF8072}{\color[HTML]{F1F1F1} $24.19$} & \cc{F1AD96}{\color[HTML]{000000} $0.849$} & \cc{D5675E}{\color[HTML]{F1F1F1} $0.869$} & \cc{D66960}{\color[HTML]{F1F1F1} $0.869$} & \cc{DD7C6E}{\color[HTML]{F1F1F1} $0.863$} & \cc{D5675E}{\color[HTML]{F1F1F1} $0.869$} & \cc{F7BCA1}{\color[HTML]{000000} $0.231$} & \cc{D5675E}{\color[HTML]{F1F1F1} $0.186$} & \cc{E58E7D}{\color[HTML]{F1F1F1} $0.207$\rlap{\textsuperscript{*}}} & \cc{EFA791}{\color[HTML]{000000} $0.220$} & \cc{E28878}{\color[HTML]{F1F1F1} $0.204$\rlap{\textsuperscript{*}}} \\
IDHFR & \cc{FDEEE6}{\color[HTML]{000000} $22.44$} & \cc{FFFBF9}{\color[HTML]{000000} $22.13$} & \cc{FAD0BB}{\color[HTML]{000000} $23.12$} & \cc{FBD5C2}{\color[HTML]{000000} $23.01$} & \cc{FBD4C1}{\color[HTML]{000000} $23.03$} & \cc{FDEAE0}{\color[HTML]{000000} $0.819$} & \cc{FEFFFF}{\color[HTML]{000000} $0.806$} & \cc{F6BA9F}{\color[HTML]{000000} $0.846$} & \cc{EFA791}{\color[HTML]{000000} $0.851$\rlap{\textsuperscript{*}}} & \cc{F1AD96}{\color[HTML]{000000} $0.849$\rlap{\textsuperscript{*}}} & \cc{FCE0D2}{\color[HTML]{000000} $0.267$} & \cc{FFFBF9}{\color[HTML]{000000} $0.295$} & \cc{EDA38D}{\color[HTML]{000000} $0.218$} & \cc{E79482}{\color[HTML]{F1F1F1} $0.210$\rlap{\textsuperscript{*}}} & \cc{E89784}{\color[HTML]{000000} $0.211$\rlap{\textsuperscript{*}}} \\
RevDGS & \cc{5C99C7}{\color[HTML]{F1F1F1} $19.58$} & \cc{7AAED2}{\color[HTML]{F1F1F1} $19.92$} & \cc{FDEFE7}{\color[HTML]{000000} $22.42$\rlap{\textsuperscript{*}}} & \cc{E8F3F8}{\color[HTML]{000000} $21.60$} & \cc{FCE7DC}{\color[HTML]{000000} $22.59$\rlap{\textsuperscript{*}}} & \cc{5E9BC8}{\color[HTML]{F1F1F1} $0.745$} & \cc{5C99C7}{\color[HTML]{F1F1F1} $0.744$} & \cc{FCE2D5}{\color[HTML]{000000} $0.824$} & \cc{FEF6F1}{\color[HTML]{000000} $0.812$} & \cc{FAD1BC}{\color[HTML]{000000} $0.834$\rlap{\textsuperscript{*}}} & \cc{649FCA}{\color[HTML]{F1F1F1} $0.410$} & \cc{5C99C7}{\color[HTML]{F1F1F1} $0.414$} & \cc{FCE0D2}{\color[HTML]{000000} $0.266$\rlap{\textsuperscript{*}}} & \cc{FEF2EC}{\color[HTML]{000000} $0.286$\rlap{\textsuperscript{*}}} & \cc{FAD0BB}{\color[HTML]{000000} $0.250$\rlap{\textsuperscript{*}}} \\
No D. & \cc{FFFFFF}{\color[HTML]{F1F1F1} \color{white} --} & \cc{FFFFFF}{\color[HTML]{F1F1F1} \color{white} --} & \cc{FBD6C4}{\color[HTML]{000000} $22.97$} & \cc{FAD1BC}{\color[HTML]{000000} $23.10$} & \cc{F8C0A4}{\color[HTML]{000000} $23.49$} & \cc{FFFFFF}{\color[HTML]{F1F1F1} \color{white} --} & \cc{FFFFFF}{\color[HTML]{F1F1F1} \color{white} --} & \cc{F0AB94}{\color[HTML]{000000} $0.850$} & \cc{EDA38D}{\color[HTML]{000000} $0.852$} & \cc{E69280}{\color[HTML]{F1F1F1} $0.857$} & \cc{FFFFFF}{\color[HTML]{F1F1F1} \color{white} --} & \cc{FFFFFF}{\color[HTML]{F1F1F1} \color{white} --} & \cc{FAD3BF}{\color[HTML]{000000} $0.252$} & \cc{F9C4AA}{\color[HTML]{000000} $0.237$} & \cc{F4B49A}{\color[HTML]{000000} $0.226$} \\
\bottomrule
\end{tabular}}
\endgroup
}
\end{subtable}
\end{table}

\section{Initialization and Training Times}
In~\cref{tab:init_times} we report the mean times needed to produce the initialization with each method across all ScanNet++ (both splits), Tanks\&Temples, and MipNerf360 scenes. Please note that these times exclude the time needed to load the dataset and store the results to disk. In~\cref{tab:practical_train_times}, we report mean training times using the evaluated initialization and densification methods. While starting with more points\footnote{Usually equal to the final number of Gaussians in the scene for this experiment, see~\cref{sec:protocol:practical_init}.} results in slightly increased training times, the differences are not drastic, and do not, in our opinion, outweigh the benefits of using dense initialization.

\begin{table}[tbp]
\captionsetup{skip=6pt}
\centering
\caption{Mean initialization times, excluding time required to load the datasets and store the output to disk. Results averaged over all used scenes on ScanNet++, MipNerf360, and Tanks\&Temples.}
\label{tab:init_times}
\begin{tabular}{lc}
\toprule
Method & Mean initialization time (s) \\
\midrule
$\text{EDGS}^*$ & 103.74 \\
Monodepth & 98.55 \\
DA3 & 71.56 \\
DA3 (no f.r.) & 66.23 \\
$\text{DA3}^\text{GS}$ & 37.12 \\
\bottomrule
\end{tabular}
\end{table}
\begin{table}[tbp]
\captionsetup{skip=6pt}
\centering
\caption{Mean training times across ScanNet++ (both splits), MipNerf360, and Tanks\&Temples using configurations and init methods evaluated in~\cref{subsec:eval:practical_init}.}
\label{tab:practical_train_times}
{\small
\begingroup \setlength{\tabcolsep}{2.5\tabcolsep}
\resizebox{0.75\columnwidth}{!}{\begin{tabular}{lcccccc}
\toprule
 \textbf{Train Time (min) ↓} & SfM & $\text{EDGS}^*$ & M. D. & DA3 & $\text{DA3}^\text{G.S.}$ & Laser \\
\midrule
AbsGS & \cc{FFFEFD}{\color[HTML]{000000} $10.8$} & \cc{FEF6F1}{\color[HTML]{000000} $10.5$} & \cc{FCFDFE}{\color[HTML]{000000} $10.9$} & \cc{F7FBFD}{\color[HTML]{000000} $11.0$} & \cc{D2E7F1}{\color[HTML]{000000} $12.0$} & \cc{FBD6C4}{\color[HTML]{000000} $9.6$} \\
INRIA & \cc{F9C8B0}{\color[HTML]{000000} $9.1$} & \cc{FDE8DD}{\color[HTML]{000000} $10.1$} & \cc{FCE7DC}{\color[HTML]{000000} $10.1$} & \cc{FDEAE0}{\color[HTML]{000000} $10.2$} & \cc{F8FBFD}{\color[HTML]{000000} $11.0$} & \cc{FACBB4}{\color[HTML]{000000} $9.2$} \\
MCMC & \cc{93C0DC}{\color[HTML]{000000} $13.3$} & \cc{B6D8E8}{\color[HTML]{000000} $12.7$} & \cc{C0DDEB}{\color[HTML]{000000} $12.5$} & \cc{B6D8E8}{\color[HTML]{000000} $12.7$} & \cc{5C99C7}{\color[HTML]{F1F1F1} $14.2$} & \cc{EDF5FA}{\color[HTML]{000000} $11.3$} \\
IDHFR & \cc{FBD5C2}{\color[HTML]{000000} $9.5$} & \cc{F8C0A4}{\color[HTML]{000000} $8.9$} & \cc{FACDB7}{\color[HTML]{000000} $9.3$} & \cc{FAD0BB}{\color[HTML]{000000} $9.4$} & \cc{FDEBE2}{\color[HTML]{000000} $10.2$} & \cc{FACDB7}{\color[HTML]{000000} $9.3$} \\
RevDGS & \cc{F5F9FC}{\color[HTML]{000000} $11.1$} & \cc{EBF4F9}{\color[HTML]{000000} $11.3$} & \cc{D7E9F3}{\color[HTML]{000000} $11.9$} & \cc{D7E9F3}{\color[HTML]{000000} $11.8$} & \cc{C5DFED}{\color[HTML]{000000} $12.3$} & \cc{FCE0D2}{\color[HTML]{000000} $9.9$} \\
No D. & \cc{FFFFFF}{\color[HTML]{F1F1F1} \color{white} --} & \cc{DC796D}{\color[HTML]{F1F1F1} $7.8$} & \cc{D5675E}{\color[HTML]{F1F1F1} $7.5$} & \cc{DC776B}{\color[HTML]{F1F1F1} $7.8$} & \cc{E08273}{\color[HTML]{F1F1F1} $7.9$} & \cc{ECA18C}{\color[HTML]{000000} $8.4$} \\
\bottomrule
\end{tabular}}
\endgroup
}
\end{table}

\section{Additional Results with Practical Init Methods}
\label{additional_practical_results}
In this section, we include results and analysis for additional experiments with practical initialization methods that could not be included in the main paper due to length constraints or scope limitations.

\PAR{Practical initialization methods at half init size.}
As an extension of our experiments in~\cref{subsec:eval:practical_init}, we evaluate what effect  initialization size has when using initialization methods other than laser scans. We do this by training with half the initialization size used in our main experiment (see~\cref{sec:protocol:practical_init}). Results are reported in~\Cref{tab:practical_half_size}.
In our experiment with laser scan evaluation (\cref{tab:laser_scan_main}), we did not see large differences when increasing the initialization size in most cases, though there was a trend of slight improvements, especially on underconstrained datasets.
The differences in~\Cref{tab:practical_half_size} are once again mostly minor and we don't see clear improvements that track across densification strategies and datasets.
Interestingly, $\text{DA3}^\text{GS}$ seems to be the method most resistant to subsampling the initialization -- the metric values stay nearly constant despite halving the initialization size.
The likely cause is that $\text{DA3}^\text{GS}$ outputs all Gaussian parameters directly, and the initial scale of the primitives is not influenced by nearest neighbor distances.
This is in contrast to the other methods tested here (Monodepth, DA3, Laser Scans), where subsampling the initialization directly affects the initial Gaussians' scales.
This agrees with our previous findings, as, in \Cref{subsec:eval:practical_init}, we have identified scale to be the most important initialization parameter besides point positions.
Another explanation could be that only a small fraction of the Gaussians output by $\text{DA3}^\text{GS}$ meaningfully contribute to the scene (thus subsampling would be unlikely to discard an ``important'' Gaussian), but we have not investigated this experimentally.
\begin{table}[tbp]
\centering
\caption{Evaluation of practical initialization methods with full and halved initialization size. ``0.5'' indicates half the number of initial points.}
\label{tab:practical_half_size}
\begin{subtable}[t]{0.75\linewidth}
\centering
\caption{ScanNet++}
{\small
\begingroup \setlength{\tabcolsep}{2.5\tabcolsep}
\resizebox{\linewidth}{!}{\begin{tabular}{lcc|cc|cc|cc}
\toprule
 \textbf{PSNR ↑} & $\text{M.D.}^{0.5}$ & M.D. & $\text{DA3}^{0.5}$ & DA3 & $\text{DA3}_\text{GS}^{0.5}$ & $\text{DA3}_\text{GS}$ & $\text{Laser}^{0.5}$ & Laser \\
\midrule
AbsGS & \cc{D0E5F0}{\color[HTML]{000000} $22.75$} & \cc{F0F7FA}{\color[HTML]{000000} $22.91$} & \cc{FCE3D6}{\color[HTML]{000000} $23.16$} & \cc{EAF3F8}{\color[HTML]{000000} $22.88$} & \cc{FACDB7}{\color[HTML]{000000} $23.30$} & \cc{F9C2A7}{\color[HTML]{000000} $23.37$} & \cc{EBF4F9}{\color[HTML]{000000} $22.88$} & \cc{FDE9DF}{\color[HTML]{000000} $23.13$} \\
IDHFR & \cc{FBD7C5}{\color[HTML]{000000} $23.23$} & \cc{F9C5AB}{\color[HTML]{000000} $23.35$} & \cc{F1AD96}{\color[HTML]{000000} $23.44$} & \cc{F9C2A7}{\color[HTML]{000000} $23.37$} & \cc{E08273}{\color[HTML]{F1F1F1} $23.57$} & \cc{D5675E}{\color[HTML]{F1F1F1} $23.65$} & \cc{FAD3BF}{\color[HTML]{000000} $23.26$} & \cc{F8BEA3}{\color[HTML]{000000} $23.39$} \\
MCMC & \cc{A3CBE2}{\color[HTML]{000000} $22.55$} & \cc{ABD1E5}{\color[HTML]{000000} $22.57$} & \cc{C7E1EE}{\color[HTML]{000000} $22.70$} & \cc{BDDBEB}{\color[HTML]{000000} $22.65$} & \cc{CAE2EF}{\color[HTML]{000000} $22.72$} & \cc{ADD2E6}{\color[HTML]{000000} $22.58$} & \cc{5C99C7}{\color[HTML]{F1F1F1} $22.32$} & \cc{A5CDE3}{\color[HTML]{000000} $22.55$} \\
\midrule
 \textbf{SSIM ↑} & $\text{M.D.}^{0.5}$ & M.D. & $\text{DA3}^{0.5}$ & DA3 & $\text{DA3}_\text{GS}^{0.5}$ & $\text{DA3}_\text{GS}$ & $\text{Laser}^{0.5}$ & Laser \\
\midrule
AbsGS & \cc{B1D5E7}{\color[HTML]{000000} $0.872$} & \cc{C0DDEB}{\color[HTML]{000000} $0.873$} & \cc{D6E9F2}{\color[HTML]{000000} $0.874$} & \cc{C2DEEC}{\color[HTML]{000000} $0.873$} & \cc{5C99C7}{\color[HTML]{F1F1F1} $0.870$} & \cc{70A7CE}{\color[HTML]{F1F1F1} $0.871$} & \cc{C8E1EE}{\color[HTML]{000000} $0.873$} & \cc{FFFDFB}{\color[HTML]{000000} $0.875$} \\
IDHFR & \cc{F9C3A8}{\color[HTML]{000000} $0.878$} & \cc{E79482}{\color[HTML]{F1F1F1} $0.879$} & \cc{E79482}{\color[HTML]{F1F1F1} $0.879$} & \cc{D97166}{\color[HTML]{F1F1F1} $0.880$} & \cc{FCE0D2}{\color[HTML]{000000} $0.876$} & \cc{FDEAE0}{\color[HTML]{000000} $0.876$} & \cc{F7BCA1}{\color[HTML]{000000} $0.878$} & \cc{D5675E}{\color[HTML]{F1F1F1} $0.880$} \\
MCMC & \cc{FDE8DD}{\color[HTML]{000000} $0.876$} & \cc{FDECE3}{\color[HTML]{000000} $0.876$} & \cc{FDEDE5}{\color[HTML]{000000} $0.876$} & \cc{FEFAF7}{\color[HTML]{000000} $0.875$} & \cc{FCE3D6}{\color[HTML]{000000} $0.876$} & \cc{CFE5F0}{\color[HTML]{000000} $0.873$} & \cc{9FC9E0}{\color[HTML]{000000} $0.872$} & \cc{FEF4EF}{\color[HTML]{000000} $0.876$} \\
\midrule
 \textbf{LPIPS ↓} & $\text{M.D.}^{0.5}$ & M.D. & $\text{DA3}^{0.5}$ & DA3 & $\text{DA3}_\text{GS}^{0.5}$ & $\text{DA3}_\text{GS}$ & $\text{Laser}^{0.5}$ & Laser \\
\midrule
AbsGS & \cc{5C99C7}{\color[HTML]{F1F1F1} $0.249$} & \cc{8DBCDA}{\color[HTML]{000000} $0.246$} & \cc{B1D5E7}{\color[HTML]{000000} $0.244$} & \cc{95C2DD}{\color[HTML]{000000} $0.245$} & \cc{FDEFE7}{\color[HTML]{000000} $0.235$} & \cc{FDECE3}{\color[HTML]{000000} $0.235$} & \cc{D1E6F1}{\color[HTML]{000000} $0.241$} & \cc{F6FAFC}{\color[HTML]{000000} $0.238$} \\
IDHFR & \cc{FEF9F6}{\color[HTML]{000000} $0.236$} & \cc{FCE3D6}{\color[HTML]{000000} $0.234$} & \cc{FBDAC9}{\color[HTML]{000000} $0.233$} & \cc{FCE1D3}{\color[HTML]{000000} $0.233$} & \cc{DA7368}{\color[HTML]{F1F1F1} $0.226$} & \cc{D5675E}{\color[HTML]{F1F1F1} $0.225$} & \cc{FAD1BC}{\color[HTML]{000000} $0.232$} & \cc{F6BA9F}{\color[HTML]{000000} $0.229$} \\
MCMC & \cc{CDE4F0}{\color[HTML]{000000} $0.241$} & \cc{CFE5F0}{\color[HTML]{000000} $0.241$} & \cc{C0DDEB}{\color[HTML]{000000} $0.243$} & \cc{DBEBF4}{\color[HTML]{000000} $0.240$} & \cc{FEF7F3}{\color[HTML]{000000} $0.236$} & \cc{F6FAFC}{\color[HTML]{000000} $0.238$} & \cc{8FBDDA}{\color[HTML]{000000} $0.246$} & \cc{FEF5F0}{\color[HTML]{000000} $0.236$} \\
\bottomrule
\end{tabular}}
\endgroup
}
\end{subtable}
\begin{subtable}[t]{0.75\linewidth}
\centering
\caption{ScanNet++ (On-Trajectory)}
{\small
\begingroup \setlength{\tabcolsep}{2.5\tabcolsep}
\resizebox{\linewidth}{!}{\begin{tabular}{lcc|cc|cc|cc}
\toprule
 \textbf{PSNR ↑} & $\text{M.D.}^{0.5}$ & M.D. & $\text{DA3}^{0.5}$ & DA3 & $\text{DA3}_\text{GS}^{0.5}$ & $\text{DA3}_\text{GS}$ & $\text{Laser}^{0.5}$ & Laser \\
\midrule
AbsGS & \cc{F3F9FB}{\color[HTML]{000000} $33.40$} & \cc{7FB2D4}{\color[HTML]{000000} $32.89$} & \cc{DDEDF5}{\color[HTML]{000000} $33.28$} & \cc{7CAFD3}{\color[HTML]{F1F1F1} $32.88$} & \cc{FEFAF7}{\color[HTML]{000000} $33.50$} & \cc{FCFDFE}{\color[HTML]{000000} $33.45$} & \cc{FFFBF9}{\color[HTML]{000000} $33.49$} & \cc{F7FBFD}{\color[HTML]{000000} $33.42$} \\
IDHFR & \cc{EB9F8A}{\color[HTML]{000000} $33.98$} & \cc{FBDDCE}{\color[HTML]{000000} $33.69$} & \cc{F3B199}{\color[HTML]{000000} $33.92$} & \cc{FBD8C7}{\color[HTML]{000000} $33.72$} & \cc{D5675E}{\color[HTML]{F1F1F1} $34.16$} & \cc{DB7569}{\color[HTML]{F1F1F1} $34.11$} & \cc{E89784}{\color[HTML]{000000} $34.01$} & \cc{D97166}{\color[HTML]{F1F1F1} $34.13$} \\
MCMC & \cc{E2EFF6}{\color[HTML]{000000} $33.31$} & \cc{6AA3CC}{\color[HTML]{F1F1F1} $32.82$} & \cc{9DC7E0}{\color[HTML]{000000} $32.99$} & \cc{5C99C7}{\color[HTML]{F1F1F1} $32.77$} & \cc{CDE4F0}{\color[HTML]{000000} $33.20$} & \cc{D2E7F1}{\color[HTML]{000000} $33.23$} & \cc{BAD9EA}{\color[HTML]{000000} $33.10$} & \cc{D1E6F1}{\color[HTML]{000000} $33.22$} \\
\midrule
 \textbf{SSIM ↑} & $\text{M.D.}^{0.5}$ & M.D. & $\text{DA3}^{0.5}$ & DA3 & $\text{DA3}_\text{GS}^{0.5}$ & $\text{DA3}_\text{GS}$ & $\text{Laser}^{0.5}$ & Laser \\
\midrule
AbsGS & \cc{E8F3F8}{\color[HTML]{000000} $0.952$} & \cc{5C99C7}{\color[HTML]{F1F1F1} $0.950$} & \cc{85B6D7}{\color[HTML]{000000} $0.950$} & \cc{629DC9}{\color[HTML]{F1F1F1} $0.950$} & \cc{B7D8E9}{\color[HTML]{000000} $0.951$} & \cc{85B6D7}{\color[HTML]{000000} $0.950$} & \cc{FFFFFE}{\color[HTML]{000000} $0.952$} & \cc{FDFEFE}{\color[HTML]{000000} $0.952$} \\
IDHFR & \cc{EB9D88}{\color[HTML]{000000} $0.954$} & \cc{FCE5D9}{\color[HTML]{000000} $0.953$} & \cc{FBD6C4}{\color[HTML]{000000} $0.953$} & \cc{FCE5D9}{\color[HTML]{000000} $0.953$} & \cc{F9C6AD}{\color[HTML]{000000} $0.953$} & \cc{FACEB8}{\color[HTML]{000000} $0.953$} & \cc{EB9D88}{\color[HTML]{000000} $0.954$} & \cc{D5675E}{\color[HTML]{F1F1F1} $0.954$} \\
MCMC & \cc{F9C6AD}{\color[HTML]{000000} $0.953$} & \cc{E6F1F7}{\color[HTML]{000000} $0.952$} & \cc{D3E7F2}{\color[HTML]{000000} $0.951$} & \cc{DDEDF5}{\color[HTML]{000000} $0.951$} & \cc{E5F1F7}{\color[HTML]{000000} $0.952$} & \cc{F5F9FC}{\color[HTML]{000000} $0.952$} & \cc{FEF5F0}{\color[HTML]{000000} $0.952$} & \cc{F9C7AE}{\color[HTML]{000000} $0.953$} \\
\midrule
 \textbf{LPIPS ↓} & $\text{M.D.}^{0.5}$ & M.D. & $\text{DA3}^{0.5}$ & DA3 & $\text{DA3}_\text{GS}^{0.5}$ & $\text{DA3}_\text{GS}$ & $\text{Laser}^{0.5}$ & Laser \\
\midrule
AbsGS & \cc{D2E7F1}{\color[HTML]{000000} $0.098$} & \cc{C5DFED}{\color[HTML]{000000} $0.099$} & \cc{649FCA}{\color[HTML]{F1F1F1} $0.102$} & \cc{C8E1EE}{\color[HTML]{000000} $0.099$} & \cc{EAF3F8}{\color[HTML]{000000} $0.097$} & \cc{ECF5F9}{\color[HTML]{000000} $0.097$} & \cc{FDEDE5}{\color[HTML]{000000} $0.095$} & \cc{FAD2BE}{\color[HTML]{000000} $0.093$} \\
IDHFR & \cc{FACCB5}{\color[HTML]{000000} $0.093$} & \cc{FBD8C7}{\color[HTML]{000000} $0.094$} & \cc{FEF7F3}{\color[HTML]{000000} $0.095$} & \cc{FBD6C4}{\color[HTML]{000000} $0.093$} & \cc{F8C0A4}{\color[HTML]{000000} $0.092$} & \cc{F2AF97}{\color[HTML]{000000} $0.092$} & \cc{EB9F8A}{\color[HTML]{000000} $0.091$} & \cc{D5675E}{\color[HTML]{F1F1F1} $0.089$} \\
MCMC & \cc{FCE7DC}{\color[HTML]{000000} $0.094$} & \cc{BFDCEB}{\color[HTML]{000000} $0.099$} & \cc{81B4D5}{\color[HTML]{000000} $0.101$} & \cc{BDDBEB}{\color[HTML]{000000} $0.099$} & \cc{5C99C7}{\color[HTML]{F1F1F1} $0.102$} & \cc{9DC7E0}{\color[HTML]{000000} $0.100$} & \cc{E7F2F8}{\color[HTML]{000000} $0.097$} & \cc{FCE3D6}{\color[HTML]{000000} $0.094$} \\
\bottomrule
\end{tabular}}
\endgroup
}
\end{subtable}

\begin{subtable}[t]{0.495\linewidth}
\centering
\caption{Mip-NeRF 360}
{\small
\begingroup \setlength{\tabcolsep}{2.5\tabcolsep}
\resizebox{\linewidth}{!}{\begin{tabular}{lcc|cc|cc}
\toprule
 \textbf{PSNR ↑} & $\text{M.D.}^{0.5}$ & M.D. & $\text{DA3}^{0.5}$ & DA3 & $\text{DA3}_\text{GS}^{0.5}$ & $\text{DA3}_\text{GS}$ \\
\midrule
AbsGS & \cc{F6BA9F}{\color[HTML]{000000} $27.64$} & \cc{F3B199}{\color[HTML]{000000} $27.67$} & \cc{76ABD1}{\color[HTML]{F1F1F1} $26.49$} & \cc{5C99C7}{\color[HTML]{F1F1F1} $26.40$} & \cc{FBDBCB}{\color[HTML]{000000} $27.43$} & \cc{FBD8C7}{\color[HTML]{000000} $27.45$} \\
IDHFR & \cc{D5675E}{\color[HTML]{F1F1F1} $27.93$} & \cc{D5675E}{\color[HTML]{F1F1F1} $27.93$} & \cc{FDE9DF}{\color[HTML]{000000} $27.32$} & \cc{FEFAF7}{\color[HTML]{000000} $27.20$} & \cc{FAD1BC}{\color[HTML]{000000} $27.50$} & \cc{FACBB4}{\color[HTML]{000000} $27.55$} \\
MCMC & \cc{DF8072}{\color[HTML]{F1F1F1} $27.85$} & \cc{EEA58F}{\color[HTML]{000000} $27.72$} & \cc{F2F8FB}{\color[HTML]{000000} $27.09$} & \cc{9BC6DF}{\color[HTML]{000000} $26.63$} & \cc{ECA18C}{\color[HTML]{000000} $27.73$} & \cc{F5B89E}{\color[HTML]{000000} $27.65$} \\
\midrule
 \textbf{SSIM ↑} & $\text{M.D.}^{0.5}$ & M.D. & $\text{DA3}^{0.5}$ & DA3 & $\text{DA3}_\text{GS}^{0.5}$ & $\text{DA3}_\text{GS}$ \\
\midrule
AbsGS & \cc{E89784}{\color[HTML]{000000} $0.827$} & \cc{E79482}{\color[HTML]{F1F1F1} $0.828$} & \cc{99C4DE}{\color[HTML]{000000} $0.807$} & \cc{6CA4CD}{\color[HTML]{F1F1F1} $0.803$} & \cc{FCE5D9}{\color[HTML]{000000} $0.820$} & \cc{FDEAE0}{\color[HTML]{000000} $0.819$} \\
IDHFR & \cc{D5675E}{\color[HTML]{F1F1F1} $0.830$} & \cc{D5675E}{\color[HTML]{F1F1F1} $0.830$} & \cc{FBDAC9}{\color[HTML]{000000} $0.821$} & \cc{FCE6DB}{\color[HTML]{000000} $0.820$} & \cc{FEF3ED}{\color[HTML]{000000} $0.818$} & \cc{FEF1EA}{\color[HTML]{000000} $0.818$} \\
MCMC & \cc{EB9D88}{\color[HTML]{000000} $0.827$} & \cc{FBD9C8}{\color[HTML]{000000} $0.821$} & \cc{F2F8FB}{\color[HTML]{000000} $0.815$} & \cc{5C99C7}{\color[HTML]{F1F1F1} $0.802$} & \cc{F9CAB2}{\color[HTML]{000000} $0.823$} & \cc{FDE8DD}{\color[HTML]{000000} $0.819$} \\
\midrule
 \textbf{LPIPS ↓} & $\text{M.D.}^{0.5}$ & M.D. & $\text{DA3}^{0.5}$ & DA3 & $\text{DA3}_\text{GS}^{0.5}$ & $\text{DA3}_\text{GS}$ \\
\midrule
AbsGS & \cc{F9C5AB}{\color[HTML]{000000} $0.138$} & \cc{F4B49A}{\color[HTML]{000000} $0.136$} & \cc{76ABD1}{\color[HTML]{F1F1F1} $0.153$} & \cc{5C99C7}{\color[HTML]{F1F1F1} $0.154$} & \cc{FACEB8}{\color[HTML]{000000} $0.138$} & \cc{F9C8B0}{\color[HTML]{000000} $0.138$} \\
IDHFR & \cc{DB7569}{\color[HTML]{F1F1F1} $0.133$} & \cc{D86D63}{\color[HTML]{F1F1F1} $0.133$} & \cc{FAD0BB}{\color[HTML]{000000} $0.139$} & \cc{FBD9C8}{\color[HTML]{000000} $0.140$} & \cc{FACEB8}{\color[HTML]{000000} $0.138$} & \cc{F9C9B1}{\color[HTML]{000000} $0.138$} \\
MCMC & \cc{D5675E}{\color[HTML]{F1F1F1} $0.133$} & \cc{E28878}{\color[HTML]{F1F1F1} $0.134$} & \cc{F5F9FC}{\color[HTML]{000000} $0.144$} & \cc{70A7CE}{\color[HTML]{F1F1F1} $0.153$} & \cc{FBD5C2}{\color[HTML]{000000} $0.139$} & \cc{FAD3BF}{\color[HTML]{000000} $0.139$} \\
\bottomrule
\end{tabular}}
\endgroup
}
\end{subtable}
\begin{subtable}[t]{0.495\linewidth}
\centering
\caption{Tanks and Temples}
{\small
\begingroup \setlength{\tabcolsep}{2.5\tabcolsep}
\resizebox{\linewidth}{!}{\begin{tabular}{lcc|cc|cc}
\toprule
 \textbf{PSNR ↑} & $\text{M.D.}^{0.5}$ & M.D. & $\text{DA3}^{0.5}$ & DA3 & $\text{DA3}_\text{GS}^{0.5}$ & $\text{DA3}_\text{GS}$ \\
\midrule
AbsGS & \cc{BDDBEB}{\color[HTML]{000000} $23.05$} & \cc{F0F7FA}{\color[HTML]{000000} $23.39$} & \cc{5C99C7}{\color[HTML]{F1F1F1} $22.61$} & \cc{9FC9E0}{\color[HTML]{000000} $22.90$} & \cc{FFFDFB}{\color[HTML]{000000} $23.52$} & \cc{FDF0E9}{\color[HTML]{000000} $23.63$} \\
IDHFR & \cc{FACEB8}{\color[HTML]{000000} $23.91$} & \cc{FAD2BE}{\color[HTML]{000000} $23.87$} & \cc{FEF6F1}{\color[HTML]{000000} $23.57$} & \cc{FEF3ED}{\color[HTML]{000000} $23.60$} & \cc{FCE0D2}{\color[HTML]{000000} $23.76$} & \cc{FBD4C1}{\color[HTML]{000000} $23.85$} \\
MCMC & \cc{E18475}{\color[HTML]{F1F1F1} $24.26$} & \cc{F4B49A}{\color[HTML]{000000} $24.08$} & \cc{FBD6C4}{\color[HTML]{000000} $23.84$} & \cc{FDE9DF}{\color[HTML]{000000} $23.69$} & \cc{D5675E}{\color[HTML]{F1F1F1} $24.38$} & \cc{DA7368}{\color[HTML]{F1F1F1} $24.33$} \\
\midrule
 \textbf{SSIM ↑} & $\text{M.D.}^{0.5}$ & M.D. & $\text{DA3}^{0.5}$ & DA3 & $\text{DA3}_\text{GS}^{0.5}$ & $\text{DA3}_\text{GS}$ \\
\midrule
AbsGS & \cc{D1E6F1}{\color[HTML]{000000} $0.827$} & \cc{FEF8F4}{\color[HTML]{000000} $0.834$} & \cc{5C99C7}{\color[HTML]{F1F1F1} $0.818$} & \cc{ADD2E6}{\color[HTML]{000000} $0.824$} & \cc{F6FAFC}{\color[HTML]{000000} $0.832$} & \cc{FEFAF7}{\color[HTML]{000000} $0.834$} \\
IDHFR & \cc{EEA58F}{\color[HTML]{000000} $0.844$} & \cc{F0A992}{\color[HTML]{000000} $0.843$} & \cc{FDE9DF}{\color[HTML]{000000} $0.836$} & \cc{FCE3D6}{\color[HTML]{000000} $0.837$} & \cc{FBD9C8}{\color[HTML]{000000} $0.838$} & \cc{FAD1BC}{\color[HTML]{000000} $0.839$} \\
MCMC & \cc{D5675E}{\color[HTML]{F1F1F1} $0.848$} & \cc{E6907F}{\color[HTML]{F1F1F1} $0.845$} & \cc{FBD9C8}{\color[HTML]{000000} $0.838$} & \cc{FDEFE7}{\color[HTML]{000000} $0.835$} & \cc{D5675E}{\color[HTML]{F1F1F1} $0.848$} & \cc{D86F65}{\color[HTML]{F1F1F1} $0.847$} \\
\midrule
 \textbf{LPIPS ↓} & $\text{M.D.}^{0.5}$ & M.D. & $\text{DA3}^{0.5}$ & DA3 & $\text{DA3}_\text{GS}^{0.5}$ & $\text{DA3}_\text{GS}$ \\
\midrule
AbsGS & \cc{9BC6DF}{\color[HTML]{000000} $0.170$} & \cc{F8FBFD}{\color[HTML]{000000} $0.157$} & \cc{5C99C7}{\color[HTML]{F1F1F1} $0.176$} & \cc{DAEBF3}{\color[HTML]{000000} $0.162$} & \cc{FCE4D8}{\color[HTML]{000000} $0.151$} & \cc{FAD1BC}{\color[HTML]{000000} $0.148$} \\
IDHFR & \cc{E58E7D}{\color[HTML]{F1F1F1} $0.140$} & \cc{E89784}{\color[HTML]{000000} $0.141$} & \cc{F9C6AD}{\color[HTML]{000000} $0.146$} & \cc{F8BEA3}{\color[HTML]{000000} $0.144$} & \cc{E38A7A}{\color[HTML]{F1F1F1} $0.140$} & \cc{DC796D}{\color[HTML]{F1F1F1} $0.138$} \\
MCMC & \cc{D5675E}{\color[HTML]{F1F1F1} $0.137$} & \cc{D5675E}{\color[HTML]{F1F1F1} $0.137$} & \cc{FAD2BE}{\color[HTML]{000000} $0.148$} & \cc{FBD6C4}{\color[HTML]{000000} $0.149$} & \cc{E99985}{\color[HTML]{000000} $0.141$} & \cc{E58E7D}{\color[HTML]{F1F1F1} $0.140$} \\
\bottomrule
\end{tabular}}
\endgroup
}
\end{subtable}
\end{table}

\PAR{Hybrid initialization with practical methods.} Besides decreasing initialization sizes, we have also evaluated using hybrid initialization (see~\cref{sec:protocol:datasets}) with Monodepth and DA3. We present the results of this experiment in~\cref{tab:practical_hybrid}. Please note that we only include methods which output point clouds, as it is not clear what the protocol for mixing point clouds and direct splat initializations should be.

On ScanNet++ we see a similar effect as with laser scans.
This is once again likely caused by the abundance of reflective surfaces -- while DA3 and Monodepth can usually reconstruct them,
they tend to predict points \textit{on} the surface of transparent objects, not behind them.
This hampers NVS performance, as the complex view-dependent effects can't be successfully encoded through spherical harmonics.
Evidently, providing points in those areas helps even if they are initially partially occluded by the dense initialization output.
The exception is Monodepth on off-trajectory ScanNet++, where we see a slight reduction in performance from including the SfM points, potentially because the harm of adding floaters (by introducing noisy SfM points) outweighs the benefits for reflective and transparent surfaces.
\begin{table}[tb]
\centering
\caption{Evaluation of practical initialization methods with and without hybrid initialization. ``+'' indicates that SfM points were included.}
\label{tab:practical_hybrid}
\begin{subtable}[t]{\linewidth}
\centering
\caption{ScanNet++}
{\small
\begingroup \setlength{\tabcolsep}{1.5\tabcolsep}
\resizebox{\linewidth}{!}{\begin{tabular}{l|cc|cc|cc|cc|cc|cc|cc|cc|cc}
\toprule
& \multicolumn{6}{c|}{\textbf{PSNR ↑}} & \multicolumn{6}{c|}{\textbf{SSIM ↑}} & \multicolumn{6}{c}{\textbf{LPIPS ↓}} \\
& M.D. & $\text{M.D.}^{+}$ & DA3 & $\text{DA3}^{+}$ & Laser & $\text{Laser}^{+}$ & M.D. & $\text{M.D.}^{+}$ & DA3 & $\text{DA3}^{+}$ & Laser & $\text{Laser}^{+}$ & M.D. & $\text{M.D.}^{+}$ & DA3 & $\text{DA3}^{+}$ & Laser & $\text{Laser}^{+}$ \\
\midrule
AbsGS & \cc{E0EEF5}{\color[HTML]{000000} $22.91$} & \cc{B2D6E7}{\color[HTML]{000000} $22.75$} & \cc{D7E9F3}{\color[HTML]{000000} $22.88$} & \cc{FACDB7}{\color[HTML]{000000} $23.24$} & \cc{DDEDF5}{\color[HTML]{000000} $22.90$} & \cc{FCE7DC}{\color[HTML]{000000} $23.13$} & \cc{85B6D7}{\color[HTML]{000000} $0.873$} & \cc{5C99C7}{\color[HTML]{F1F1F1} $0.872$} & \cc{89B9D8}{\color[HTML]{000000} $0.873$} & \cc{FEFAF7}{\color[HTML]{000000} $0.876$} & \cc{ABD1E5}{\color[HTML]{000000} $0.873$} & \cc{E6F1F7}{\color[HTML]{000000} $0.875$} & \cc{9DC7E0}{\color[HTML]{000000} $0.246$} & \cc{5C99C7}{\color[HTML]{F1F1F1} $0.249$} & \cc{A5CDE3}{\color[HTML]{000000} $0.245$} & \cc{FFFFFE}{\color[HTML]{000000} $0.239$} & \cc{ABD1E5}{\color[HTML]{000000} $0.245$} & \cc{FDEFE7}{\color[HTML]{000000} $0.238$} \\
IDHFR & \cc{EFA791}{\color[HTML]{000000} $23.35$} & \cc{F6FAFC}{\color[HTML]{000000} $22.99$} & \cc{EB9F8A}{\color[HTML]{000000} $23.37$} & \cc{D5675E}{\color[HTML]{F1F1F1} $23.49$} & \cc{F2AF97}{\color[HTML]{000000} $23.33$} & \cc{E79482}{\color[HTML]{F1F1F1} $23.39$} & \cc{EB9F8A}{\color[HTML]{000000} $0.879$} & \cc{CAE2EF}{\color[HTML]{000000} $0.874$} & \cc{DA7368}{\color[HTML]{F1F1F1} $0.880$} & \cc{D97166}{\color[HTML]{F1F1F1} $0.880$} & \cc{EDA38D}{\color[HTML]{000000} $0.879$} & \cc{D5675E}{\color[HTML]{F1F1F1} $0.880$} & \cc{F9C3A8}{\color[HTML]{000000} $0.234$} & \cc{EAF3F8}{\color[HTML]{000000} $0.241$} & \cc{F9C1A6}{\color[HTML]{000000} $0.233$} & \cc{E28878}{\color[HTML]{F1F1F1} $0.231$} & \cc{F9C3A8}{\color[HTML]{000000} $0.234$} & \cc{D5675E}{\color[HTML]{F1F1F1} $0.229$} \\
MCMC & \cc{649FCA}{\color[HTML]{F1F1F1} $22.57$} & \cc{66A0CB}{\color[HTML]{F1F1F1} $22.57$} & \cc{87B8D7}{\color[HTML]{000000} $22.65$} & \cc{CDE4F0}{\color[HTML]{000000} $22.84$} & \cc{609CC8}{\color[HTML]{F1F1F1} $22.56$} & \cc{5C99C7}{\color[HTML]{F1F1F1} $22.55$} & \cc{FFFFFE}{\color[HTML]{000000} $0.876$} & \cc{E7F2F8}{\color[HTML]{000000} $0.875$} & \cc{EBF4F9}{\color[HTML]{000000} $0.875$} & \cc{F8C0A4}{\color[HTML]{000000} $0.878$} & \cc{F1F7FB}{\color[HTML]{000000} $0.875$} & \cc{F3F9FB}{\color[HTML]{000000} $0.876$} & \cc{E3F0F6}{\color[HTML]{000000} $0.241$} & \cc{FEFFFF}{\color[HTML]{000000} $0.239$} & \cc{F2F8FB}{\color[HTML]{000000} $0.240$} & \cc{F9C8B0}{\color[HTML]{000000} $0.234$} & \cc{E0EEF5}{\color[HTML]{000000} $0.241$} & \cc{FBD9C8}{\color[HTML]{000000} $0.236$} \\
\bottomrule
\end{tabular}}
\endgroup
}
\end{subtable}
\begin{subtable}[t]{\linewidth}
\centering
\caption{ScanNet++ (On-Trajectory)}
{\small
\begingroup \setlength{\tabcolsep}{1.5\tabcolsep}
\resizebox{\linewidth}{!}{\begin{tabular}{l|cc|cc|cc|cc|cc|cc|cc|cc|cc}
\toprule
& \multicolumn{6}{c|}{\textbf{PSNR ↑}} & \multicolumn{6}{c|}{\textbf{SSIM ↑}} & \multicolumn{6}{c}{\textbf{LPIPS ↓}} \\
& M.D. & $\text{M.D.}^{+}$ & DA3 & $\text{DA3}^{+}$ & Laser & $\text{Laser}^{+}$ & M.D. & $\text{M.D.}^{+}$ & DA3 & $\text{DA3}^{+}$ & Laser & $\text{Laser}^{+}$ & M.D. & $\text{M.D.}^{+}$ & DA3 & $\text{DA3}^{+}$ & Laser & $\text{Laser}^{+}$ \\
\midrule
AbsGS & \cc{81B4D5}{\color[HTML]{000000} $32.89$} & \cc{ECF5F9}{\color[HTML]{000000} $33.35$} & \cc{7DB1D4}{\color[HTML]{F1F1F1} $32.88$} & \cc{FFFCFA}{\color[HTML]{000000} $33.47$} & \cc{87B8D7}{\color[HTML]{000000} $32.91$} & \cc{F9FCFD}{\color[HTML]{000000} $33.42$} & \cc{5C99C7}{\color[HTML]{F1F1F1} $0.950$} & \cc{E2EFF6}{\color[HTML]{000000} $0.952$} & \cc{629DC9}{\color[HTML]{F1F1F1} $0.950$} & \cc{D7E9F3}{\color[HTML]{000000} $0.951$} & \cc{6CA4CD}{\color[HTML]{F1F1F1} $0.950$} & \cc{FDFEFE}{\color[HTML]{000000} $0.952$} & \cc{6CA4CD}{\color[HTML]{F1F1F1} $0.099$} & \cc{B2D6E7}{\color[HTML]{000000} $0.097$} & \cc{74AAD0}{\color[HTML]{F1F1F1} $0.099$} & \cc{81B4D5}{\color[HTML]{000000} $0.098$} & \cc{76ABD1}{\color[HTML]{F1F1F1} $0.099$} & \cc{FCE6DB}{\color[HTML]{000000} $0.093$} \\
IDHFR & \cc{FBD9C8}{\color[HTML]{000000} $33.69$} & \cc{DE7E70}{\color[HTML]{F1F1F1} $34.05$} & \cc{FBD4C1}{\color[HTML]{000000} $33.72$} & \cc{E08273}{\color[HTML]{F1F1F1} $34.04$} & \cc{FBD8C7}{\color[HTML]{000000} $33.70$} & \cc{D5675E}{\color[HTML]{F1F1F1} $34.13$} & \cc{FCE5D9}{\color[HTML]{000000} $0.953$} & \cc{E58E7D}{\color[HTML]{F1F1F1} $0.954$} & \cc{FCE5D9}{\color[HTML]{000000} $0.953$} & \cc{F1AD96}{\color[HTML]{000000} $0.954$} & \cc{FCE2D5}{\color[HTML]{000000} $0.953$} & \cc{D5675E}{\color[HTML]{F1F1F1} $0.954$} & \cc{FDEFE7}{\color[HTML]{000000} $0.094$} & \cc{FACFBA}{\color[HTML]{000000} $0.092$} & \cc{FDECE3}{\color[HTML]{000000} $0.093$} & \cc{FDE8DD}{\color[HTML]{000000} $0.093$} & \cc{FDF0E9}{\color[HTML]{000000} $0.094$} & \cc{D5675E}{\color[HTML]{F1F1F1} $0.089$} \\
MCMC & \cc{6CA4CD}{\color[HTML]{F1F1F1} $32.82$} & \cc{D1E6F1}{\color[HTML]{000000} $33.21$} & \cc{5C99C7}{\color[HTML]{F1F1F1} $32.77$} & \cc{DDEDF5}{\color[HTML]{000000} $33.27$} & \cc{5E9BC8}{\color[HTML]{F1F1F1} $32.77$} & \cc{D3E7F2}{\color[HTML]{000000} $33.22$} & \cc{E6F1F7}{\color[HTML]{000000} $0.952$} & \cc{FACEB8}{\color[HTML]{000000} $0.953$} & \cc{DDEDF5}{\color[HTML]{000000} $0.951$} & \cc{FBD9C8}{\color[HTML]{000000} $0.953$} & \cc{DAEBF3}{\color[HTML]{000000} $0.951$} & \cc{F9C7AE}{\color[HTML]{000000} $0.953$} & \cc{609CC8}{\color[HTML]{F1F1F1} $0.099$} & \cc{FBD9C8}{\color[HTML]{000000} $0.093$} & \cc{5C99C7}{\color[HTML]{F1F1F1} $0.099$} & \cc{E3F0F6}{\color[HTML]{000000} $0.095$} & \cc{609CC8}{\color[HTML]{F1F1F1} $0.099$} & \cc{FFFCFA}{\color[HTML]{000000} $0.094$} \\
\bottomrule
\end{tabular}}
\endgroup
}
\end{subtable}
\begin{subtable}[t]{0.8\linewidth}
\centering
\caption{Mip-NeRF 360}
{\small
\begingroup \setlength{\tabcolsep}{2.5\tabcolsep}
\resizebox{\linewidth}{!}{\begin{tabular}{l|cc|cc|cc|cc|cc|cc}
\toprule
& \multicolumn{4}{c|}{\textbf{PSNR ↑}} & \multicolumn{4}{c|}{\textbf{SSIM ↑}} & \multicolumn{4}{c}{\textbf{LPIPS ↓}} \\
& M.D. & $\text{M.D.}^{+}$ & DA3 & $\text{DA3}^{+}$ & M.D. & $\text{M.D.}^{+}$ & DA3 & $\text{DA3}^{+}$ & M.D. & $\text{M.D.}^{+}$ & DA3 & $\text{DA3}^{+}$ \\
\midrule
AbsGS & \cc{F4B69C}{\color[HTML]{000000} $27.67$} & \cc{F4B49A}{\color[HTML]{000000} $27.67$} & \cc{5C99C7}{\color[HTML]{F1F1F1} $26.40$} & \cc{FCE2D5}{\color[HTML]{000000} $27.39$} & \cc{E89784}{\color[HTML]{000000} $0.828$} & \cc{E69280}{\color[HTML]{F1F1F1} $0.828$} & \cc{6CA4CD}{\color[HTML]{F1F1F1} $0.803$} & \cc{FBD4C1}{\color[HTML]{000000} $0.822$} & \cc{F3B199}{\color[HTML]{000000} $0.136$} & \cc{EDA38D}{\color[HTML]{000000} $0.136$} & \cc{5C99C7}{\color[HTML]{F1F1F1} $0.154$} & \cc{F9C3A8}{\color[HTML]{000000} $0.137$} \\
IDHFR & \cc{D76B61}{\color[HTML]{F1F1F1} $27.93$} & \cc{D5675E}{\color[HTML]{F1F1F1} $27.94$} & \cc{FFFBF9}{\color[HTML]{000000} $27.20$} & \cc{ECA18C}{\color[HTML]{000000} $27.74$} & \cc{D66960}{\color[HTML]{F1F1F1} $0.830$} & \cc{D5675E}{\color[HTML]{F1F1F1} $0.831$} & \cc{FCE7DC}{\color[HTML]{000000} $0.820$} & \cc{EFA791}{\color[HTML]{000000} $0.827$} & \cc{D66960}{\color[HTML]{F1F1F1} $0.133$} & \cc{D5675E}{\color[HTML]{F1F1F1} $0.133$} & \cc{FBD8C7}{\color[HTML]{000000} $0.140$} & \cc{E38A7A}{\color[HTML]{F1F1F1} $0.135$} \\
MCMC & \cc{EFA791}{\color[HTML]{000000} $27.72$} & \cc{EB9D88}{\color[HTML]{000000} $27.76$} & \cc{9BC6DF}{\color[HTML]{000000} $26.63$} & \cc{F9C5AB}{\color[HTML]{000000} $27.59$} & \cc{FBDAC9}{\color[HTML]{000000} $0.821$} & \cc{FBD4C1}{\color[HTML]{000000} $0.822$} & \cc{5C99C7}{\color[HTML]{F1F1F1} $0.802$} & \cc{FCE6DB}{\color[HTML]{000000} $0.820$} & \cc{E18475}{\color[HTML]{F1F1F1} $0.134$} & \cc{DB7569}{\color[HTML]{F1F1F1} $0.134$} & \cc{70A7CE}{\color[HTML]{F1F1F1} $0.153$} & \cc{F8C0A4}{\color[HTML]{000000} $0.137$} \\
\bottomrule
\end{tabular}}
\endgroup
}
\end{subtable}
\begin{subtable}[t]{0.8\linewidth}
\centering
\caption{Tanks and Temples}
{\small
\begingroup \setlength{\tabcolsep}{2.5\tabcolsep}
\resizebox{\linewidth}{!}{\begin{tabular}{l|cc|cc|cc|cc|cc|cc}
\toprule
& \multicolumn{4}{c|}{\textbf{PSNR ↑}} & \multicolumn{4}{c|}{\textbf{SSIM ↑}} & \multicolumn{4}{c}{\textbf{LPIPS ↓}} \\
& M.D. & $\text{M.D.}^{+}$ & DA3 & $\text{DA3}^{+}$ & M.D. & $\text{M.D.}^{+}$ & DA3 & $\text{DA3}^{+}$ & M.D. & $\text{M.D.}^{+}$ & DA3 & $\text{DA3}^{+}$ \\
\midrule
AbsGS & \cc{E5F1F7}{\color[HTML]{000000} $23.39$} & \cc{D2E7F1}{\color[HTML]{000000} $23.31$} & \cc{5C99C7}{\color[HTML]{F1F1F1} $22.90$} & \cc{89B9D8}{\color[HTML]{000000} $23.03$} & \cc{F2F8FB}{\color[HTML]{000000} $0.834$} & \cc{F7FBFD}{\color[HTML]{000000} $0.834$} & \cc{5C99C7}{\color[HTML]{F1F1F1} $0.824$} & \cc{CCE3EF}{\color[HTML]{000000} $0.831$} & \cc{A3CBE2}{\color[HTML]{000000} $0.157$} & \cc{BDDBEB}{\color[HTML]{000000} $0.155$} & \cc{5C99C7}{\color[HTML]{F1F1F1} $0.162$} & \cc{93C0DC}{\color[HTML]{000000} $0.158$} \\
IDHFR & \cc{F9C4AA}{\color[HTML]{000000} $23.87$} & \cc{F9C5AB}{\color[HTML]{000000} $23.87$} & \cc{FEF3ED}{\color[HTML]{000000} $23.60$} & \cc{F9CAB2}{\color[HTML]{000000} $23.83$} & \cc{EDA38D}{\color[HTML]{000000} $0.843$} & \cc{EDA38D}{\color[HTML]{000000} $0.843$} & \cc{FDEEE6}{\color[HTML]{000000} $0.837$} & \cc{F2AF97}{\color[HTML]{000000} $0.843$} & \cc{F9C8B0}{\color[HTML]{000000} $0.141$} & \cc{F7BCA1}{\color[HTML]{000000} $0.139$} & \cc{FCE3D6}{\color[HTML]{000000} $0.144$} & \cc{F5B89E}{\color[HTML]{000000} $0.139$} \\
MCMC & \cc{DF8072}{\color[HTML]{F1F1F1} $24.08$} & \cc{D76B61}{\color[HTML]{F1F1F1} $24.13$} & \cc{FCE3D6}{\color[HTML]{000000} $23.69$} & \cc{D5675E}{\color[HTML]{F1F1F1} $24.15$} & \cc{DF8072}{\color[HTML]{F1F1F1} $0.845$} & \cc{D5675E}{\color[HTML]{F1F1F1} $0.847$} & \cc{FFFFFE}{\color[HTML]{000000} $0.835$} & \cc{DC776B}{\color[HTML]{F1F1F1} $0.846$} & \cc{E48C7B}{\color[HTML]{F1F1F1} $0.137$} & \cc{D5675E}{\color[HTML]{F1F1F1} $0.134$} & \cc{FBFDFE}{\color[HTML]{000000} $0.149$} & \cc{E08273}{\color[HTML]{F1F1F1} $0.136$} \\
\bottomrule
\end{tabular}}
\endgroup
}
\end{subtable}
\end{table}

On MipNerf360 and Tanks\&Temples, on the other hand, only DA3 benefits from using hybrid initialization, while Monodepth maintains very similar performance. 
This behavior is investigated in more detail in \Cref{init_precision_no_gt}, where we analyze Monodepth and DA3 outputs on these datasets.
In short, we find that the DA3 output quality on these datasets is often significantly worse, as it struggles to predict geometry which aligns with the SfM point clouds\footnote{While noisy and sparse, the SfM point clouds are our best indicator of true scene geometry. In all cases we've seen, the number of outliers was low enough to visually infer the scene structure.}, especially in scene regions far from the input cameras. Thus we believe that DA3 benefits from including the SfM points because its outputs are often not correctly aligned with the underlying scene geometry. Adding accurate initialization seeds in those regions thus improves NVS quality. Conversely, as Monodepth explicitly aligns the depth predictions to the SfM points, including them doesn't have much effect in the absence of reflective or transparent surfaces.

\section{Initialization Precision for Datasets Without GT}\label{init_precision_no_gt}

\begin{figure}[tbp]
    \centering
    \renewcommand{\arraystretch}{1.0} 
    
    \begin{tabular}{c m{0.4\textwidth} m{0.4\textwidth}}
        
        SfM \hspace{1.5em}& 
        \includegraphics[width=\linewidth]{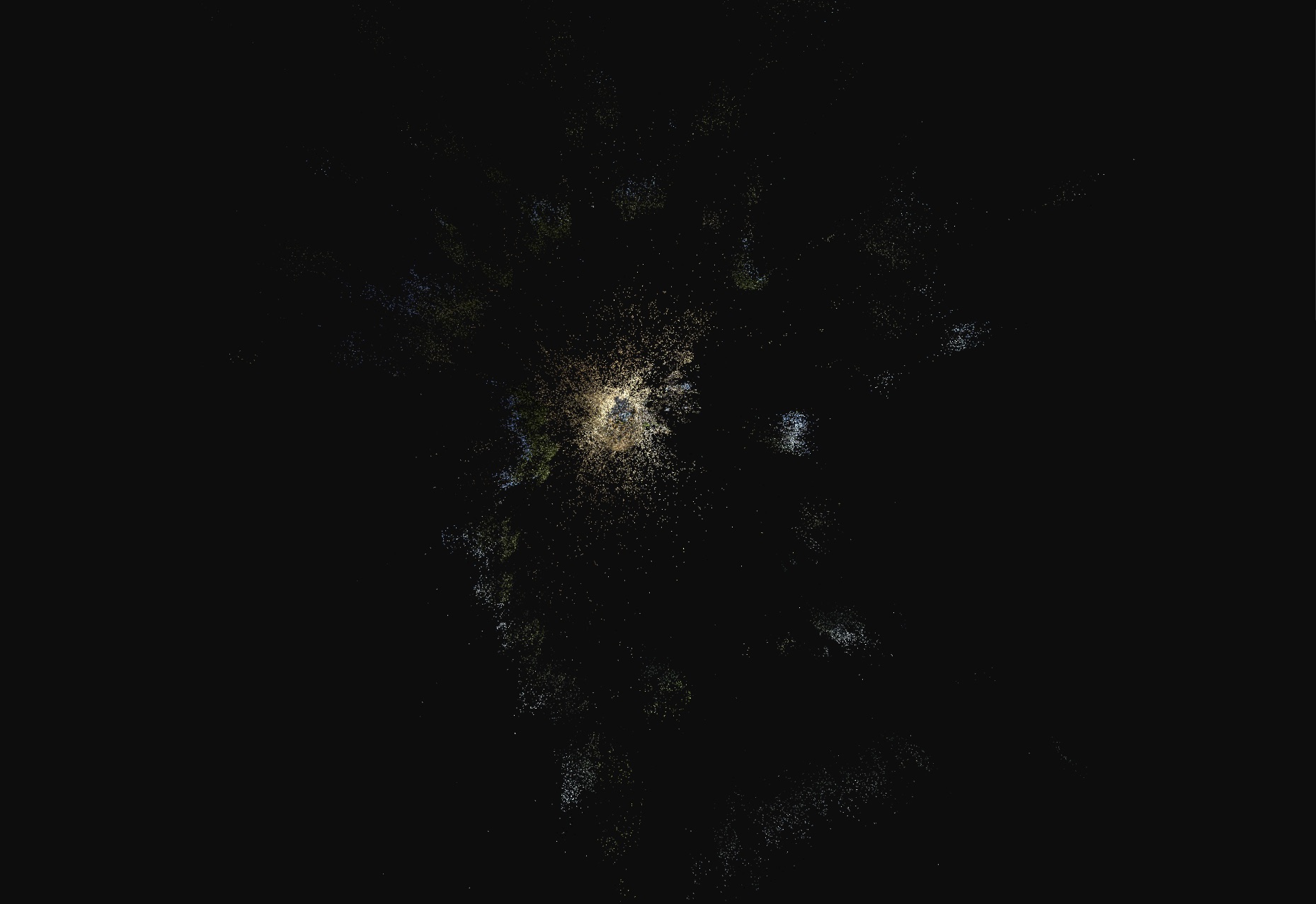} & 
        \includegraphics[width=\linewidth]{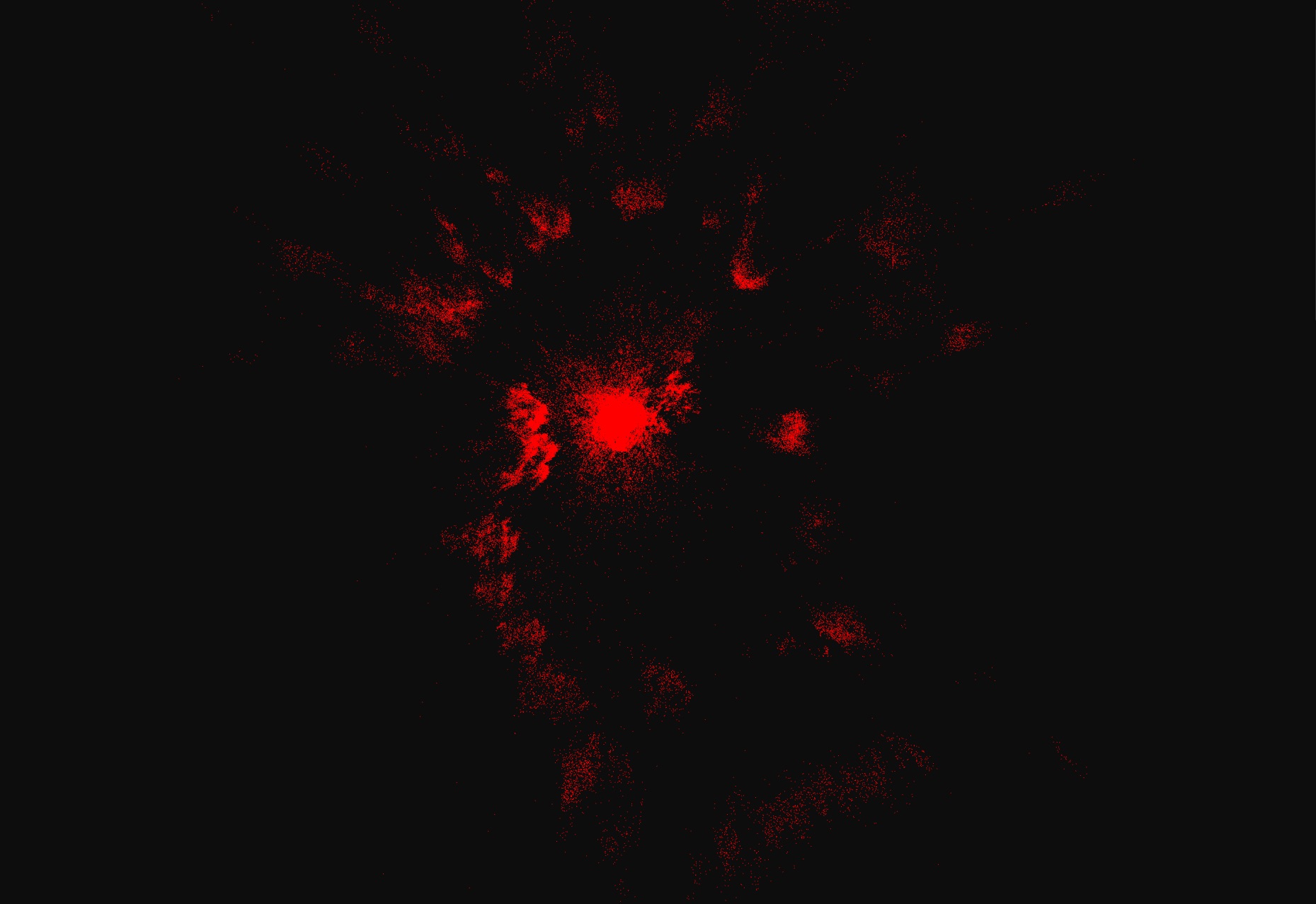} \\
        
        EDGS \hspace{1em}& 
        \includegraphics[width=\linewidth]{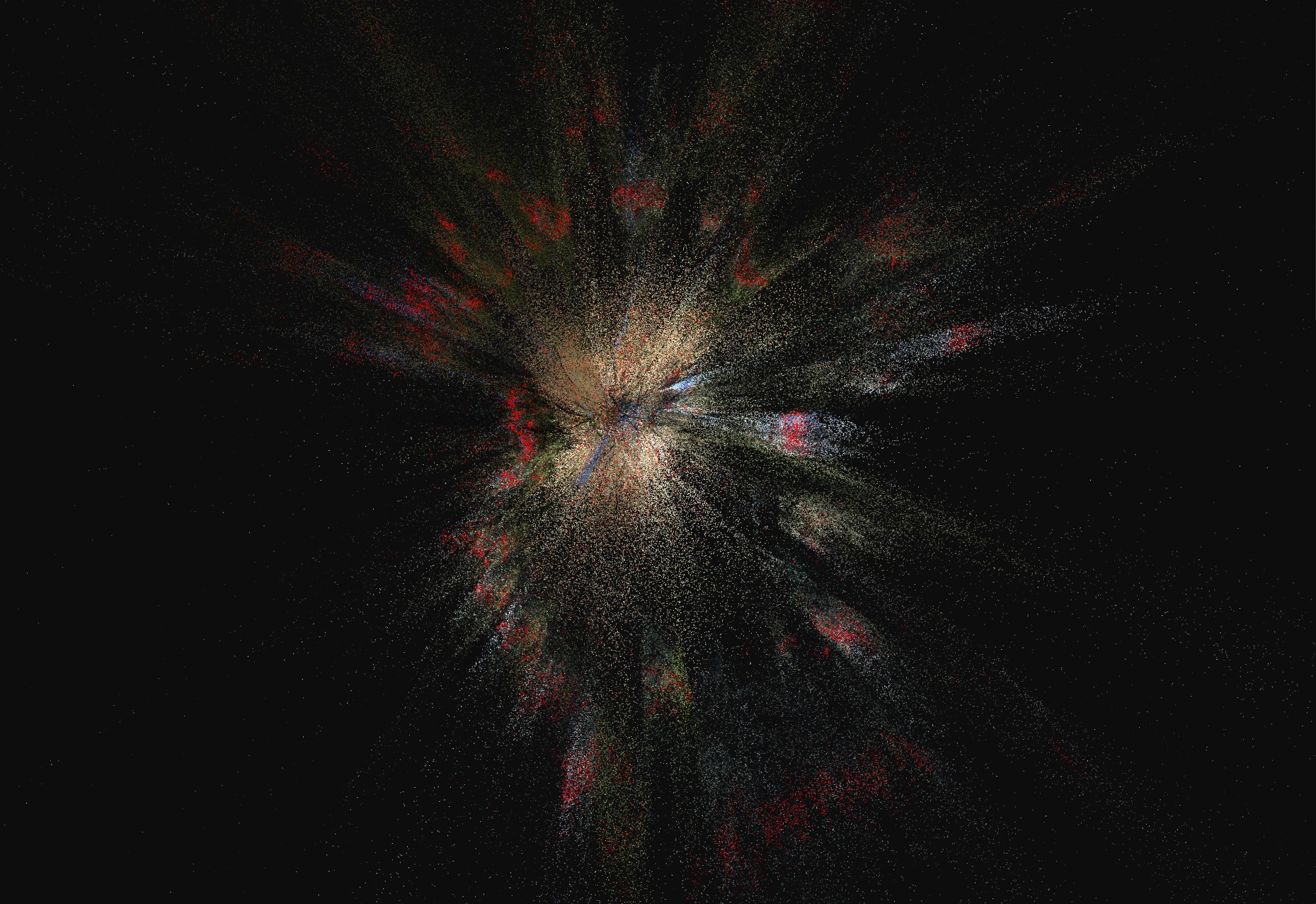} & 
        \includegraphics[width=\linewidth]{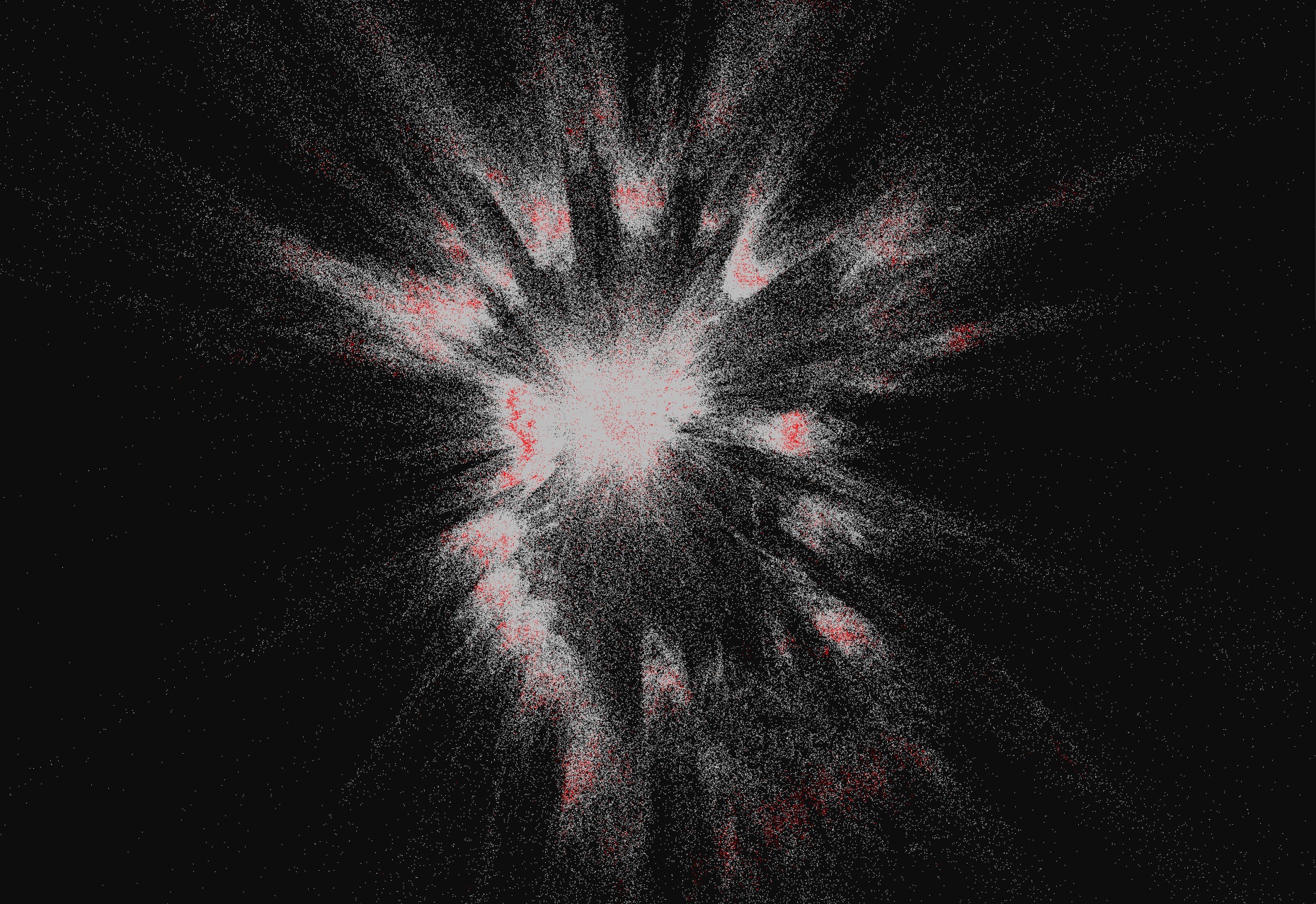} \\
        
        M.D. \hspace{1em}& 
        \includegraphics[width=\linewidth]{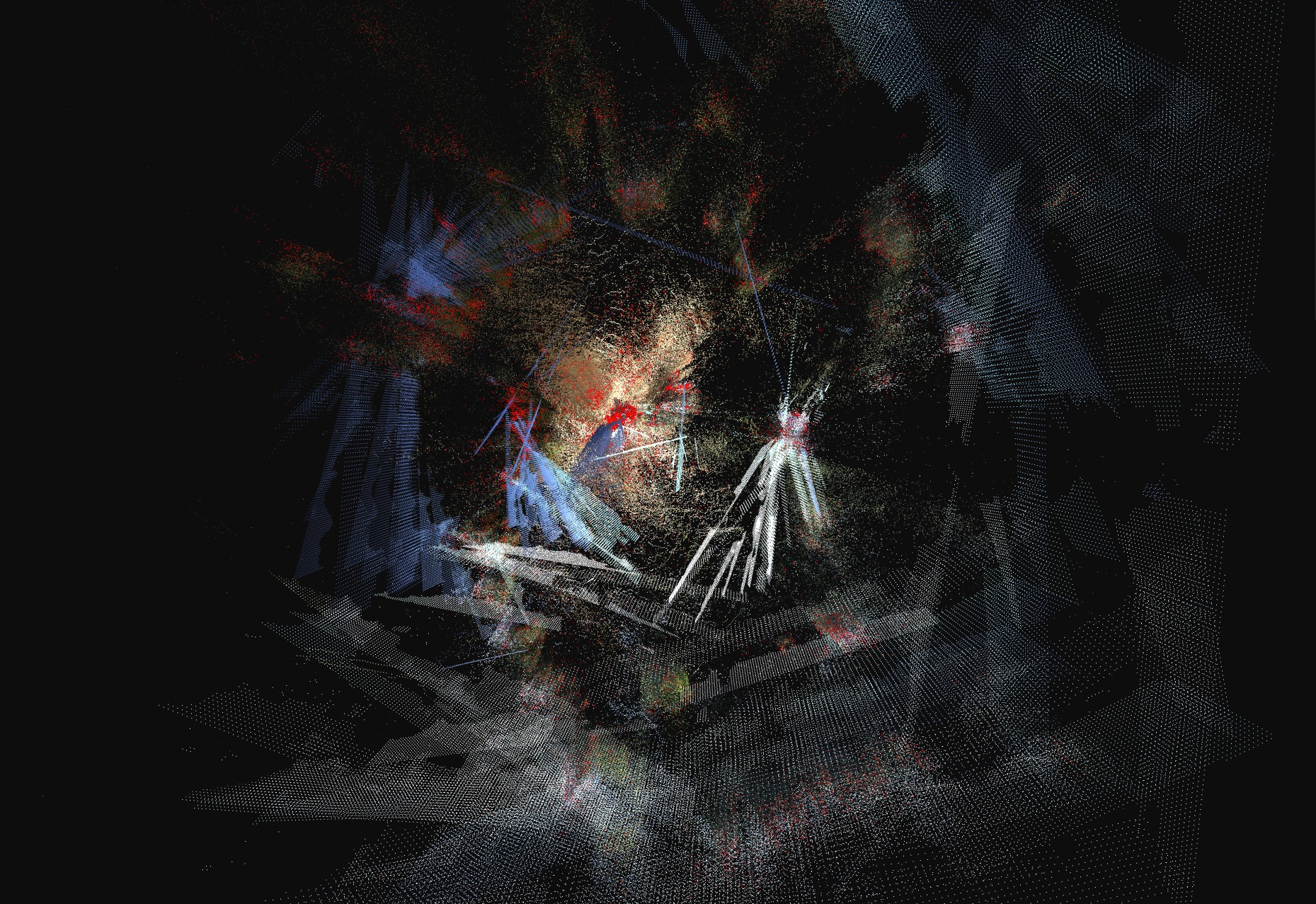} & 
        \includegraphics[width=\linewidth]{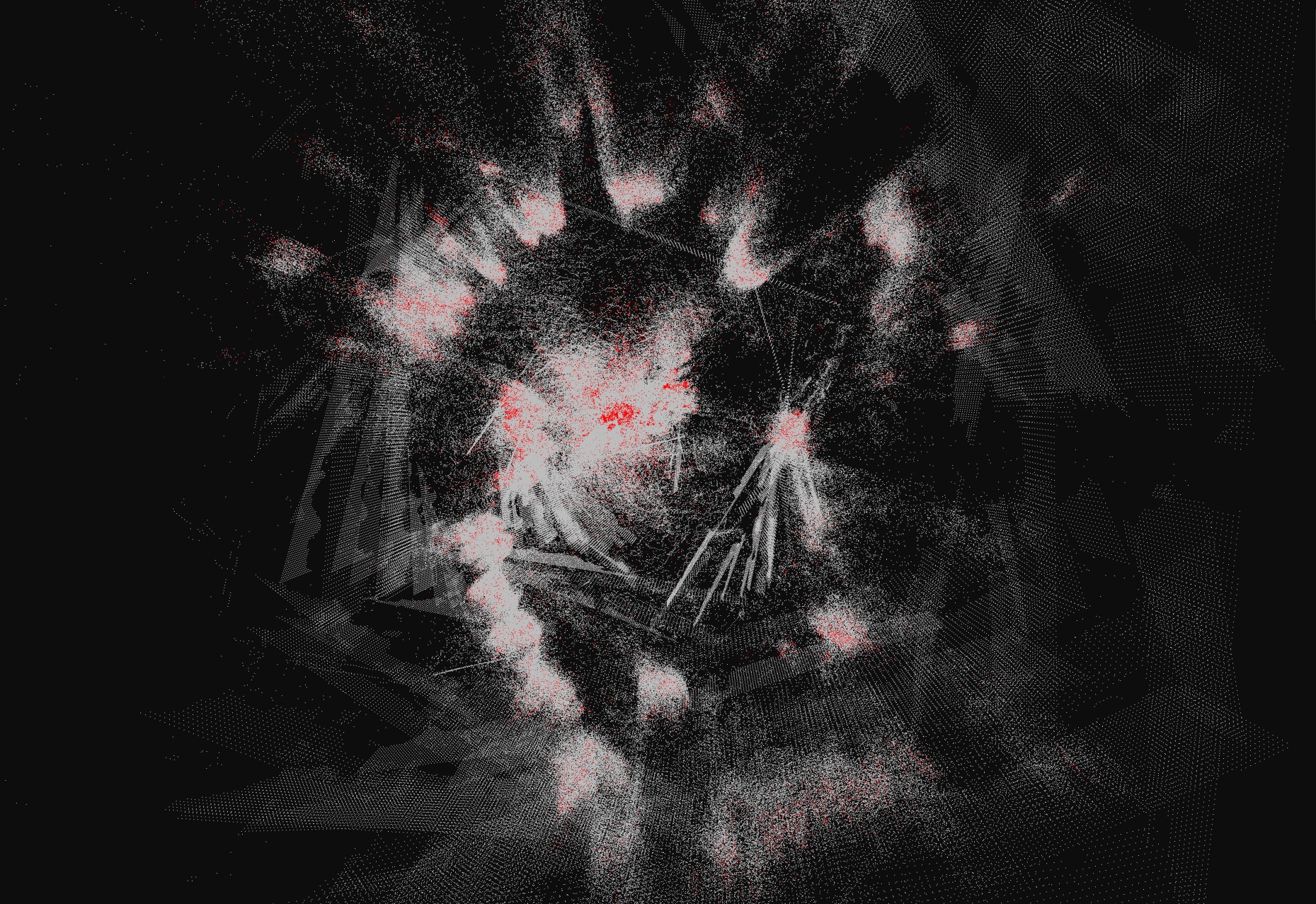} \\
        
        DA3 \hspace{1em}& 
        \includegraphics[width=\linewidth]{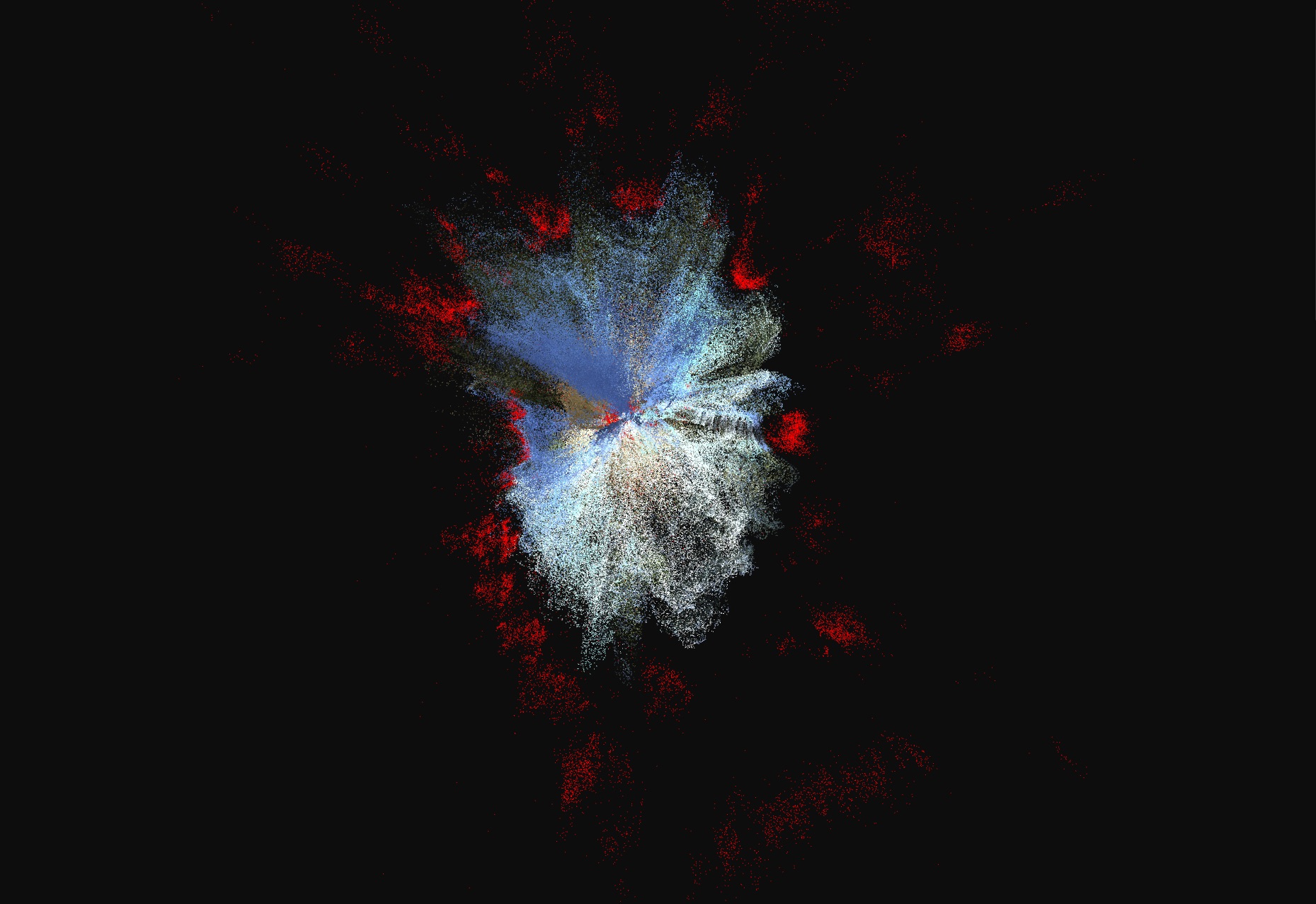} & 
        \includegraphics[width=\linewidth]{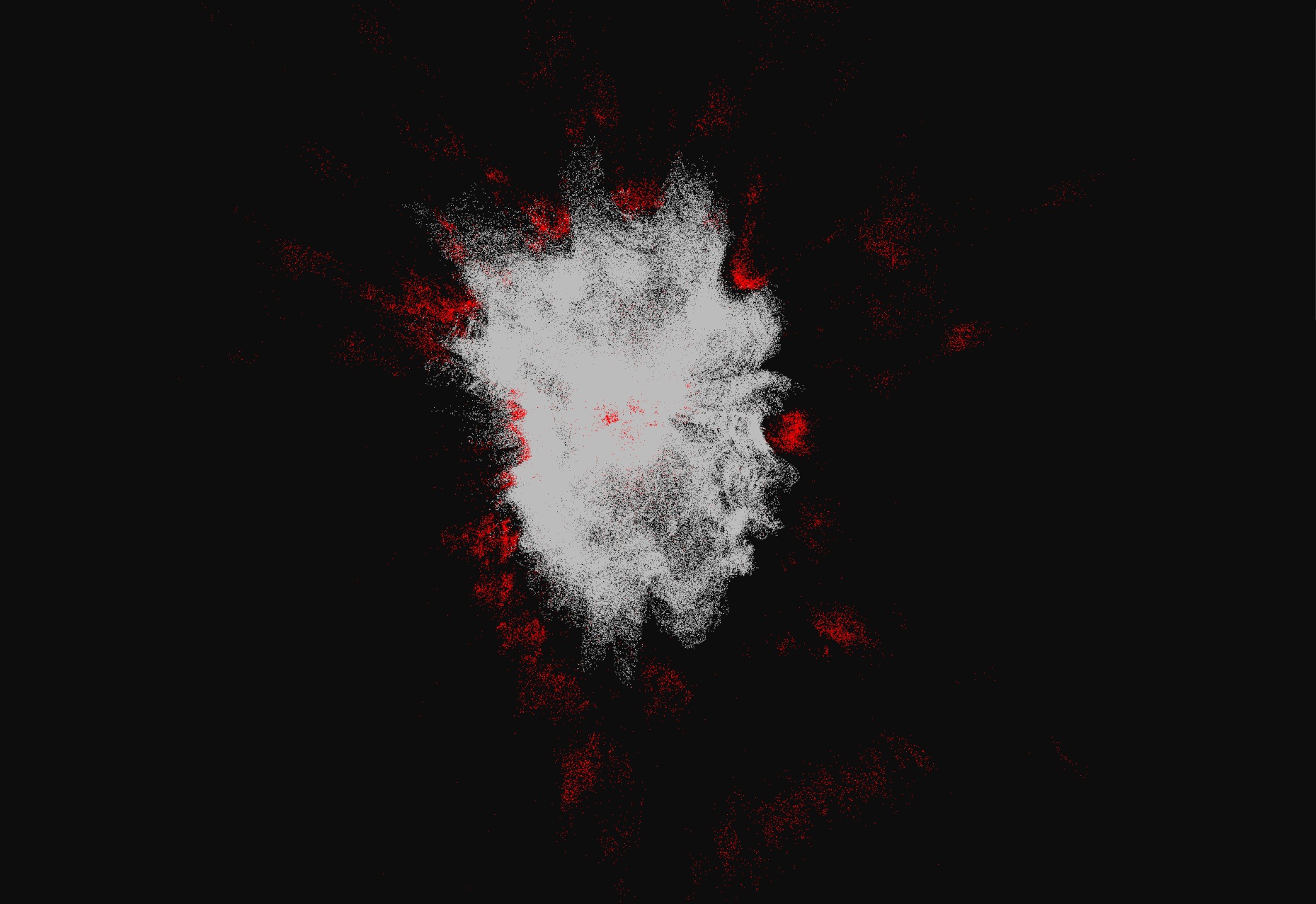} \\
        
        $\text{DA3}^\text{GS}$ \hspace{1em}& 
        \includegraphics[width=\linewidth]{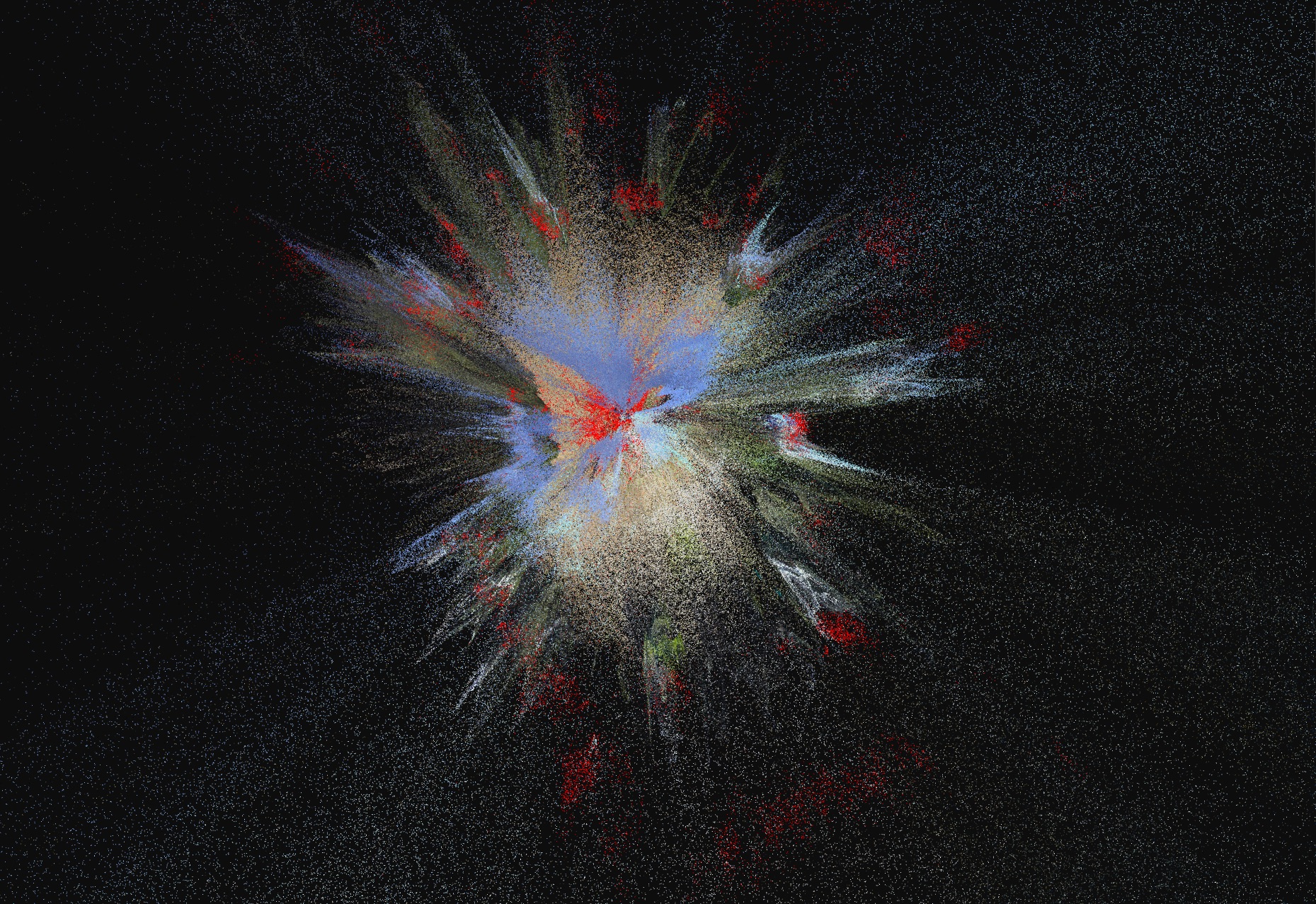} & 
        \includegraphics[width=\linewidth]{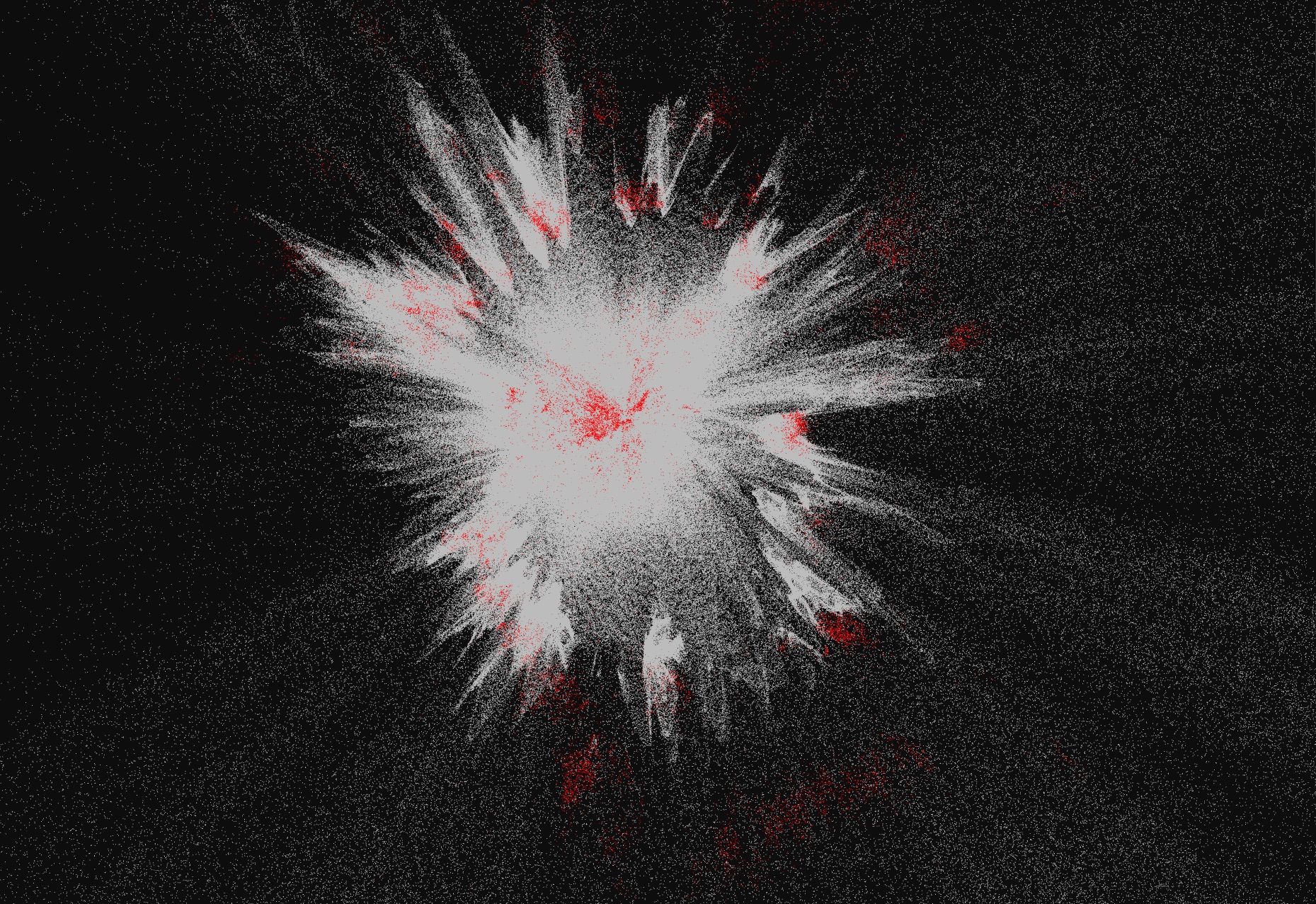} \\
        
    \end{tabular}
    \caption{Visualization of initialization outputs on the ``Caterpillar'' scene from the Tanks\&Temples dataset~\cite{tanksandtemples}. SfM points shown in red. We see that DA3 output is not aligned to the SfM point cloud. Surprisingly, \textbf{the DA3 GS head actually provides points that are more consistent with the SfM point cloud} (at the cost of noise).}
    \label{fig:caterpillar_point_clouds}
\end{figure}

\begin{figure}[tbp]
    \centering
    
    \includegraphics[width=0.9\linewidth]{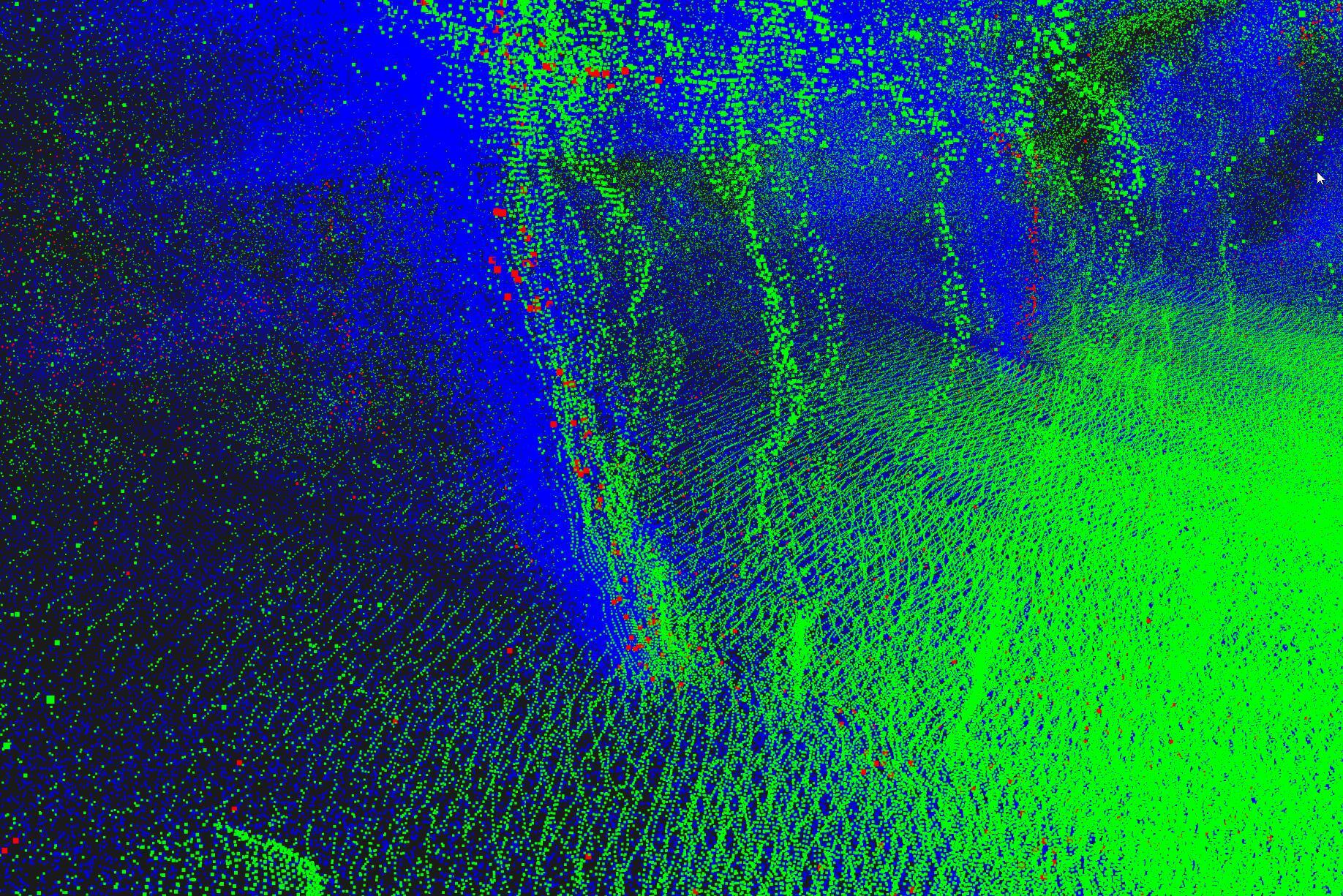} 
    
    \caption{An example of DA3's GS head improving alignment of points produced by the main model on the ``Bicycle'' scene from the MipNerf360 dataset~\cite{mipnerf360}. SfM points are depicted in red, DA3 (base) points in green, and $\text{DA3}^\text{GS}$ points in blue.}
    \label{fig:bicycle_da3_da3gs}
\end{figure}

As mentioned in the previous section, the discrepancy in the effect of using hybrid initialization between DA3 and Monodepth has prompted us to initiate an additional investigation into the outputs of these methods. 
After visual inspection of the initialization outputs, we have identified several cases where the DA3 output, while cleaner, is not aligned to the SfM point clouds.
This effect seems to be exaggerated for parts of the scene that are far from the input cameras.
We show an example of this on the ``Caterpillar'' scene in~\cref{fig:caterpillar_point_clouds}, where we
see DA3 output severely ``undershooting'' the distances to the trees in the background of the scene, whereas the other methods successfully place points in the general vicinity of the SfM clusters.

Surprisingly, the output of $\text{DA3}^\text{GS}$ aligns with the SfM point cloud much better than the output of base DA3,\footnote{The GS head has a depth offset output, which allows it to shift depth values predicted by the DA3 backbone.} which we find very interesting, considering the GS head was trained for feed-forward NVS\footnote{The DA3 tech report~\cite{depthanything3} specifies that besides photometric losses, the scale-shift-invariant depth loss was also used for GS head supervision.}
on the DL3DV-10K dataset~\cite{dl3dv10k}, which has no overlap with Tanks\&Temples.
We also saw this and similar behaviors on other Tanks\&Temples and MipNerf360 scenes, such as the examples in~\cref{fig:bicycle_da3_da3gs,fig:da3_da3_gs_bonsai}, where we see the GS head unifying disparate depth predictions from different images into a cluster aligned with the SfM points. While it is true that we used fewer images when running DA3 with the GS head in our experiments, we have performed preliminary testing at identical image counts, and saw the same results. We propose two possible causes for this behavior, though it is plausible that the real cause is a combination of both: 
\begin{itemize}
    \item The fact of training on DL3DV-10K itself improved DA3 generalization despite the main model being frozen, by forcing the GS head to predict depth offsets that improved upon the base model's predictions. 
    \item The improvement does not come just from more training data, but from training for the feed-forward 3DGS task itself. As 3DGS representations that succeed at NVS require geometric consistency (unless under-constrained by the input views), it is conceivable that successfully performing this task requires the model to improve its geometric consistency.
\end{itemize}

While we did not manually review the data for each scene, we have attempted to quantify some geometric properties of the produced initializations w.r.t. the SfM point clouds, which are our best approximation of ground truth data.\footnote{Tanks\&Temples has GT point clouds, but for only some of the scenes, and they often don't cover far away scene regions, where we see the most difference in output across the initialization methods.}
As the SfM point clouds are sparse, we can't compute $F_1$ scores, so we compute completeness w.r.t the SfM point clouds, using the following procedure.
We take the point clouds for all the dense init methods (we use Gaussian centers for $\text{EDGS}^*$ and $\text{DA3}^\text{GS}$), and for SfM, and perform the following steps:
\begin{enumerate}
    \item We apply scene scale normalization based on the input cameras and the SfM point cloud, using the method implemented in gsplat's reference implementation, and then apply voxel subsampling to all point clouds, with a voxel size of 0.02.
    \item For a given scene and a dense init method $M$, we find the nearest dense points $p^M_i$ for all SfM points $p^\mathit{SfM}_i$.
    \item We compute L2 distances $d^M_i$ between these point pairs, and filter out pairs with SfM points, for which at least one of the distances $d^M_i$ is above the 90th percentile of all distances (across points for all evaluated methods). This is done in an attempt to filter out SfM outliers. Though this approach is not robust, it does reduce the influence of incorrectly positioned SfM points.
    \item Finally, for each init method, we simply sum over all distances $d^M_i$ to obtain a measure of recall relative to the sparse SfM reconstruction.
\end{enumerate}

While these scores are %
approximate, and only useful for relative comparisons on the same data, we think they serve our purpose here sufficiently. We computed this metric for all MipNerf360 scenes, and the 14 Tanks\&Temples scenes we use in our evaluation, and obtained the following mean results: \textbf{821 for $\text{EDGS}^*$, 942 for Monodepth, 8080 for DA3, and 1747 for $\text{DA3}^\text{GS}$}. This perfectly aligns with our qualitative assessment of the outputs -- EDGS and Monodepth score similarly, while DA3 gets a significantly worse result. Introducing the GS head significantly improves the base DA3 results.

\section{Full $\text{DA3}^\text{GS}$ Components Ablation}\label{da3_gs_ablation}
\begin{table}[tbp]
\centering
\caption{Ablation on $\text{DA3}^\text{GS}$ initialization components.}
\label{tab:da3_gs_components_ablation_full}
\begin{subtable}[t]{0.9\linewidth}
\centering
\caption{ScanNet++}
\label{tab:da3_gs_components_ablation_full_scannet++}
{\small
\begingroup \setlength{\tabcolsep}{2.5\tabcolsep}
\resizebox{\linewidth}{!}{\begin{tabular}{l|ccc|ccc|ccc}
\toprule
& \multicolumn{3}{c|}{\textbf{PSNR ↑}} & \multicolumn{3}{c|}{\textbf{SSIM ↑}} & \multicolumn{3}{c}{\textbf{LPIPS ↓}} \\
& AbsGS & MCMC & IDHFR & AbsGS & MCMC & IDHFR & AbsGS & MCMC & IDHFR \\
\midrule
Base & \cc{FFFFFE}{\color[HTML]{000000} $23.37$} & \cc{FFFFFE}{\color[HTML]{000000} $22.58$} & \cc{FFFFFE}{\color[HTML]{000000} $23.65$} & \cc{FFFFFE}{\color[HTML]{000000} $0.871$} & \cc{FFFFFE}{\color[HTML]{000000} $0.873$} & \cc{FFFFFE}{\color[HTML]{000000} $0.876$} & \cc{FFFFFE}{\color[HTML]{000000} $0.235$} & \cc{FFFFFE}{\color[HTML]{000000} $0.238$} & \cc{FFFFFE}{\color[HTML]{000000} $0.225$} \\
\midrule
As point cloud& \cc{7AAED2}{\color[HTML]{F1F1F1} $-0.25$} & \cc{D5675E}{\color[HTML]{F1F1F1} $+0.29$} & \cc{C0DDEB}{\color[HTML]{000000} $-0.14$} & \cc{F9C9B1}{\color[HTML]{000000} $+0.003$} & \cc{F9C9B1}{\color[HTML]{000000} $+0.003$} & \cc{F0AB94}{\color[HTML]{000000} $+0.004$} & \cc{7DB1D4}{\color[HTML]{F1F1F1} $+0.007$} & \cc{FBD9C8}{\color[HTML]{000000} $-0.003$} & \cc{C0DDEB}{\color[HTML]{000000} $+0.004$} \\
kNN scale & \cc{4086BC}{\color[HTML]{F1F1F1} $-0.33$} & \cc{D36159}{\color[HTML]{F1F1F1} $+0.30$} & \cc{A5CDE3}{\color[HTML]{000000} $-0.19$} & \cc{F9C9B1}{\color[HTML]{000000} $+0.003$} & \cc{F9C9B1}{\color[HTML]{000000} $+0.003$} & \cc{F9C9B1}{\color[HTML]{000000} $+0.003$} & \cc{7DB1D4}{\color[HTML]{F1F1F1} $+0.007$} & \cc{FBD9C8}{\color[HTML]{000000} $-0.003$} & \cc{B1D5E7}{\color[HTML]{000000} $+0.005$} \\
Isotropic scale & \cc{C0DDEB}{\color[HTML]{000000} $-0.14$} & \cc{C43B3C}{\color[HTML]{F1F1F1} $+0.35$} & \cc{FFFFFE}{\color[HTML]{000000} $+0.00$} & \cc{FBDBCB}{\color[HTML]{000000} $+0.002$} & \cc{C43B3C}{\color[HTML]{F1F1F1} $+0.007$} & \cc{F0AB94}{\color[HTML]{000000} $+0.004$} & \cc{C0DDEB}{\color[HTML]{000000} $+0.004$} & \cc{F8C0A4}{\color[HTML]{000000} $-0.005$} & \cc{D0E5F0}{\color[HTML]{000000} $+0.003$} \\
Uniform $\alpha=0.1$ & \cc{D7E9F3}{\color[HTML]{000000} $-0.09$} & \cc{F7BCA1}{\color[HTML]{000000} $+0.18$} & \cc{C5DFED}{\color[HTML]{000000} $-0.13$} & \cc{D2E7F1}{\color[HTML]{000000} $-0.002$} & \cc{F9C9B1}{\color[HTML]{000000} $+0.003$} & \cc{E8F3F8}{\color[HTML]{000000} $-0.001$} & \cc{E0EEF5}{\color[HTML]{000000} $+0.002$} & \cc{FBD9C8}{\color[HTML]{000000} $-0.003$} & \cc{E0EEF5}{\color[HTML]{000000} $+0.002$} \\
Rotation noise 45° & \cc{D7E9F3}{\color[HTML]{000000} $-0.09$} & \cc{F4B49A}{\color[HTML]{000000} $+0.19$} & \cc{FFFFFE}{\color[HTML]{000000} $+0.00$} & \cc{FFFFFE}{\color[HTML]{000000} $+0.000$} & \cc{F9C9B1}{\color[HTML]{000000} $+0.003$} & \cc{FFFFFE}{\color[HTML]{000000} $+0.000$} & \cc{E0EEF5}{\color[HTML]{000000} $+0.002$} & \cc{FACCB5}{\color[HTML]{000000} $-0.004$} & \cc{FFFFFE}{\color[HTML]{000000} $+0.000$} \\
Color noise 0.5 & \cc{6CA4CD}{\color[HTML]{F1F1F1} $-0.27$} & \cc{97C3DD}{\color[HTML]{000000} $-0.21$} & \cc{E0EEF5}{\color[HTML]{000000} $-0.07$} & \cc{D2E7F1}{\color[HTML]{000000} $-0.002$} & \cc{FFFFFE}{\color[HTML]{000000} $+0.000$} & \cc{FDEDE5}{\color[HTML]{000000} $+0.001$} & \cc{327CB7}{\color[HTML]{F1F1F1} $+0.010$} & \cc{C0DDEB}{\color[HTML]{000000} $+0.004$} & \cc{B1D5E7}{\color[HTML]{000000} $+0.005$} \\
\bottomrule
\end{tabular}}
\endgroup
}
\end{subtable}
\begin{subtable}[t]{0.9\linewidth}
\centering
\caption{ScanNet++ (On-Trajectory)}
\label{tab:da3_gs_components_ablation_full_eval_on_train_set_scannet++}
{\small
\begingroup \setlength{\tabcolsep}{2.5\tabcolsep}
\resizebox{\linewidth}{!}{\begin{tabular}{l|ccc|ccc|ccc}
\toprule
& \multicolumn{3}{c|}{\textbf{PSNR ↑}} & \multicolumn{3}{c|}{\textbf{SSIM ↑}} & \multicolumn{3}{c}{\textbf{LPIPS ↓}} \\
& AbsGS & MCMC & IDHFR & AbsGS & MCMC & IDHFR & AbsGS & MCMC & IDHFR \\
\midrule
Base & \cc{FFFFFE}{\color[HTML]{000000} $33.45$} & \cc{FFFFFE}{\color[HTML]{000000} $33.23$} & \cc{FFFFFE}{\color[HTML]{000000} $34.11$} & \cc{FFFFFE}{\color[HTML]{000000} $0.950$} & \cc{FFFFFE}{\color[HTML]{000000} $0.952$} & \cc{FFFFFE}{\color[HTML]{000000} $0.953$} & \cc{FFFFFE}{\color[HTML]{000000} $0.097$} & \cc{FFFFFE}{\color[HTML]{000000} $0.100$} & \cc{FFFFFE}{\color[HTML]{000000} $0.092$} \\
\midrule
As point cloud& \cc{A5CDE3}{\color[HTML]{000000} $-0.13$} & \cc{CBE3EF}{\color[HTML]{000000} $-0.08$} & \cc{327CB7}{\color[HTML]{F1F1F1} $-0.24$} & \cc{F8BEA3}{\color[HTML]{000000} $+0.001$} & \cc{FFFFFE}{\color[HTML]{000000} $+0.000$} & \cc{FFFFFE}{\color[HTML]{000000} $+0.000$} & \cc{AFD4E6}{\color[HTML]{000000} $+0.003$} & \cc{FBD5C2}{\color[HTML]{000000} $-0.002$} & \cc{CBE3EF}{\color[HTML]{000000} $+0.002$} \\
kNN scale & \cc{E5F1F7}{\color[HTML]{000000} $-0.04$} & \cc{ECF5F9}{\color[HTML]{000000} $-0.03$} & \cc{327CB7}{\color[HTML]{F1F1F1} $-0.24$} & \cc{F8BEA3}{\color[HTML]{000000} $+0.001$} & \cc{FFFFFE}{\color[HTML]{000000} $+0.000$} & \cc{FFFFFE}{\color[HTML]{000000} $+0.000$} & \cc{AFD4E6}{\color[HTML]{000000} $+0.003$} & \cc{F8BEA3}{\color[HTML]{000000} $-0.003$} & \cc{CBE3EF}{\color[HTML]{000000} $+0.002$} \\
Isotropic scale & \cc{F8FBFD}{\color[HTML]{000000} $-0.01$} & \cc{FDEAE0}{\color[HTML]{000000} $+0.04$} & \cc{FBD5C2}{\color[HTML]{000000} $+0.08$} & \cc{F8BEA3}{\color[HTML]{000000} $+0.001$} & \cc{FFFFFE}{\color[HTML]{000000} $+0.000$} & \cc{F8BEA3}{\color[HTML]{000000} $+0.001$} & \cc{FFFFFE}{\color[HTML]{000000} $+0.000$} & \cc{FDEAE0}{\color[HTML]{000000} $-0.001$} & \cc{FDEAE0}{\color[HTML]{000000} $-0.001$} \\
Uniform $\alpha=0.1$ & \cc{B7D8E9}{\color[HTML]{000000} $-0.11$} & \cc{D1E6F1}{\color[HTML]{000000} $-0.07$} & \cc{D7E9F3}{\color[HTML]{000000} $-0.06$} & \cc{FFFFFE}{\color[HTML]{000000} $+0.000$} & \cc{FFFFFE}{\color[HTML]{000000} $+0.000$} & \cc{FFFFFE}{\color[HTML]{000000} $+0.000$} & \cc{FFFFFE}{\color[HTML]{000000} $+0.000$} & \cc{E5F1F7}{\color[HTML]{000000} $+0.001$} & \cc{FDEAE0}{\color[HTML]{000000} $-0.001$} \\
Rotation noise 45° & \cc{FEF5F0}{\color[HTML]{000000} $+0.02$} & \cc{FFFFFE}{\color[HTML]{000000} $+0.00$} & \cc{FFFFFE}{\color[HTML]{000000} $+0.00$} & \cc{FFFFFE}{\color[HTML]{000000} $+0.000$} & \cc{FFFFFE}{\color[HTML]{000000} $+0.000$} & \cc{FFFFFE}{\color[HTML]{000000} $+0.000$} & \cc{FFFFFE}{\color[HTML]{000000} $+0.000$} & \cc{E5F1F7}{\color[HTML]{000000} $+0.001$} & \cc{FFFFFE}{\color[HTML]{000000} $+0.000$} \\
Color noise 0.5 & \cc{C4DFED}{\color[HTML]{000000} $-0.09$} & \cc{468ABF}{\color[HTML]{F1F1F1} $-0.22$} & \cc{FEFAF7}{\color[HTML]{000000} $+0.01$} & \cc{FFFFFE}{\color[HTML]{000000} $+0.000$} & \cc{327CB7}{\color[HTML]{F1F1F1} $-0.002$} & \cc{FFFFFE}{\color[HTML]{000000} $+0.000$} & \cc{5C99C7}{\color[HTML]{F1F1F1} $+0.005$} & \cc{327CB7}{\color[HTML]{F1F1F1} $+0.006$} & \cc{85B6D7}{\color[HTML]{000000} $+0.004$} \\
\bottomrule
\end{tabular}}
\endgroup
}
\end{subtable}
\begin{subtable}[t]{0.9\linewidth}
\centering
\caption{Mip-NeRF 360}
\label{tab:da3_gs_components_ablation_full_mipnerf360}
{\small
\begingroup \setlength{\tabcolsep}{2.5\tabcolsep}
\resizebox{\linewidth}{!}{\begin{tabular}{l|ccc|ccc|ccc}
\toprule
& \multicolumn{3}{c|}{\textbf{PSNR ↑}} & \multicolumn{3}{c|}{\textbf{SSIM ↑}} & \multicolumn{3}{c}{\textbf{LPIPS ↓}} \\
& AbsGS & MCMC & IDHFR & AbsGS & MCMC & IDHFR & AbsGS & MCMC & IDHFR \\
\midrule
Base & \cc{FFFFFE}{\color[HTML]{000000} $27.45$} & \cc{FFFFFE}{\color[HTML]{000000} $27.65$} & \cc{FFFFFE}{\color[HTML]{000000} $27.55$} & \cc{FFFFFE}{\color[HTML]{000000} $0.819$} & \cc{FFFFFE}{\color[HTML]{000000} $0.819$} & \cc{FFFFFE}{\color[HTML]{000000} $0.818$} & \cc{FFFFFE}{\color[HTML]{000000} $0.138$} & \cc{FFFFFE}{\color[HTML]{000000} $0.139$} & \cc{FFFFFE}{\color[HTML]{000000} $0.138$} \\
\midrule
As point cloud& \cc{F9C5AB}{\color[HTML]{000000} $+0.16$} & \cc{3880BA}{\color[HTML]{F1F1F1} $-0.34$} & \cc{D5675E}{\color[HTML]{F1F1F1} $+0.29$} & \cc{F0AB94}{\color[HTML]{000000} $+0.008$} & \cc{327CB7}{\color[HTML]{F1F1F1} $-0.014$} & \cc{D36159}{\color[HTML]{F1F1F1} $+0.012$} & \cc{FFFFFE}{\color[HTML]{000000} $+0.000$} & \cc{327CB7}{\color[HTML]{F1F1F1} $+0.007$} & \cc{F0AB94}{\color[HTML]{000000} $-0.004$} \\
kNN scale & \cc{F4B49A}{\color[HTML]{000000} $+0.19$} & \cc{6CA4CD}{\color[HTML]{F1F1F1} $-0.27$} & \cc{C43B3C}{\color[HTML]{F1F1F1} $+0.35$} & \cc{E99985}{\color[HTML]{000000} $+0.009$} & \cc{7AAED2}{\color[HTML]{F1F1F1} $-0.010$} & \cc{CB4E4A}{\color[HTML]{F1F1F1} $+0.013$} & \cc{FDEDE5}{\color[HTML]{000000} $-0.001$} & \cc{7AAED2}{\color[HTML]{F1F1F1} $+0.005$} & \cc{E18676}{\color[HTML]{F1F1F1} $-0.005$} \\
Isotropic scale & \cc{EDF5FA}{\color[HTML]{000000} $-0.04$} & \cc{FBFDFE}{\color[HTML]{000000} $-0.01$} & \cc{F6FAFC}{\color[HTML]{000000} $-0.02$} & \cc{FEF6F1}{\color[HTML]{000000} $+0.001$} & \cc{FFFFFE}{\color[HTML]{000000} $+0.000$} & \cc{FCE4D8}{\color[HTML]{000000} $+0.003$} & \cc{FFFFFE}{\color[HTML]{000000} $+0.000$} & \cc{FFFFFE}{\color[HTML]{000000} $+0.000$} & \cc{FFFFFE}{\color[HTML]{000000} $+0.000$} \\
Uniform $\alpha=0.1$ & \cc{F2F8FB}{\color[HTML]{000000} $-0.03$} & \cc{FFFCFA}{\color[HTML]{000000} $+0.01$} & \cc{F6FAFC}{\color[HTML]{000000} $-0.02$} & \cc{FFFFFE}{\color[HTML]{000000} $+0.000$} & \cc{FFFFFE}{\color[HTML]{000000} $+0.000$} & \cc{FFFFFE}{\color[HTML]{000000} $+0.000$} & \cc{FFFFFE}{\color[HTML]{000000} $+0.000$} & \cc{FFFFFE}{\color[HTML]{000000} $+0.000$} & \cc{FFFFFE}{\color[HTML]{000000} $+0.000$} \\
Rotation noise 45° & \cc{FFFFFE}{\color[HTML]{000000} $+0.00$} & \cc{FFFFFE}{\color[HTML]{000000} $+0.00$} & \cc{FEF8F4}{\color[HTML]{000000} $+0.02$} & \cc{FFFFFE}{\color[HTML]{000000} $+0.000$} & \cc{FEF6F1}{\color[HTML]{000000} $+0.001$} & \cc{FEF6F1}{\color[HTML]{000000} $+0.001$} & \cc{FFFFFE}{\color[HTML]{000000} $+0.000$} & \cc{FFFFFE}{\color[HTML]{000000} $+0.000$} & \cc{FFFFFE}{\color[HTML]{000000} $+0.000$} \\
Color noise 0.5 & \cc{D2E7F1}{\color[HTML]{000000} $-0.10$} & \cc{FFFCFA}{\color[HTML]{000000} $+0.01$} & \cc{F6FAFC}{\color[HTML]{000000} $-0.02$} & \cc{FFFFFE}{\color[HTML]{000000} $+0.000$} & \cc{FDEDE5}{\color[HTML]{000000} $+0.002$} & \cc{FDEDE5}{\color[HTML]{000000} $+0.002$} & \cc{5695C5}{\color[HTML]{F1F1F1} $+0.006$} & \cc{9DC7E0}{\color[HTML]{000000} $+0.004$} & \cc{D2E7F1}{\color[HTML]{000000} $+0.002$} \\
\bottomrule
\end{tabular}}
\endgroup
}
\end{subtable}
\begin{subtable}[t]{0.9\linewidth}
\centering
\caption{Tanks and Temples}
\label{tab:da3_gs_components_ablation_full_tanksandtemples}
{\small
\begingroup \setlength{\tabcolsep}{2.5\tabcolsep}
\resizebox{\linewidth}{!}{\begin{tabular}{l|ccc|ccc|ccc}
\toprule
& \multicolumn{3}{c|}{\textbf{PSNR ↑}} & \multicolumn{3}{c|}{\textbf{SSIM ↑}} & \multicolumn{3}{c}{\textbf{LPIPS ↓}} \\
& AbsGS & MCMC & IDHFR & AbsGS & MCMC & IDHFR & AbsGS & MCMC & IDHFR \\
\midrule
Base & \cc{FFFFFE}{\color[HTML]{000000} $23.63$} & \cc{FFFFFE}{\color[HTML]{000000} $24.33$} & \cc{FFFFFE}{\color[HTML]{000000} $23.85$} & \cc{FFFFFE}{\color[HTML]{000000} $0.834$} & \cc{FFFFFE}{\color[HTML]{000000} $0.847$} & \cc{FFFFFE}{\color[HTML]{000000} $0.839$} & \cc{FFFFFE}{\color[HTML]{000000} $0.148$} & \cc{FFFFFE}{\color[HTML]{000000} $0.140$} & \cc{FFFFFE}{\color[HTML]{000000} $0.138$} \\
\midrule
As point cloud& \cc{B4D6E8}{\color[HTML]{000000} $-0.14$} & \cc{327CB7}{\color[HTML]{F1F1F1} $-0.29$} & \cc{FDE9DF}{\color[HTML]{000000} $+0.05$} & \cc{FDEDE5}{\color[HTML]{000000} $+0.001$} & \cc{327CB7}{\color[HTML]{F1F1F1} $-0.007$} & \cc{F0AB94}{\color[HTML]{000000} $+0.004$} & \cc{B1D5E7}{\color[HTML]{000000} $+0.005$} & \cc{E0EEF5}{\color[HTML]{000000} $+0.002$} & \cc{E0EEF5}{\color[HTML]{000000} $+0.002$} \\
kNN scale & \cc{EEF6FA}{\color[HTML]{000000} $-0.03$} & \cc{66A0CB}{\color[HTML]{F1F1F1} $-0.23$} & \cc{FEF2EC}{\color[HTML]{000000} $+0.03$} & \cc{FBDBCB}{\color[HTML]{000000} $+0.002$} & \cc{7AAED2}{\color[HTML]{F1F1F1} $-0.005$} & \cc{F0AB94}{\color[HTML]{000000} $+0.004$} & \cc{D0E5F0}{\color[HTML]{000000} $+0.003$} & \cc{F0F7FA}{\color[HTML]{000000} $+0.001$} & \cc{F0F7FA}{\color[HTML]{000000} $+0.001$} \\
Isotropic scale & \cc{CFE5F0}{\color[HTML]{000000} $-0.09$} & \cc{EAF3F8}{\color[HTML]{000000} $-0.04$} & \cc{F5F9FC}{\color[HTML]{000000} $-0.02$} & \cc{FFFFFE}{\color[HTML]{000000} $+0.000$} & \cc{FFFFFE}{\color[HTML]{000000} $+0.000$} & \cc{FBDBCB}{\color[HTML]{000000} $+0.002$} & \cc{F0F7FA}{\color[HTML]{000000} $+0.001$} & \cc{FFFFFE}{\color[HTML]{000000} $+0.000$} & \cc{E0EEF5}{\color[HTML]{000000} $+0.002$} \\
Uniform $\alpha=0.1$ & \cc{EEF6FA}{\color[HTML]{000000} $-0.03$} & \cc{EEF6FA}{\color[HTML]{000000} $-0.03$} & \cc{FFFBF9}{\color[HTML]{000000} $+0.01$} & \cc{E8F3F8}{\color[HTML]{000000} $-0.001$} & \cc{FFFFFE}{\color[HTML]{000000} $+0.000$} & \cc{FFFFFE}{\color[HTML]{000000} $+0.000$} & \cc{FEF3ED}{\color[HTML]{000000} $-0.001$} & \cc{F0F7FA}{\color[HTML]{000000} $+0.001$} & \cc{FFFFFE}{\color[HTML]{000000} $+0.000$} \\
Rotation noise 45° & \cc{F9FCFD}{\color[HTML]{000000} $-0.01$} & \cc{EAF3F8}{\color[HTML]{000000} $-0.04$} & \cc{FFFFFE}{\color[HTML]{000000} $+0.00$} & \cc{FFFFFE}{\color[HTML]{000000} $+0.000$} & \cc{FDEDE5}{\color[HTML]{000000} $+0.001$} & \cc{FFFFFE}{\color[HTML]{000000} $+0.000$} & \cc{FFFFFE}{\color[HTML]{000000} $+0.000$} & \cc{FFFFFE}{\color[HTML]{000000} $+0.000$} & \cc{F0F7FA}{\color[HTML]{000000} $+0.001$} \\
Color noise 0.5 & \cc{B4D6E8}{\color[HTML]{000000} $-0.14$} & \cc{F5F9FC}{\color[HTML]{000000} $-0.02$} & \cc{FFFFFE}{\color[HTML]{000000} $+0.00$} & \cc{BCDBEA}{\color[HTML]{000000} $-0.003$} & \cc{FFFFFE}{\color[HTML]{000000} $+0.000$} & \cc{E8F3F8}{\color[HTML]{000000} $-0.001$} & \cc{327CB7}{\color[HTML]{F1F1F1} $+0.010$} & \cc{C0DDEB}{\color[HTML]{000000} $+0.004$} & \cc{B1D5E7}{\color[HTML]{000000} $+0.005$} \\
\bottomrule
\end{tabular}}
\endgroup
}
\end{subtable}
\end{table}

In our main investigation, we saw that $\text{DA3}^\text{GS}$ provided significantly improved NVS results compared to using base DA3. Though we have later found this to be caused mostly by differences in point positions (see \cref{init_precision_no_gt}), we have also performed other ablations on parameters of $\text{DA3}^\text{GS}$-initialized Gaussians, which we report in~\cref{tab:da3_gs_components_ablation_full}.
We describe the specifics of all these ablations below:
\begin{description}
    \item[As point cloud] As mentioned in the main paper, for this test, we simply use the Gaussians' centers and colors to form a point cloud, and apply the traditional initialization heuristics from~\cite{3dgs}.
    \item[kNN scale] Here, we isolate scale anisotropy \textit{and} absolute scale, as we compute isotropic scales following~\cite{3dgs}.
    \item[Isotropic scale] To disentangle overall size from the effects of anisotropic initialization, given a Gaussian with scales $s_x, s_y, s_z$, we set all to $\frac{1}{3}(s_x + s_y + s_z)$.
    \item[Uniform opacity] We initialize all opacities to 0.1, following~\cite{3dgs}.
    \item[Rotation noise 45°] To investigate the effect of initializing the rotation component of the Gaussians' covariances, we perturb the rotations with noise. We sample a random rotation axis uniformly, and sample rotation angles from a normal distribution with $\sigma=45\text{°}$. We then compose existing rotations with these random offsets.
    \item[Color noise 0.5] We apply normally distributed noise with $\sigma = 0.5$ to all RGB values (in range $[0, 1]$), clamping those that land outside the valid range.
\end{description}

\begin{figure}[tbp]
    \centering
    \includegraphics[width=\linewidth]{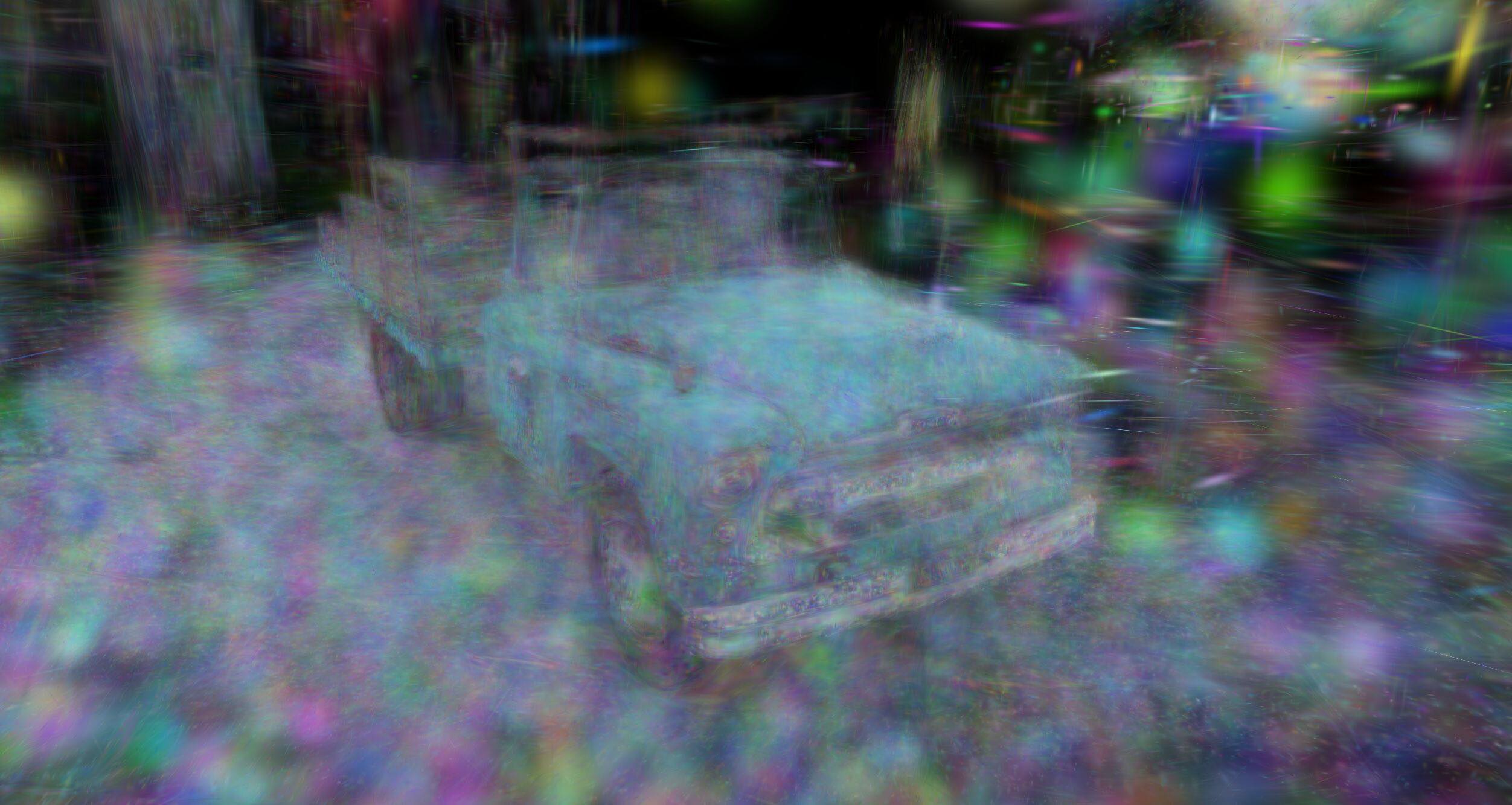} 
    \caption{An example of $\text{DA3}^\text{GS}$ initial Gaussians with applied color noise on the ``Truck'' scene from Tanks\&Temples.}
    \label{fig:da3_gs_color_noise}
\end{figure}

Looking at the results in~\cref{tab:da3_gs_components_ablation_full}, we see that the other ablation components that were omitted from the main tables do not have a significant effect as scales. Adding rotation noise seems to have a very limited effect, especially on MipNerf360 and Tanks\&Temples. Adding color noise does result in some changes, though they are not as significant as when adjusting scales, and, as demonstrated in~\cref{fig:da3_gs_color_noise}, the level of color noise we apply is beyond anything that can be expected from a regular initialization.

\section{Full Description of ``Monodepth'' Initialization}\label{monodepth_details}
\begin{figure}
    \centering
    \setlength{\tabcolsep}{1.5pt}
    \begin{tabular}{ccc}
        \includegraphics[width=0.32\textwidth]{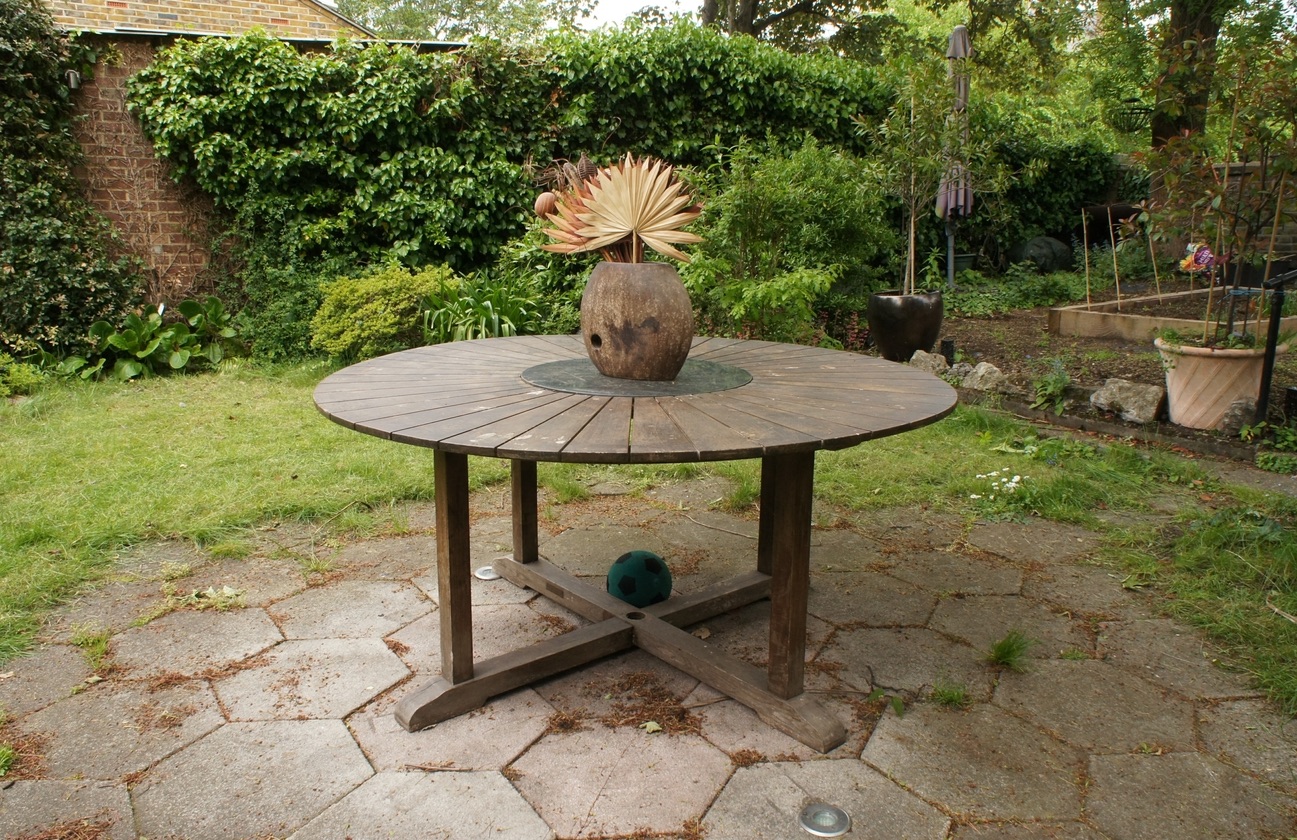} & 
        \includegraphics[width=0.32\textwidth]{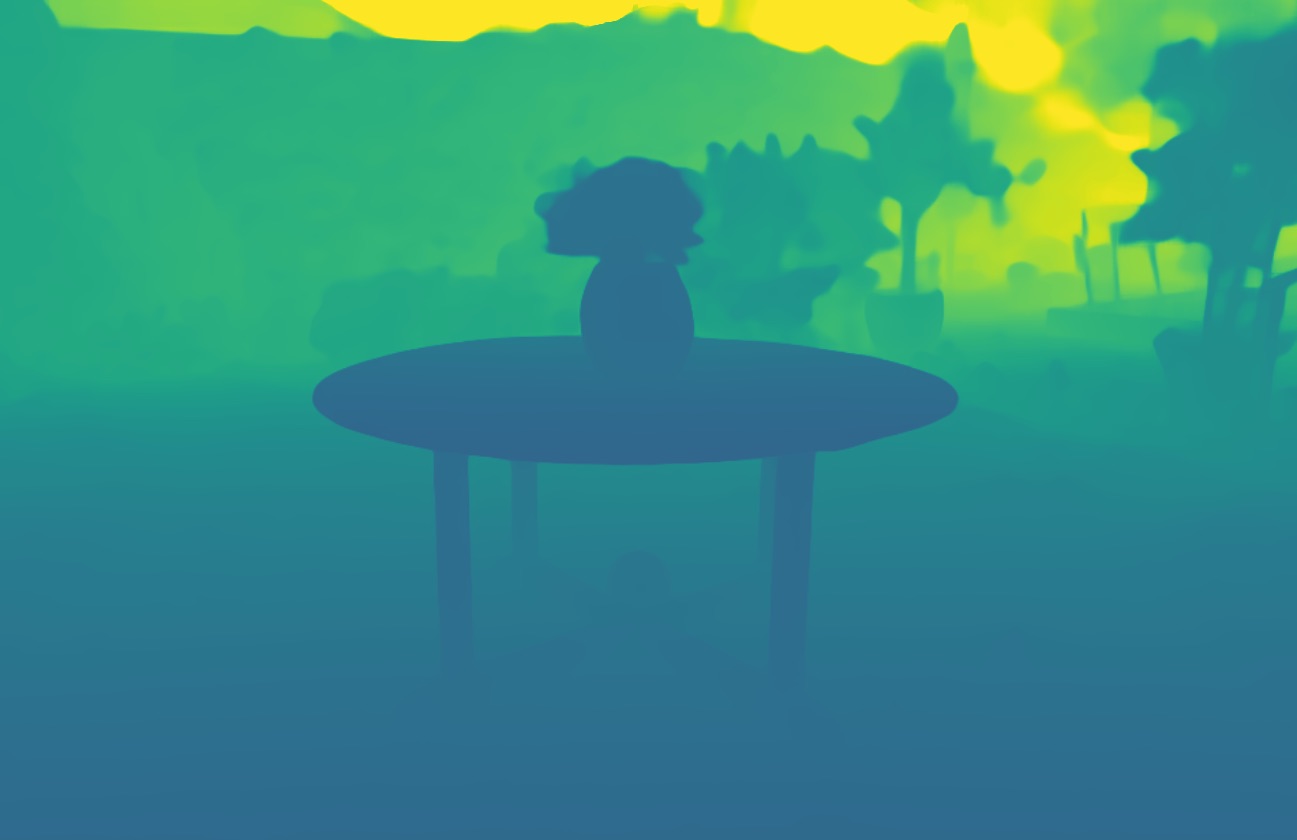} & 
        \includegraphics[width=0.32\textwidth]{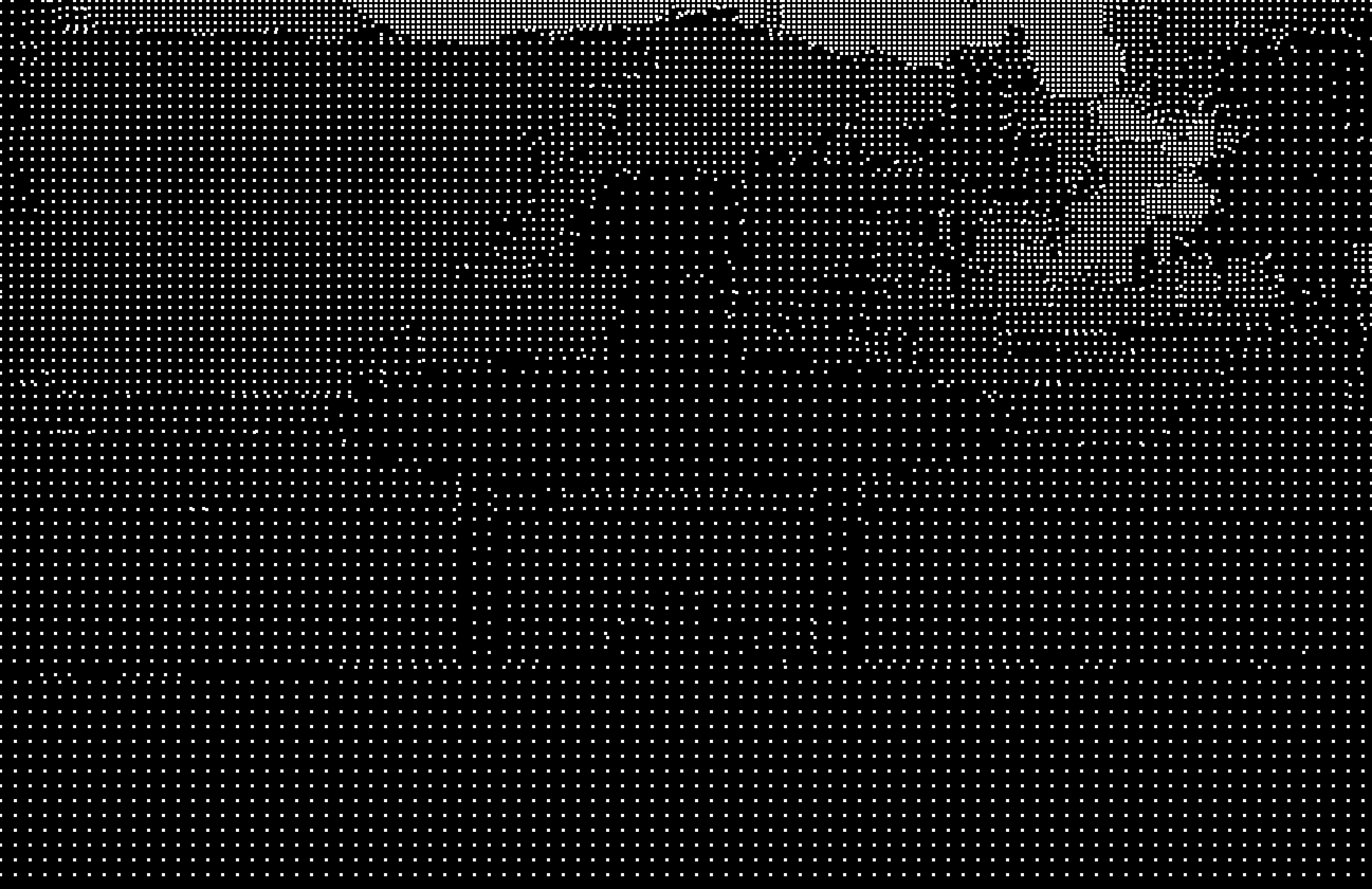} \\
        RGB & Predicted Depth & Subsampling Mask \\
    \end{tabular}
    \caption{Example of an adaptive subsampling mask constructed based on depth values. Selected pixels in the subsampling mask are visualized as squares spanning multiple pixels for better visibility.
    }\label{fig:adaptive_subsample_mask}
\end{figure}

Our monocular depth initialization pipeline begins by selecting up to 300 images for depth prediction, alignment, and inverse projection of points to world space. If there are fewer than 300 images, all images are used, otherwise, we select 300 cameras using K-Means clustering of their $\mathbb{R}^{4\times4}$ extrinsic calibration matrices. This simple approach follows the current (at the time of writing) public code release for EDGS~\cite{edgs}, and works quite well in practice. After this, the following steps are performed for each input image independently:
\begin{enumerate}
    \item The depth predictor is invoked. In this case we use Metric3D~V2~\cite{metric3dv2}, with the DINOv2-reg ViT Large backbone~\cite{vit, dinov2, vits_need_registers}. 
    \item The predicted depth map is then aligned to the SfM point cloud using LO-RANSAC~\cite{lo-ransac}. For a given set of sample SfM points that lie in the image, scale and shift that minimize error in the least squares sense are estimated using a closed form solution~\cite{ranftl2020robustmonoculardepthestimation}. We use 4 samples per iteration, a confidence threshold of 0.999, an inlier threshold of 0.01, and limit the algorithm to at most 2500 iterations.
    \item While this coarse alignment serves as a good estimate in most cases, we observed that estimating scale and shift for the whole image is not enough, as the relative alignment of the monocular depth prediction w.r.t. the SfM depths may vary across different objects and depth levels in the image. 
    To this end, we employ a post-alignment approach, used by Yehe Liu and Aly El Hakie in their open source monocular depth initialization implementation~\cite{depth_densifier_github}. Given a list of SfM point depths $d_i^*$ and a list of predicted depths $d_i$ at corresponding pixels, this method computes corrected depth values for all predicted depths in the image. For each input depth $d_i$ (not just those with known $d_i^*$), a new depth value $\hat{d_i}$ is calculated by first finding the indices $k, k+1$ of depth measurements with known SfM counterparts, such that $d_k \leq d_i < d_{k+1}$, and then performing linear interpolation of the corresponding SfM depth values: 
    \begin{align}
    t_i &= \frac{d_i - d_k}{d_{k+1} - d_k}\space{}, \\
    \hat{d}_i &= d_k^* + t_i\left(d_{k+1}^* - d_k^*\right) \space{}.
    \end{align}
    This approach relies on the assumption that, when sorted, both the (unknown) ground truth depth values, and the predicted depth values change linearly inside the intervals between known $d_i$ and $d_i^*$. 
    While this assumption is not strictly true in practice, this provides a good approximation, especially in cases when the number of SfM points in the image is large. Of course, incorrect alignment can occur in practice, especially for images with few known SfM depths, or when the depth predictor's output produces different depths for objects located at the same distance from the camera. But overall, this method is 
    an improvement over image-global alignment. An evolved version of this approach could utilize the available information about the spatial distribution of pixels with known $d_i$ and $d_i^*$.
    \item To select which image points should be used to create world-space points, we use adaptive sampling of the image based on the depth values.
    The idea is to skew the output point distribution in a way that compensates for the effects of perspective projection and the camera trajectory characteristics of typical outside-in captures, both of which would result in much higher point density close to the cameras, if uniform subsampling was to be used. To perform the depth-guided subsampling, the depth map is transformed into a per-pixel map of target subsample factors from the range $[D_{\text{min}}, D_{\text{max}}]$ using linear interpolation driven by the depth values normalized to $[0, 1]$ range. In our experiments we use $D_{\text{min}} = 5$ and $D_{\text{max}} = 15$.
    Additionally, to provide robustness to outliers, the depth map values are clamped to the range $[Q_1 - 1.5\,\mathrm{IQR},\; Q_3 + 1.5\,\mathrm{IQR}]$, where $\mathrm{IQR}$ denotes the inter-quartile range.
    After subsample factors $S_{i,j}$ are calculated for each pixel, the final pixel mask 
    used to select pixels for inverse projection is calculated as follows:
    \begin{equation}
        M_{i,j} = \begin{cases}
            1 & \text{if } (i \bmod \lfloor S_{i,j} \rfloor =0)
                            \land (j \bmod \lfloor S_{i,j} \rfloor=0),\\
            0 & \text{otherwise}.
            \end{cases}
    \end{equation}\label{eq:subsample_mask}
    An example of the produced mask can be seen in \Cref{fig:adaptive_subsample_mask}.
    \item We additionally mask out pixels where the depth gradient (approximated via finite differences) is above a certain threshold to reduce noise from unprojecting points at object boundaries.
    \item World-space points are created for the selected image points using inverse projection with the known camera parameters.
\end{enumerate}
Finally, we apply a version of the floater removal method implemented in~\cite{depth_densifier_github} to filter out noise in front of the cameras. This method works by iterating over all input cameras and counting the number of floater votes for each point. A camera votes for a point to be considered a floater, if the point is significantly in front of the predicted depth map for that camera (post alignment).
In the original implementation, a point is removed if it receives at least $T_\text{floater}$ votes from different cameras. We use an adjusted formulation where we collect both floater and non-floater votes, to account for the number of images a given point is visible in. Specifically, we keep points for which the fraction of floater votes was greater than or equal to $0.4$. Furthermore, we always keep points that receive less than 3 total votes.

\section{Verification of RevDGS Implementation}

To verify the validity of our RevDGS implementation in the absence of official reference code, we compared our NVS results on MipNerf360 with the values reported in~\cite{revdgs}. With our implementation PSNR decreased by 0.2, SSIM decreased by 0.01, and LPIPS (VGG)\footnote{All other results in this paper/supplementary use the AlexNet LPIPS backbone as it is recommended for use as a forward metric on the official LPIPS GitHub, but we compare results with VGG backbone here.} increased (which is worse) by 0.01.   
These differences are not drastic and are likely due to the (unreleased) original RevDGS code using a different renderer (we use gsplat).

In our experiments, we saw RevDGS perform especially poorly using SfM on ETH3D (\cref{tab:laser_scan_main}), which prompted us to further investigate that case. Upon inspecting the trained splats, we find that RevDGS tends to severely overfit to the few training views on ETH3D when using SfM initialization. However, we think that this is unlikely to be a fault specific to our implementation. Our source code is publicly available on https://github.com/deivse/ivd\_splat and we welcome third-party validation of our implementation.

\section{Examples of ScanNet++ Camera Positions}\label{scannetpp_cameras}

\begin{figure}[tbp]
    \centering
    \renewcommand{\arraystretch}{1.0} 
    
    \begin{tabular}{>{\centering\arraybackslash}m{0.45\textwidth} >{\centering\arraybackslash}m{0.45\textwidth}}
            
        Default Split\vspace{0.75em}& 
        On-Trajectory\vspace{0.75em}\\

        \includegraphics[width=\linewidth]{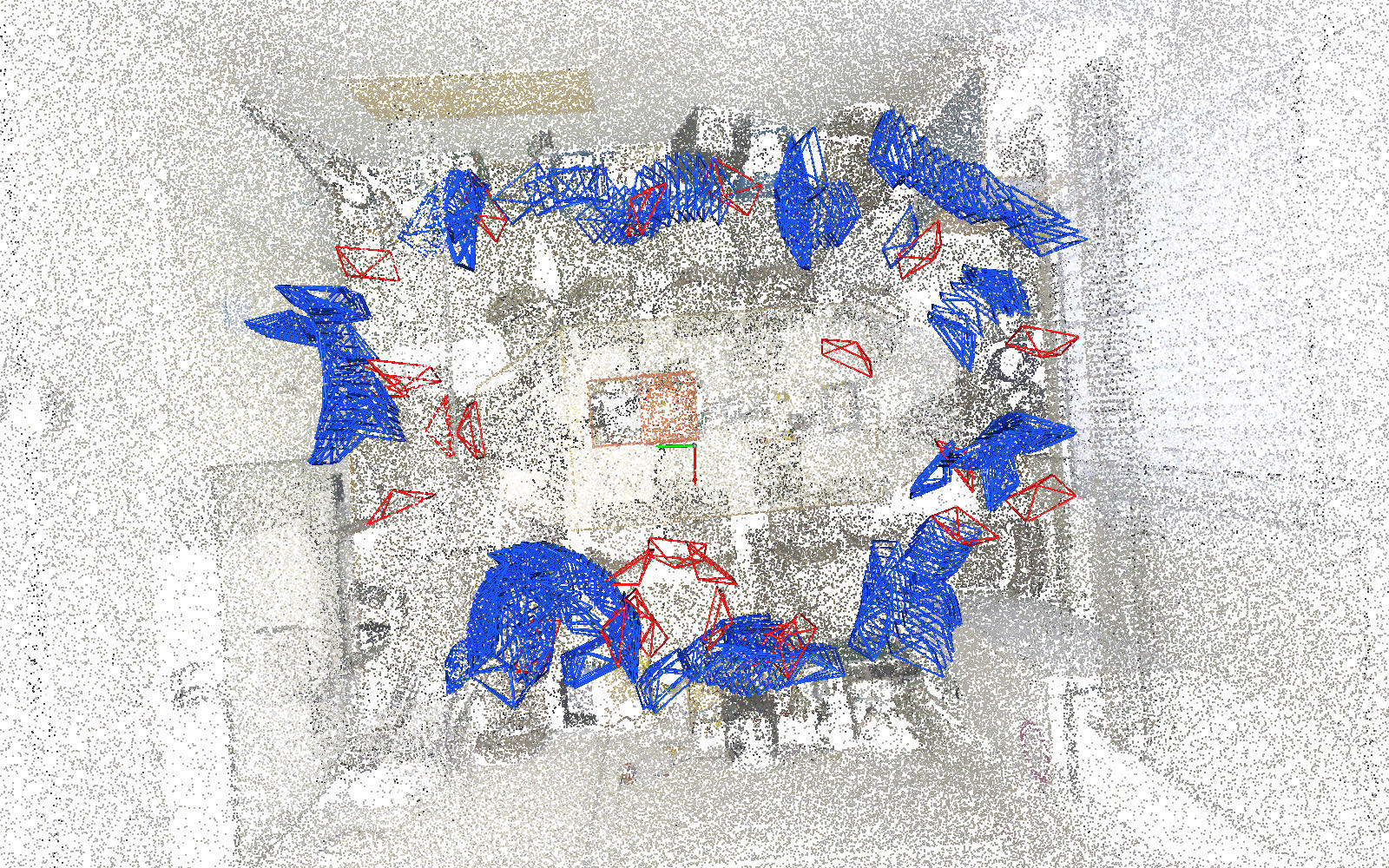} & 
        \includegraphics[width=\linewidth]{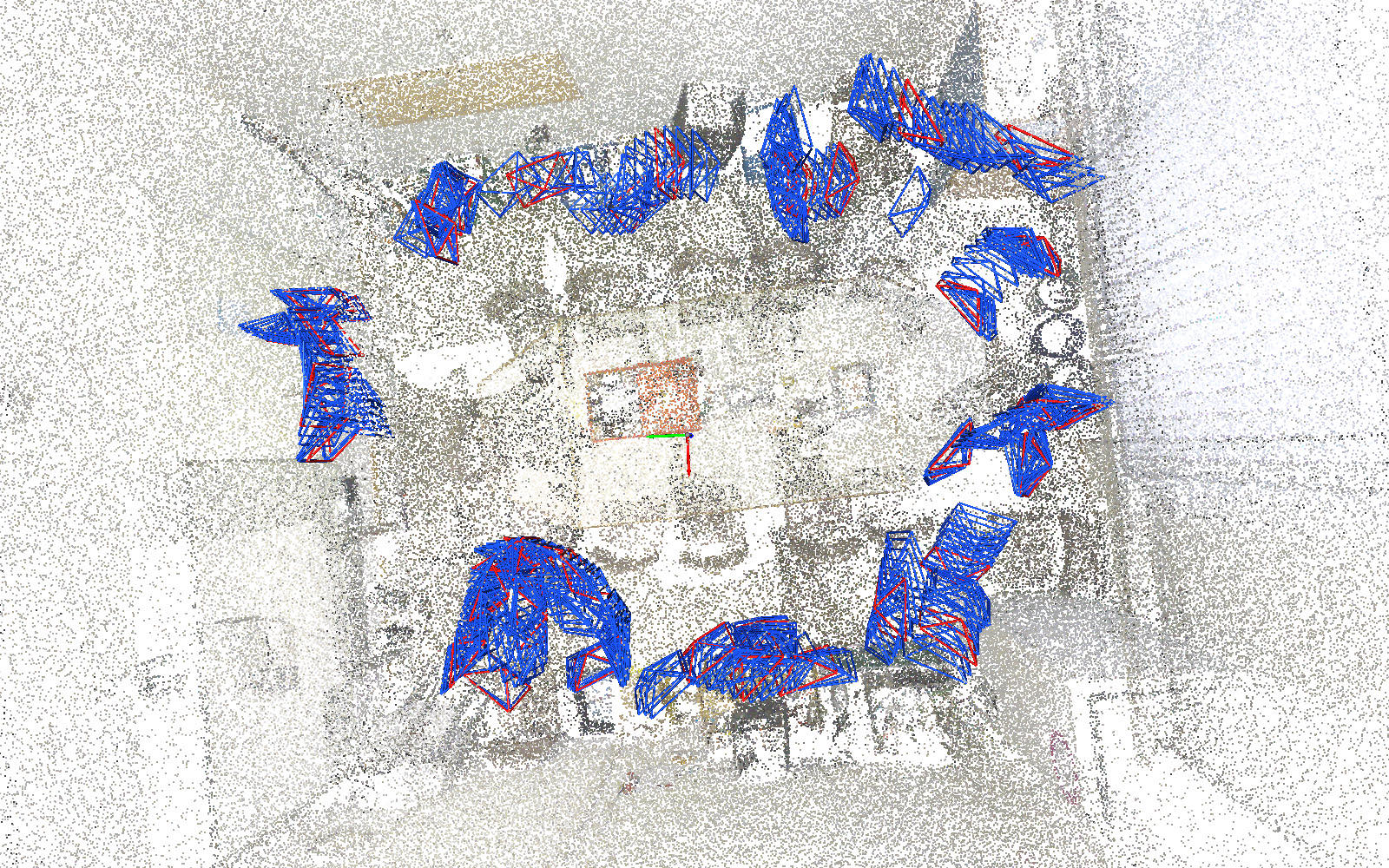} \\        

        \includegraphics[width=\linewidth]{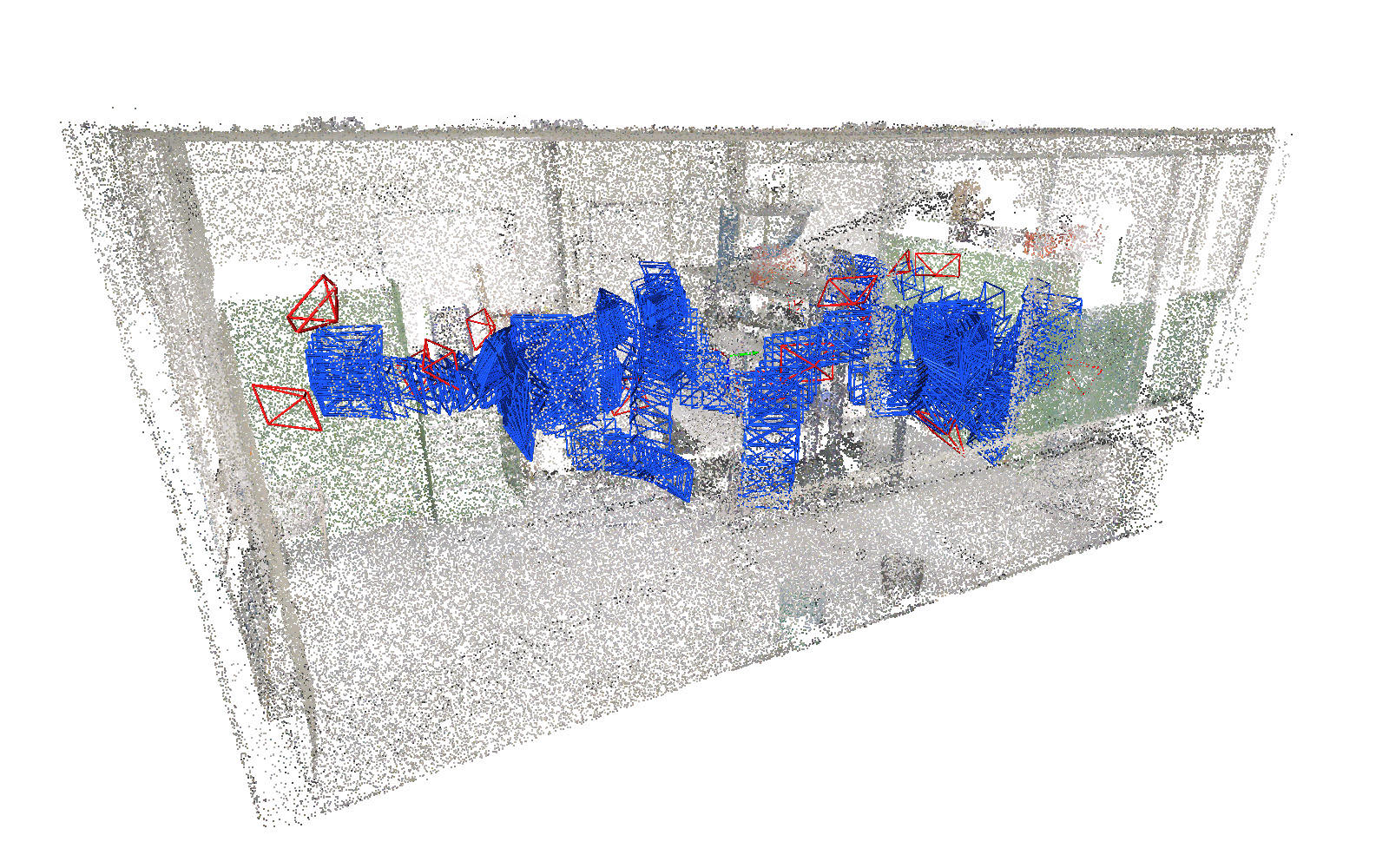} & 
        \includegraphics[width=\linewidth]{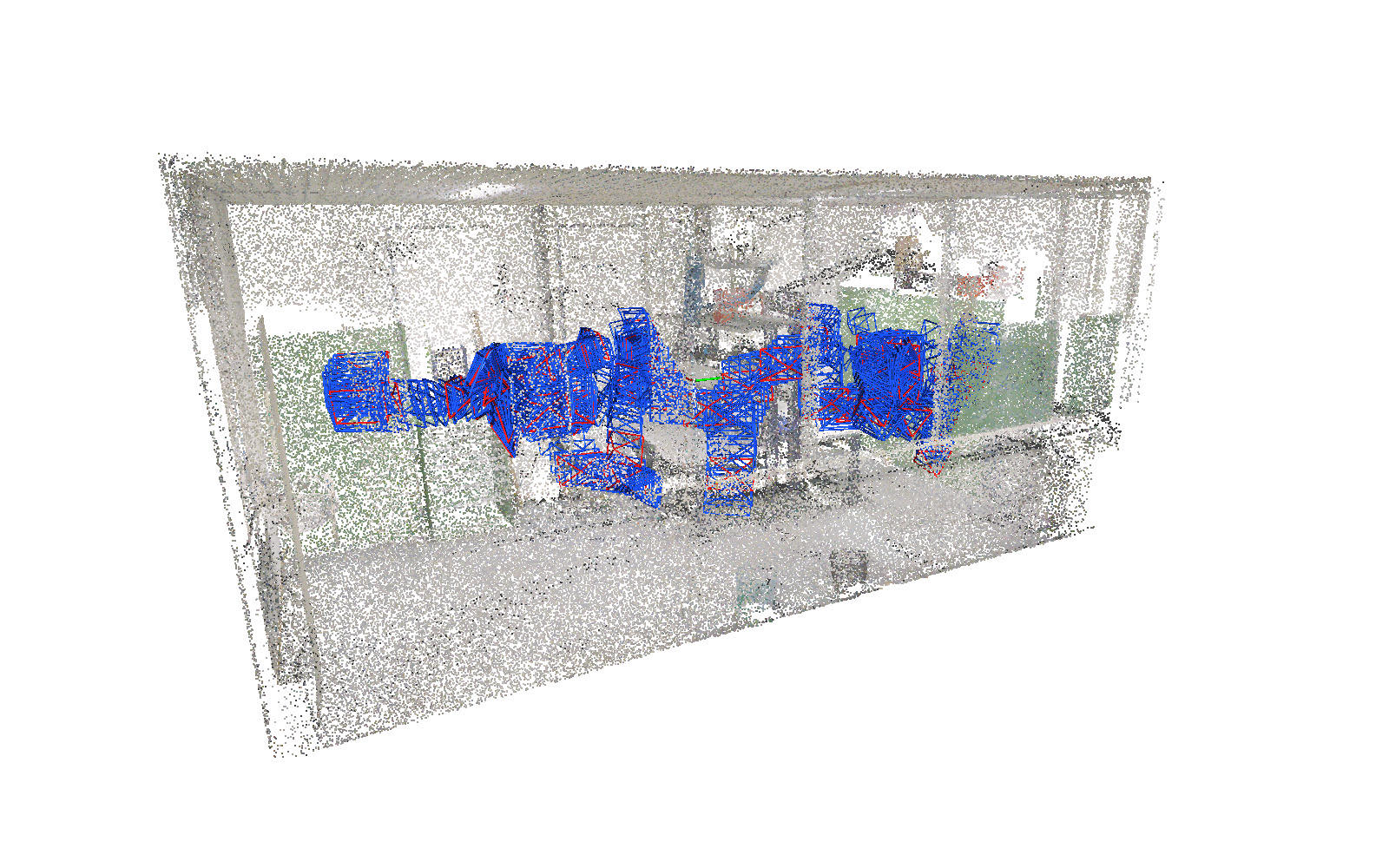} \\
        
        \includegraphics[width=\linewidth]{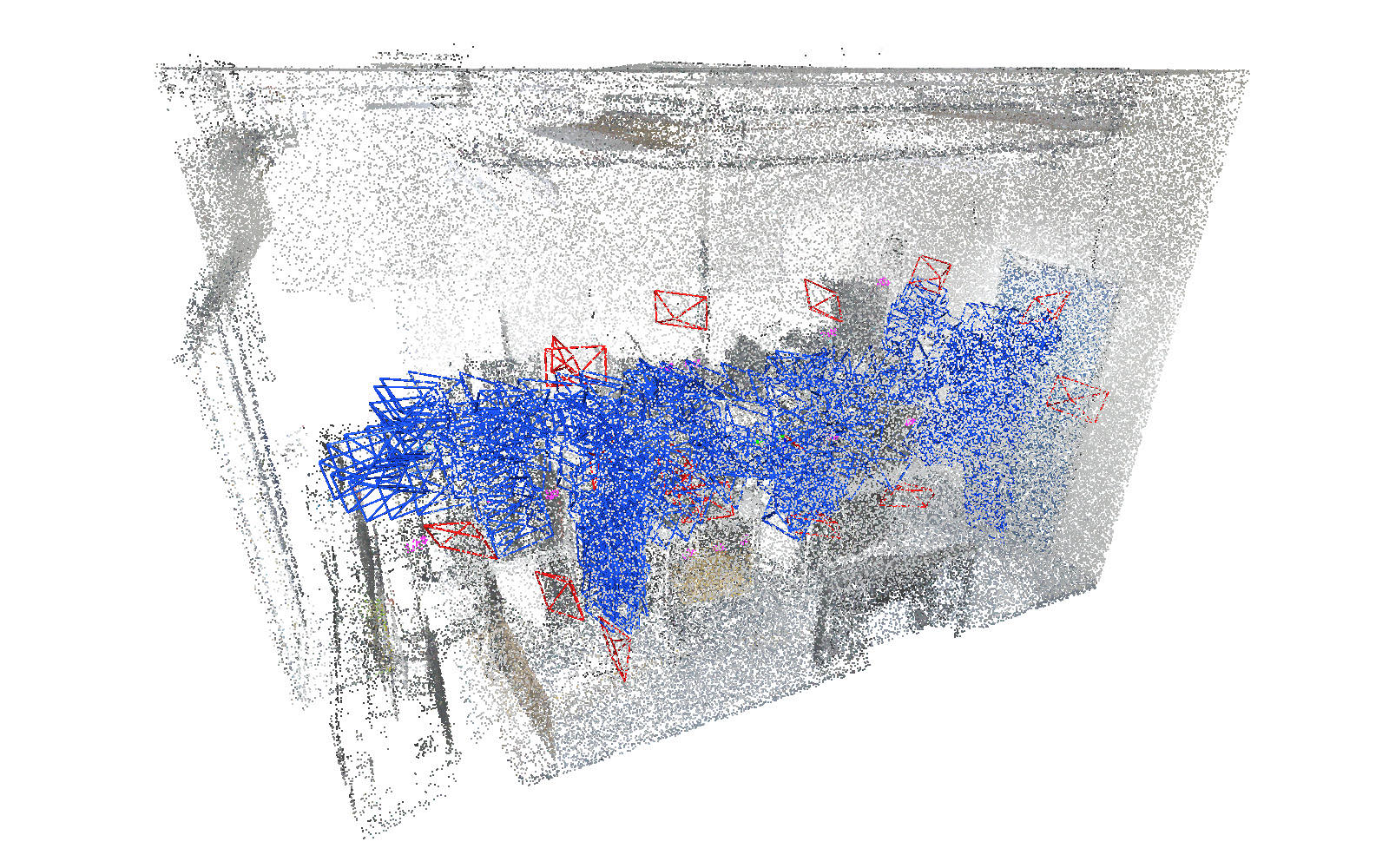} & 
        \includegraphics[width=\linewidth]{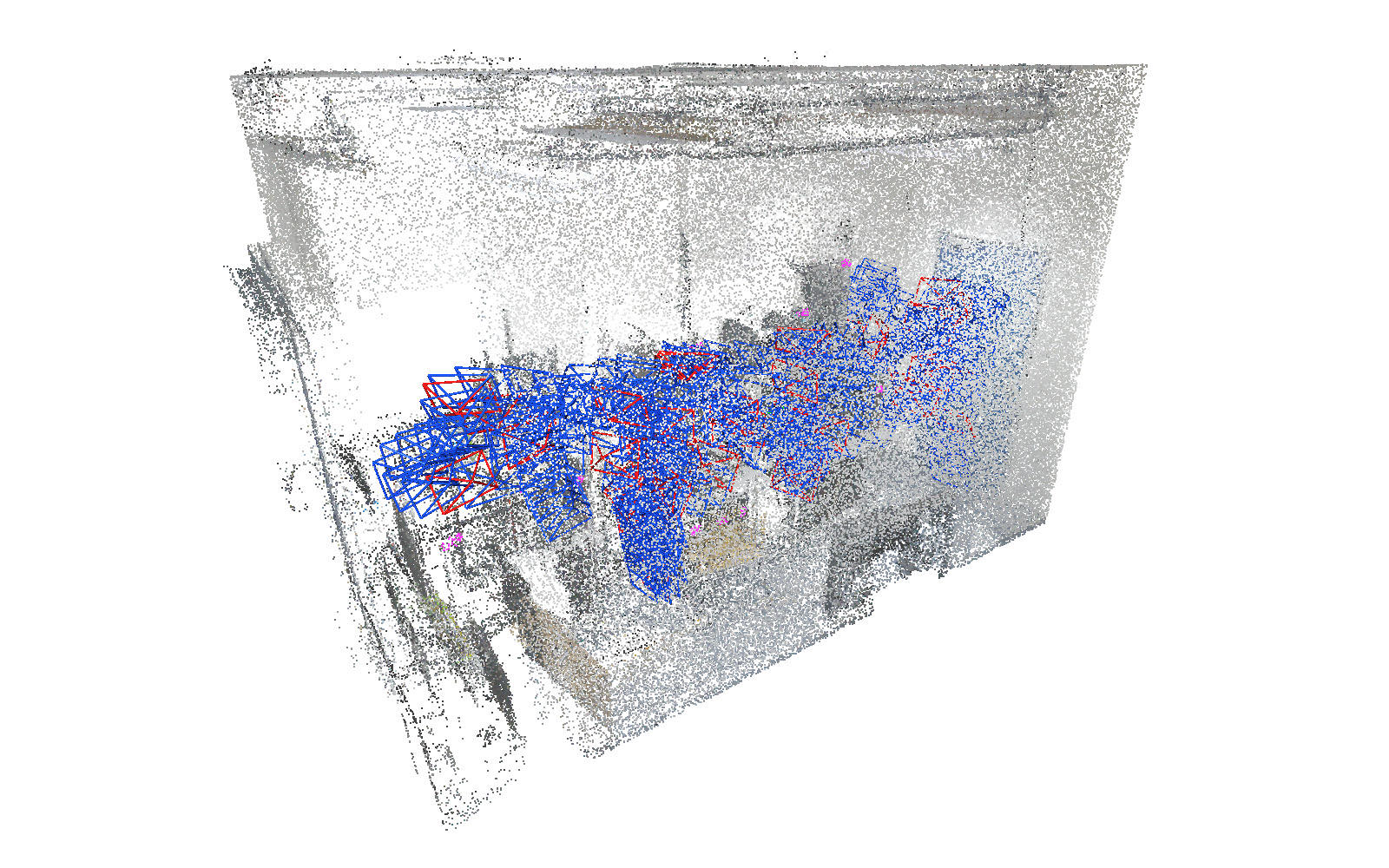} \\
    \end{tabular}
    \caption{Visualization of camera trajectories on 3 ScanNet++ scenes for the default split, and using our ``on-trajectory'' split,
    which we use to compare with default to test generalization. Training cameras are depicted in blue, and test cameras in red. Depicted scenes are (top to bottom): ``f3d64c30f8'', ``c5439f4607'', ``b0a08200c9''.}
    \label{fig:scannetpp_trajectories}
\end{figure}

As described in~\cref{sec:protocol:datasets} of the paper, we divide evaluation on ScanNet++ into two tracks -- using the default split provided by the dataset authors, and building our own test set out of every 8th training image (naturally, these images are removed from the training set). We do this, as the default test images on ScanNet++ are intentionally captured with camera positions far from the training trajectory~\cite{scannetpp}.  Thus, building our own second split, which uses images close to the training views, allows us to separately evaluate NVS generalization and on-trajectory view synthesis performance. In~\cref{fig:scannetpp_trajectories}, we provide examples on three scenes, illustrating both the default camera trajectories, and our ``on-trajectory'' splits.

\section{Examples of Initialization Perturbed by Noise}\label{noise_examples}
\begin{figure}[t]
    \centering
    \includegraphics[width=0.33\linewidth]{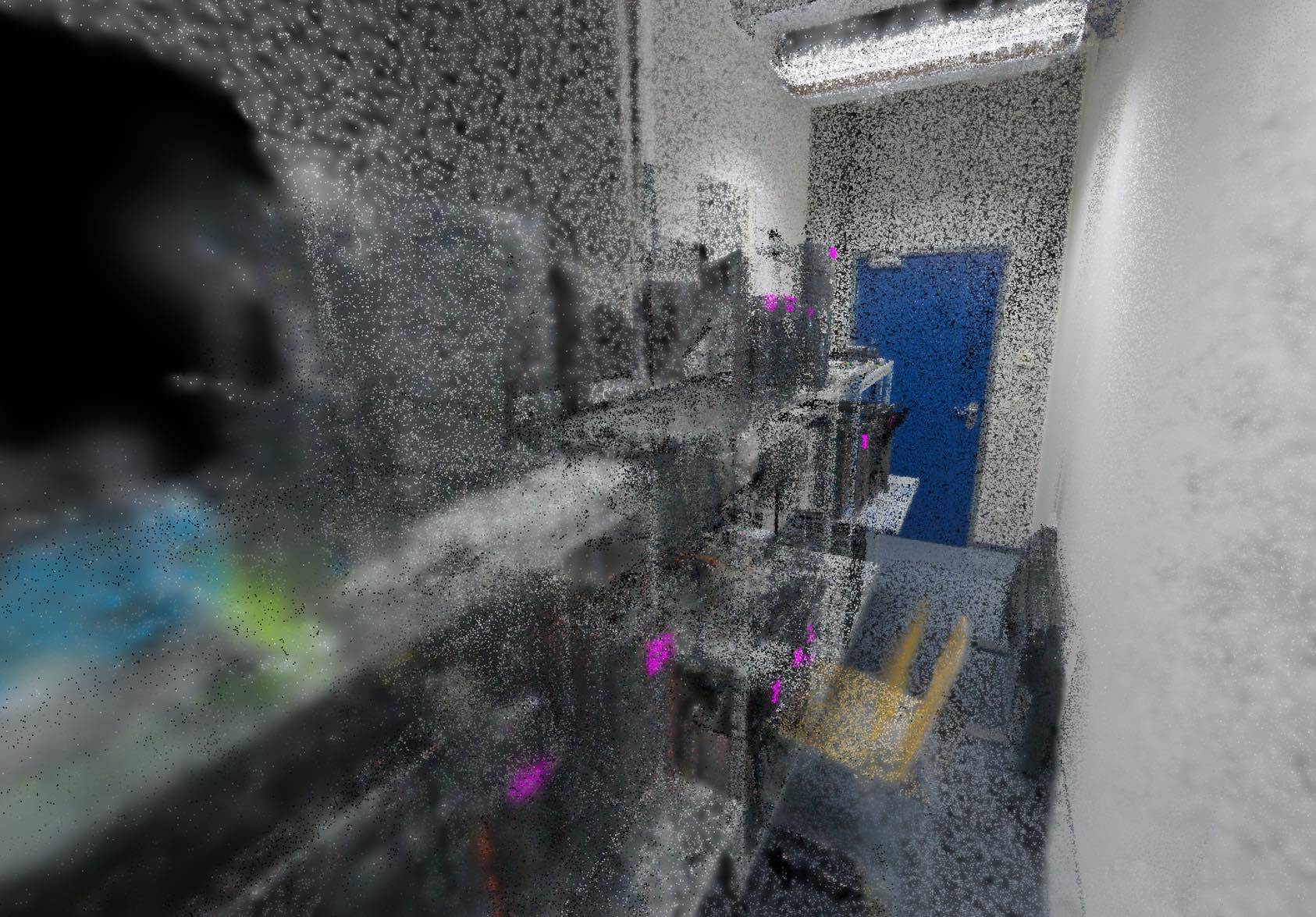}\hfill
    \includegraphics[width=0.33\linewidth]{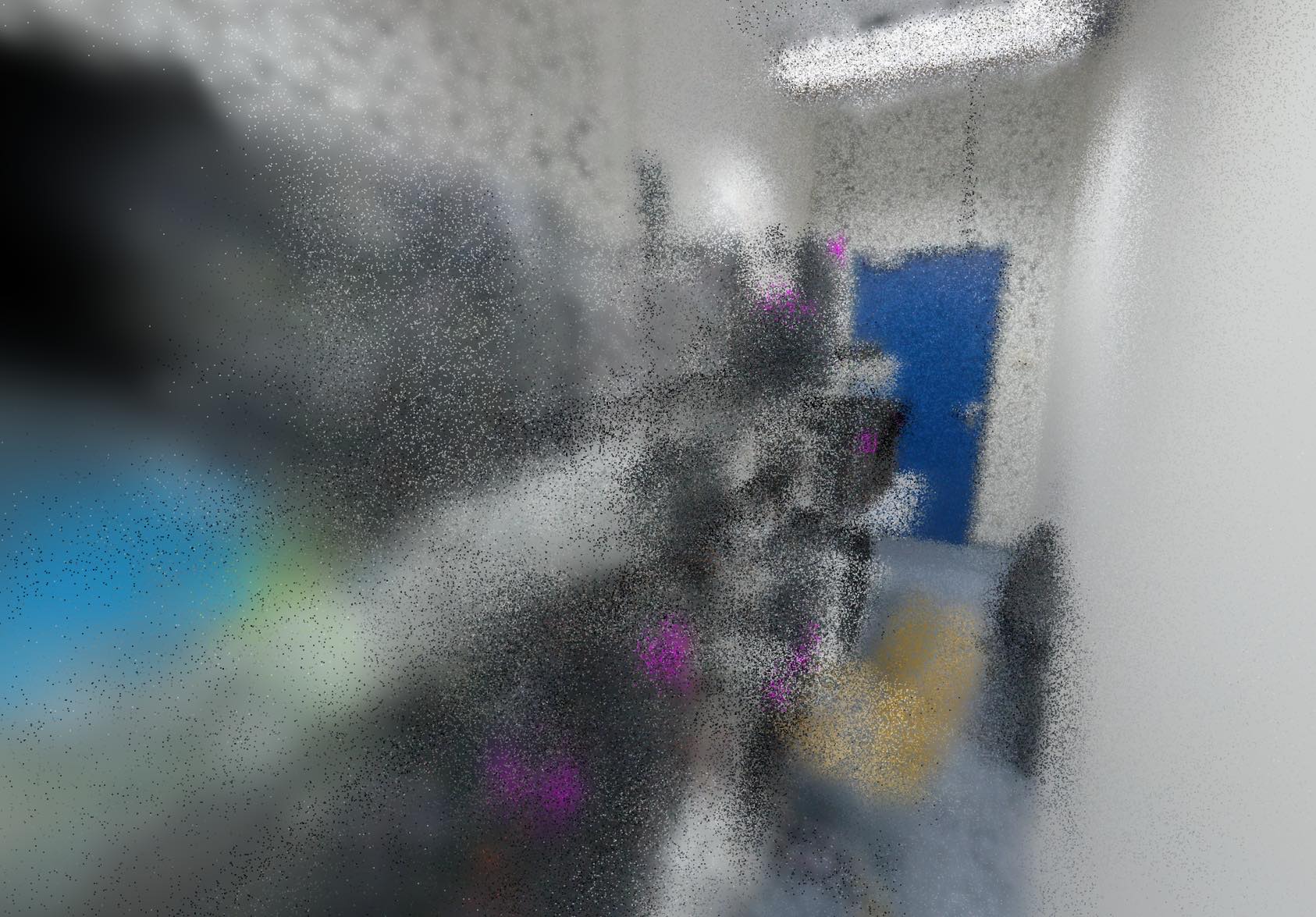}\hfill
    \includegraphics[width=0.33\linewidth]{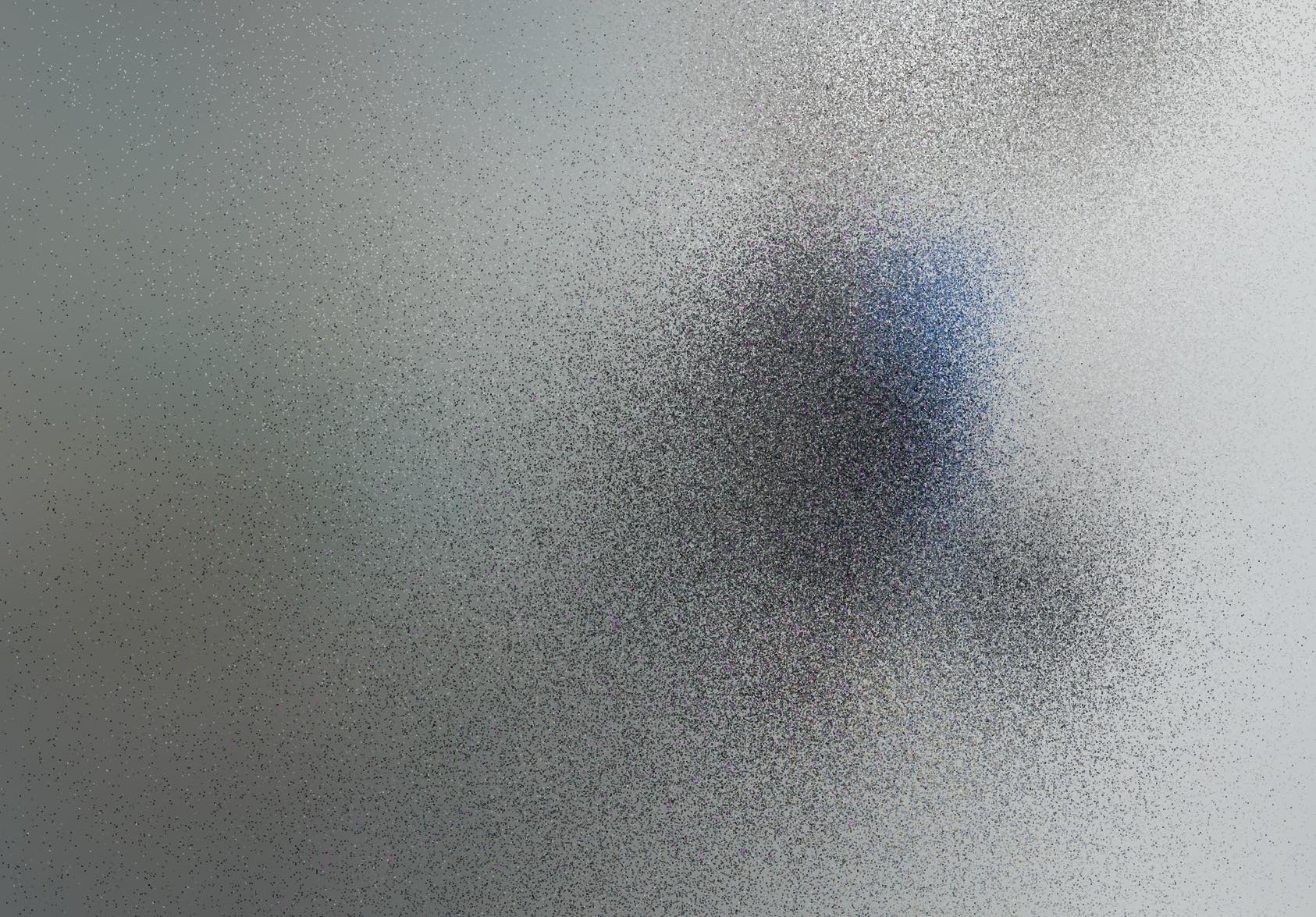}
    \caption{Examples of initial Gaussians produced from laser scan point clouds on the ``b0a08200c9'' ScanNet++ scene. Gaussian centers are visualized with colored dots. Noise levels, from left to right: no noise, $\sigma=0.01S$, $\sigma=0.1S$.}
    \label{fig:noise_examples}
\end{figure}

In \Cref{fig:noise_examples}, we provide an example of initial Gaussians produced from laser scan point clouds perturbed by normal noise at standard deviations equal to $1\%$ and $10\%$ of the scene extent $S$, as described in~\cref{subsec:eval:laser_init}.

\section{Statistical Analysis}\label{statistical_analysis}
\begin{table}[t]
\centering
\caption{Mean delta between SfM and dense initialization. In parentheses: (1) root-mean-square (RMS) of per-scene pooled standard deviations across the 3 seeds for SfM, (2) RMS of per-scene pooled standard deviations across the 3 seeds with the given initialization method, (3) Hedges' $g$ (effect size) averaged over scenes. (\textbf{ScanNet++})}
\label{tab:scannet++_dense_improvement}
{\small 
\begingroup \setlength{\tabcolsep}{1.5\tabcolsep}
\resizebox{0.95\columnwidth}{!}{\begin{tabular}{l|ccc}
\toprule
\textbf{Strategy} & \textbf{$\Delta$PSNR $\uparrow$} & \textbf{$\Delta$SSIM $\uparrow$} & \textbf{$\Delta$LPIPS $\downarrow$} \\
\midrule
& \multicolumn{3}{c}{\textbf{$\text{EDGS}^*$}} \\
\midrule
AbsGS & \cc{FCE1D3}{\color[HTML]{000000} $0.28 (.209, .118, 2.76)$} & \cc{76ABD1}{\color[HTML]{F1F1F1} $-0.007 (.002, .003, -5.01)$} & \cc{DBEBF4}{\color[HTML]{000000} $0.005 (.003, .003, 2.61)$} \\
INRIA & \cc{FBDAC9}{\color[HTML]{000000} $0.35 (.262, .142, 1.70)$} & \cc{91BFDB}{\color[HTML]{000000} $-0.006 (.002, .001, -5.55)$} & \cc{F1F7FB}{\color[HTML]{000000} $0.002 (.003, .003, 1.17)$} \\
MCMC & \cc{F6BA9F}{\color[HTML]{000000} $0.62 (.471, .215, 1.41)$} & \cc{F6BA9F}{\color[HTML]{000000} $0.005 (.008, .002, .469)$} & \cc{F9CAB2}{\color[HTML]{000000} $-0.009 (.011, .003, 0.624)$} \\
IDHFR & \cc{FFFDFB}{\color[HTML]{000000} $0.02 (.216, .138, .352)$} & \cc{ABD1E5}{\color[HTML]{000000} $-0.005 (.002, .001, -3.53)$} & \cc{BDDBEB}{\color[HTML]{000000} $0.009 (.003, .002, 3.88)$} \\
RevDGS & \cc{F5F9FC}{\color[HTML]{000000} $-0.08 (.159, .405, 0.471)$} & \cc{5C99C7}{\color[HTML]{F1F1F1} $-0.008 (.002, .006, -5.21)$} & \cc{D3E7F2}{\color[HTML]{000000} $0.006 (.003, .010, 1.44)$} \\
\midrule
& \multicolumn{3}{c}{\textbf{Monodepth}} \\
\midrule
AbsGS & \cc{F9C4AA}{\color[HTML]{000000} $0.55 (.209, .127, 2.52)$} & \cc{FBD7C5}{\color[HTML]{000000} $0.003 (.002, .002, 2.09)$} & \cc{FDE8DD}{\color[HTML]{000000} $-0.004 (.003, .003, 0.301)$} \\
INRIA & \cc{F7BCA1}{\color[HTML]{000000} $0.61 (.262, .160, 3.24)$} & \cc{FBD7C5}{\color[HTML]{000000} $0.003 (.002, .003, 2.34)$} & \cc{FBD6C4}{\color[HTML]{000000} $-0.007 (.003, .002, -2.47)$} \\
MCMC & \cc{F9C5AB}{\color[HTML]{000000} $0.54 (.471, .183, 1.37)$} & \cc{EB9D88}{\color[HTML]{000000} $0.006 (.008, .003, .966)$} & \cc{FAD0BB}{\color[HTML]{000000} $-0.008 (.011, .004, -1.98)$} \\
IDHFR & \cc{FBDDCE}{\color[HTML]{000000} $0.32 (.216, .127, 1.40)$} & \cc{FCE5D9}{\color[HTML]{000000} $0.002 (.002, .002, 1.69)$} & \cc{FFFFFE}{\color[HTML]{000000} $0.000 (.003, .002, 0.048)$} \\
RevDGS & \cc{FACEB8}{\color[HTML]{000000} $0.46 (.159, .146, 1.99)$} & \cc{FCE5D9}{\color[HTML]{000000} $0.002 (.002, .002, 1.47)$} & \cc{FBDCCC}{\color[HTML]{000000} $-0.006 (.003, .003, -2.05)$} \\
\midrule
& \multicolumn{3}{c}{\textbf{DA3}} \\
\midrule
AbsGS & \cc{F9C1A6}{\color[HTML]{000000} $0.58 (.209, .191, 2.59)$} & \cc{FBD7C5}{\color[HTML]{000000} $0.003 (.002, .002, 2.25)$} & \cc{FDE8DD}{\color[HTML]{000000} $-0.004 (.003, .003, -1.72)$} \\
INRIA & \cc{F7BCA1}{\color[HTML]{000000} $0.61 (.262, .141, 2.98)$} & \cc{F9CAB2}{\color[HTML]{000000} $0.004 (.002, .002, 2.56)$} & \cc{FBD6C4}{\color[HTML]{000000} $-0.007 (.003, .003, -2.31)$} \\
MCMC & \cc{F9C5AB}{\color[HTML]{000000} $0.54 (.471, .227, 1.17)$} & \cc{F6BA9F}{\color[HTML]{000000} $0.005 (.008, .002, .924)$} & \cc{FAD0BB}{\color[HTML]{000000} $-0.008 (.011, .004, -2.55)$} \\
IDHFR & \cc{FBDDCE}{\color[HTML]{000000} $0.32 (.216, .165, 1.11)$} & \cc{FCE5D9}{\color[HTML]{000000} $0.002 (.002, .002, 1.49)$} & \cc{F8FBFD}{\color[HTML]{000000} $0.001 (.003, .002, .496)$} \\
RevDGS & \cc{FACFBA}{\color[HTML]{000000} $0.45 (.159, .157, 1.26)$} & \cc{FCE5D9}{\color[HTML]{000000} $0.002 (.002, .001, 1.79)$} & \cc{FBDCCC}{\color[HTML]{000000} $-0.006 (.003, .002, -1.87)$} \\
\midrule
& \multicolumn{3}{c}{\textbf{$\text{DA3}^\text{G.S.}$}} \\
\midrule
AbsGS & \cc{D66960}{\color[HTML]{F1F1F1} $0.98 (.209, .128, 6.03)$} & \cc{FFFFFE}{\color[HTML]{000000} $0.000 (.002, .002, -1.74)$} & \cc{E99985}{\color[HTML]{000000} $-0.014 (.003, .002, -5.29)$} \\
INRIA & \cc{D86F65}{\color[HTML]{F1F1F1} $0.95 (.262, .110, 5.39)$} & \cc{EEF6FA}{\color[HTML]{000000} $-0.001 (.002, .001, -4.88)$} & \cc{E48C7B}{\color[HTML]{F1F1F1} $-0.015 (.003, .002, -5.88)$} \\
MCMC & \cc{EFA791}{\color[HTML]{000000} $0.70 (.471, .378, 1.63)$} & \cc{F6BA9F}{\color[HTML]{000000} $0.005 (.008, .005, .721)$} & \cc{EDA38D}{\color[HTML]{000000} $-0.013 (.011, .008, -1.79)$} \\
IDHFR & \cc{F7BCA1}{\color[HTML]{000000} $0.61 (.216, .150, 2.88)$} & \cc{EEF6FA}{\color[HTML]{000000} $-0.001 (.002, .002, -2.23)$} & \cc{FAD0BB}{\color[HTML]{000000} $-0.008 (.003, .002, -3.14)$} \\
RevDGS & \cc{D5675E}{\color[HTML]{F1F1F1} $0.99 (.159, .131, 5.55)$} & \cc{FFFFFE}{\color[HTML]{000000} $-0.000 (.002, .001, -2.87)$} & \cc{D5675E}{\color[HTML]{F1F1F1} $-0.018 (.003, .001, -8.48)$} \\
\midrule
& \multicolumn{3}{c}{\textbf{$1.0\text  {G}_\mathit{m}$ $\text{{Laser}}^+$}} \\
\midrule
AbsGS & \cc{F0A992}{\color[HTML]{000000} $0.69 (.209, .204, 4.13)$} & \cc{F9CAB2}{\color[HTML]{000000} $0.004 (.002, .003, 2.70)$} & \cc{F7BCA1}{\color[HTML]{000000} $-0.011 (.003, .004, -3.25)$} \\
INRIA & \cc{EFA791}{\color[HTML]{000000} $0.70 (.262, .139, 3.93)$} & \cc{F9CAB2}{\color[HTML]{000000} $0.004 (.002, .002, 4.20)$} & \cc{EDA38D}{\color[HTML]{000000} $-0.013 (.003, .003, -5.46)$} \\
MCMC & \cc{FAD2BE}{\color[HTML]{000000} $0.42 (.471, .413, 1.11)$} & \cc{F9CAB2}{\color[HTML]{000000} $0.004 (.008, .006, .857)$} & \cc{F9C4AA}{\color[HTML]{000000} $-0.010 (.011, .010, -2.46)$} \\
IDHFR & \cc{FBDDCE}{\color[HTML]{000000} $0.32 (.216, .162, 1.51)$} & \cc{FCE5D9}{\color[HTML]{000000} $0.002 (.002, .002, 2.11)$} & \cc{FDEEE6}{\color[HTML]{000000} $-0.003 (.003, .002, -1.81)$} \\
RevDGS & \cc{F9C4AA}{\color[HTML]{000000} $0.55 (.159, .161, 2.77)$} & \cc{F9CAB2}{\color[HTML]{000000} $0.004 (.002, .002, 3.24)$} & \cc{F9C4AA}{\color[HTML]{000000} $-0.010 (.003, .003, -3.98)$} \\
\bottomrule
\end{tabular}}
\endgroup
}
\end{table}

\begin{table}[t]
\centering
\caption{Mean delta between SfM and dense initialization. In parentheses: (1) root-mean-square (RMS) of per-scene pooled standard deviations across the 3 seeds for SfM, (2) RMS of per-scene pooled standard deviations across the 3 seeds with the given initialization method, (3) Hedges' $g$ (effect size) averaged over scenes. (\textbf{ScanNet++ (On-Trajectory)})}
\label{tab:eval_on_train_set_scannet++_dense_improvement}
{\small 
\begingroup \setlength{\tabcolsep}{1.5\tabcolsep}
\resizebox{0.95\columnwidth}{!}{\begin{tabular}{l|ccc}
\toprule
\textbf{Strategy} & \textbf{$\Delta$PSNR $\uparrow$} & \textbf{$\Delta$SSIM $\uparrow$} & \textbf{$\Delta$LPIPS $\downarrow$} \\
\midrule
& \multicolumn{3}{c}{\textbf{$\text{EDGS}^*$}} \\
\midrule
AbsGS & \cc{85B6D7}{\color[HTML]{000000} $-0.33 (.135, .125, -4.79)$} & \cc{5C99C7}{\color[HTML]{F1F1F1} $-0.002 (.001, .000, -7.21)$} & \cc{FCE0D2}{\color[HTML]{000000} $-0.005 (.001, .001, -7.45)$} \\
INRIA & \cc{93C0DC}{\color[HTML]{000000} $-0.30 (.112, .105, -4.51)$} & \cc{5C99C7}{\color[HTML]{F1F1F1} $-0.002 (.000, .001, -3.14)$} & \cc{E58E7D}{\color[HTML]{F1F1F1} $-0.014 (.001, .001, -21.5)$} \\
MCMC & \cc{FCE0D2}{\color[HTML]{000000} $0.12 (.286, .152, .788)$} & \cc{F9CAB2}{\color[HTML]{000000} $0.001 (.003, .001, 1.85)$} & \cc{FACDB7}{\color[HTML]{000000} $-0.008 (.006, .003, -6.92)$} \\
IDHFR & \cc{FFFDFB}{\color[HTML]{000000} $0.01 (.371, .067, -1.89)$} & \cc{FFFFFE}{\color[HTML]{000000} $-0.000 (.001, .000, 0.718)$} & \cc{FCE6DB}{\color[HTML]{000000} $-0.004 (.001, .000, -8.53)$} \\
RevDGS & \cc{5C99C7}{\color[HTML]{F1F1F1} $-0.41 (.189, .082, -6.24)$} & \cc{5C99C7}{\color[HTML]{F1F1F1} $-0.002 (.001, .001, -7.82)$} & \cc{FCE0D2}{\color[HTML]{000000} $-0.005 (.003, .003, -2.33)$} \\
\midrule
& \multicolumn{3}{c}{\textbf{Monodepth}} \\
\midrule
AbsGS & \cc{89B9D8}{\color[HTML]{000000} $-0.32 (.135, .070, -5.23)$} & \cc{BDDBEB}{\color[HTML]{000000} $-0.001 (.001, .000, -4.18)$} & \cc{FEF3ED}{\color[HTML]{000000} $-0.002 (.001, .001, -3.90)$} \\
INRIA & \cc{85B6D7}{\color[HTML]{000000} $-0.33 (.112, .106, -4.46)$} & \cc{BDDBEB}{\color[HTML]{000000} $-0.001 (.000, .001, -1.02)$} & \cc{F4B49A}{\color[HTML]{000000} $-0.011 (.001, .001, -15.9)$} \\
MCMC & \cc{C8E1EE}{\color[HTML]{000000} $-0.17 (.286, .114, -1.54)$} & \cc{FFFFFE}{\color[HTML]{000000} $0.000 (.003, .000, .272)$} & \cc{FCE0D2}{\color[HTML]{000000} $-0.005 (.006, .001, -3.83)$} \\
IDHFR & \cc{EEF6FA}{\color[HTML]{000000} $-0.05 (.371, .106, -2.10)$} & \cc{FFFFFE}{\color[HTML]{000000} $-0.000 (.001, .000, -1.49)$} & \cc{FEF9F6}{\color[HTML]{000000} $-0.001 (.001, .001, -3.19)$} \\
RevDGS & \cc{D8EAF3}{\color[HTML]{000000} $-0.12 (.189, .092, -4.35)$} & \cc{FFFFFE}{\color[HTML]{000000} $0.000 (.001, .000, -2.17)$} & \cc{FCE0D2}{\color[HTML]{000000} $-0.005 (.003, .001, -2.77)$} \\
\midrule
& \multicolumn{3}{c}{\textbf{DA3}} \\
\midrule
AbsGS & \cc{85B6D7}{\color[HTML]{000000} $-0.33 (.135, .083, -5.36)$} & \cc{BDDBEB}{\color[HTML]{000000} $-0.001 (.001, .000, -4.37)$} & \cc{FEF3ED}{\color[HTML]{000000} $-0.002 (.001, .001, -3.55)$} \\
INRIA & \cc{89B9D8}{\color[HTML]{000000} $-0.32 (.112, .096, -4.00)$} & \cc{FFFFFE}{\color[HTML]{000000} $-0.000 (.000, .000, 0.883)$} & \cc{F4B49A}{\color[HTML]{000000} $-0.011 (.001, .001, -17.0)$} \\
MCMC & \cc{C8E1EE}{\color[HTML]{000000} $-0.17 (.286, .112, -1.62)$} & \cc{FFFFFE}{\color[HTML]{000000} $0.000 (.003, .001, .516)$} & \cc{FCE0D2}{\color[HTML]{000000} $-0.005 (.006, .001, -3.97)$} \\
IDHFR & \cc{F2F8FB}{\color[HTML]{000000} $-0.04 (.371, .101, -2.07)$} & \cc{FFFFFE}{\color[HTML]{000000} $-0.000 (.001, .000, -1.85)$} & \cc{FEF9F6}{\color[HTML]{000000} $-0.001 (.001, .001, -3.42)$} \\
RevDGS & \cc{D8EAF3}{\color[HTML]{000000} $-0.12 (.189, .098, -3.42)$} & \cc{FFFFFE}{\color[HTML]{000000} $0.000 (.001, .001, -2.65)$} & \cc{FCE0D2}{\color[HTML]{000000} $-0.005 (.003, .001, -3.83)$} \\
\midrule
& \multicolumn{3}{c}{\textbf{$\text{DA3}^\text{G.S.}$}} \\
\midrule
AbsGS & \cc{F7BCA1}{\color[HTML]{000000} $0.25 (.135, .112, 1.48)$} & \cc{BDDBEB}{\color[HTML]{000000} $-0.001 (.001, .000, -2.67)$} & \cc{FCE6DB}{\color[HTML]{000000} $-0.004 (.001, .001, -4.99)$} \\
INRIA & \cc{FBD6C4}{\color[HTML]{000000} $0.16 (.112, .110, 1.13)$} & \cc{5C99C7}{\color[HTML]{F1F1F1} $-0.002 (.000, .000, -2.30)$} & \cc{F9C1A6}{\color[HTML]{000000} $-0.010 (.001, .001, -15.8)$} \\
MCMC & \cc{ECA18C}{\color[HTML]{000000} $0.30 (.286, .137, 1.98)$} & \cc{F9CAB2}{\color[HTML]{000000} $0.001 (.003, .001, .468)$} & \cc{FCE6DB}{\color[HTML]{000000} $-0.004 (.006, .001, 0.934)$} \\
IDHFR & \cc{E08273}{\color[HTML]{F1F1F1} $0.36 (.371, .088, 1.65)$} & \cc{FFFFFE}{\color[HTML]{000000} $0.000 (.001, .000, 0.752)$} & \cc{FDEDE5}{\color[HTML]{000000} $-0.003 (.001, .001, -5.31)$} \\
RevDGS & \cc{FACEB8}{\color[HTML]{000000} $0.19 (.189, .219, .531)$} & \cc{FFFFFE}{\color[HTML]{000000} $-0.000 (.001, .002, -3.39)$} & \cc{FBDAC9}{\color[HTML]{000000} $-0.006 (.003, .002, -1.85)$} \\
\midrule
& \multicolumn{3}{c}{\textbf{$1.0\text  {G}_\mathit{m}$ $\text{{Laser}}^+$}} \\
\midrule
AbsGS & \cc{F9C6AD}{\color[HTML]{000000} $0.22 (.135, .081, 1.76)$} & \cc{F9CAB2}{\color[HTML]{000000} $0.001 (.001, .000, 2.55)$} & \cc{FACDB7}{\color[HTML]{000000} $-0.008 (.001, .001, -11.7)$} \\
INRIA & \cc{F9C9B1}{\color[HTML]{000000} $0.21 (.112, .081, 2.09)$} & \cc{F9CAB2}{\color[HTML]{000000} $0.001 (.000, .001, 6.90)$} & \cc{D5675E}{\color[HTML]{F1F1F1} $-0.017 (.001, .001, -24.5)$} \\
MCMC & \cc{FAD1BC}{\color[HTML]{000000} $0.18 (.286, .236, .660)$} & \cc{D5675E}{\color[HTML]{F1F1F1} $0.002 (.003, .001, 1.59)$} & \cc{F9C7AE}{\color[HTML]{000000} $-0.009 (.006, .003, -5.21)$} \\
IDHFR & \cc{E48C7B}{\color[HTML]{F1F1F1} $0.34 (.371, .174, 1.46)$} & \cc{F9CAB2}{\color[HTML]{000000} $0.001 (.001, .001, 3.92)$} & \cc{FCE0D2}{\color[HTML]{000000} $-0.005 (.001, .001, -8.32)$} \\
RevDGS & \cc{F9C9B1}{\color[HTML]{000000} $0.21 (.189, .111, .902)$} & \cc{D5675E}{\color[HTML]{F1F1F1} $0.002 (.001, .001, 3.69)$} & \cc{F9C1A6}{\color[HTML]{000000} $-0.010 (.003, .001, -9.07)$} \\
\bottomrule
\end{tabular}}
\endgroup
}
\end{table}

\begin{table}[t]
\centering
\caption{Mean delta between SfM and dense initialization. In parentheses: (1) root-mean-square (RMS) of per-scene pooled standard deviations across the 3 seeds for SfM, (2) RMS of per-scene pooled standard deviations across the 3 seeds with the given initialization method, (3) Hedges' $g$ (effect size) averaged over scenes. (\textbf{ETH3D})}
\label{tab:eth3d_dense_improvement}
{\small 
\begingroup \setlength{\tabcolsep}{1.5\tabcolsep}
\resizebox{0.95\columnwidth}{!}{\begin{tabular}{l|ccc}
\toprule
\textbf{Strategy} & \textbf{$\Delta$PSNR $\uparrow$} & \textbf{$\Delta$SSIM $\uparrow$} & \textbf{$\Delta$LPIPS $\downarrow$} \\
\midrule
& \multicolumn{3}{c}{\textbf{$\text{EDGS}^*$}} \\
\midrule
AbsGS & \cc{FBD7C5}{\color[HTML]{000000} $1.20 (.514, .426, 2.62)$} & \cc{FBD8C7}{\color[HTML]{000000} $0.030 (.010, .011, 7.87)$} & \cc{FCDECF}{\color[HTML]{000000} $-0.049 (.013, .030, -7.26)$} \\
INRIA & \cc{FCDECF}{\color[HTML]{000000} $0.99 (.711, .327, 1.45)$} & \cc{FDECE3}{\color[HTML]{000000} $0.015 (.013, .007, .243)$} & \cc{FBD7C5}{\color[HTML]{000000} $-0.058 (.021, .007, -15.6)$} \\
MCMC & \cc{FCE2D5}{\color[HTML]{000000} $0.87 (.958, .304, .137)$} & \cc{FCE3D6}{\color[HTML]{000000} $0.022 (.017, .002, 2.25)$} & \cc{FCDECF}{\color[HTML]{000000} $-0.049 (.057, .003, -4.00)$} \\
IDHFR & \cc{FDEDE5}{\color[HTML]{000000} $0.55 (.627, .398, .693)$} & \cc{FEF5F0}{\color[HTML]{000000} $0.008 (.006, .007, .903)$} & \cc{FDEDE5}{\color[HTML]{000000} $-0.027 (.016, .009, -3.77)$} \\
RevDGS & \cc{EDA38D}{\color[HTML]{000000} $2.31 (.744, .891, 2.78)$} & \cc{EFA791}{\color[HTML]{000000} $0.058 (.015, .038, 8.61)$} & \cc{E58E7D}{\color[HTML]{F1F1F1} $-0.127 (.028, .048, -6.63)$} \\
\midrule
& \multicolumn{3}{c}{\textbf{Monodepth}} \\
\midrule
AbsGS & \cc{FAD1BC}{\color[HTML]{000000} $1.38 (.514, .487, 3.84)$} & \cc{FBD8C7}{\color[HTML]{000000} $0.030 (.010, .006, 9.52)$} & \cc{FCDFD1}{\color[HTML]{000000} $-0.047 (.013, .015, -9.39)$} \\
INRIA & \cc{FCE6DB}{\color[HTML]{000000} $0.76 (.711, .673, 1.82)$} & \cc{FDEDE5}{\color[HTML]{000000} $0.014 (.013, .014, 4.86)$} & \cc{FCE4D8}{\color[HTML]{000000} $-0.039 (.021, .038, -8.92)$} \\
MCMC & \cc{FDE8DD}{\color[HTML]{000000} $0.70 (.958, .302, .466)$} & \cc{FDEFE7}{\color[HTML]{000000} $0.013 (.017, .002, 1.06)$} & \cc{FCE4D8}{\color[HTML]{000000} $-0.040 (.057, .005, -1.93)$} \\
IDHFR & \cc{FDECE3}{\color[HTML]{000000} $0.59 (.627, .604, 1.37)$} & \cc{FDEDE5}{\color[HTML]{000000} $0.014 (.006, .014, 3.72)$} & \cc{FEF4EF}{\color[HTML]{000000} $-0.016 (.016, .040, -4.42)$} \\
RevDGS & \cc{E28878}{\color[HTML]{F1F1F1} $2.70 (.744, .853, 5.21)$} & \cc{E79482}{\color[HTML]{F1F1F1} $0.065 (.015, .007, 9.50)$} & \cc{E28878}{\color[HTML]{F1F1F1} $-0.131 (.028, .012, -8.00)$} \\
\midrule
& \multicolumn{3}{c}{\textbf{DA3}} \\
\midrule
AbsGS & \cc{FBD6C4}{\color[HTML]{000000} $1.24 (.514, .446, 3.93)$} & \cc{FDEFE7}{\color[HTML]{000000} $0.013 (.010, .016, -4.87)$} & \cc{FCE3D6}{\color[HTML]{000000} $-0.041 (.013, .031, 0.386)$} \\
INRIA & \cc{FCE6DB}{\color[HTML]{000000} $0.76 (.711, .304, 1.51)$} & \cc{FEFAF7}{\color[HTML]{000000} $0.004 (.013, .008, -7.55)$} & \cc{FCDFD1}{\color[HTML]{000000} $-0.047 (.021, .005, -3.16)$} \\
MCMC & \cc{FEF3ED}{\color[HTML]{000000} $0.38 (.958, .490, -1.98)$} & \cc{FEF9F6}{\color[HTML]{000000} $0.005 (.017, .005, -9.96)$} & \cc{FCE4D8}{\color[HTML]{000000} $-0.039 (.057, .007, 2.82)$} \\
IDHFR & \cc{FFFBF9}{\color[HTML]{000000} $0.14 (.627, .330, .913)$} & \cc{FBFDFE}{\color[HTML]{000000} $-0.003 (.006, .010, -1.87)$} & \cc{FEF7F3}{\color[HTML]{000000} $-0.012 (.016, .012, .361)$} \\
RevDGS & \cc{F0A992}{\color[HTML]{000000} $2.23 (.744, .874, 3.34)$} & \cc{F9C4AA}{\color[HTML]{000000} $0.046 (.015, .013, 3.79)$} & \cc{E38A7A}{\color[HTML]{F1F1F1} $-0.129 (.028, .011, -6.04)$} \\
\midrule
& \multicolumn{3}{c}{\textbf{$\text{DA3}^\text{G.S.}$}} \\
\midrule
AbsGS & \cc{F2AF97}{\color[HTML]{000000} $2.13 (.514, .166, 7.23)$} & \cc{FCE4D8}{\color[HTML]{000000} $0.021 (.010, .004, .549)$} & \cc{FBDBCB}{\color[HTML]{000000} $-0.052 (.013, .016, -4.07)$} \\
INRIA & \cc{FAD1BC}{\color[HTML]{000000} $1.40 (.711, .716, 2.45)$} & \cc{FCFDFE}{\color[HTML]{000000} $-0.002 (.013, .012, -7.83)$} & \cc{FCE4D8}{\color[HTML]{000000} $-0.040 (.021, .007, -5.74)$} \\
MCMC & \cc{FCE4D8}{\color[HTML]{000000} $0.83 (.958, .457, .180)$} & \cc{FDF0E9}{\color[HTML]{000000} $0.012 (.017, .005, -5.39)$} & \cc{FCDFD1}{\color[HTML]{000000} $-0.047 (.057, .008, 1.70)$} \\
IDHFR & \cc{FDE9DF}{\color[HTML]{000000} $0.67 (.627, .441, 1.21)$} & \cc{F2F8FB}{\color[HTML]{000000} $-0.008 (.006, .007, -3.92)$} & \cc{FEF4EF}{\color[HTML]{000000} $-0.017 (.016, .011, 1.75)$} \\
RevDGS & \cc{D5675E}{\color[HTML]{F1F1F1} $3.20 (.744, .883, 5.59)$} & \cc{F9C5AB}{\color[HTML]{000000} $0.045 (.015, .026, 4.11)$} & \cc{E18475}{\color[HTML]{F1F1F1} $-0.133 (.028, .021, -6.29)$} \\
\midrule
& \multicolumn{3}{c}{\textbf{$1.0\text  {G}_\mathit{m}$ $\text{{Laser}}$}} \\
\midrule
AbsGS & \cc{F7BCA1}{\color[HTML]{000000} $1.97 (.514, .485, 4.32)$} & \cc{F4B69C}{\color[HTML]{000000} $0.053 (.010, .019, 18.8)$} & \cc{F9C4AA}{\color[HTML]{000000} $-0.086 (.013, .022, -13.4)$} \\
INRIA & \cc{FAD1BC}{\color[HTML]{000000} $1.39 (.711, .478, 2.16)$} & \cc{FBD7C5}{\color[HTML]{000000} $0.031 (.013, .006, 5.68)$} & \cc{F9C7AE}{\color[HTML]{000000} $-0.081 (.021, .017, -19.5)$} \\
MCMC & \cc{FCE1D3}{\color[HTML]{000000} $0.91 (.958, .332, .964)$} & \cc{FCE0D2}{\color[HTML]{000000} $0.024 (.017, .003, 4.47)$} & \cc{FBD7C5}{\color[HTML]{000000} $-0.058 (.057, .005, -5.55)$} \\
IDHFR & \cc{FBDBCB}{\color[HTML]{000000} $1.08 (.627, .582, 2.33)$} & \cc{FAD1BC}{\color[HTML]{000000} $0.036 (.006, .010, 7.62)$} & \cc{FAD0BB}{\color[HTML]{000000} $-0.068 (.016, .008, -11.0)$} \\
RevDGS & \cc{D97166}{\color[HTML]{F1F1F1} $3.05 (.744, 1.06, 4.19)$} & \cc{D5675E}{\color[HTML]{F1F1F1} $0.082 (.015, .036, 12.1)$} & \cc{D5675E}{\color[HTML]{F1F1F1} $-0.154 (.028, .035, -10.4)$} \\
\bottomrule
\end{tabular}}
\endgroup
}
\end{table}

\begin{table}[t]
\centering
\caption{Mean delta between SfM and dense initialization. In parentheses: (1) root-mean-square (RMS) of per-scene pooled standard deviations across the 3 seeds for SfM, (2) RMS of per-scene pooled standard deviations across the 3 seeds with the given initialization method, (3) Hedges' $g$ (effect size) averaged over scenes. (\textbf{Mip-NeRF 360})}
\label{tab:mipnerf360_dense_improvement}
{\small 
\begingroup \setlength{\tabcolsep}{1.5\tabcolsep}
\resizebox{0.95\columnwidth}{!}{\begin{tabular}{l|ccc}
\toprule
\textbf{Strategy} & \textbf{$\Delta$PSNR $\uparrow$} & \textbf{$\Delta$SSIM $\uparrow$} & \textbf{$\Delta$LPIPS $\downarrow$} \\
\midrule
& \multicolumn{3}{c}{\textbf{$\text{EDGS}^*$}} \\
\midrule
AbsGS & \cc{F7FBFD}{\color[HTML]{000000} $-0.09 (.114, .036, -2.65)$} & \cc{FFFFFE}{\color[HTML]{000000} $-0.000 (.001, .001, 0.129)$} & \cc{FDEFE7}{\color[HTML]{000000} $-0.006 (.001, .001, -6.82)$} \\
INRIA & \cc{FEFFFF}{\color[HTML]{000000} $-0.01 (.137, .058, 0.612)$} & \cc{FBDDCE}{\color[HTML]{000000} $0.010 (.002, .000, 12.6)$} & \cc{D5675E}{\color[HTML]{F1F1F1} $-0.039 (.003, .001, -45.5)$} \\
MCMC & \cc{E6F1F7}{\color[HTML]{000000} $-0.28 (.107, .057, -6.79)$} & \cc{E1EFF6}{\color[HTML]{000000} $-0.007 (.001, .000, -8.09)$} & \cc{FCE1D3}{\color[HTML]{000000} $-0.011 (.001, .001, -13.0)$} \\
IDHFR & \cc{F1F7FB}{\color[HTML]{000000} $-0.16 (.071, .046, -3.17)$} & \cc{FFFFFE}{\color[HTML]{000000} $-0.000 (.002, .001, 1.03)$} & \cc{FEF7F3}{\color[HTML]{000000} $-0.003 (.001, .001, -5.16)$} \\
RevDGS & \cc{FFFDFB}{\color[HTML]{000000} $0.03 (.063, .108, 1.83)$} & \cc{FCE1D3}{\color[HTML]{000000} $0.009 (.001, .001, 10.4)$} & \cc{FBD9C8}{\color[HTML]{000000} $-0.014 (.001, .001, -26.6)$} \\
\midrule
& \multicolumn{3}{c}{\textbf{Monodepth}} \\
\midrule
AbsGS & \cc{FEF1EA}{\color[HTML]{000000} $0.20 (.114, .044, 1.80)$} & \cc{FEF2EC}{\color[HTML]{000000} $0.004 (.001, .001, 3.75)$} & \cc{FEF5F0}{\color[HTML]{000000} $-0.004 (.001, .001, -4.60)$} \\
INRIA & \cc{FDEDE5}{\color[HTML]{000000} $0.25 (.137, .067, 3.62)$} & \cc{FBDDCE}{\color[HTML]{000000} $0.010 (.002, .001, 13.1)$} & \cc{EEA58F}{\color[HTML]{000000} $-0.028 (.003, .002, -24.2)$} \\
MCMC & \cc{E5F1F7}{\color[HTML]{000000} $-0.29 (.107, .037, -8.93)$} & \cc{CCE3EF}{\color[HTML]{000000} $-0.012 (.001, .000, -15.5)$} & \cc{FCE7DC}{\color[HTML]{000000} $-0.009 (.001, .001, -9.23)$} \\
IDHFR & \cc{FFFBF9}{\color[HTML]{000000} $0.06 (.071, .022, 1.84)$} & \cc{FEF9F6}{\color[HTML]{000000} $0.002 (.002, .001, 4.57)$} & \cc{FFFDFB}{\color[HTML]{000000} $-0.001 (.001, .000, -2.84)$} \\
RevDGS & \cc{FEFAF7}{\color[HTML]{000000} $0.07 (.063, .087, 0.197)$} & \cc{FEF2EC}{\color[HTML]{000000} $0.004 (.001, .001, .406)$} & \cc{FDEFE7}{\color[HTML]{000000} $-0.006 (.001, .001, -6.27)$} \\
\midrule
& \multicolumn{3}{c}{\textbf{DA3}} \\
\midrule
AbsGS & \cc{83B5D6}{\color[HTML]{000000} $-1.18 (.114, .265, -5.20)$} & \cc{99C4DE}{\color[HTML]{000000} $-0.022 (.001, .003, -6.94)$} & \cc{CAE2EF}{\color[HTML]{000000} $0.016 (.001, .006, .347)$} \\
INRIA & \cc{A5CDE3}{\color[HTML]{000000} $-0.95 (.137, .098, -8.73)$} & \cc{D5E8F2}{\color[HTML]{000000} $-0.010 (.002, .001, -4.53)$} & \cc{FAD1BC}{\color[HTML]{000000} $-0.017 (.003, .002, -15.8)$} \\
MCMC & \cc{6AA3CC}{\color[HTML]{F1F1F1} $-1.35 (.107, .102, -14.5)$} & \cc{5C99C7}{\color[HTML]{F1F1F1} $-0.031 (.001, .002, -24.8)$} & \cc{DDEDF5}{\color[HTML]{000000} $0.010 (.001, .003, 2.72)$} \\
IDHFR & \cc{C1DDEC}{\color[HTML]{000000} $-0.69 (.071, .094, -7.10)$} & \cc{D5E8F2}{\color[HTML]{000000} $-0.010 (.002, .002, -8.74)$} & \cc{EBF4F9}{\color[HTML]{000000} $0.006 (.001, .002, 2.40)$} \\
RevDGS & \cc{5C99C7}{\color[HTML]{F1F1F1} $-1.45 (.063, .181, -6.19)$} & \cc{9FC9E0}{\color[HTML]{000000} $-0.021 (.001, .003, -5.58)$} & \cc{C6E0ED}{\color[HTML]{000000} $0.017 (.001, .003, -4.03)$} \\
\midrule
& \multicolumn{3}{c}{\textbf{$\text{DA3}^\text{G.S.}$}} \\
\midrule
AbsGS & \cc{FBFDFE}{\color[HTML]{000000} $-0.05 (.114, .055, -2.33)$} & \cc{EAF3F8}{\color[HTML]{000000} $-0.005 (.001, .001, -5.34)$} & \cc{FEFAF7}{\color[HTML]{000000} $-0.002 (.001, .001, -2.62)$} \\
INRIA & \cc{F6FAFC}{\color[HTML]{000000} $-0.10 (.137, .053, -4.07)$} & \cc{F2F8FB}{\color[HTML]{000000} $-0.003 (.002, .001, -3.93)$} & \cc{F0AB94}{\color[HTML]{000000} $-0.027 (.003, .001, -28.0)$} \\
MCMC & \cc{E0EEF5}{\color[HTML]{000000} $-0.35 (.107, .048, -8.93)$} & \cc{C4DFED}{\color[HTML]{000000} $-0.014 (.001, .001, -16.3)$} & \cc{FEF5F0}{\color[HTML]{000000} $-0.004 (.001, .001, -2.01)$} \\
IDHFR & \cc{E2EFF6}{\color[HTML]{000000} $-0.32 (.071, .028, -10.9)$} & \cc{D5E8F2}{\color[HTML]{000000} $-0.010 (.002, .001, -18.8)$} & \cc{F5F9FC}{\color[HTML]{000000} $0.003 (.001, .001, 4.84)$} \\
RevDGS & \cc{F5F9FC}{\color[HTML]{000000} $-0.12 (.063, .053, -2.08)$} & \cc{EAF3F8}{\color[HTML]{000000} $-0.005 (.001, .001, -9.87)$} & \cc{FEF2EC}{\color[HTML]{000000} $-0.005 (.001, .001, -2.91)$} \\
\bottomrule
\end{tabular}}
\endgroup
}
\end{table}

\begin{table}[t]
\centering
\caption{Mean delta between SfM and dense initialization. In parentheses: (1) root-mean-square (RMS) of per-scene pooled standard deviations across the 3 seeds for SfM, (2) RMS of per-scene pooled standard deviations across the 3 seeds with the given initialization method, (3) Hedges' $g$ (effect size) averaged over scenes. (\textbf{Tanks and Temples})}
\label{tab:tanksandtemples_dense_improvement}
{\small 
\begingroup \setlength{\tabcolsep}{1.5\tabcolsep}
\resizebox{0.95\columnwidth}{!}{\begin{tabular}{l|ccc}
\toprule
\textbf{Strategy} & \textbf{$\Delta$PSNR $\uparrow$} & \textbf{$\Delta$SSIM $\uparrow$} & \textbf{$\Delta$LPIPS $\downarrow$} \\
\midrule
& \multicolumn{3}{c}{\textbf{$\text{EDGS}^*$}} \\
\midrule
AbsGS & \cc{FCDFD1}{\color[HTML]{000000} $0.24 (.221, .174, 0.343)$} & \cc{F9C7AE}{\color[HTML]{000000} $0.008 (.005, .002, 1.31)$} & \cc{D5675E}{\color[HTML]{F1F1F1} $-0.026 (.010, .004, -9.27)$} \\
INRIA & \cc{DDEDF5}{\color[HTML]{000000} $-0.21 (.125, .137, -1.39)$} & \cc{FEF8F4}{\color[HTML]{000000} $0.001 (.001, .001, 2.50)$} & \cc{EDA38D}{\color[HTML]{000000} $-0.019 (.002, .002, -15.7)$} \\
MCMC & \cc{B7D8E9}{\color[HTML]{000000} $-0.44 (.120, .093, -4.96)$} & \cc{DCECF4}{\color[HTML]{000000} $-0.004 (.002, .001, -4.09)$} & \cc{FDEBE2}{\color[HTML]{000000} $-0.005 (.002, .001, -6.03)$} \\
IDHFR & \cc{EDF5FA}{\color[HTML]{000000} $-0.11 (.123, .097, -1.97)$} & \cc{F6FAFC}{\color[HTML]{000000} $-0.001 (.002, .001, -3.10)$} & \cc{FDEFE7}{\color[HTML]{000000} $-0.004 (.001, .001, -6.19)$} \\
RevDGS & \cc{F6FAFC}{\color[HTML]{000000} $-0.06 (.134, .175, -1.57)$} & \cc{FBDCCC}{\color[HTML]{000000} $0.005 (.003, .004, 2.15)$} & \cc{F9C2A7}{\color[HTML]{000000} $-0.015 (.003, .005, -10.8)$} \\
\midrule
& \multicolumn{3}{c}{\textbf{Monodepth}} \\
\midrule
AbsGS & \cc{EDA38D}{\color[HTML]{000000} $0.58 (.221, .126, 3.52)$} & \cc{F2AF97}{\color[HTML]{000000} $0.010 (.005, .002, 3.42)$} & \cc{F7BCA1}{\color[HTML]{000000} $-0.016 (.010, .004, -4.55)$} \\
INRIA & \cc{FCE6DB}{\color[HTML]{000000} $0.19 (.125, .172, 2.17)$} & \cc{FCE3D6}{\color[HTML]{000000} $0.004 (.001, .002, 5.04)$} & \cc{FCDFD1}{\color[HTML]{000000} $-0.008 (.002, .002, -8.47)$} \\
MCMC & \cc{D1E6F1}{\color[HTML]{000000} $-0.28 (.120, .120, -3.87)$} & \cc{D3E7F2}{\color[HTML]{000000} $-0.005 (.002, .001, -5.66)$} & \cc{FDEFE7}{\color[HTML]{000000} $-0.004 (.002, .001, -4.12)$} \\
IDHFR & \cc{FDEBE2}{\color[HTML]{000000} $0.15 (.123, .109, 1.31)$} & \cc{FEF1EA}{\color[HTML]{000000} $0.002 (.002, .001, 2.03)$} & \cc{FFFBF9}{\color[HTML]{000000} $-0.001 (.001, .002, -1.41)$} \\
RevDGS & \cc{FFFEFD}{\color[HTML]{000000} $0.01 (.134, .132, -1.46)$} & \cc{FFFFFE}{\color[HTML]{000000} $-0.000 (.003, .003, 0.437)$} & \cc{F0F7FA}{\color[HTML]{000000} $0.003 (.003, .005, 1.22)$} \\
\midrule
& \multicolumn{3}{c}{\textbf{DA3}} \\
\midrule
AbsGS & \cc{FEF7F3}{\color[HTML]{000000} $0.06 (.221, .155, -2.67)$} & \cc{F6FAFC}{\color[HTML]{000000} $-0.001 (.005, .003, -5.00)$} & \cc{FBDBCB}{\color[HTML]{000000} $-0.009 (.010, .004, -2.67)$} \\
INRIA & \cc{CCE3EF}{\color[HTML]{000000} $-0.31 (.125, .129, -4.61)$} & \cc{D3E7F2}{\color[HTML]{000000} $-0.005 (.001, .001, -4.35)$} & \cc{FDEFE7}{\color[HTML]{000000} $-0.004 (.002, .001, -6.60)$} \\
MCMC & \cc{6EA6CE}{\color[HTML]{F1F1F1} $-0.73 (.120, .104, -15.6)$} & \cc{5C99C7}{\color[HTML]{F1F1F1} $-0.015 (.002, .001, -22.5)$} & \cc{D2E7F1}{\color[HTML]{000000} $0.009 (.002, .001, 9.85)$} \\
IDHFR & \cc{DFEDF5}{\color[HTML]{000000} $-0.20 (.123, .095, -3.70)$} & \cc{D3E7F2}{\color[HTML]{000000} $-0.005 (.002, .001, -7.23)$} & \cc{F0F7FA}{\color[HTML]{000000} $0.003 (.001, .001, 2.22)$} \\
RevDGS & \cc{CBE3EF}{\color[HTML]{000000} $-0.32 (.134, .232, -5.88)$} & \cc{B1D5E7}{\color[HTML]{000000} $-0.009 (.003, .006, -6.10)$} & \cc{E1EFF6}{\color[HTML]{000000} $0.006 (.003, .008, 1.85)$} \\
\midrule
& \multicolumn{3}{c}{\textbf{$\text{DA3}^\text{G.S.}$}} \\
\midrule
AbsGS & \cc{D5675E}{\color[HTML]{F1F1F1} $0.80 (.221, .097, 5.50)$} & \cc{F8BEA3}{\color[HTML]{000000} $0.009 (.005, .001, -1.13)$} & \cc{DC776B}{\color[HTML]{F1F1F1} $-0.024 (.010, .002, -5.68)$} \\
INRIA & \cc{FDEAE0}{\color[HTML]{000000} $0.16 (.125, .115, 1.32)$} & \cc{F6FAFC}{\color[HTML]{000000} $-0.001 (.001, .001, -1.44)$} & \cc{FACEB8}{\color[HTML]{000000} $-0.012 (.002, .002, -10.1)$} \\
MCMC & \cc{F2F8FB}{\color[HTML]{000000} $-0.08 (.120, .136, -1.47)$} & \cc{E5F1F7}{\color[HTML]{000000} $-0.003 (.002, .002, -4.18)$} & \cc{FFFFFE}{\color[HTML]{000000} $-0.000 (.002, .001, .753)$} \\
IDHFR & \cc{FEF3ED}{\color[HTML]{000000} $0.09 (.123, .084, 0.729)$} & \cc{EDF5FA}{\color[HTML]{000000} $-0.002 (.002, .001, -8.71)$} & \cc{FEF3ED}{\color[HTML]{000000} $-0.003 (.001, .001, -1.14)$} \\
RevDGS & \cc{FCE2D5}{\color[HTML]{000000} $0.22 (.134, .126, .631)$} & \cc{FEF8F4}{\color[HTML]{000000} $0.001 (.003, .002, -1.69)$} & \cc{FAD2BE}{\color[HTML]{000000} $-0.011 (.003, .002, -4.21)$} \\
\bottomrule
\end{tabular}}
\endgroup
}
\end{table}

In this section, we provide details of the statistical methodology used to mark statistically significant improvements in \cref{tab:laser_scan_main,tab:practical_main,tab:final_recon_fscore} in the main text, and some of the tables in the supplementary material.
Furthermore, in
\Cref{tab:scannet++_dense_improvement,tab:eval_on_train_set_scannet++_dense_improvement,tab:eth3d_dense_improvement,tab:mipnerf360_dense_improvement,tab:tanksandtemples_dense_improvement},
we provide per-scene results for mean NVS improvements over SfM with each initialization method, including pooled (across scenes) standard deviations across random seeds (separately for results with SfM and with the given initialization), as well as Hedges' $g$ effect size values for improvements over SfM.

\PAR{Statistical Evaluation Protocol.} We perform non-parametric statistical tests following Demšar's recommendations~\cite{demsar_stat_comparisons_2006}. We divide our data into families for family-wise error rate (FWER) control by grouping results by densification strategy and metric. For example, all PSNR values using MCMC densification are processed by the pipeline described below independently from results using other metric or other densification strategies. Within each family, the statistical unit is the scene, and the initializations are the methods being compared. Consequently, our significance conclusions are specific to a given metric and densification strategy; we do not use these tests to make global statistical claims across different densification strategies or metrics.

We perform the following steps within each densification/metric family (as defined above):
\begin{enumerate}
    \item We first take the mean of per-seed metric values for each data point (scene, initialization). This produces one observation per scene--initialization pair, so scenes, not individual random seeds, are the experimental units. To characterize seed-level variability we provide pooled seed-level standard deviations and dataset-mean effect sizes (\cref{tab:scannet++_dense_improvement,tab:eval_on_train_set_scannet++_dense_improvement,tab:eth3d_dense_improvement,tab:mipnerf360_dense_improvement,tab:tanksandtemples_dense_improvement}), as well as plots of per-seed data (\cref{fig:practical_nvs_all_data_absgs,fig:practical_nvs_all_data_inria,fig:practical_nvs_all_data_idhfr,fig:practical_nvs_all_data_mcmc,fig:practical_nvs_all_data_revdgs,fig:practical_nvs_all_data_no_d}).
    \item We then perform the Friedman test with the Iman-Davenport correction. The null hypothesis is that the initialization methods have equal mean ranks across scenes. If the omnibus null hypothesis is rejected at $p < 0.05$, we proceed to the post-hoc comparisons, otherwise none of the cells are marked as having significant improvements.
    \item Finally, we perform post-hoc tests to determine which initializations produce significant improvements ($p<0.05$) over SfM, using the following test statistic~\cite{demsar_stat_comparisons_2006}:
    \begin{equation}
        z = (R_i - R_{\text{SfM}}) /  \sqrt{\frac{k(k+1)}{6N}}\, ,
    \end{equation}
    where $R_i$ and $R_\text{SfM}$ are the average ranks of the given initialization method and the SfM control across scenes, $N$ is the number of scenes, and $k$ is the number of methods (initializations). We use a one-sided test because our hypothesis concerns improvements over SfM, with lower average rank corresponding to better performance.\footnote{For metrics where greater values indicate better performance, data is multiplied by $-1$ prior to ranking.}
    We then use Holm's step-down procedure to control FWER at $\alpha=.05$.
\end{enumerate}

\noindent{\bf{Limitations:}}
\begin{itemize}
    \item Because our datasets contain relatively few scenes (7 to 15), the power of the test is somewhat limited. This is especially true on ETH3D, which only has 7 scenes, thus we see fewer statistically significant improvements there. Because of this, non-statistically-significant results should not be interpreted as an indication of equivalent performance among the methods.
    \item As mentioned above, our choice of families for FWER control limits the scope of statistical guarantees to improvements over SfM using a given metric and densification strategy. Any claims in the paper spanning across densification methods and/or metrics should thus be treated as observational and not based on this statistical analysis.
\end{itemize}

\section{Per-Scene Data Analysis}\label{per_scene_data_analysis}
\begin{figure}[t]
    \centering
    \includegraphics[width=0.48\linewidth]{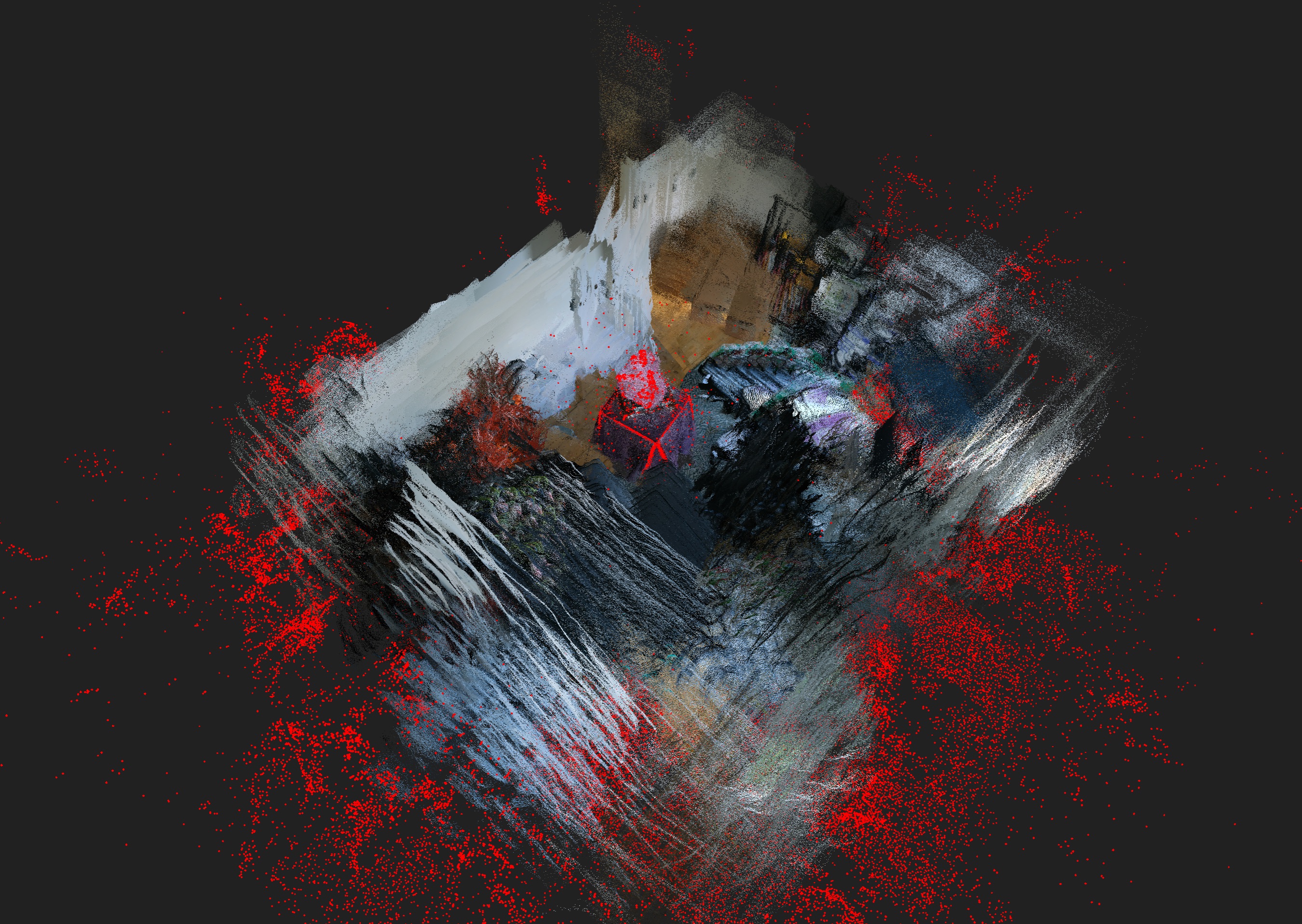}\hfill
    \includegraphics[width=0.48\linewidth]{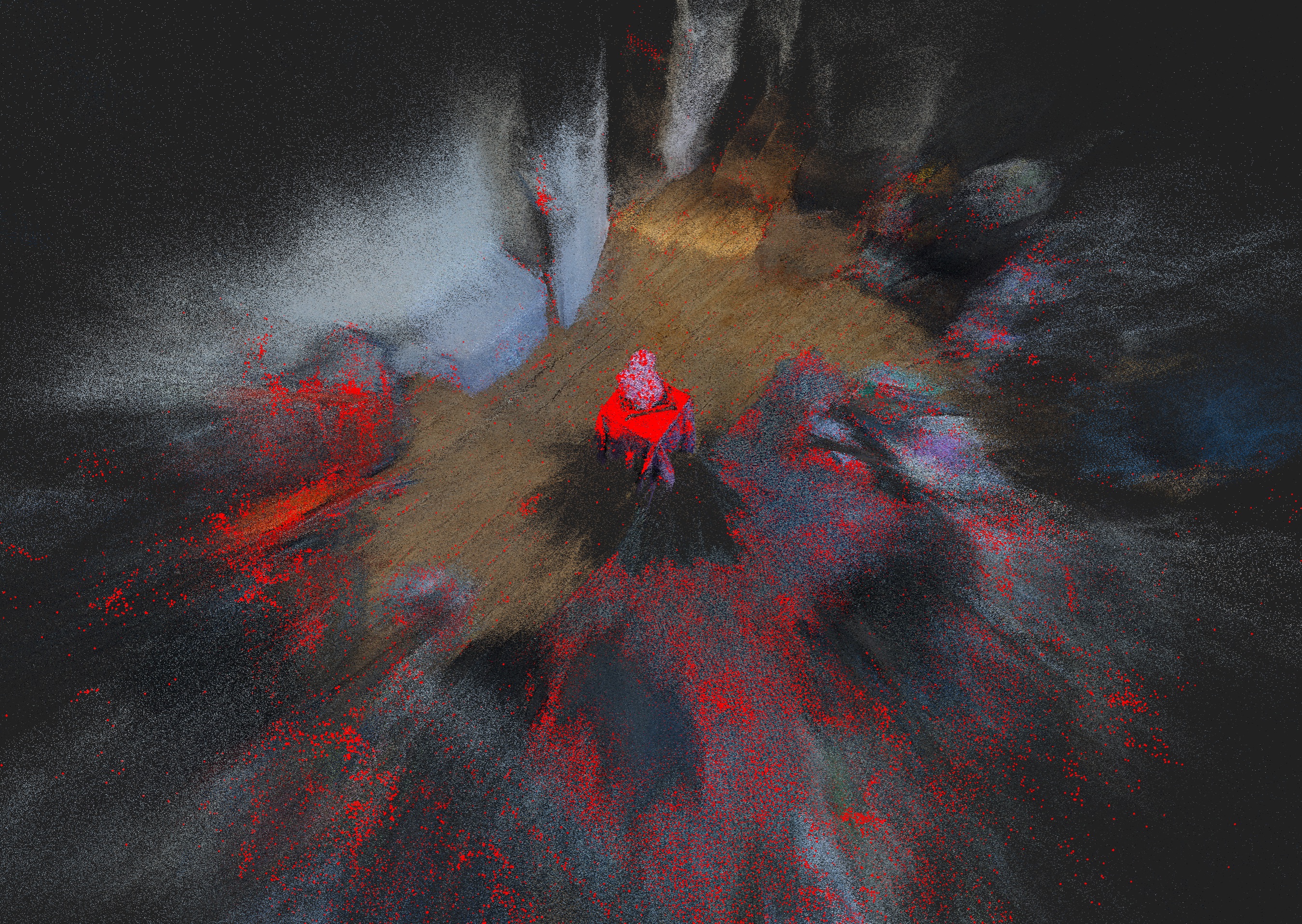}
    \caption{Comparison of the point cloud produced by DA3 (left) and the centers of Gaussians predicted by $\text{DA3}^\text{GS}$ on the ``Bonsai'' scene from MipNerf360. SfM points are visualized in \textbf{red} for reference.}
    \label{fig:da3_da3_gs_bonsai}
\end{figure}

As statistical analysis and aggregation may hide some useful patterns in the data, we also plot all our data points (for each scene, random seed, initialization and densification) for our laser scan 
(\cref{fig:laser_scan_nvs_all_data_absgs,fig:laser_scan_nvs_all_data_inria,fig:laser_scan_nvs_all_data_idhfr,fig:laser_scan_nvs_all_data_mcmc,fig:laser_scan_nvs_all_data_revdgs,fig:laser_scan_nvs_all_data_no_d})
and practical initialization
(\cref{fig:practical_nvs_all_data_absgs,fig:practical_nvs_all_data_inria,fig:practical_nvs_all_data_idhfr,fig:practical_nvs_all_data_mcmc,fig:practical_nvs_all_data_revdgs,fig:practical_nvs_all_data_no_d}) experiments.
Visual analysis of these plots leads us to the following observations:
\begin{itemize}
    \item In the vast majority of cases, it is clear that the change in NVS metrics correlates with the choice of initialization strategy, and can't be attributed to random variance across seeds alone.
    
    \item The impact of using dense initialization (whether with laser scan data, or one of the other methods) is not uniform across all scenes.
    For example, on ScanNet++, there are several scenes (\eg ``bcd2436daf'') where the improvement is much more pronounced than on others.
    These are usually scenes with few textured surfaces, for example largely consisting of white walls.
    This is expected, as such scenes are a worst-case scenario for 3DGS with SfM -- the lack of texture detail contributes both to low SfM point density, and, in many cases, makes photometric constraints insufficient to establish correct geometry during optimization.
    Dense initialization alleviates this issue by providing correctly positioned splats from the start, resolving the geometric ambiguity (though as we've seen in \Cref{sec:protocol:geometric,excluded_eth3d_results}, accuracy post-optimization is still usually reduced).
    
    \item Finally, there are some outliers among the data, where one of the initialization methods performs significantly worse on a given scene. The most notable one is DA3 on the ``Bonsai'' MipNerf360 scene, where we see a stark difference to the performance of other methods, including $\text{DA3}^\text{GS}$. As seen in \cref{fig:da3_da3_gs_bonsai}, the DA3-predicted point cloud on this scene consists of multiple misaligned predictions, none of which are consistent with the SfM points. On the other hand, the $\text{DA3}^\text{GS}$ output, while noisy, aligns with the SfM point cloud much better. This is consistent with the effects discussed in \cref{init_precision_no_gt}.  
\end{itemize}

\newcommand{\laserScanAllDataPlotFigure}[2]{
    \begin{figure}[htbp]
        \centering
        \includeinkscape[width=1.0\textwidth]{laser_scan_analysis_#2_svg-tex.pdf_tex} 
        \caption{Per-scene plot of NVS metrics with each random seed for the laser scan initialization experiment using #1 densification. Hybrid initialization is used on ScanNet++ (except for experiments using same init size as SfM), but not ETH3D. The dots represent raw metric values for each of the 3 random seeds. Faint gray lines are the individual data points for SfM initialization, and we also plot mean values (across seeds) using SfM initialization (black line) for clarity. Dense initialization runs for a given seed are connected by faint colored lines (color varies per seed).}
        \label{fig:laser_scan_nvs_all_data_#2}
    \end{figure}
}

\laserScanAllDataPlotFigure{AbsGS}{absgs}
\laserScanAllDataPlotFigure{INRIA}{inria}
\laserScanAllDataPlotFigure{IDHFR}{idhfr}
\laserScanAllDataPlotFigure{MCMC}{mcmc}
\laserScanAllDataPlotFigure{RevDGS}{revdgs}
\laserScanAllDataPlotFigure{no}{no_d}

\newcommand{\practicalAllDataPlotFigure}[2]{
    \begin{figure}[htbp]
        \centering
        \includeinkscape[width=0.85\textwidth]{practical_analysis_#2_svg-tex.pdf_tex} 
        \caption{Per-scene plot of NVS metrics with each random seed for the practical initialization experiment using #1 densification. Hybrid initialization is used for laser scan on ScanNet++ (except for experiments using same init size as SfM), but not ETH3D. The dots represent raw metric values for each of the 3 random seeds. Faint gray lines are the individual data points for SfM initialization, and we also plot mean values (across seeds) using SfM initialization (black line) for clarity. Dense initialization runs for a given seed are connected by faint colored lines (color varies per seed).}
        \label{fig:practical_nvs_all_data_#2}
    \end{figure}
}

\practicalAllDataPlotFigure{AbsGS}{absgs}
\practicalAllDataPlotFigure{INRIA}{inria}
\practicalAllDataPlotFigure{IDHFR}{idhfr}
\practicalAllDataPlotFigure{MCMC}{mcmc}
\practicalAllDataPlotFigure{RevDGS}{revdgs}
\practicalAllDataPlotFigure{no}{no_d}

\section{Additional Qualitative Results}

\begin{figure}[tbp]
  \centering

  \includegraphics[width=0.2485\linewidth]{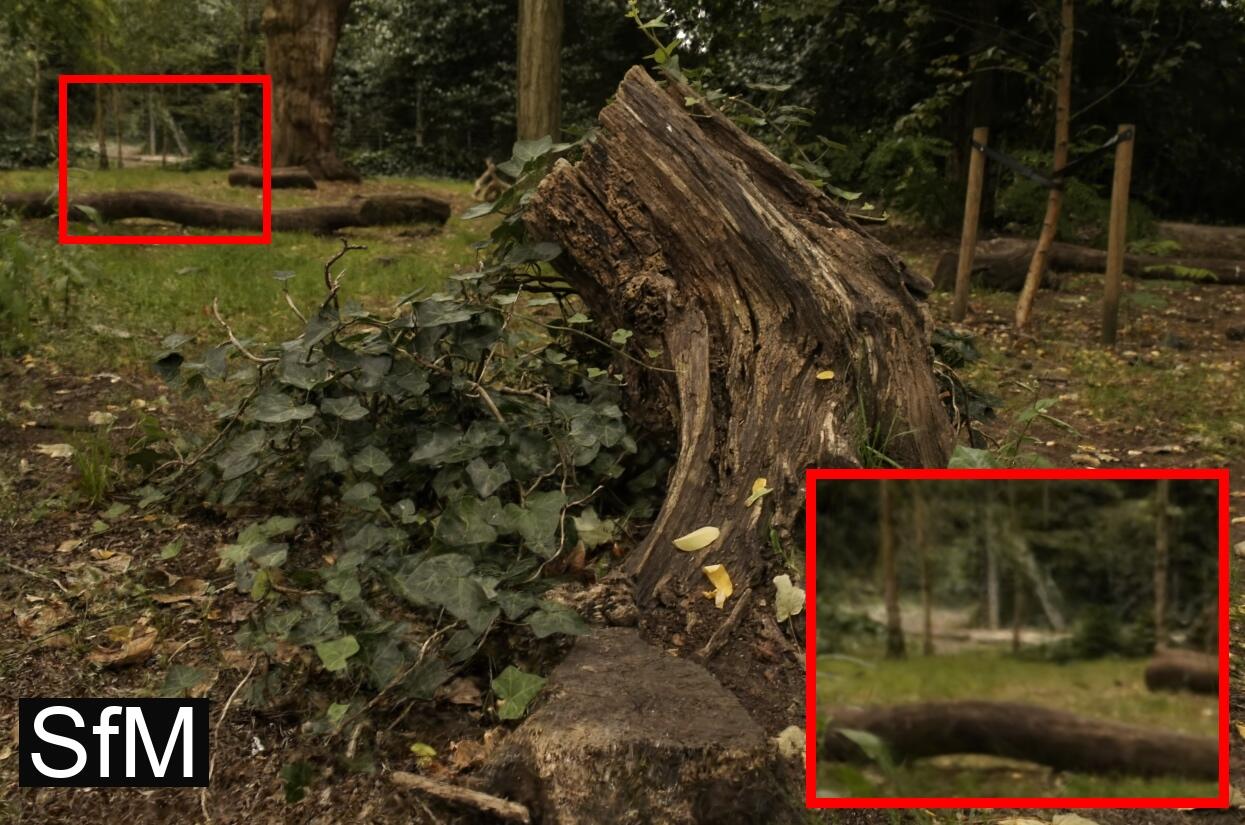}\hfill
  \includegraphics[width=0.2485\linewidth]{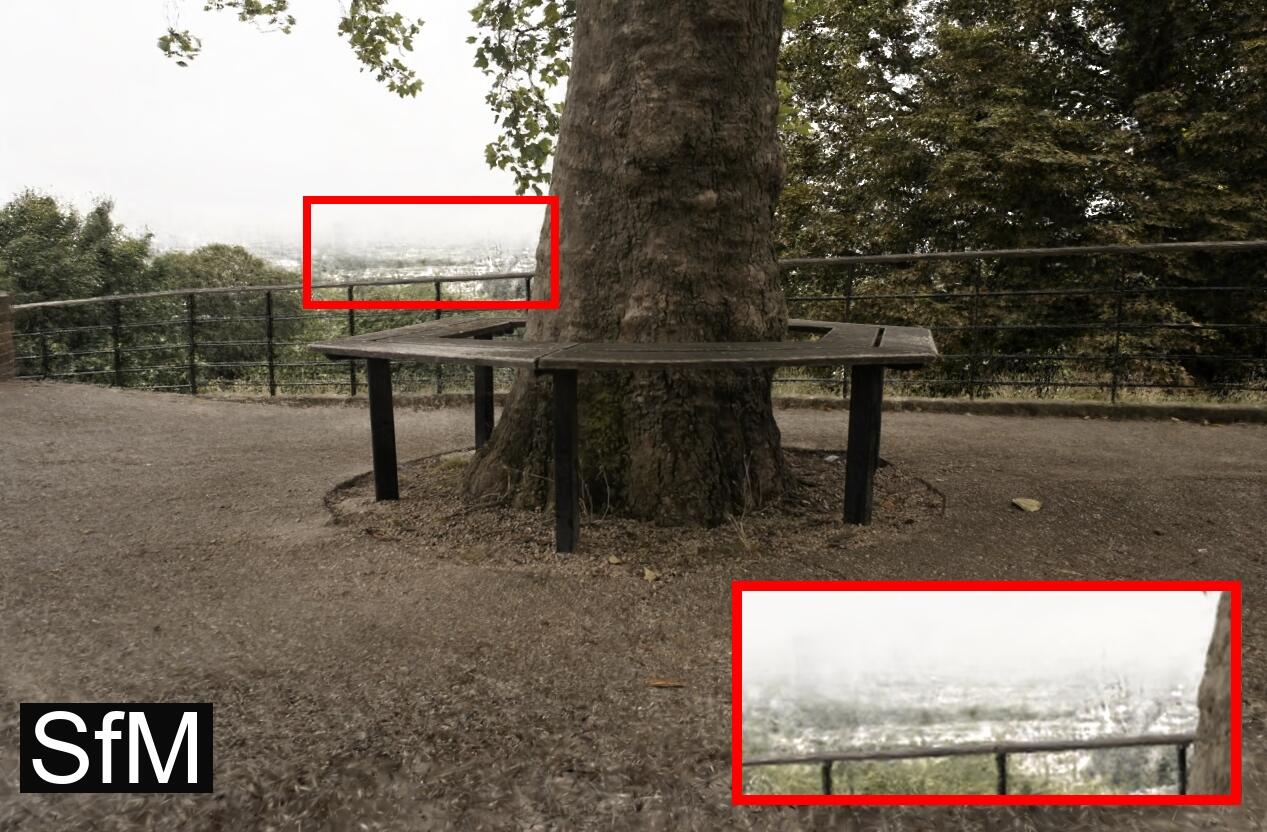}\hfill
  \includegraphics[width=0.2485\linewidth]{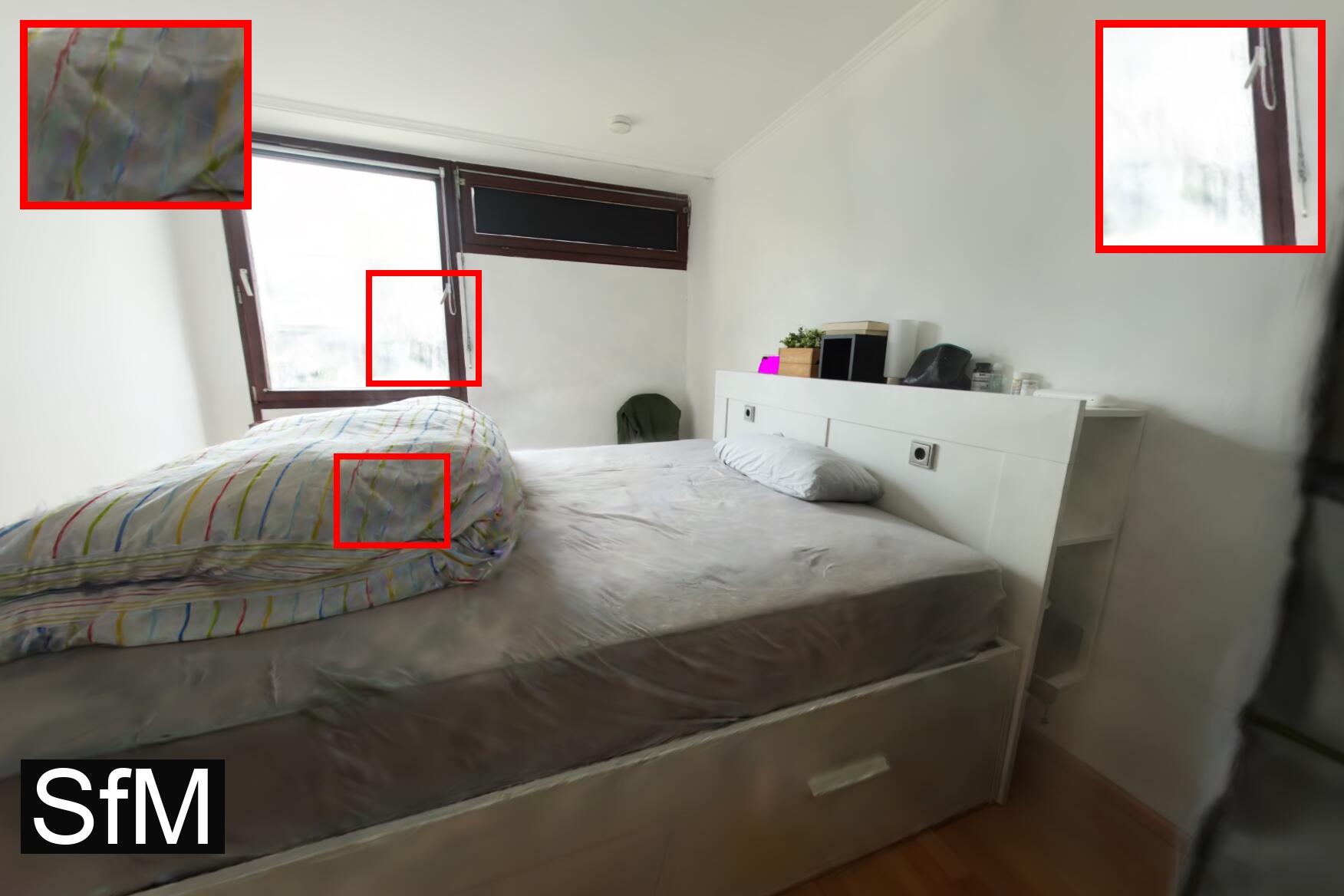}\hfill
  \includegraphics[width=0.2485\linewidth]{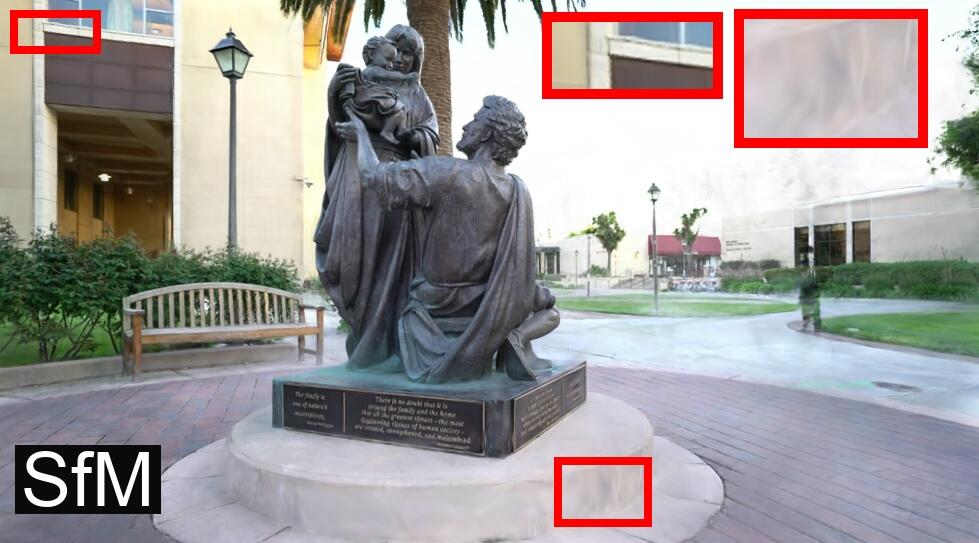}\hfill

  \includegraphics[width=0.2485\linewidth]{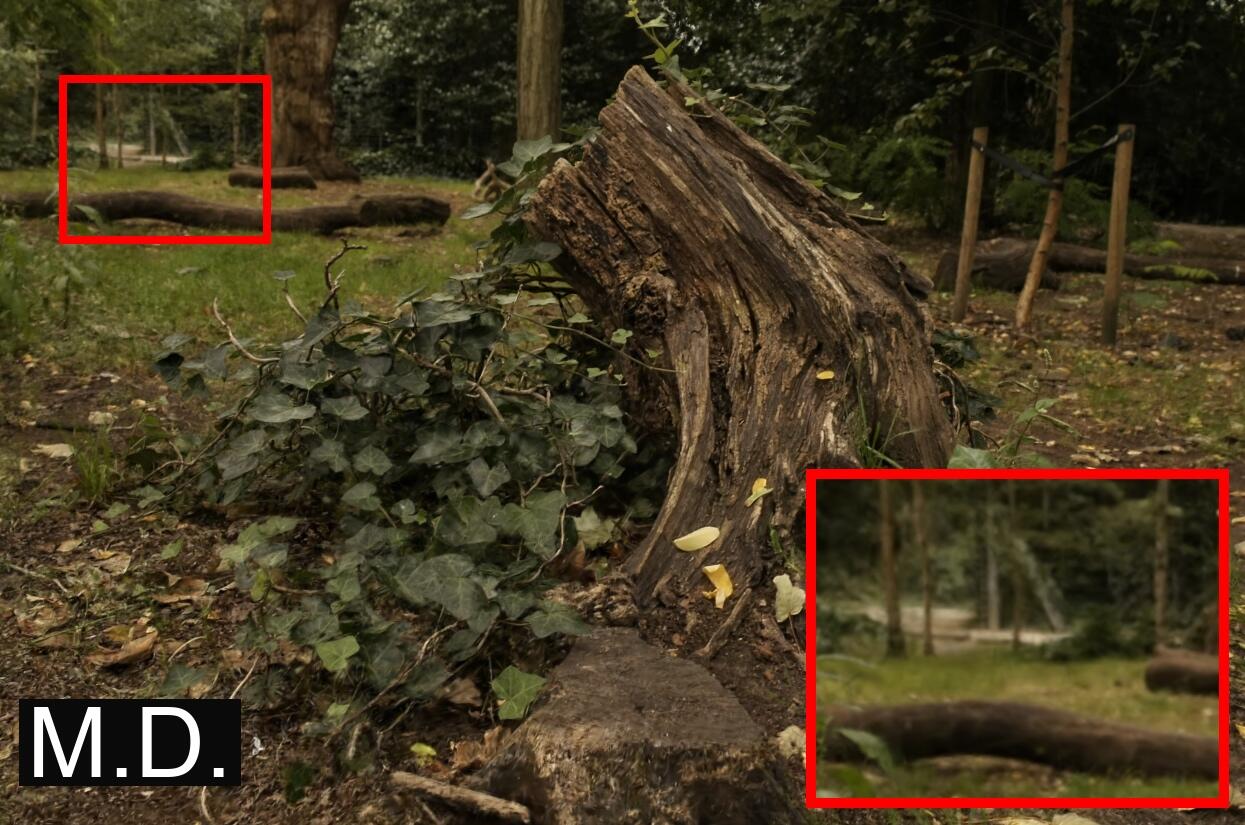}\hfill
  \includegraphics[width=0.2485\linewidth]{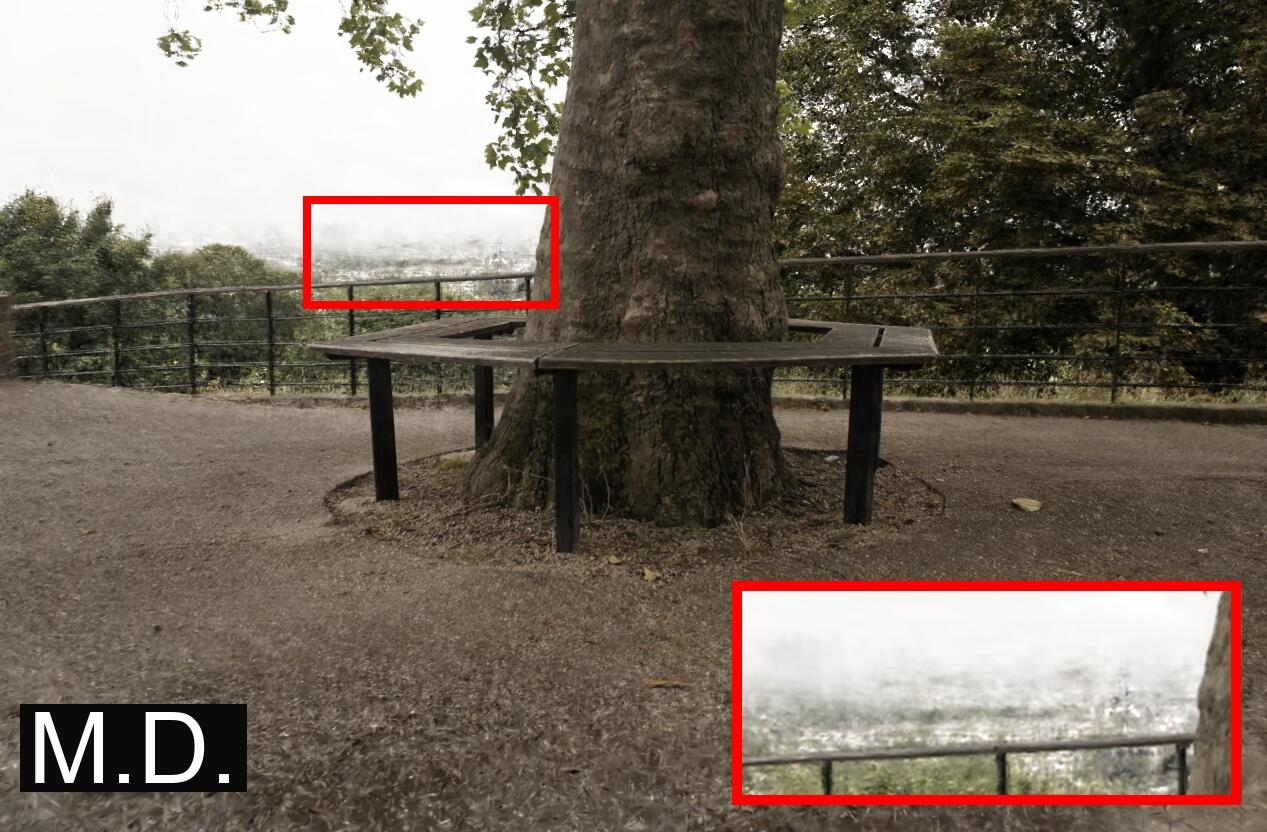}\hfill
  \includegraphics[width=0.2485\linewidth]{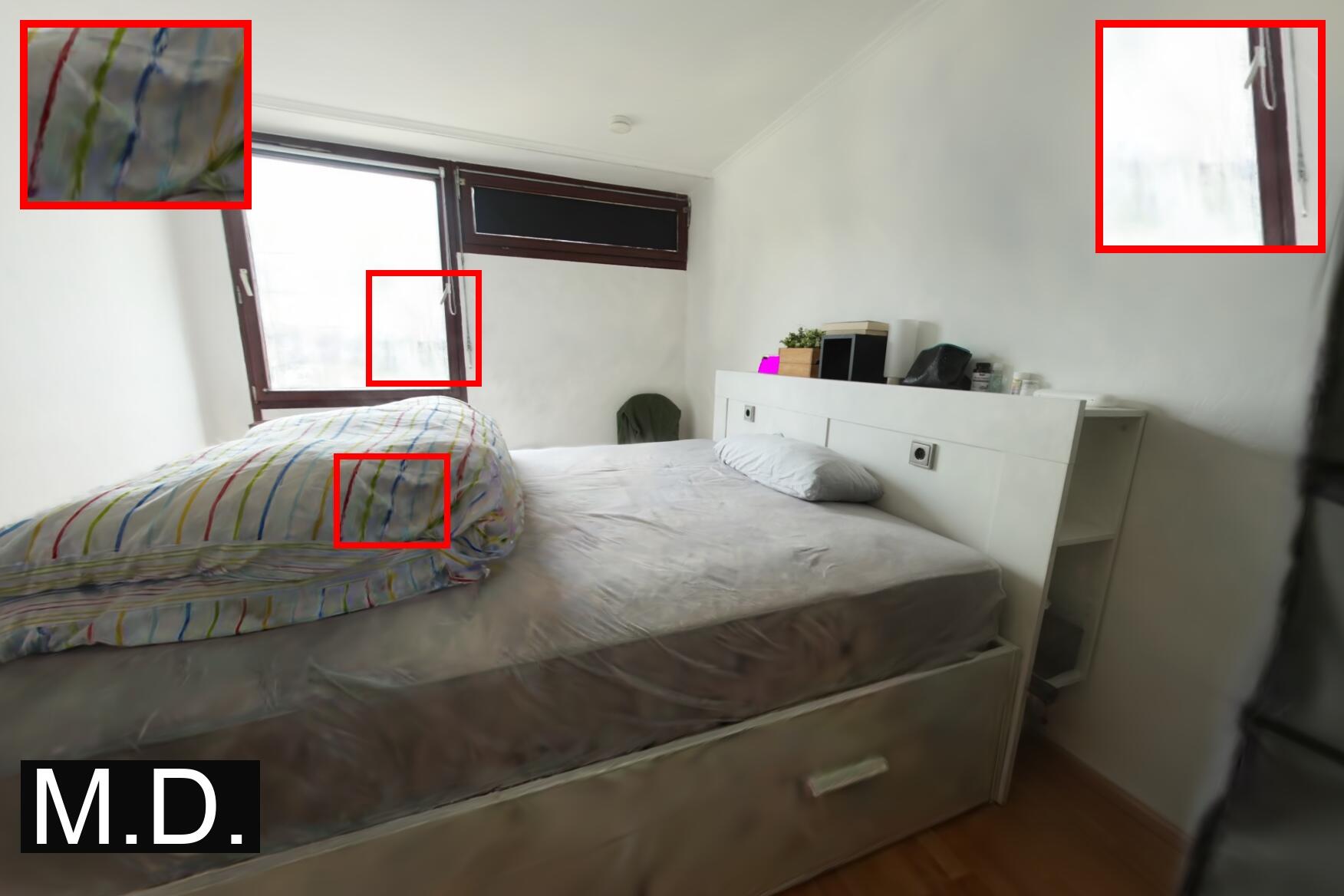}\hfill
  \includegraphics[width=0.2485\linewidth]{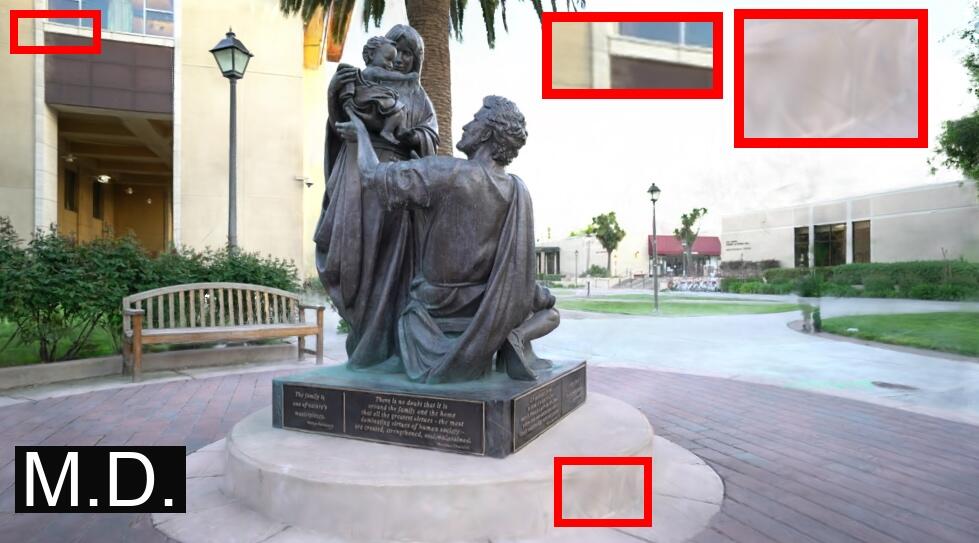}\hfill

  \includegraphics[width=0.2485\linewidth]{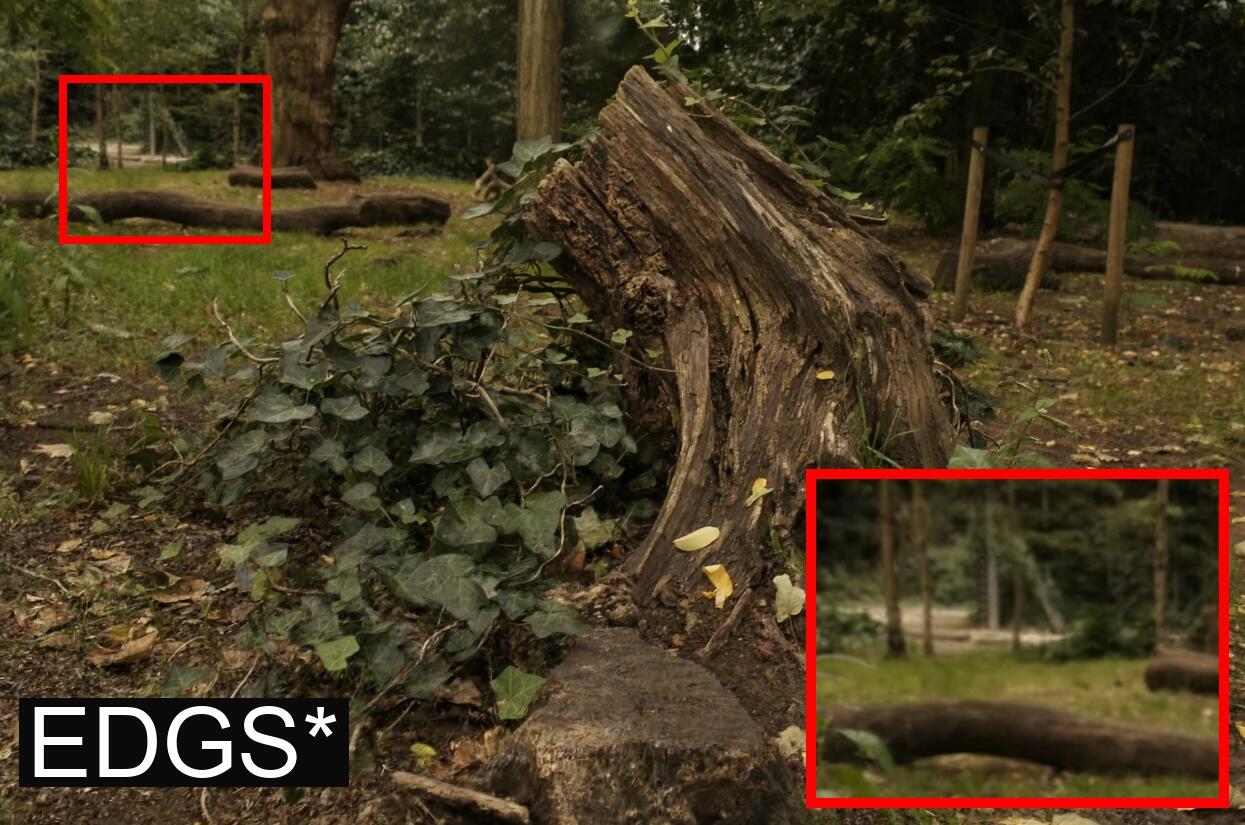}\hfill
  \includegraphics[width=0.2485\linewidth]{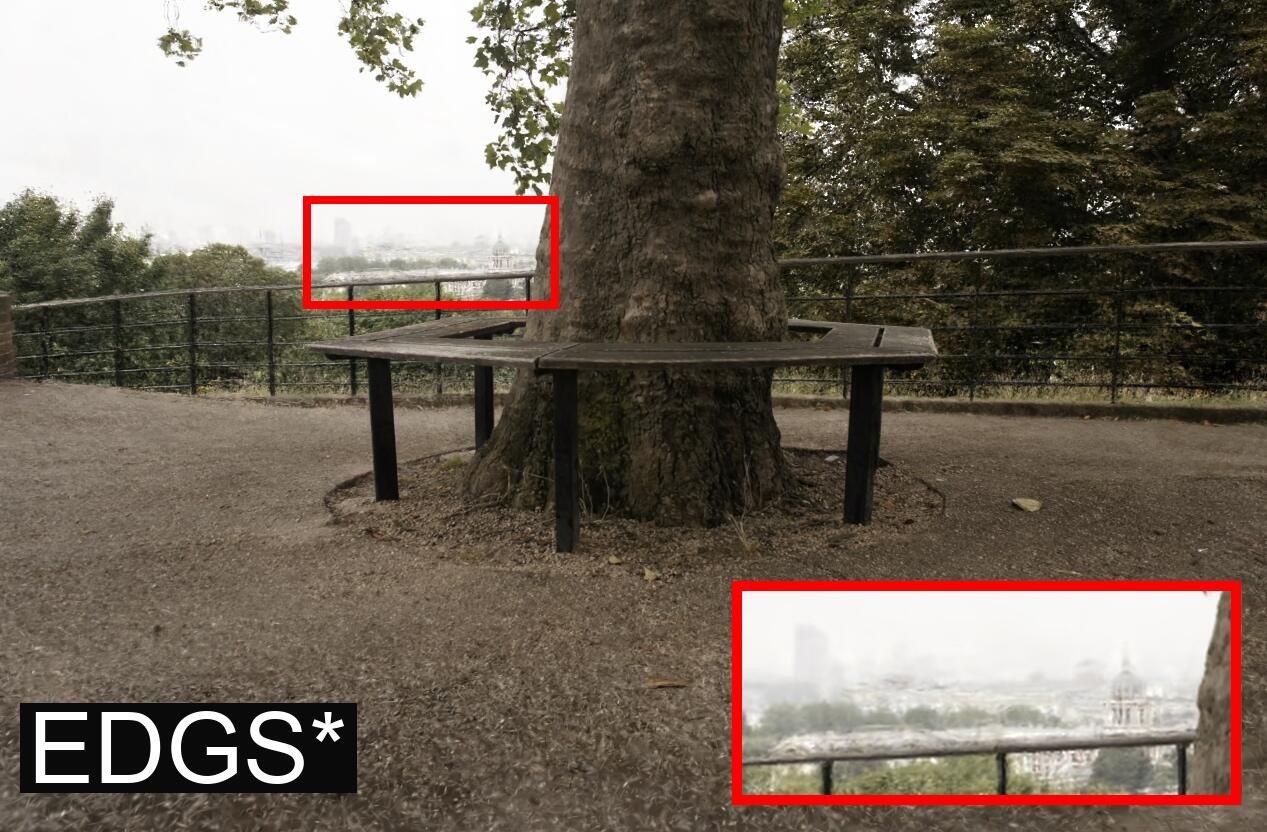}\hfill
  \includegraphics[width=0.2485\linewidth]{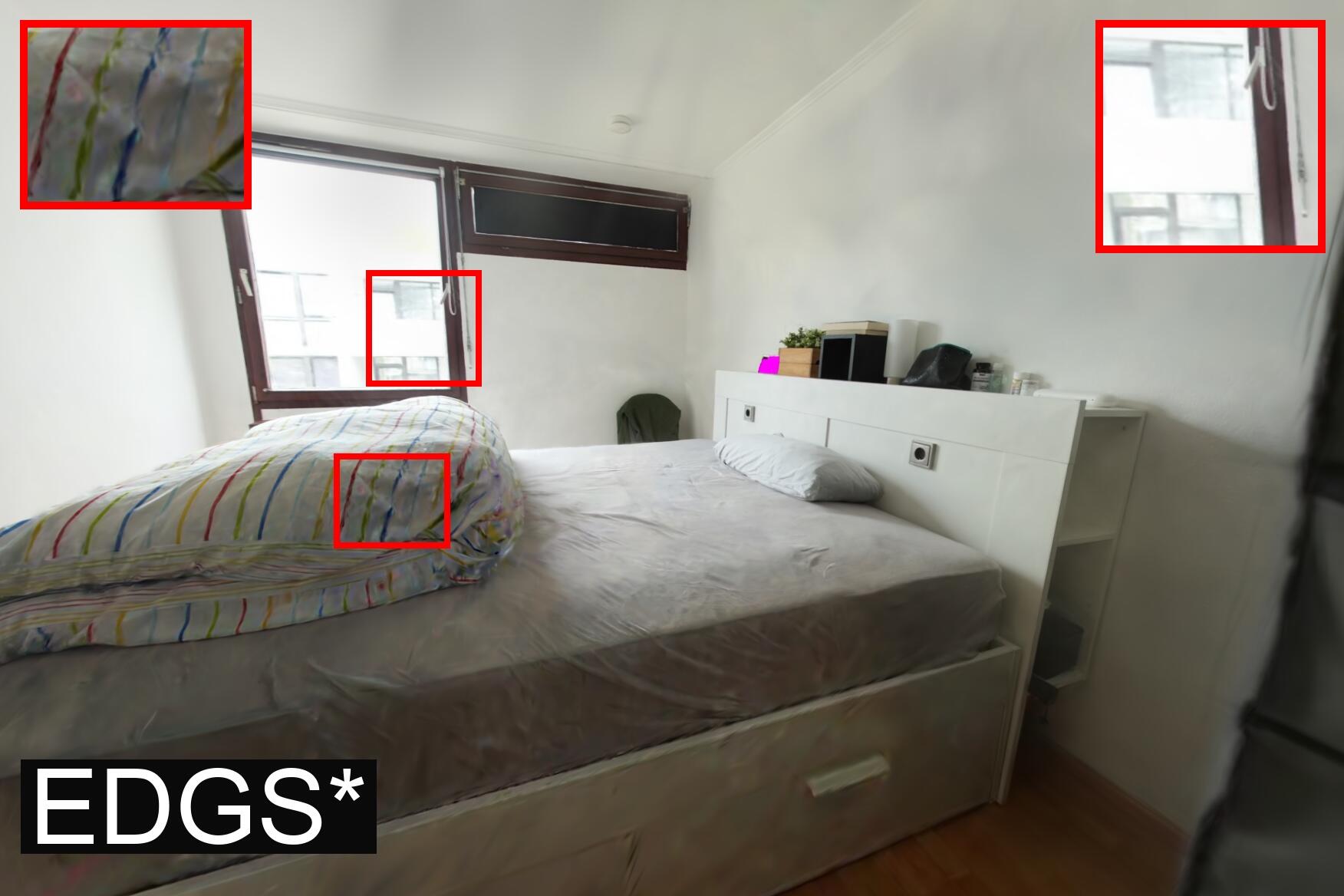}\hfill
  \includegraphics[width=0.2485\linewidth]{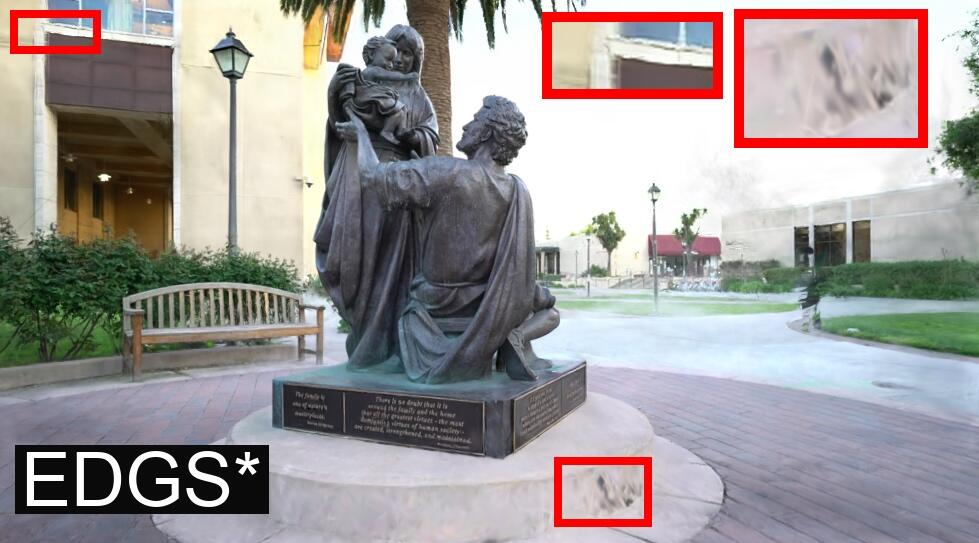}\hfill

  \includegraphics[width=0.2485\linewidth]{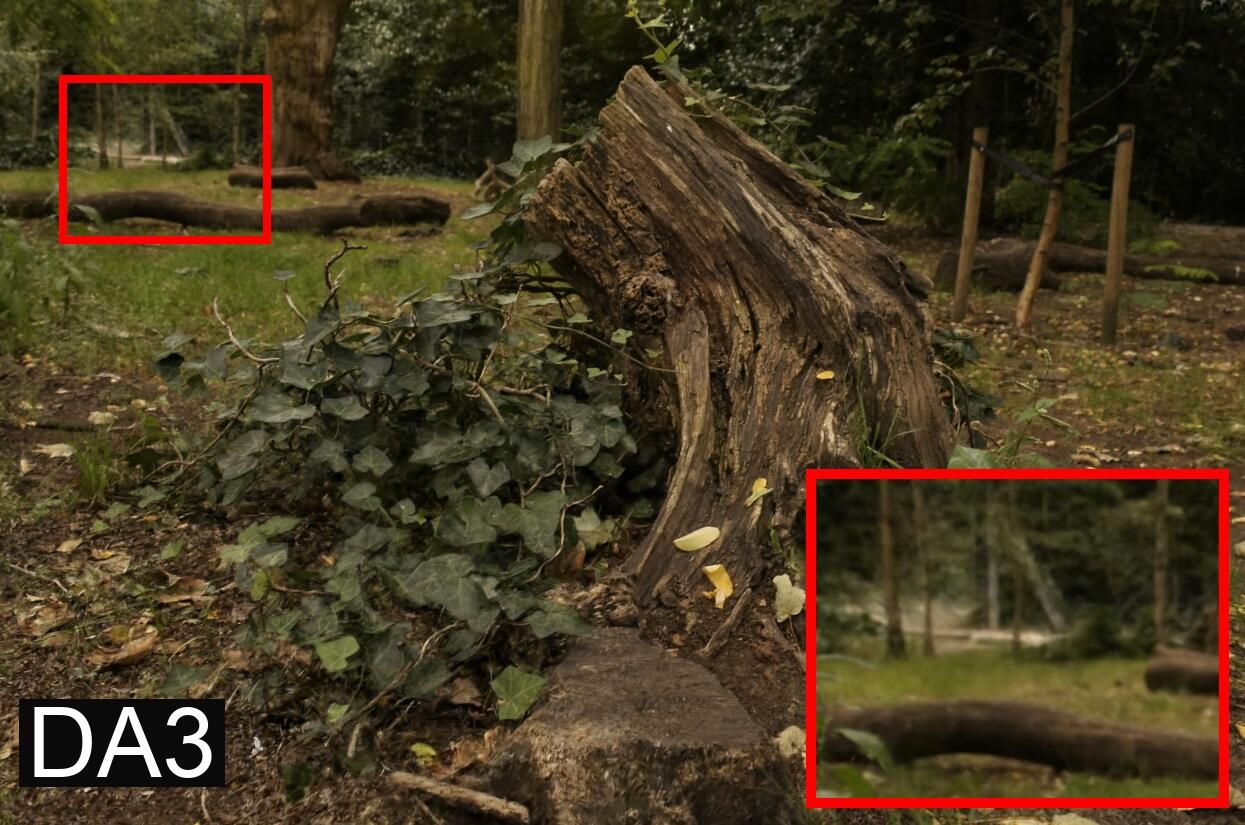}\hfill
  \includegraphics[width=0.2485\linewidth]{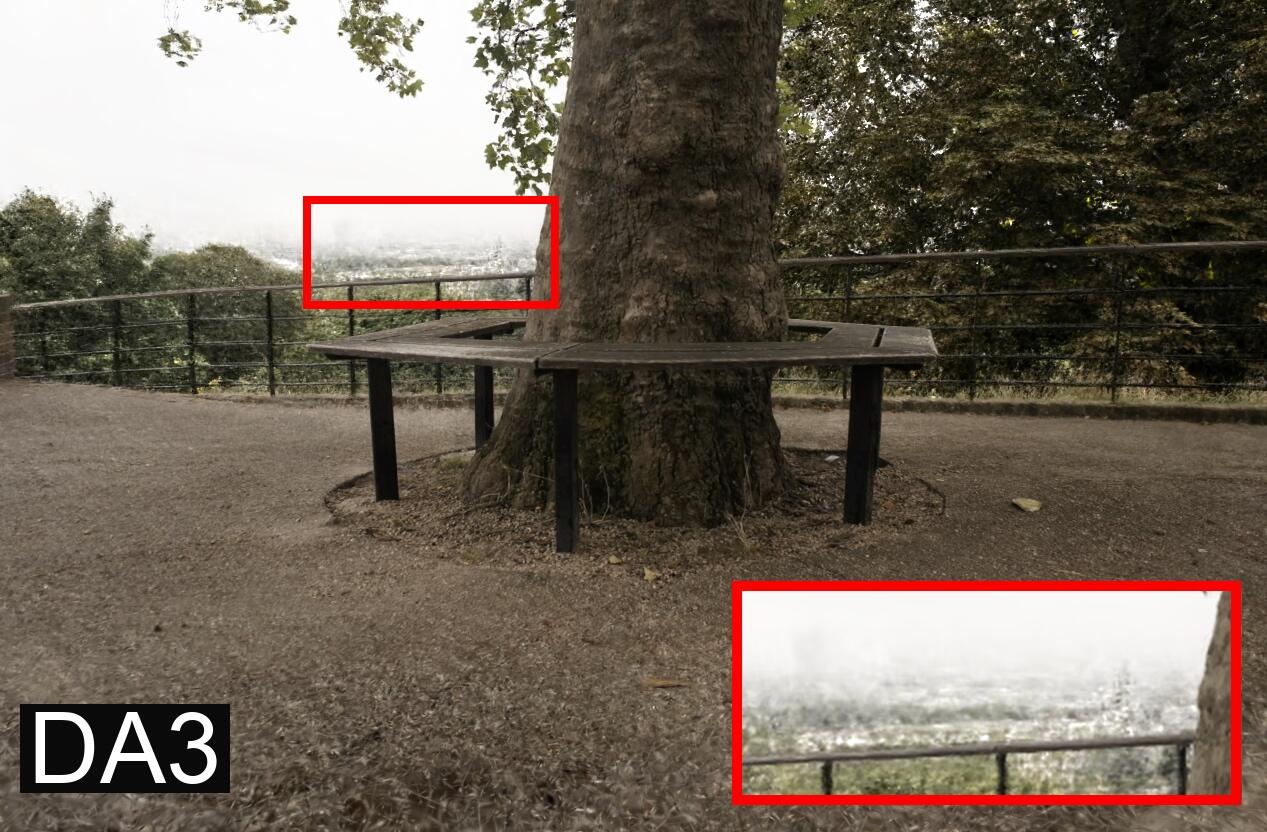}\hfill
  \includegraphics[width=0.2485\linewidth]{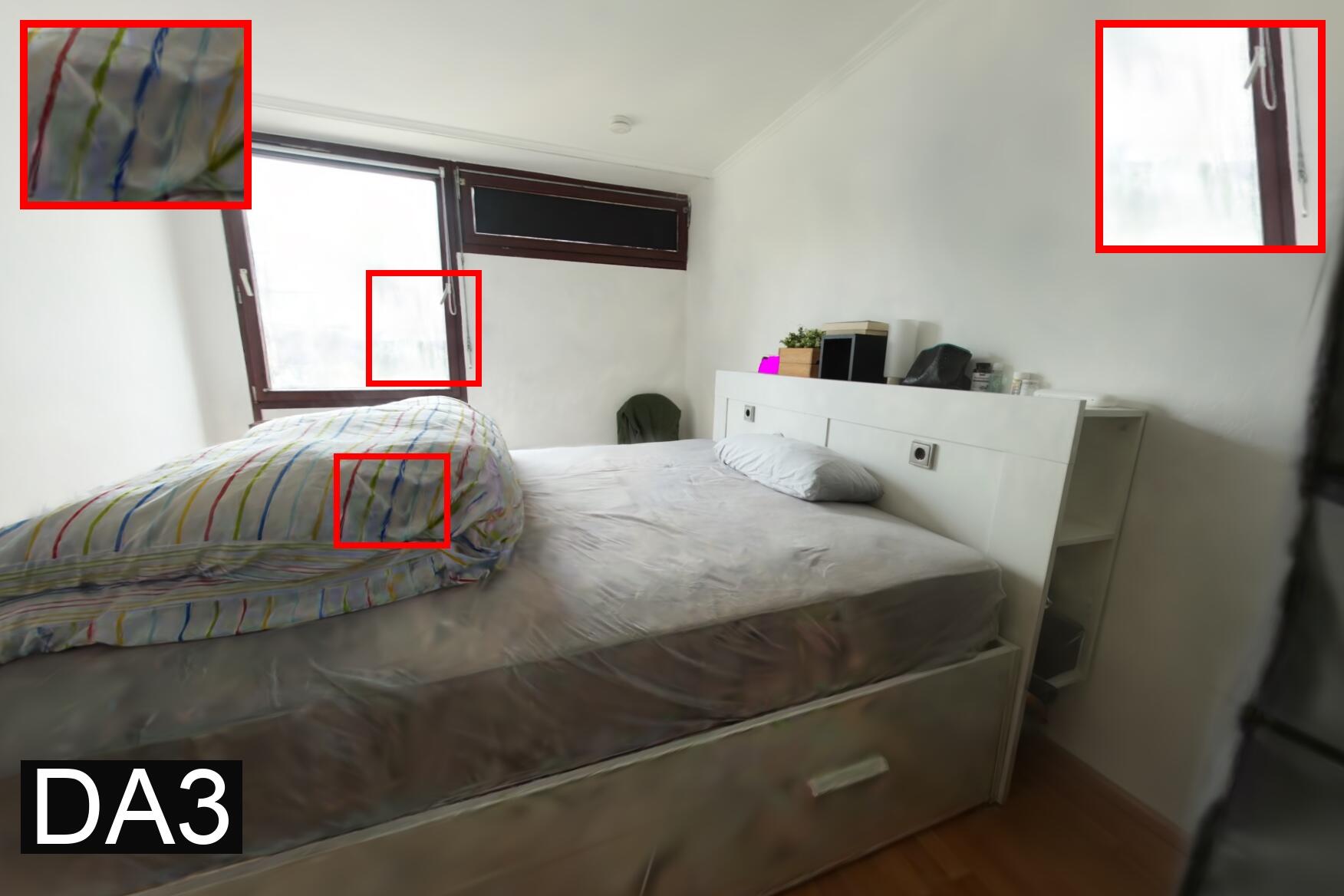}\hfill
  \includegraphics[width=0.2485\linewidth]{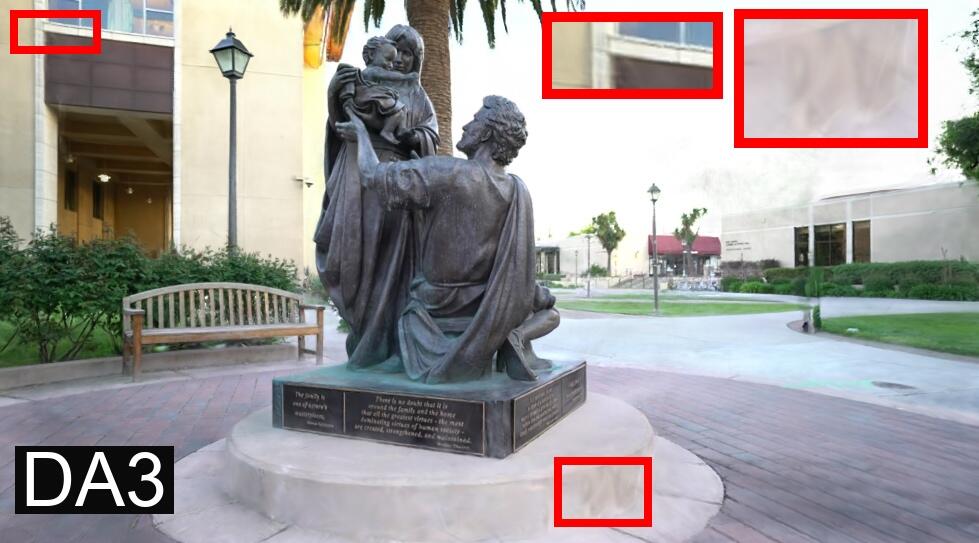}\hfill

  \includegraphics[width=0.2485\linewidth]{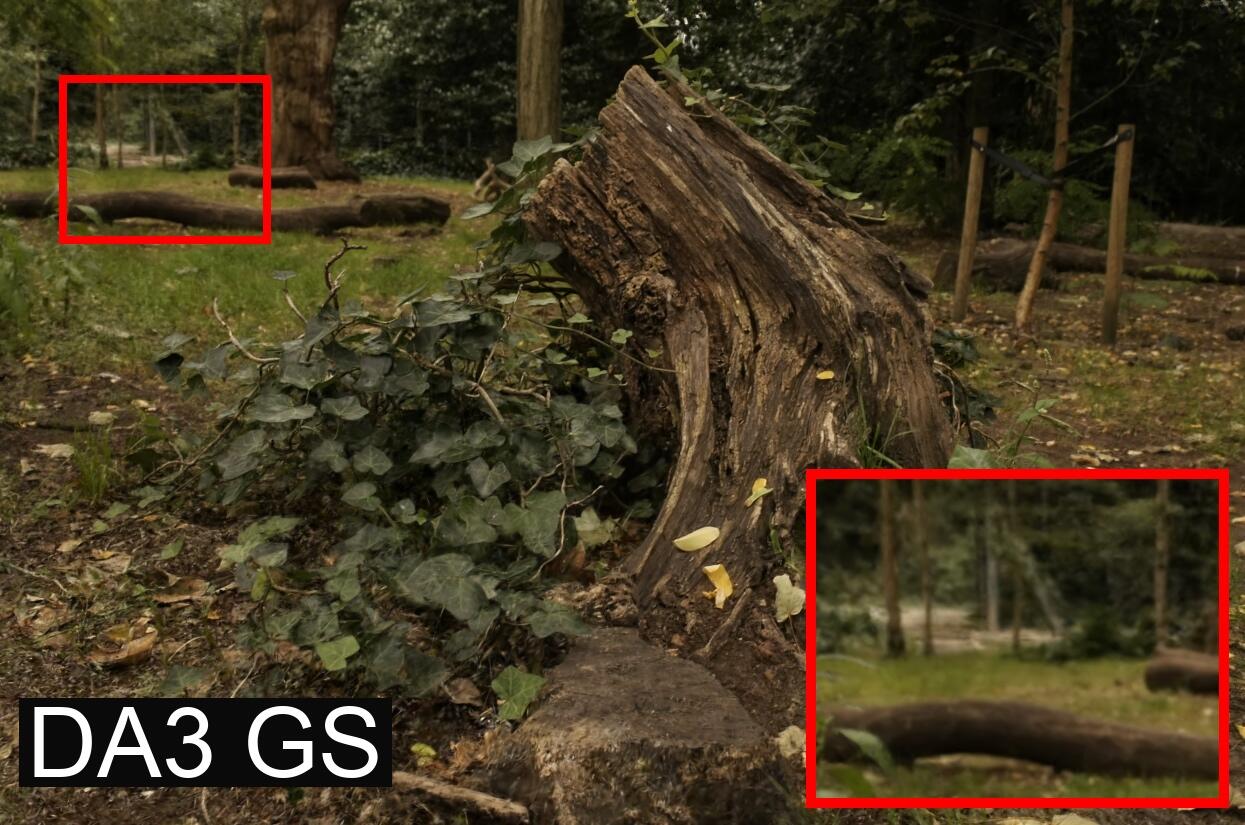}\hfill
  \includegraphics[width=0.2485\linewidth]{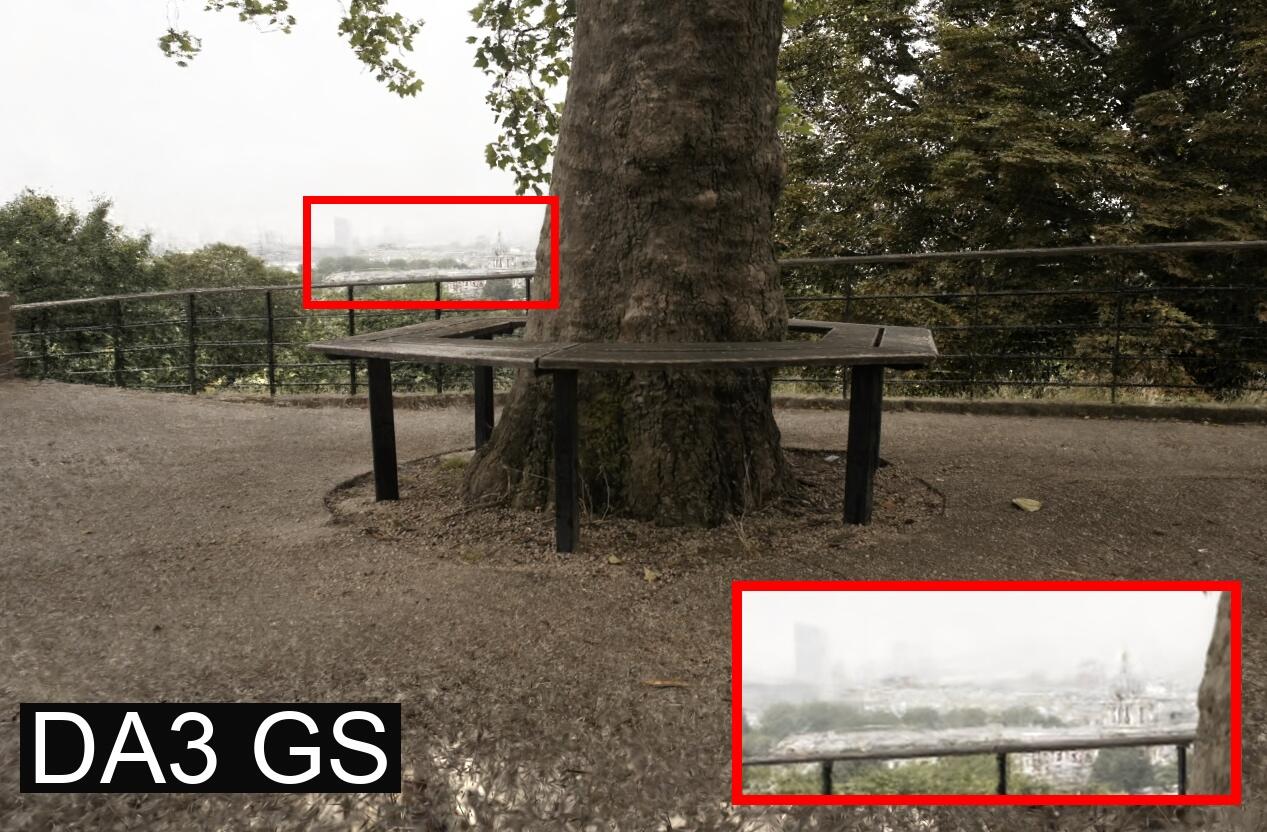}\hfill
  \includegraphics[width=0.2485\linewidth]{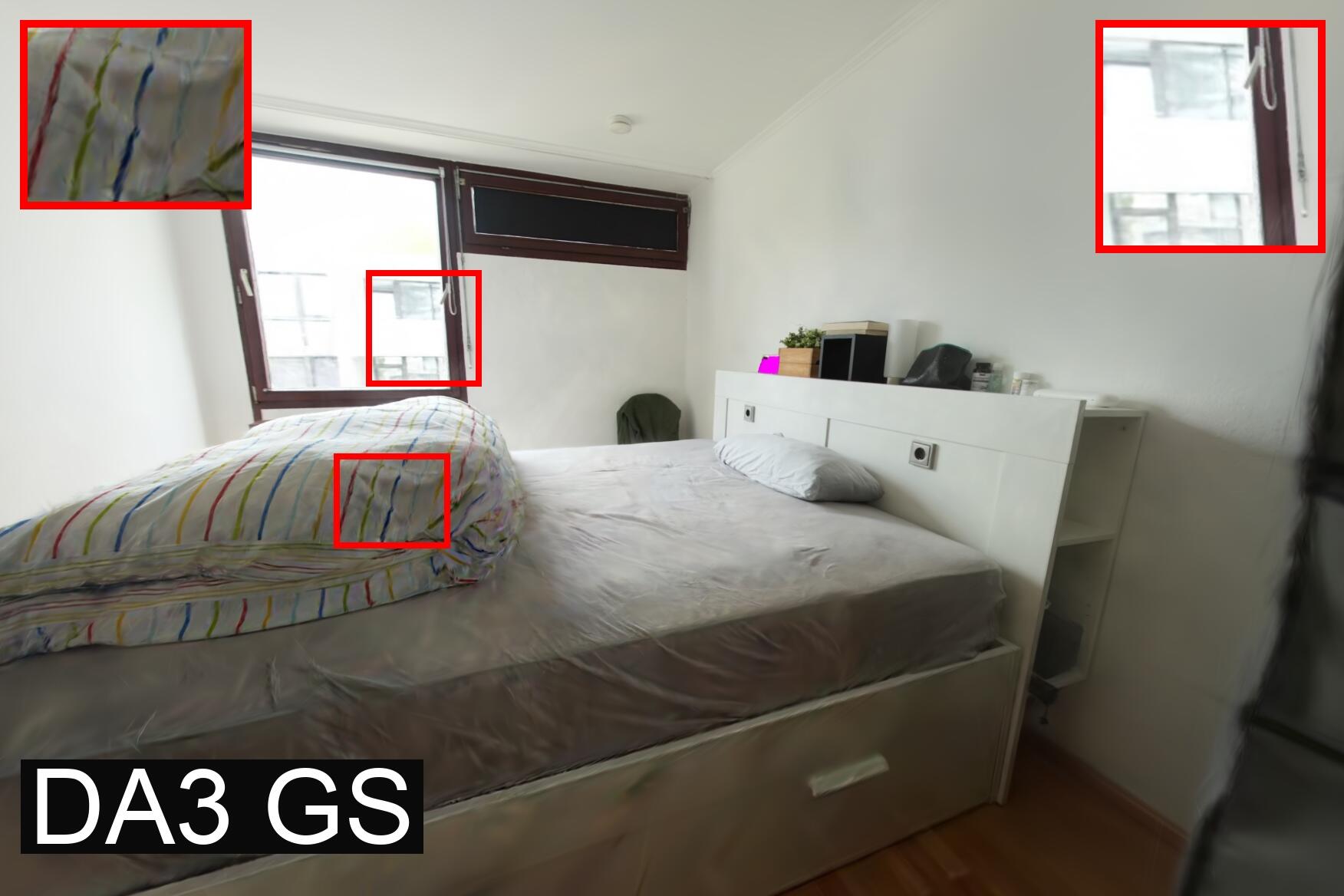}\hfill
  \includegraphics[width=0.2485\linewidth]{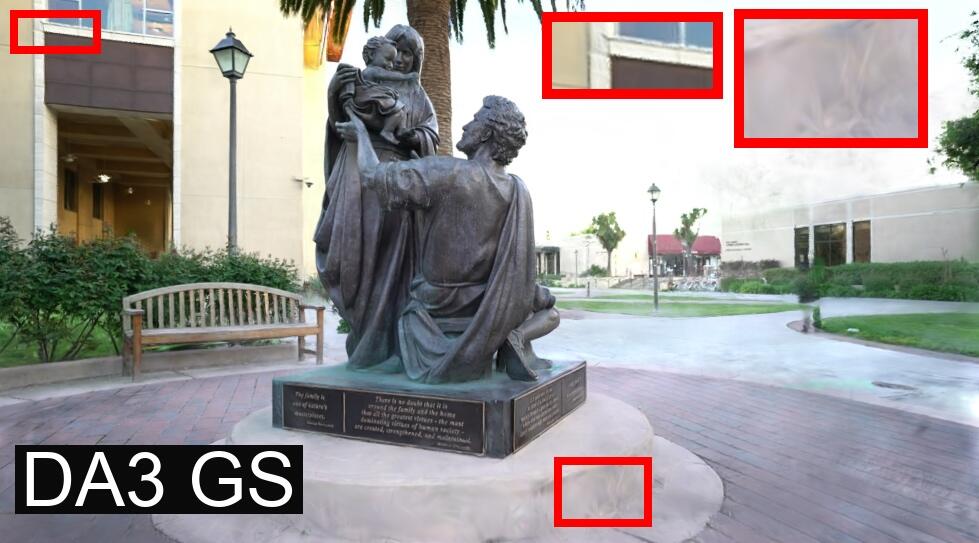}\hfill

  \includegraphics[width=0.2485\linewidth]{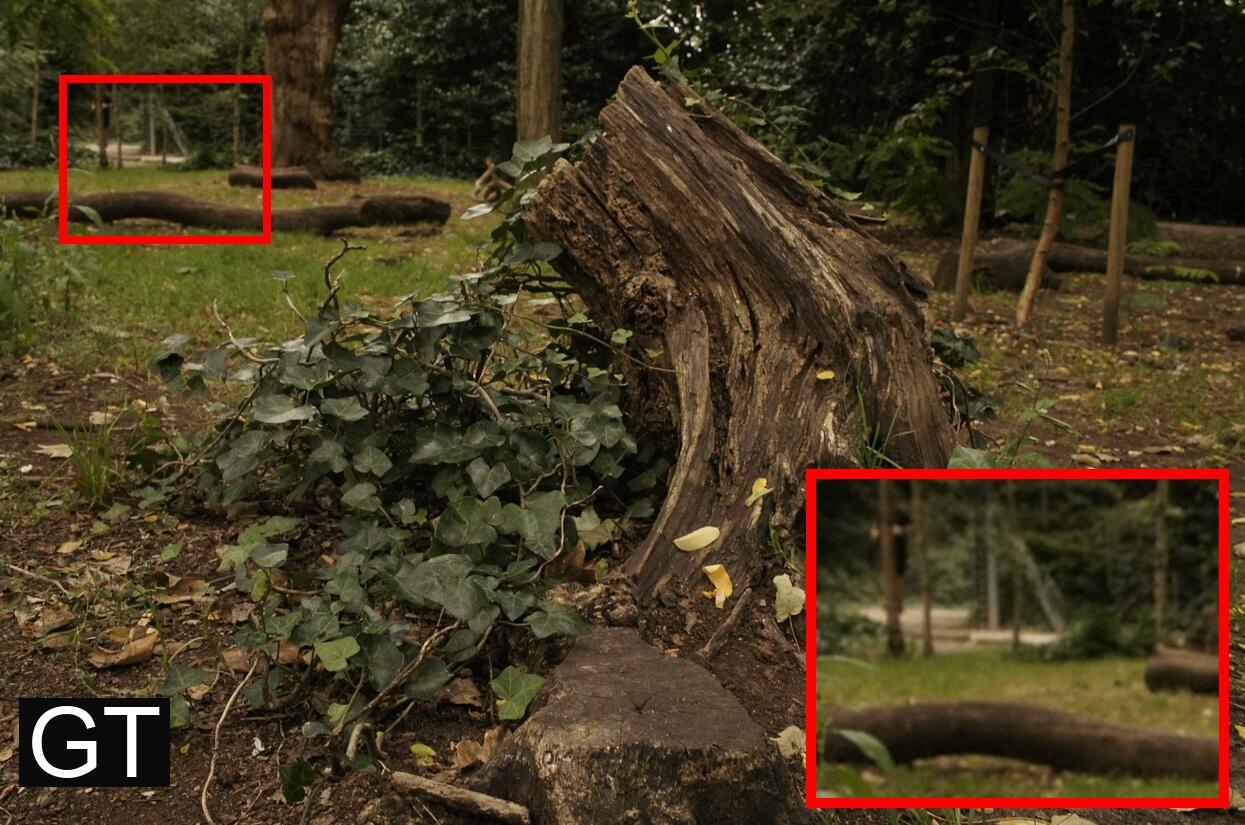}\hfill
  \includegraphics[width=0.2485\linewidth]{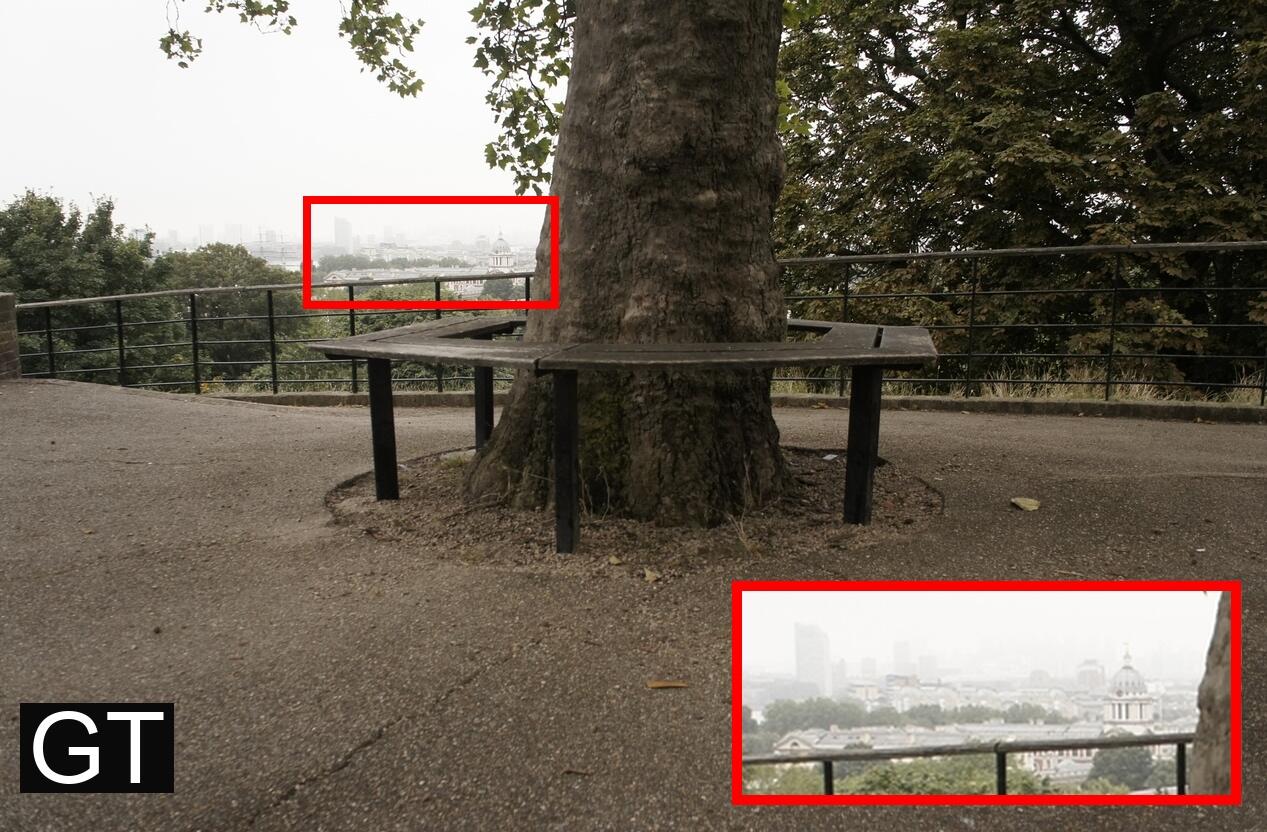}\hfill
  \includegraphics[width=0.2485\linewidth]{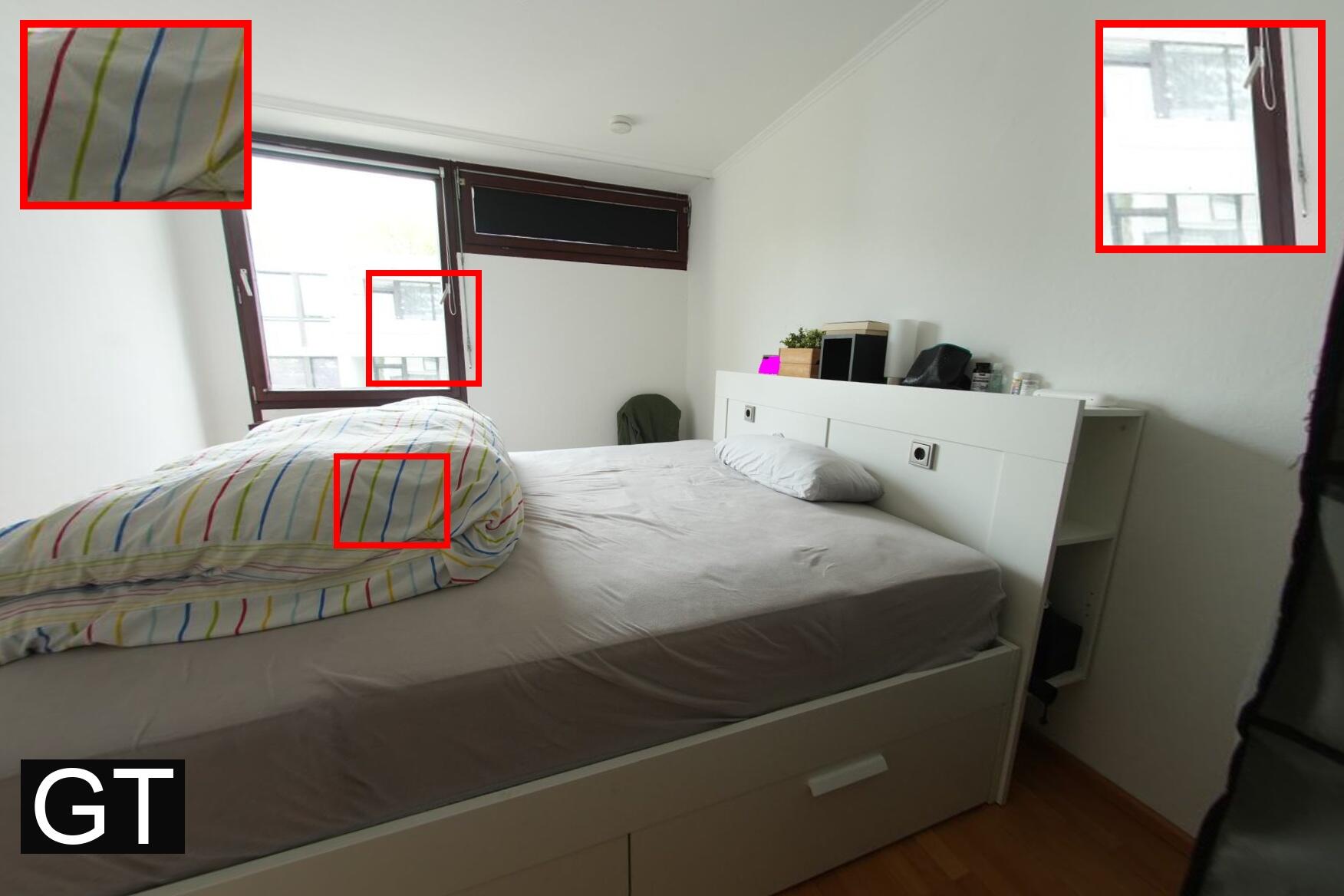}\hfill
  \includegraphics[width=0.2485\linewidth]{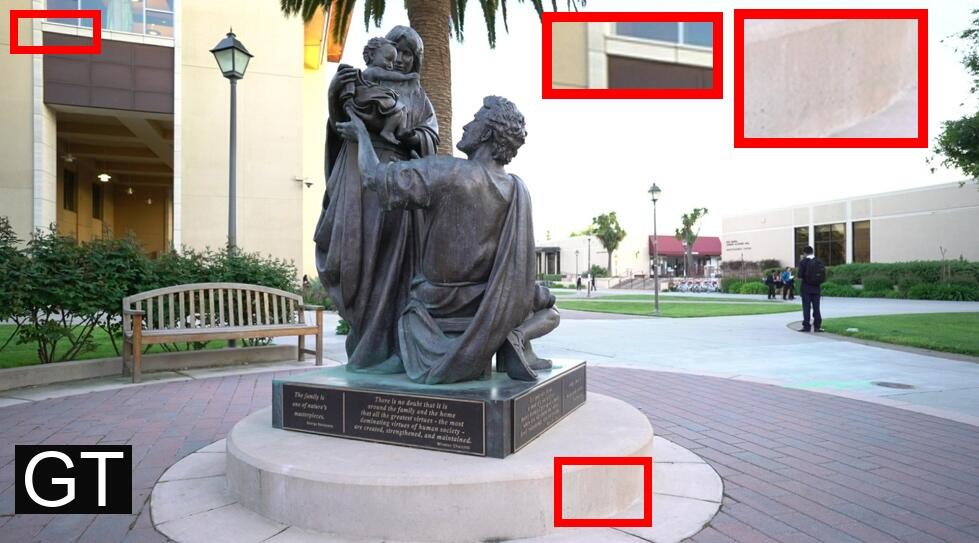}\hfill

  \caption{Qualitative results using IDHFR densification and the practical initialization methods evaluated in the paper. The depicted scenes are (in columns, left to right):
  (1) MipNerf360 - ``Stump'', (2) MipNerf360 - ``Treehill'', (3) ScanNet++ (default split) - ``bcd2436daf'', (4) Tanks \& Temples - ``Family''.
  }
  \label{fig:qualitative_init_idhfr}
\end{figure}

\begin{figure}[tbp]
  \centering

  \includegraphics[width=0.2485\linewidth]{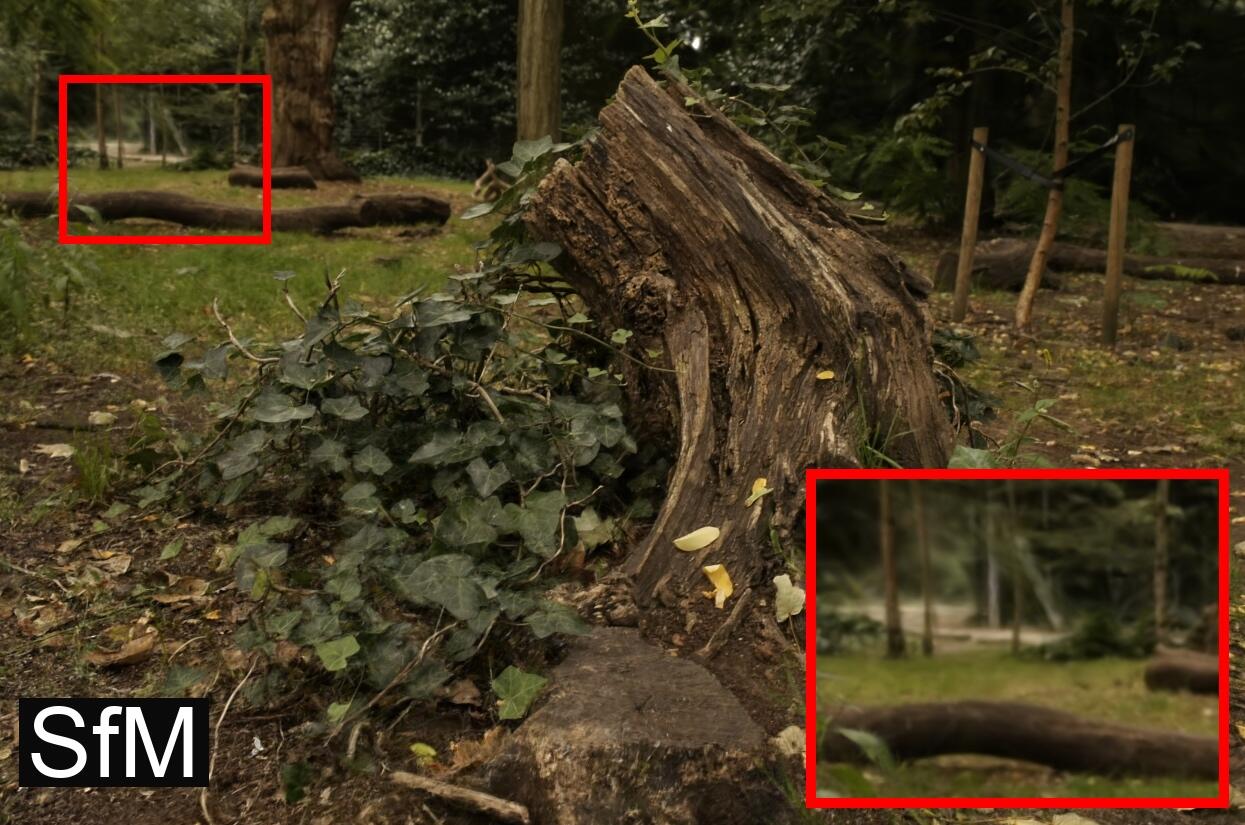}\hfill
  \includegraphics[width=0.2485\linewidth]{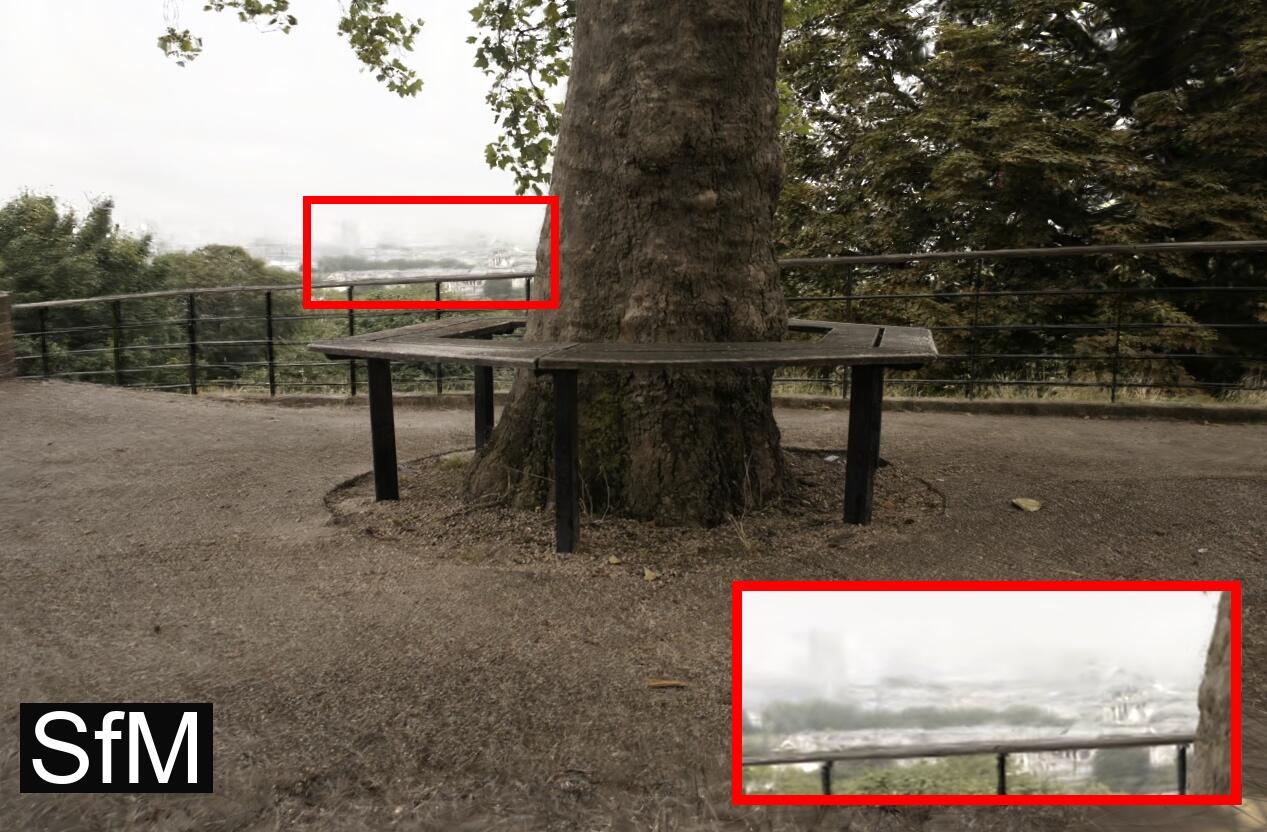}\hfill
  \includegraphics[width=0.2485\linewidth]{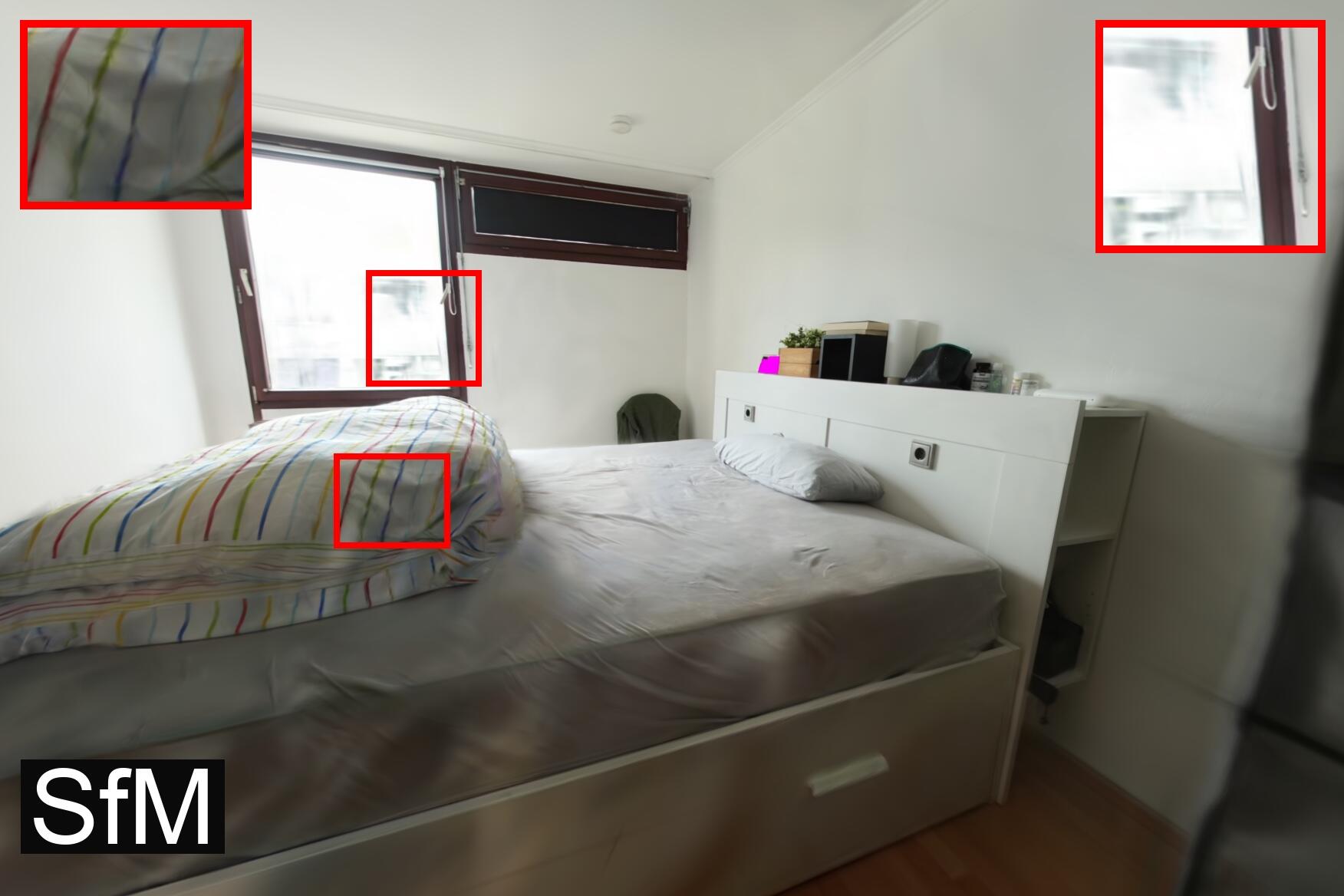}\hfill
  \includegraphics[width=0.2485\linewidth]{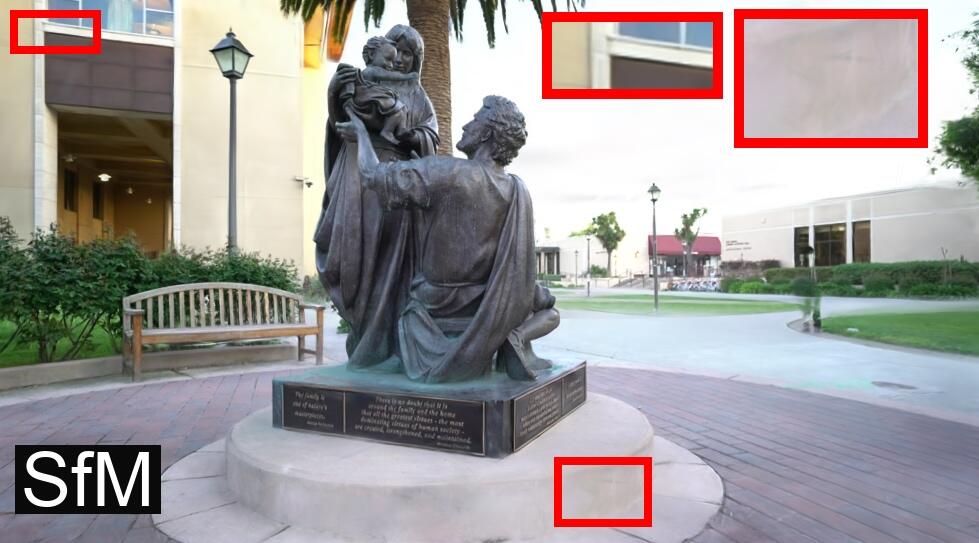}\hfill

  \includegraphics[width=0.2485\linewidth]{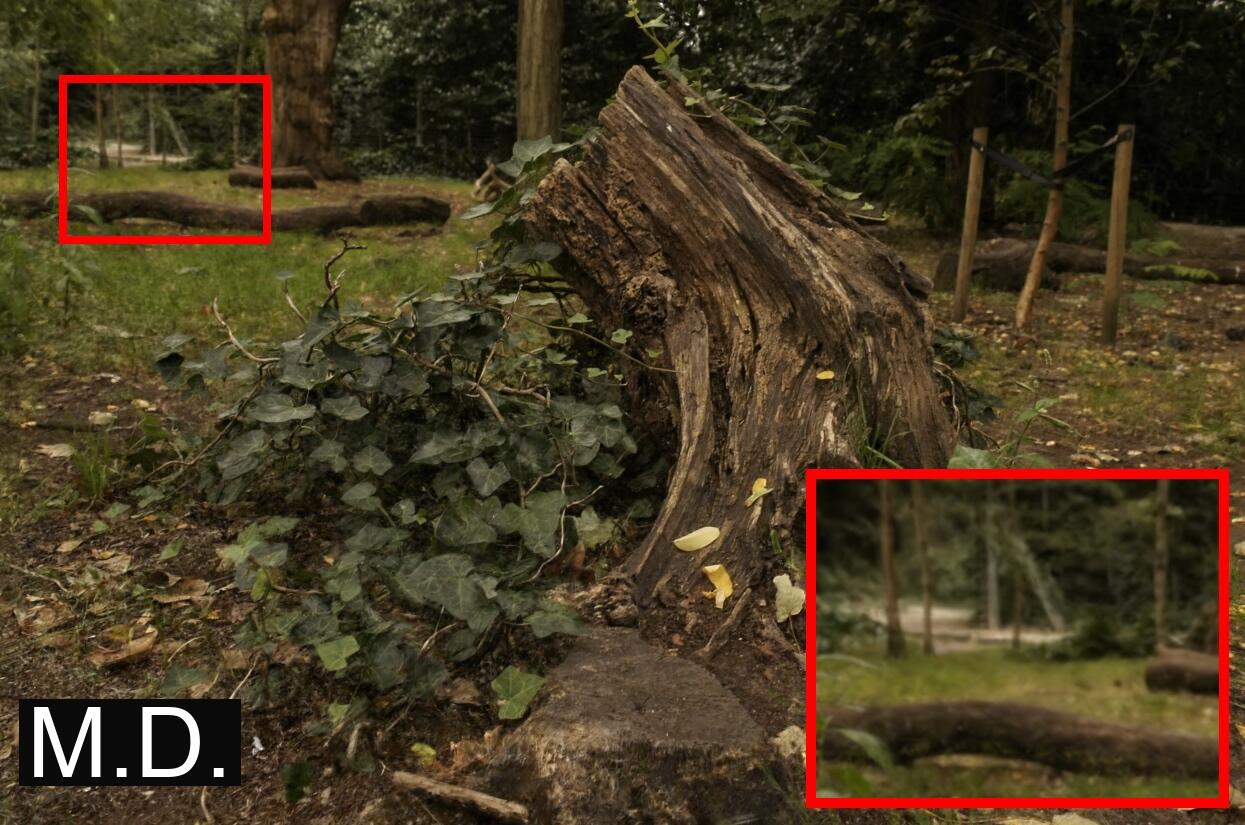}\hfill
  \includegraphics[width=0.2485\linewidth]{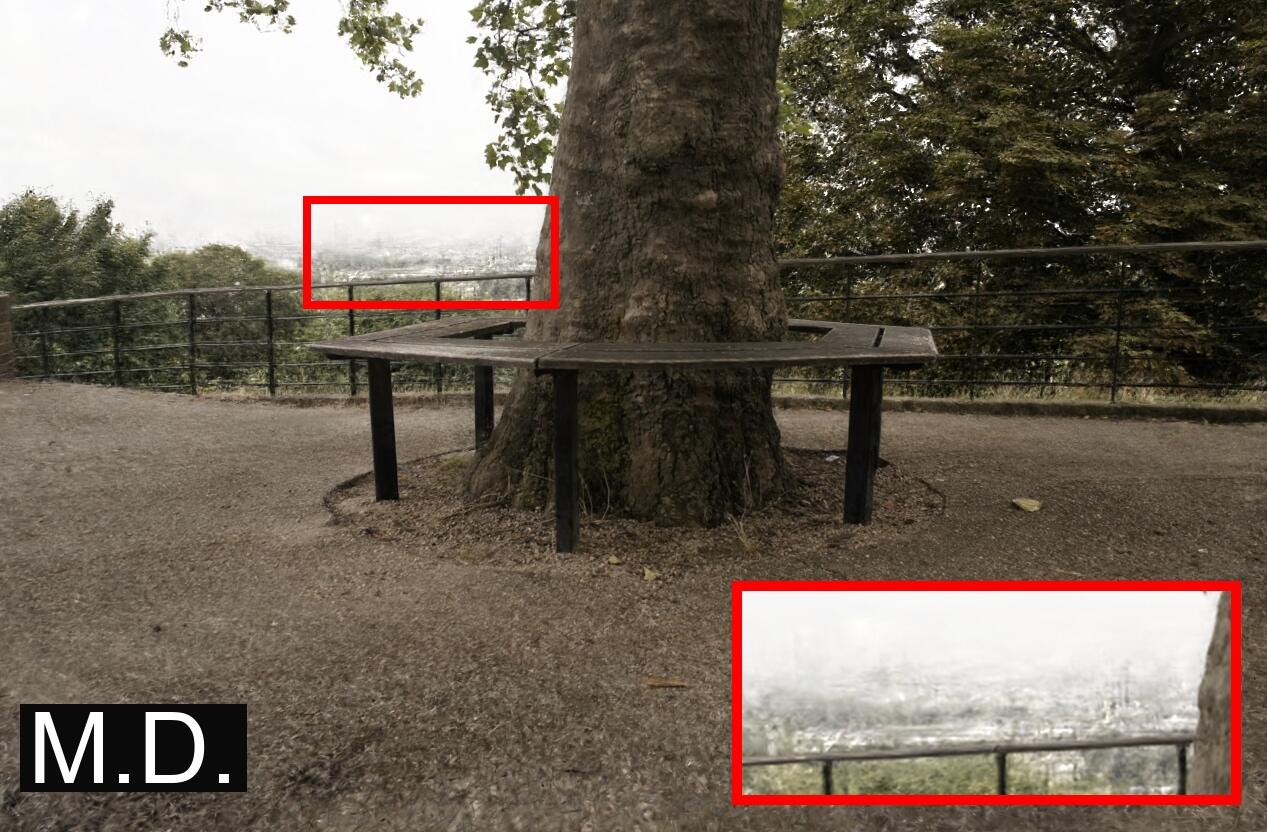}\hfill
  \includegraphics[width=0.2485\linewidth]{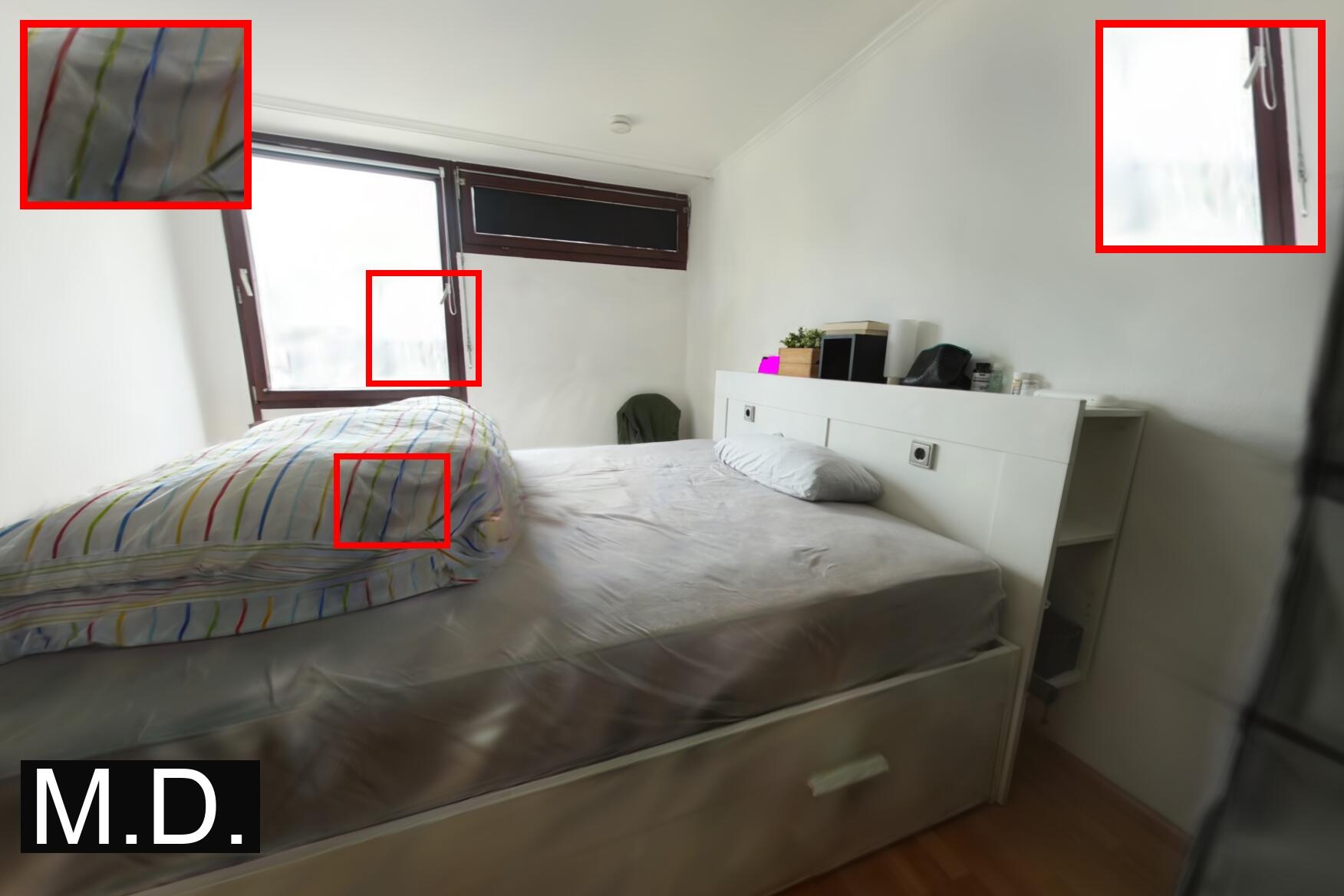}\hfill
  \includegraphics[width=0.2485\linewidth]{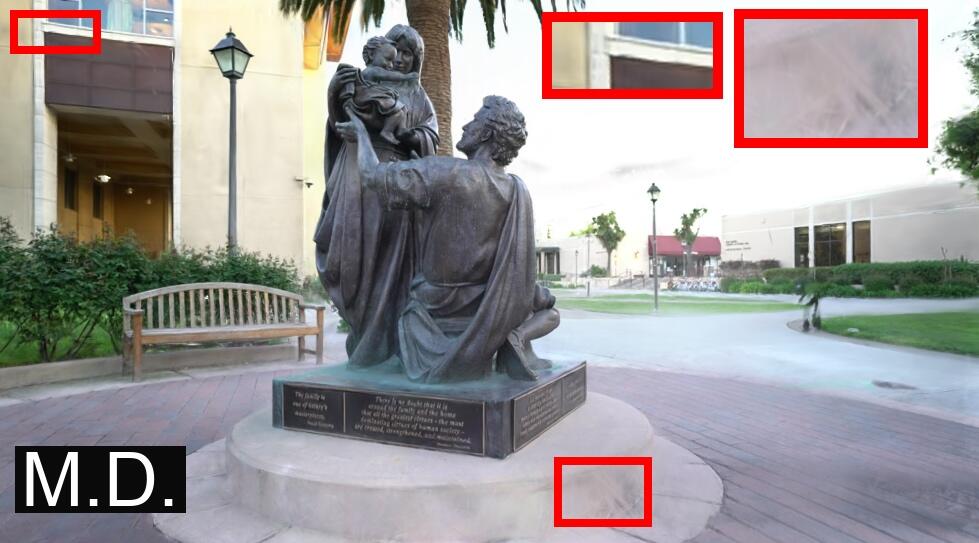}\hfill

  \includegraphics[width=0.2485\linewidth]{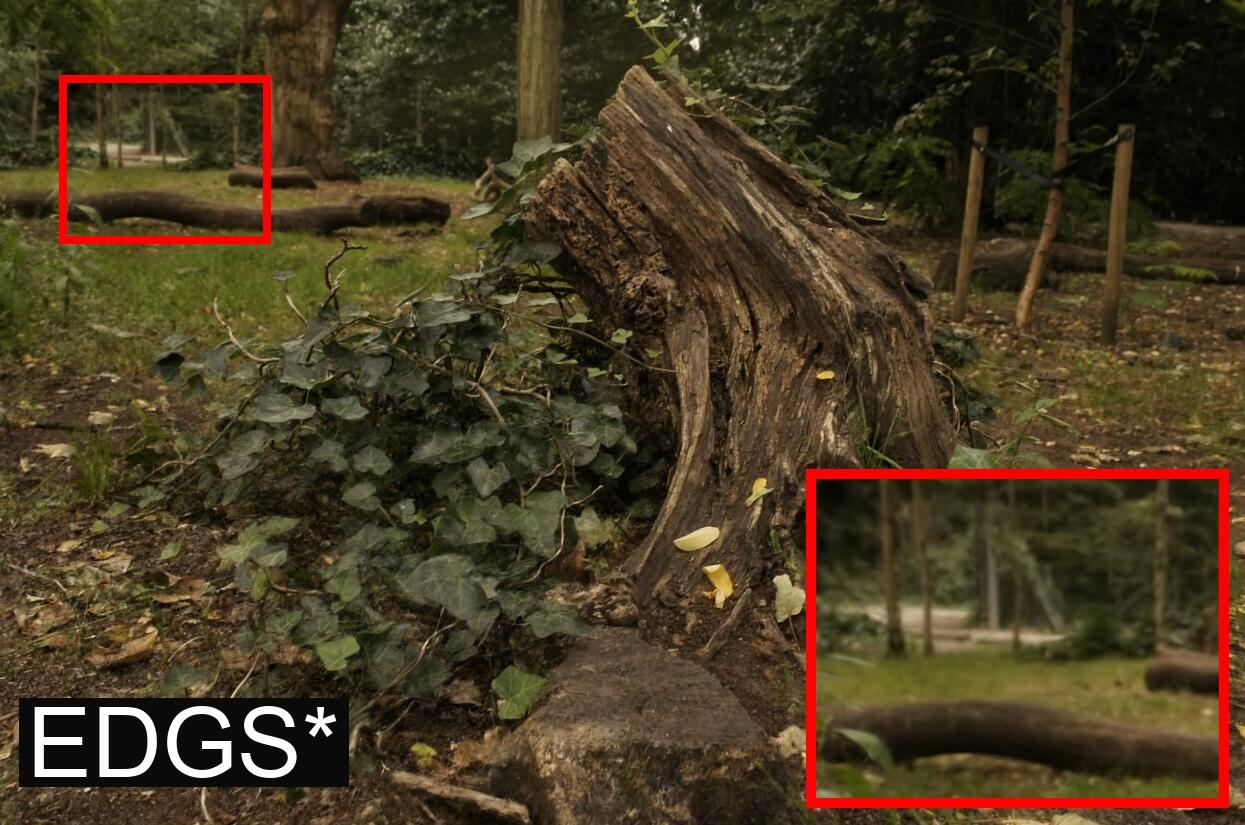}\hfill
  \includegraphics[width=0.2485\linewidth]{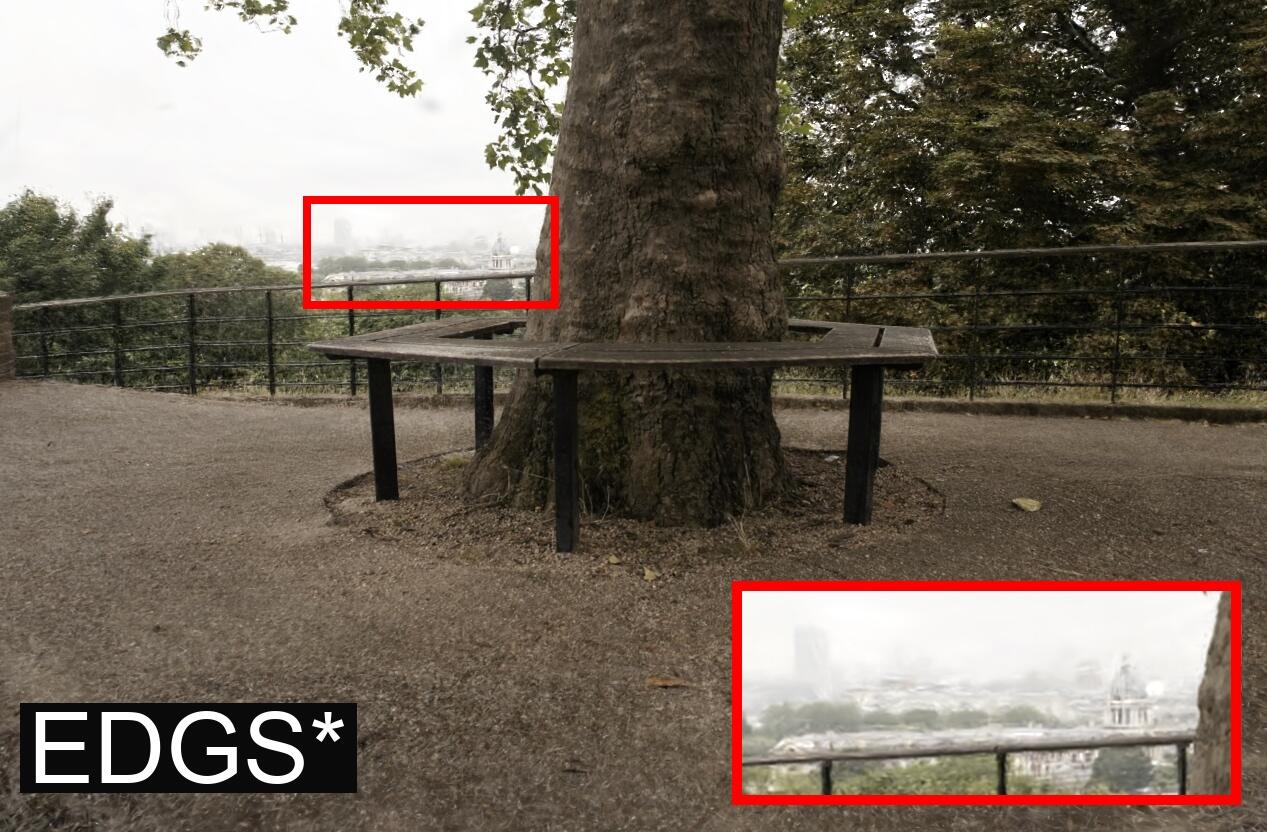}\hfill
  \includegraphics[width=0.2485\linewidth]{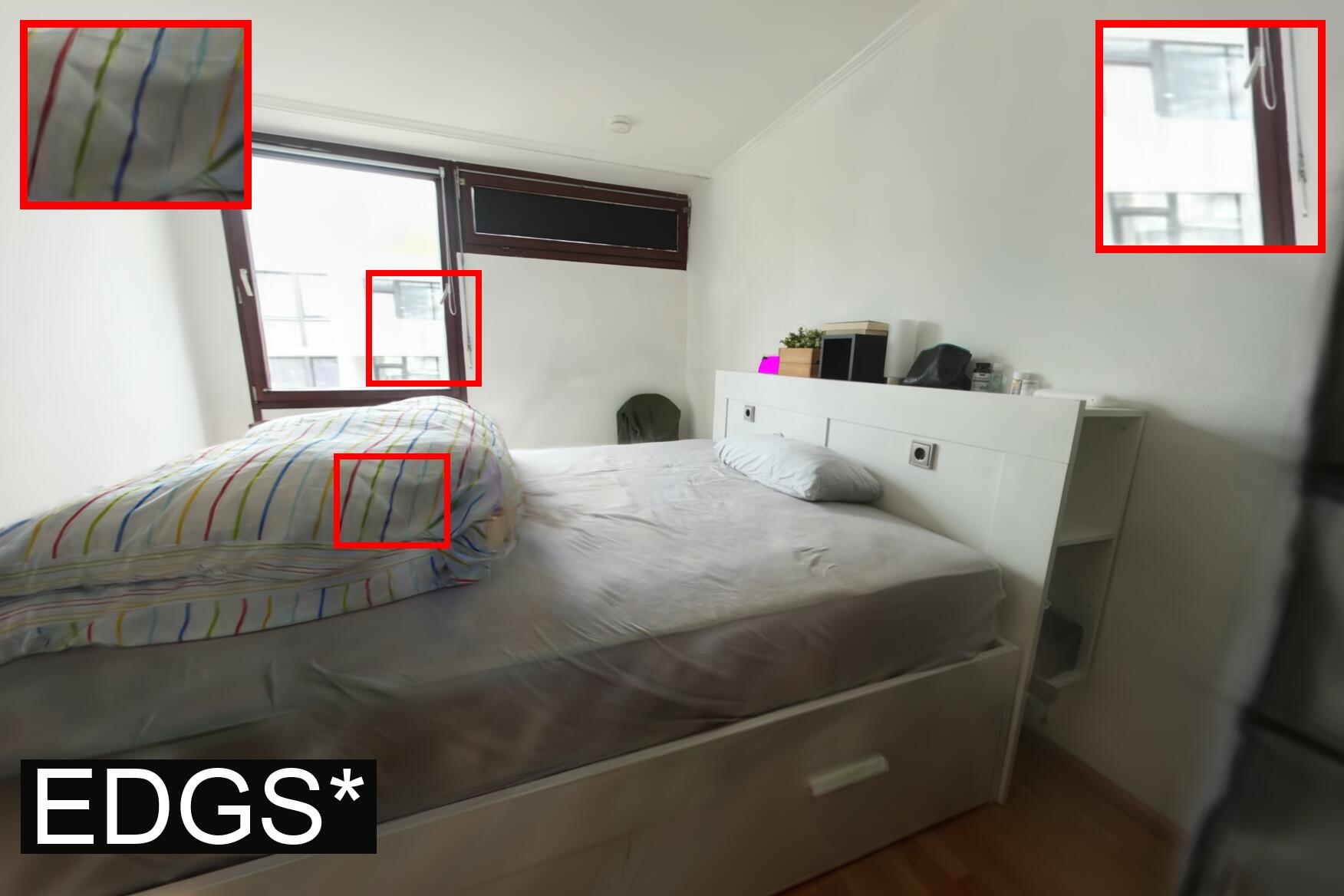}\hfill
  \includegraphics[width=0.2485\linewidth]{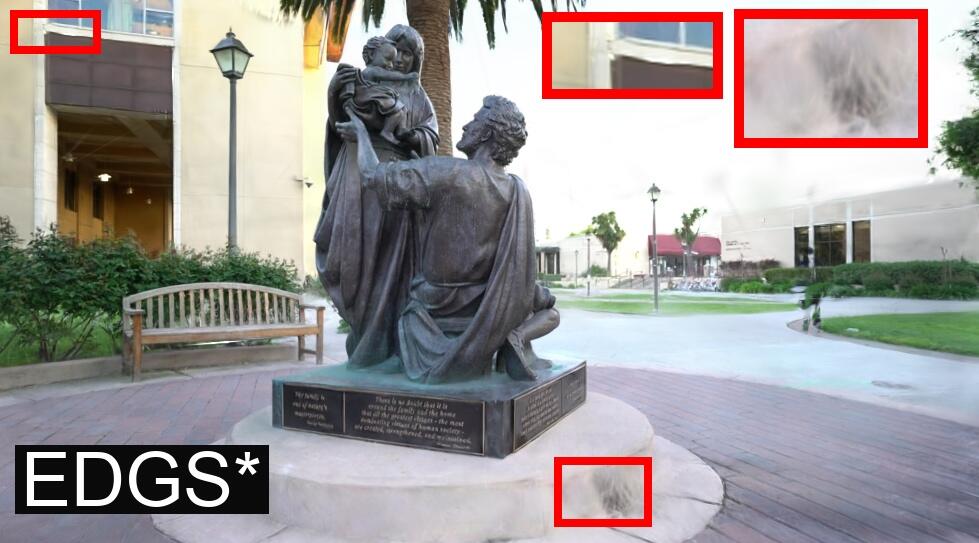}\hfill

  \includegraphics[width=0.2485\linewidth]{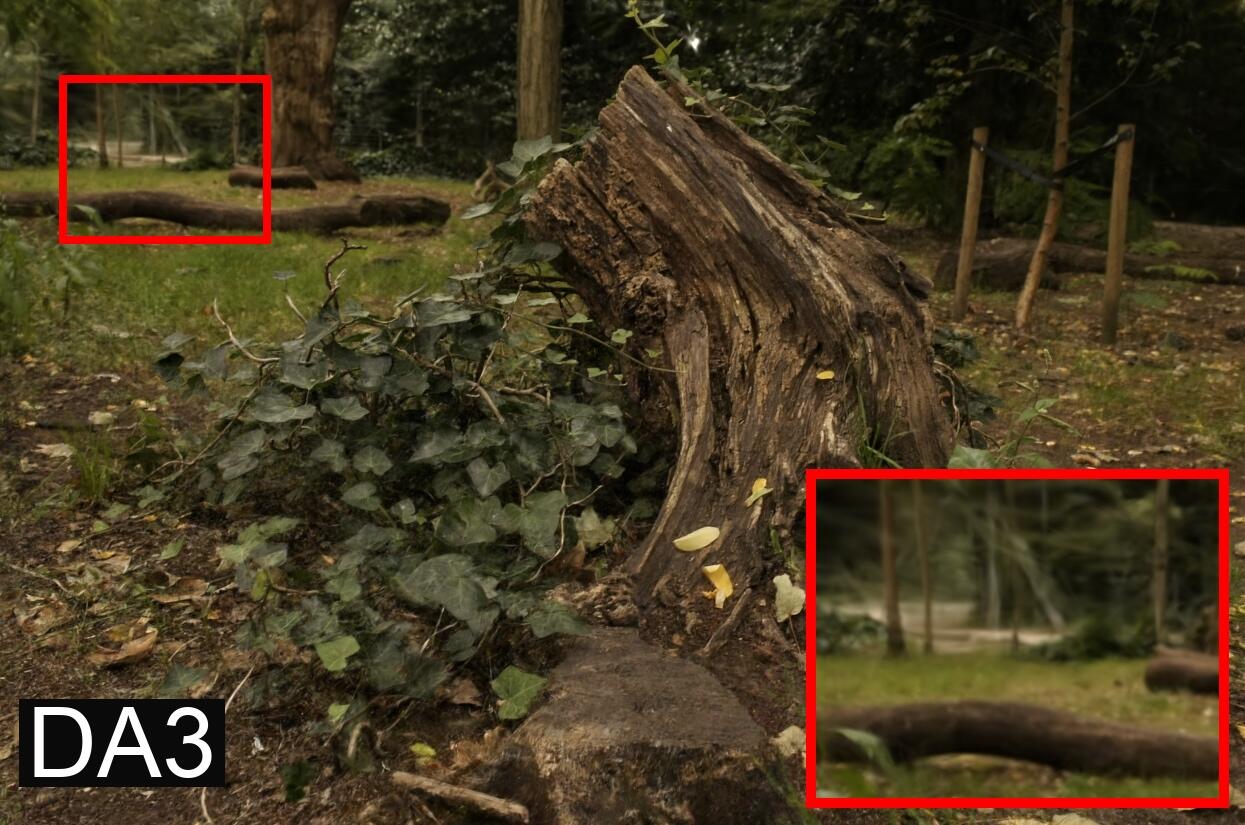}\hfill
  \includegraphics[width=0.2485\linewidth]{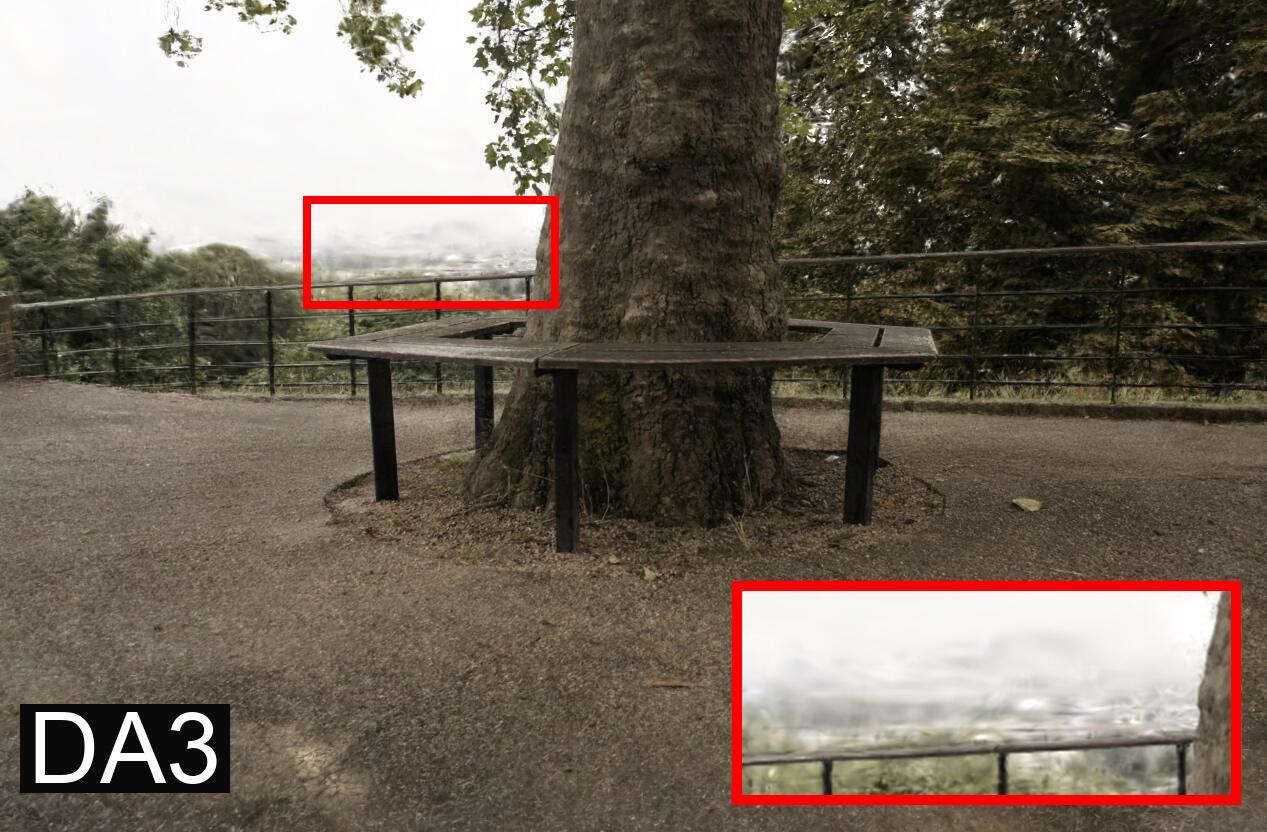}\hfill
  \includegraphics[width=0.2485\linewidth]{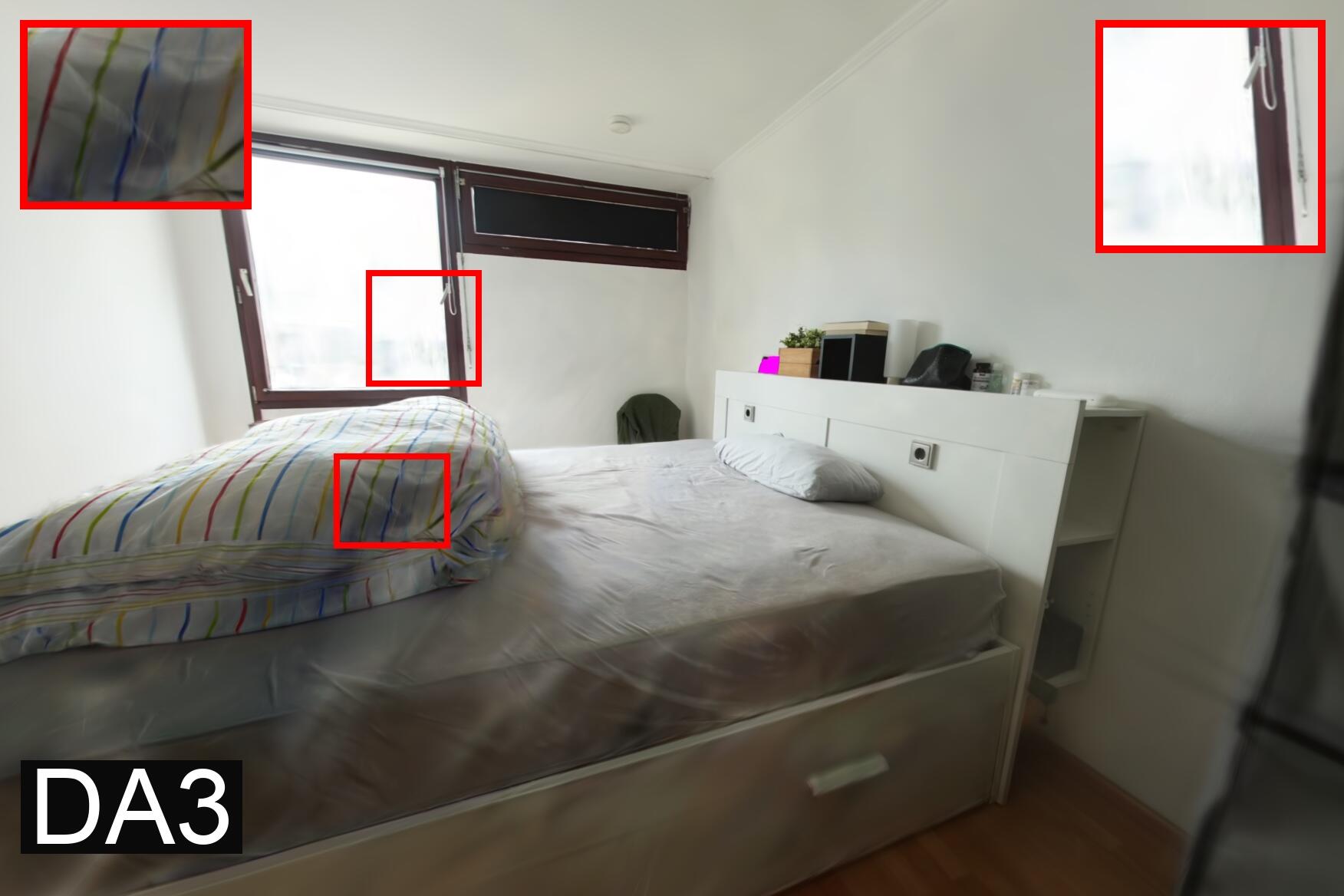}\hfill
  \includegraphics[width=0.2485\linewidth]{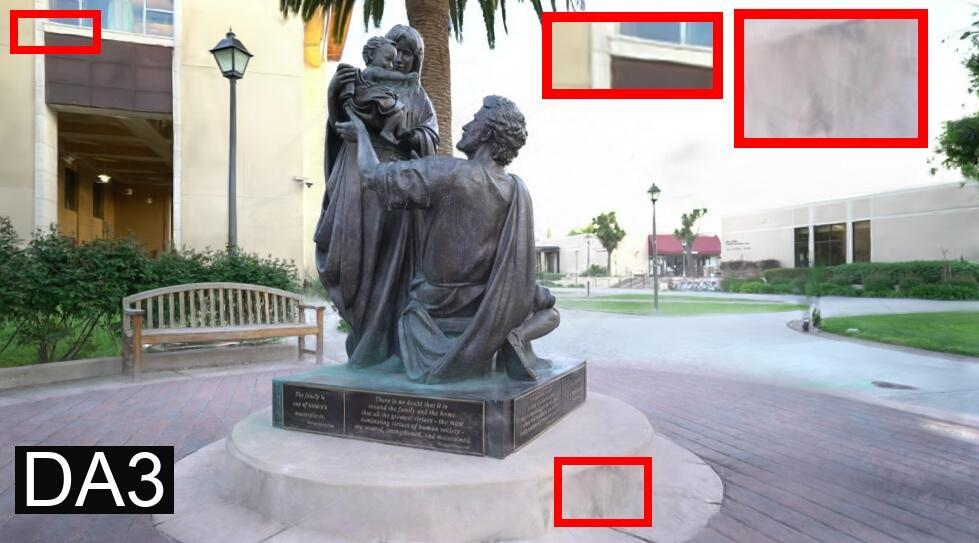}\hfill

  \includegraphics[width=0.2485\linewidth]{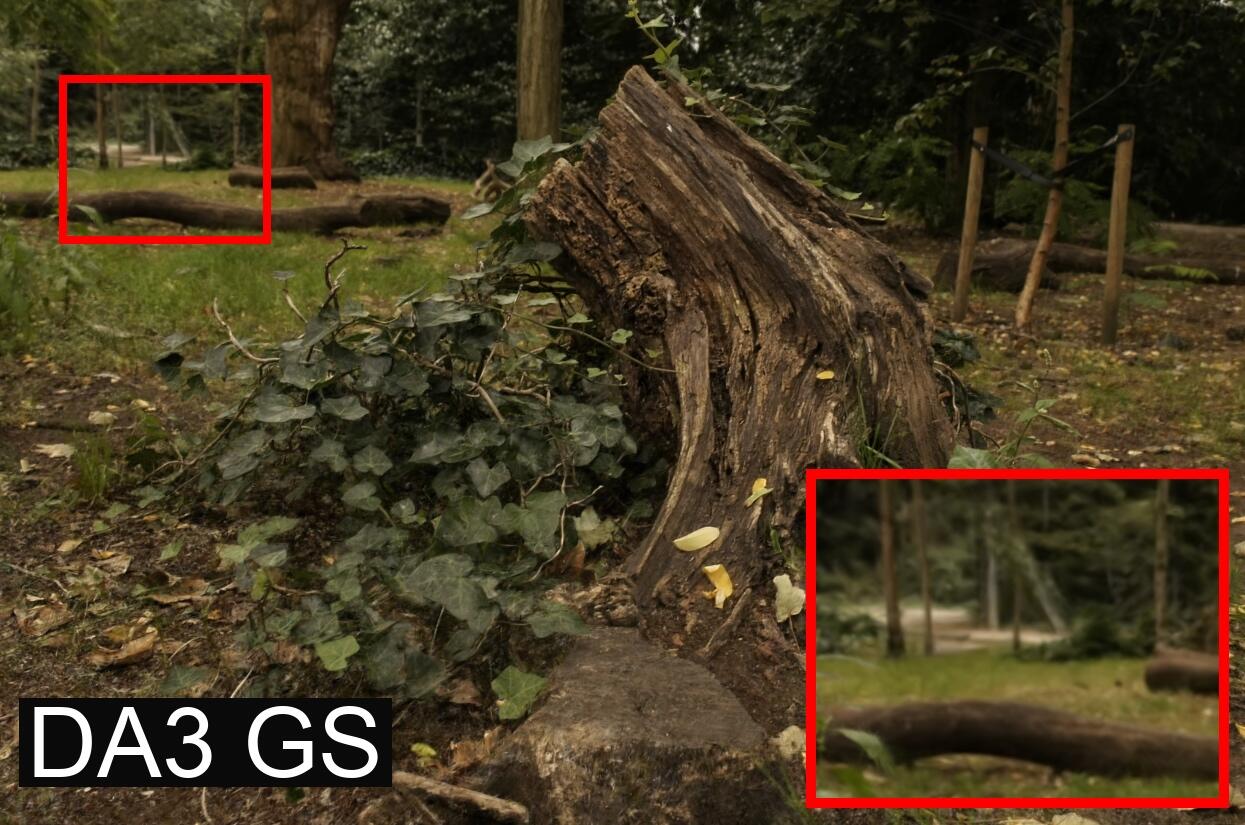}\hfill
  \includegraphics[width=0.2485\linewidth]{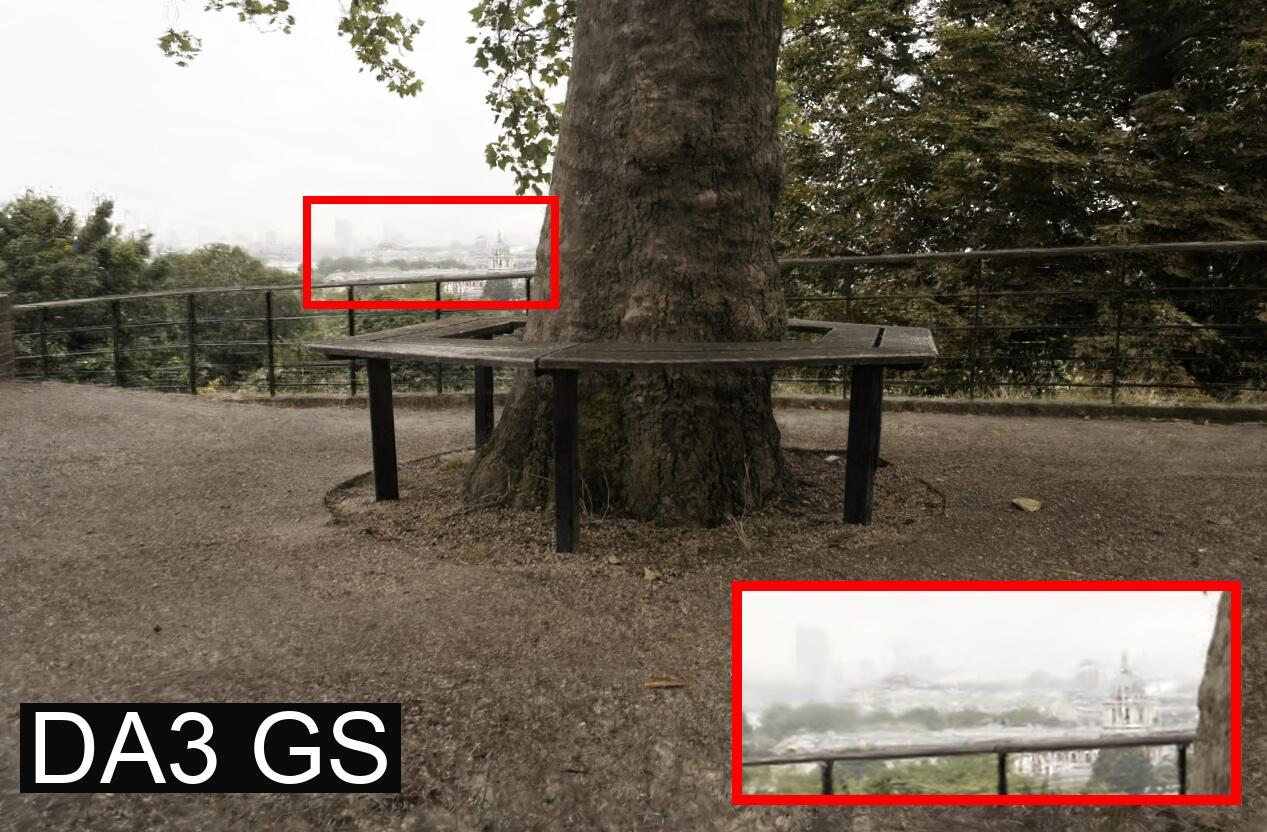}\hfill
  \includegraphics[width=0.2485\linewidth]{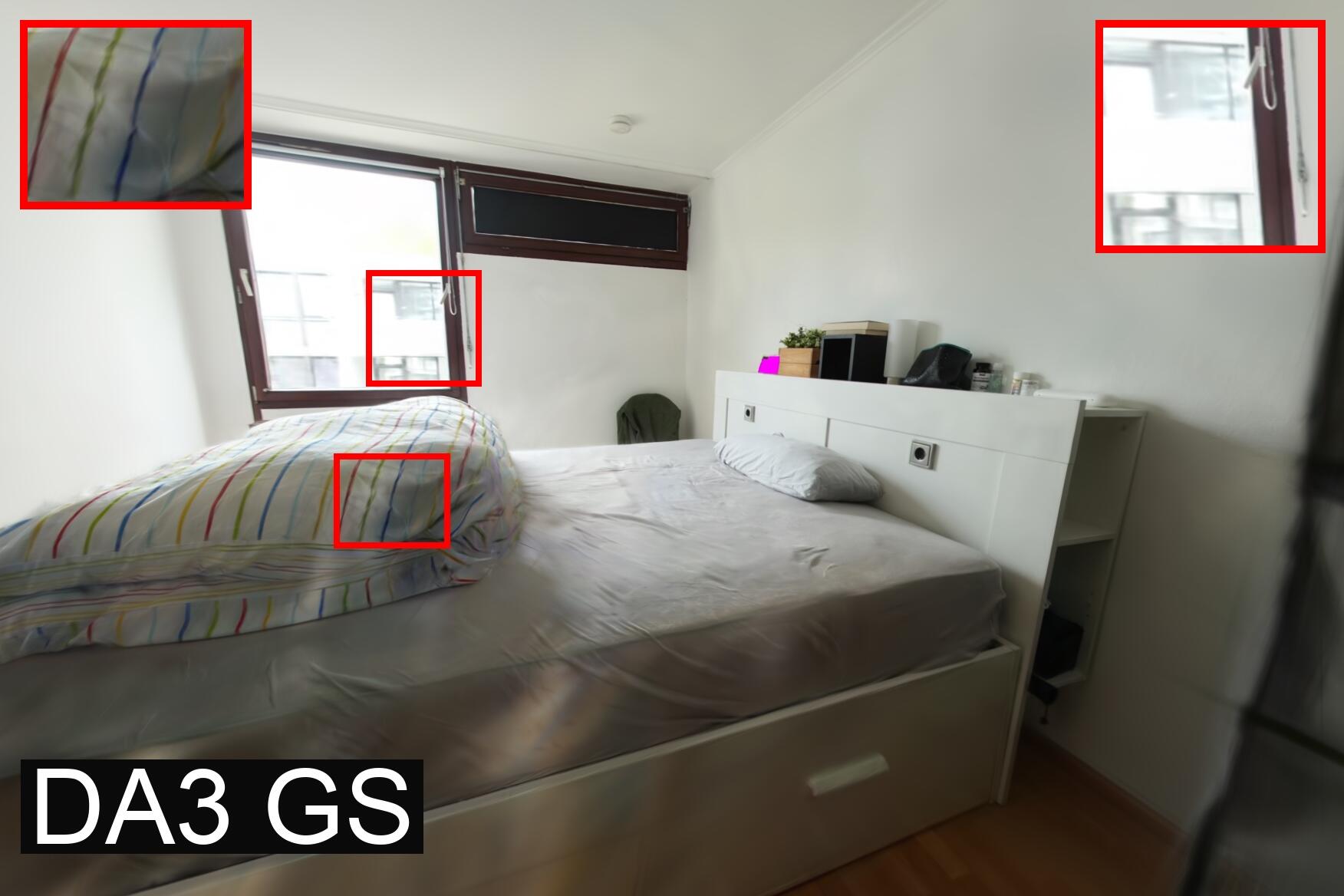}\hfill
  \includegraphics[width=0.2485\linewidth]{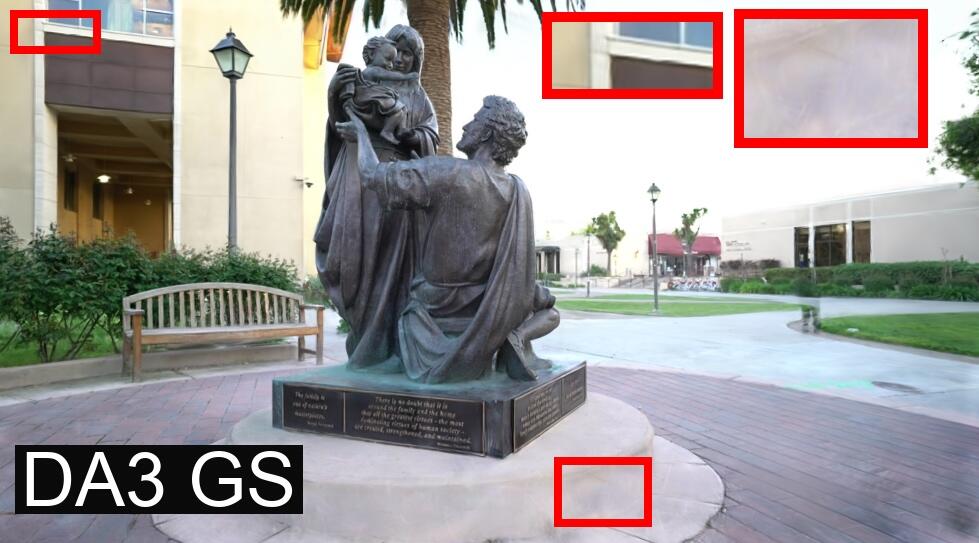}\hfill

  \includegraphics[width=0.2485\linewidth]{images/supp/init_methods_jpeg/mn360_stump/zoomed_text_gt-color.jpg}\hfill
  \includegraphics[width=0.2485\linewidth]{images/supp/init_methods_jpeg/mn360_treehill/zoomed_text_gt-color.jpg}\hfill
  \includegraphics[width=0.2485\linewidth]{images/supp/init_methods_jpeg/scannetpp_bedroom/zoomed_zoomed_text_gt-color.jpg}\hfill
  \includegraphics[width=0.2485\linewidth]{images/supp/init_methods_jpeg/tnt_family/zoomed_zoomed_text_gt-color.jpg}\hfill

  \caption{Qualitative results using MCMC densification and the practical initialization methods evaluated in the paper. The depicted scenes are (in columns, left to right):
  (1) MipNerf360 - ``Stump'', (2) MipNerf360 - ``Treehill'', (3) ScanNet++ (default split) - ``bcd2436daf'', (4) Tanks \& Temples - ``Family''.
  }
  \label{fig:qualitative_init_mcmc}
\end{figure}

In \Cref{fig:qualitative_init_idhfr,fig:qualitative_init_mcmc}, we provide additional qualitative results achieved using 
the best performing densification strategies in our experiments -- IDHFR and MCMC, and all of the practical initialization methods evaluated in the paper.
The differences are mostly visible in regions of the scene that are not often visible in training views,\footnote{For outside-in captures this includes far away background areas, as they are often visible in only a fraction of the views.}
where having more initial points reduces the burden on densification and improves detail recovery.

\section{Lists of Scenes Per Dataset}
In this section we provide the exact list of scenes used for evaluation with each dataset. On \textbf{MipNerf360}, all scenes are used, including the ``Flowers'' and ``Treehill'' scenes.
\begin{description}
    \item[ScanNet++:] c5439f4607, bcd2436daf, b0a08200c9, 6115eddb86, f3d64c30f8, 3f15a9266d, 5eb31827b7, 3db0a1c8f3, 40aec5fffa, 9071e139d9, e7af285f7d, bde1e\-479ad, 5748ce6f01, 825d228aec, 7831862f02.
    \item[Tanks \& Temples:] auditorium, ballroom, palace, temple, family, horse, lighthouse, m60, train, barn, caterpillar, church, meetingroom, truck.
    \item[ETH3D:] pipes, kicker, terrace, relief, relief\_2, terrains, office.
\end{description}

\end{document}